\documentclass[preprint,review,12pt]{elsarticle}
\usepackage{epsfig}
\usepackage{verbatim}
\usepackage{longtable}
\usepackage{amssymb}
\usepackage{amsfonts}
\usepackage{amstext}
\usepackage{amsmath}
\usepackage{epstopdf}
\usepackage{graphicx}
\usepackage{subfigure}
\usepackage{float}
\usepackage{IEEEtrantools}
\usepackage{multirow,color}
\usepackage{gensymb}
\usepackage{lmodern}
\usepackage{titlesec}
\usepackage{url}
\newbox{\bigpicturebox}
\usepackage{lineno}
\usepackage{setspace}
\usepackage{ulem}
\usepackage{scalerel}
\usepackage[export]{adjustbox}
\usepackage[dvipsnames]{xcolor}
\usepackage{lineno}

\usepackage{rotating}

\newcommand{\FEI}
{{\sf \uppercase{F}\kern-0.20em\uppercase{E}\kern-0.20em\uppercase{I}}}

\newcommand{\x}{\mathbf{x}}

\newcommand{\Teta}{\boldsymbol{\Theta}}

\journal{Engineering Structures}

\titleformat{\paragraph}
{\normalfont\normalsize\bfseries}{\theparagraph}{1em}{}
\titlespacing*{\paragraph}
{0pt}{3.25ex plus 1ex minus .2ex}{1.5ex plus .2ex}

\begin{document}

\begin{frontmatter}


\title{Machine learning based surrogate modeling with SVD enabled training for nonlinear civil structures subject to dynamic loading}

\author[label1]{Siddharth S. Parida\corref{cor1}}
\ead{paridas@erau.edu}

\author[label2]{Supratik Bose}
\ead{supratik@buffalo.edu}

\author[label1]{Megan Butcher}
\ead{butcherm3@my.erau.edu}

\author[label3]{Georgios Apostolakis}
\ead{georgios.apostolakis@ucf.edu }

\author[label4]{Prashant Shekhar}
\ead{shekharp@erau.edu}


\cortext[cor1]{corresponding author}

\address[label1]{Department of Civil Engineering, Embry-Riddle Aeroautical University, Daytona Beach, FL, USA}

\address[label2]{SK Ghosh and Associates (International Code Council)Springfield, IL, USA}

\address[label3]{Department of Civil, Environmental and Construction Engineering, University of Central Florida,Orlando, Fl, USA}

\address[label4]{Department of Mathematics, Embry-Riddle Aeroautical University, Daytona Beach, FL, USA}


\tableofcontents


\begin{abstract}
The computationally expensive estimation of engineering demand parameters (EDPs) via finite element (FE) models, while considering earthquake and parameter uncertainty limits the use of the Performance Based Earthquake Engineering framework. Attempts have been made to substitute FE models with surrogate models, however, most of these models are a function of building parameters only. This necessitates re-training for earthquakes not previously seen by the surrogate. In this paper, the authors propose a machine learning based surrogate model framework, which considers both these uncertainties in order to predict for unseen earthquakes. Accordingly, earthquakes are characterized by their projections on an orthonormal basis, computed using SVD of a representative ground motion suite. This enables one to generate large varieties of earthquakes by randomly sampling these weights and multiplying them with the basis. The weights along with the constitutive parameters serve as inputs to a machine learning model with EDPs as the desired output. Four competing machine learning models were tested and it was observed that a deep neural network (DNN) gave the most accurate prediction. The framework is validated by using it to successfully predict the peak response of one-story and three-story buildings represented using stick models, subjected to unseen far-field ground motions.\newline
\vspace{0.5cm}

\end{abstract}


\end{frontmatter}


\newpage

\section{Introduction}\label{section1}
The performance-based earthquake engineering (PBEE) framework, developed at \cite{Cornell:2000}, is a powerful tool that can express performance of civil infrastructures during future earthquakes in terms of many useful decision variables (DVs), such as economic losses, downtime, casualties etc. that facilitates communication between the stakeholders and design professionals. This framework relies on four successive steps that include hazard analysis, structural analysis (SA) using finite element (FE) simulation, mapping engineering demand parameters (EDPs) as obtained from SA to damage measures (DMs), and finally, decision-making by translating DMs to various decision variables (DVs). Despite being an excellent framework for performance assessment,  its use is limited in complex civil infrastructures due to high computational costs, that arises from non-linear probabilistic FE simulation of the structure, considering uncertainties in future earthquakes that require incremental dynamic analysis  and constitutive material parameters \cite{Parida2020,ZAKERESTEGHAMATI2021112971}. To alleviate this computational burden, researchers often use surrogate models to compute the EDPs. In its most common form, a surrogate model aims to derive a low computational cost mathematical function that directly maps the inputs to the finite element model to its output, by-passing the computationally expensive non-linear dynamic analysis.  Often, this mapping function is developed based on fundamental structural dynamic theory in conjunction with Newtonian mechanics. Such physics based approaches have found wide use in earthquake engineering practice due to underlying physical robustness and intuitiveness \cite{Guan_2021}. However, these methods often rely on many inherent simplistic assumptions, which can reduce the accuracy of response estimates. Contrast to physics-based approaches, data driven approaches develop the underlying mathematical mapping function based on training on a large set of input-output data.

Recently, machine learning models like logistic regression (LR),  decision trees (DT), random forest (RF), support vector regression (SVR), artificial neural networks (ANN) among others are finding  traction in the civil engineering community as viable data-driven surrogate models to predict EDPs for their ease of implementation and ability to capture high degree of non-linearity \cite{Padgett_ML_review}.  LR, RF and SVR based surrogate models have been used in Ref.\cite{ATAEI2015203} for fragility assessment of the deck unseating mode of failure for coastal bridges.  Mangalathu et al.\cite{Mang_RF} used RF to generate and update bridge specific fragility curves and ANN \cite{Mang_ANN} to derive multidimensional seismic fragility of single and two column bent, box-girder bridges.  Segura et al. \cite{Segura} used a polynomial response surface surrogate model to construct fragility surface  for efficient seismic assessment of gravity dams.  In Ref.\cite{Mang_boost} machine learning techniques with boosting algorithms is used for response prediction of reinforced concrete frame buildings.  In the aforementioned studies and many other similar works \cite{MOLLER2009432, kocamaz_binici_tuncay_2021,WU_2019, PEREZRAMIREZ2019603, AHMED2022103737, KIANI2019108},  it has been clearly demonstrated that  once the machine learning model is properly trained, it can replicate a non-linear finite element model output accurately.  It must be noted that a considerable computational effort is expended in the initial training of the machine models to replicate the non-linear behavior of the structure,  however, the value in these methods is drawn from the fact once trained subsequent estimation of non-linear prediction are computationally trivial.  Most studies in literature, however, train the data driven surrogate model for a particular suite of earthquakes.  Changing the earthquake suite would necessitate retraining the surrogate model for a new set of earthquakes, which would lead to more computational expense due to repeated training.  A few studies like in Ref. \cite{Taflanidis} use Kriging surrogate model in conjunction with stochastic ground motion models to predict non-linear structural response.  Uncertain parameters of the ground motion models and finite element models are used as inputs to the surrogate model to predict the structural response. Even though this approach is an effective solution to the aforementioned problem associated with re-training, it utilizes stochastic ground motion models which are based on phenomenologically developed empirical equations to generate the excitation that may limit its use for increased complexity, size, and nonlinearity of structures. Additionally,  the performance of Kriging models is heavily dependent on the specific variogram selected for capturing the spatial structure \cite{cressie2015statistics}, making them less robust for modeling complex  non-linear behaviors. The authors believe that in order for the surrogate model to have any meaningful application and good scalability, one needs to develop a method that uses characteristics from historical earthquake data to train the surrogate models, along with accounting for high degrees of non-linearity that may arise during an earthquake.

This research paper aims to propose a machine learning based framework that would address the problem associated with repeated training, along with being highly accurate in capturing larger degrees of non-linearity and therefore, can be easily integrated into the PBEE framework for computationally inexpensive decision making.  In effect, this is akin to solving three simultaneous problems a) identifying an exhaustive set of input parameters that are a good representation of historic earthquake data and can be used in training machine learning models, b) identifying a set of input parameters that capture sensitivity of finite element model to its parameters and c) identifying the best machine learning model that is not only able to capture the non-linearity but also variation in response that may arise due to variation in earthquakes and material parameters.  To this end, this paper first discuses a systematic framework for  development of the surrogate models in Section \ref{section2}. Different challenges that might arise for training a surrogate model in the context of earthquake engineering are discussed along with proposing an ideal workflow needed for development of surrogate model. Next, in Section \ref{section3}, a singular value decomposition based earthquake characterization technique is put forward that aims to characterize earthquakes in a way meaningful for generating training data for the surrogate model.  This is followed by a brief discussion of various machine learning models relevant for the proposed framework in Section \ref{section4}.  In Section \ref{section5}, the framework is applied to train machine learning models to replicate non-linear FE model prediction of one-story and three-story buildings subject to earthquake excitation.  The framework is validated using unseen earthquakes and material parameters which have not been used in training. 


\section{Systematic Development of the Surrogate model}\label{section2}
In this section, the authors propose a machine learning based surrogate modeling tool that can be easily integrated into the PBEE framework for computationally inexpensive decision making. The objective of the machine learning model would be to predict the EDPs as can be obtained from a FE simulation  for a given earthquake and a set of constitutive parameter values.  Since machine learning models are data-driven surrogate models, they need to be trained on a large set of data consisting of input-output ordered pair obtained from the finite element model they intend to replicate. Therefore, a systematic surrogate model framework development should incorporate the following:
\begin{itemize}
\item Choosing unique earthquake features that can be used to generate earthquake time histories needed for finite element model simulation.
\item Simulating the FE model for the constructed time histories and constitutive material parameters -- by taking a large number of combinations of values for earthquake features and material parameters -- to create a training data set.
\item Train a ``suitable'' machine learning model till a pre-selected error tolerance is achieved.
\item Validating with unseen earthquakes and material parameters that have not been used in training prior to deployment. Only if it predicts EDPs with reasonable accuracy then the machine learning model can be deployed.  
\end{itemize}
In summary,  the success of the machine learning model depends on three vital components a) selection of suitable features and generation of training data b) selection of suitable machine learning model and c) thorough validation of the model prior to deployment.  However, these steps are not trivial and have been expanded upon in the following.

\subsection{Selection of suitable features and generation of training data }\label{subsec1}
In order to train the model, one needs to create a training set of input features that should include characteristics of earthquake time history and building model parameters. Once the set of input variables are decided, the finite element model is simulated for different realizations of the input variables and the corresponding EDPs are stored. The machine learning model is trained using input-output data sets till satisfactory convergence is attained. Such a supervised learning method is quite common and has been carried out by \cite{MOLLER2009432, Mang_boost, ZAKERESTEGHAMATI2021112971}. When this supervised learning technique is extended to account for earthquake and building material uncertainty, two important problems need to be addressed. First, the input features chosen should be exhaustive for predicting EDPs with certain degree of accuracy and second, it should be possible to draw realizations of those input features and simulate the finite element model corresponding to them, in order to obtain training output. Traditionally, earthquakes have been characterized using single valued parameters like peak frequency, corresponding Fourier amplitude, peak ground acceleration (PGA), peak ground velocity (PGV), pseudo-spectral acceleration, Arias intensity, spectral moment, among others \cite{book:Kramer, Bose:2019}. From a training point of view, such characterization parameters pose a problem, since it is impossible to generate earthquake time histories for different realizations of these parameters i.e.  no inverse relationship exists between these intensity measures and earthquake time histories. This implies that such parameters are not unique to an earthquake. Not only selection of features is important, but also the existence of an inverse relation, so that enough training data can be generated, is imperative. Therefore, instead of using traditional intensity measures of earthquake characterization the authors have proposed an alternative technique for ground motion characterization as described in Section \ref{section3}.

\subsection{Selection of suitable machine learning model}\label{subsec2}
Machine learning models have been widely used in literature of earthquake engineering to tackle different problems, including surrogate modeling \cite{ATAEI2015203,MOLLER2009432, kocamaz_binici_tuncay_2021,WU_2019, PEREZRAMIREZ2019603, AHMED2022103737, KIANI2019108} . Depending on the available data,  objective and the complexity of the problem,  available computational expenditure and desired level of accuracy, various machine learning models, like the simplistic multiple linear regression,  to the more complex artificial neural networks are available to choose from. Moreover,  even for a machine learning model of choice, one would need to tune its hyperparameters that in turn affect their accuracy and efficiency drastically \cite{shekhar2020hierarchical,shekhar2020hierarchical1}. For example, for a neural network, the number of layers and neurons in each layer can be adjusted to affect its accuracy and efficiency \cite{Papadrakakis}. One can also think of the number of polynomial terms and their orders as hyperparmeters for a simple linear regression problem.  Selection of machine learning models and selection of hyper-parameters need to be systematically handled and have been discussed further in Section \ref{section4}.

\subsection{Validation of model and deployment}
Inherently machine learning is different from traditional statistics due to its focus on generalization \cite{mohri2018foundations}. Hence, instead of fitting closely to the training data, machine learning is more concerned with performing well on unseen samples. This enables deployment of such models in a real world setting. Following the same idea,  machine learning models after being trained on training data set is validated further on a data set of unseen samples, called validation dataset,  to demonstrate the effectiveness of the learnt model.  In order to effectively assess the performance of the machine learning model, the experimental setup needs to be carefully designed to restrict any information leak between the training and the validation set. 

The aforementioned discussion outlines the various steps and challenges encountered in the development of a surrogate model.  In the following section, the authors discuss a singular value decomposition based tool to solve the problem associated with earthquake feature extraction as discussed in Section \ref{section3}. This approach,  unlike traditional earthquake  characterization techniques, would help in generating a large data set of representative time histories that can be used in training machine learning models for a wide range of structural systems.  

\section{Singular value decomposition tool for ground motion characterization}\label{section3}
\begin{figure}[h!]
\begin{center}
\subfigure[\label{svd1} ]{\includegraphics[width=0.32\textwidth]{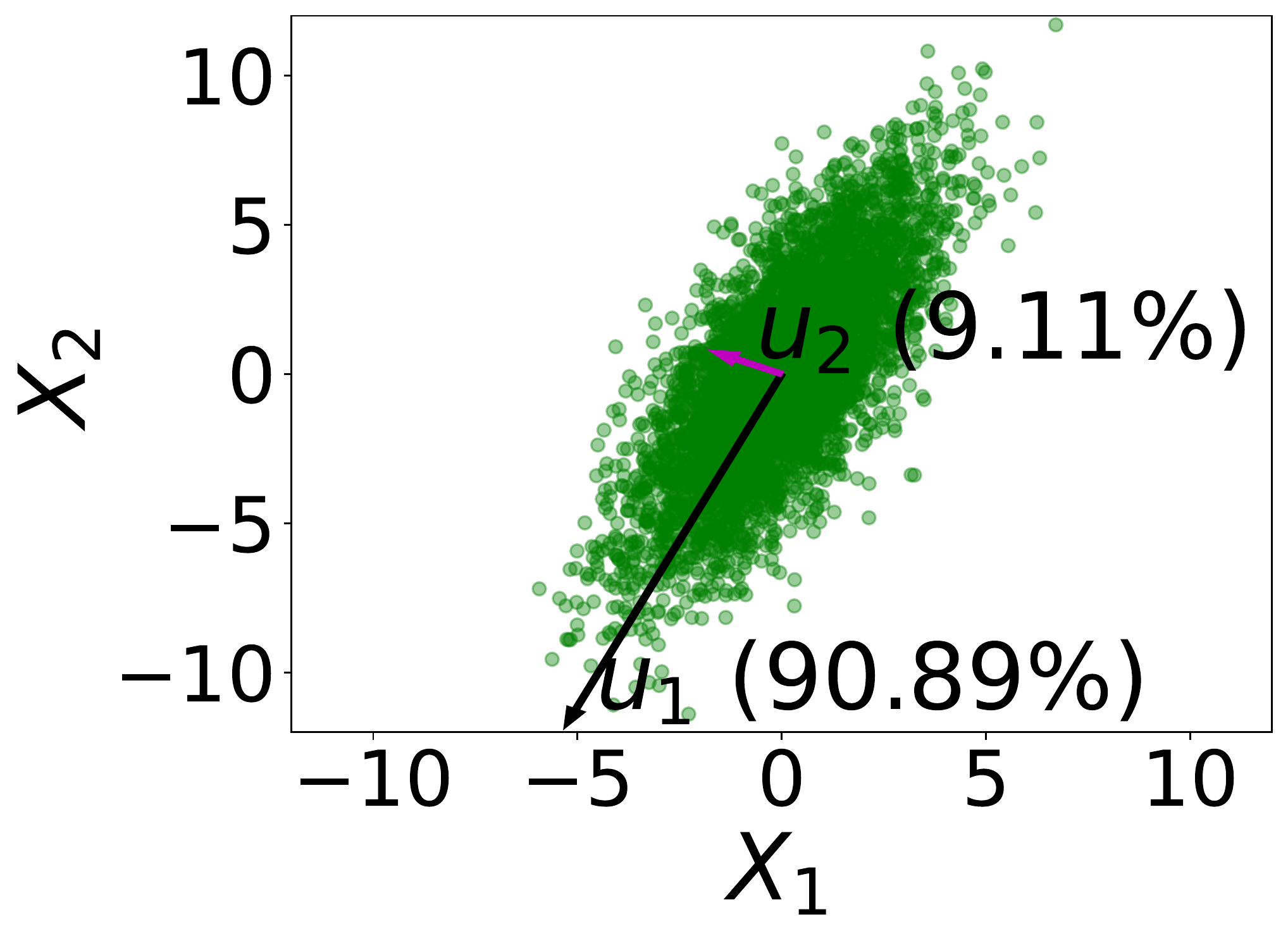}} 
\subfigure[\label{svd2} ]{\includegraphics[width=0.32\textwidth]{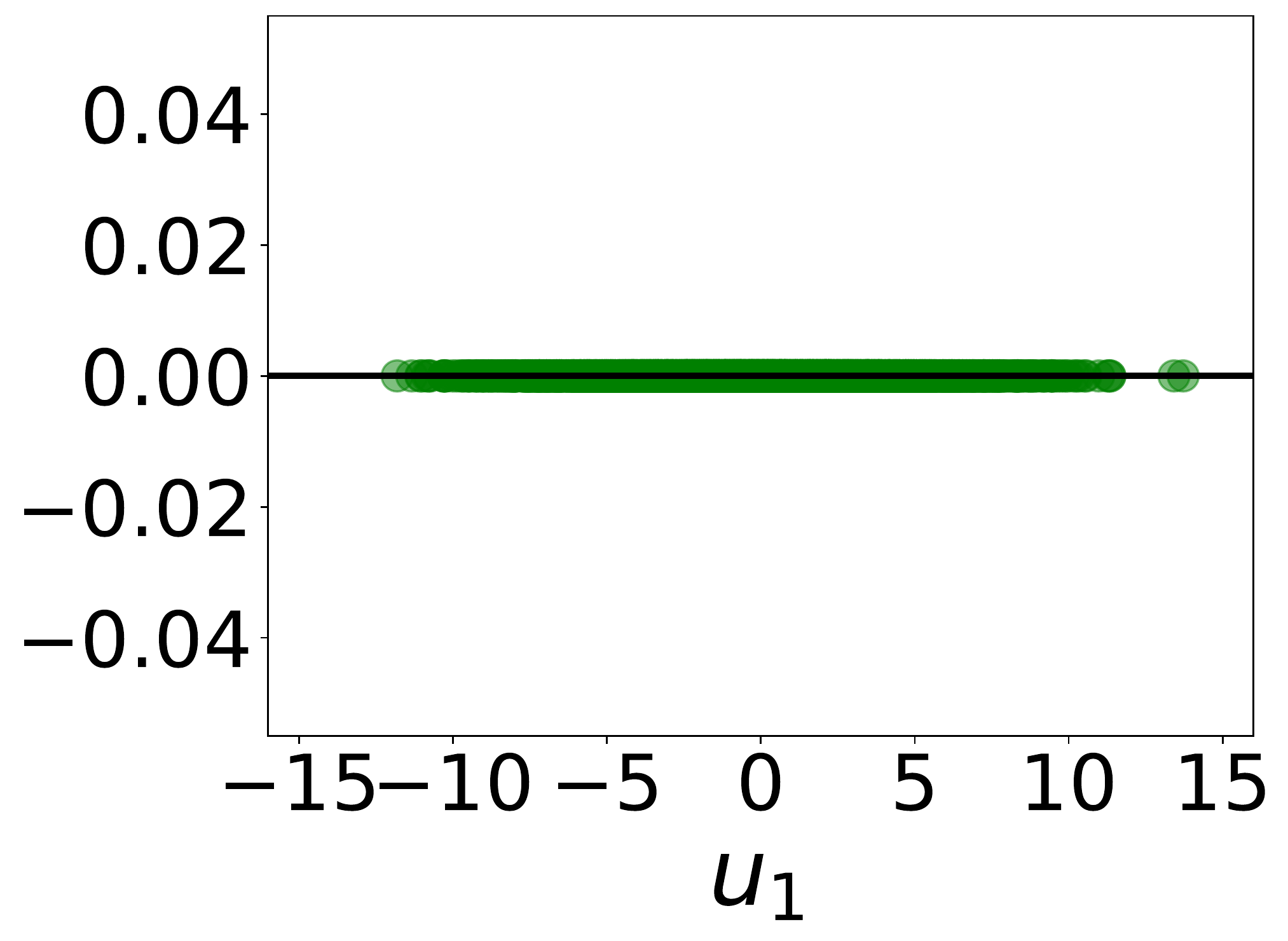}} 
\subfigure[\label{svd3} ]{\includegraphics[width=0.32\textwidth]{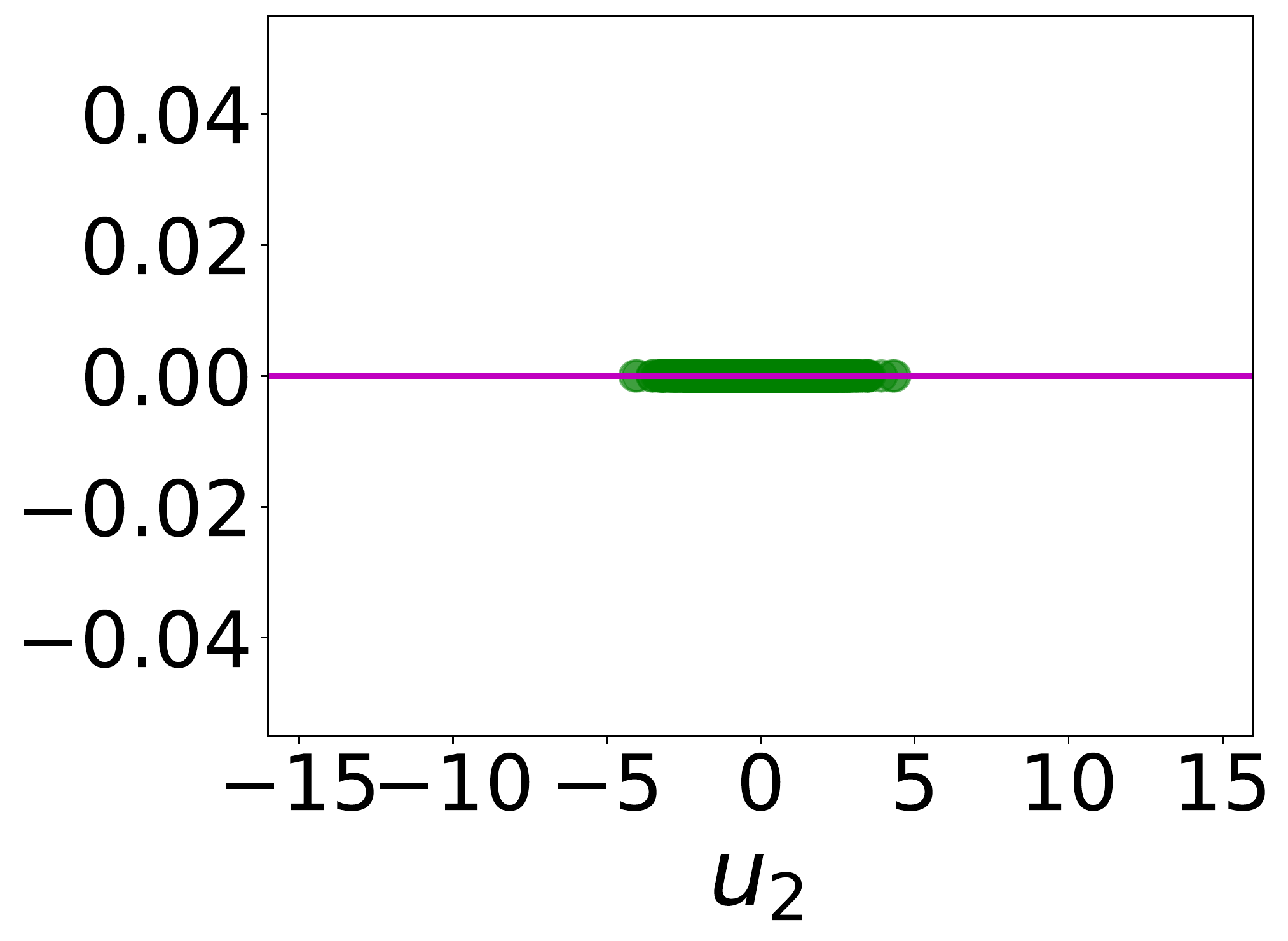}} 
\caption{(a)The directions of first 2 columns of $U$ (represented as $u_1$ and $u_2$) capturing the entire variance of a random bivariate dataset. Projection of the data set on (a) $u_1$ preserves maximum variation of the dataset while (b) $u_2$ captures the remaining small variation.}
\label{SVD}
\end{center}
\end{figure}

Consider a suite of $m$ earthquakes with $n$ times steps each. Following the theory of singular value decomposition (SVD) \cite{brunton2019data} one can represent the $n \times m$ matrix $A_{n \times m}$ of $m$ earthquakes as:
\begin{equation}\label{PCA1}
A_{n \times m}=U_{n \times m} S_{m \times m} V_{m \times m}^T
\end{equation}
where, $U_{n \times m}$ and $V_{m \times m }$ are orthogonal matrices and $S_{m \times m}$ is a square matrix containing singular values on the diagonal. The columns of $U$ form an orthogonal basis for the column space of matrix $A$. Here, multiplying matrices $S$ and $V^T$, we obtain an alternate representation for columns of $A$ when projected on columns in $U$ ( which form a $m$-dimensional basis).
\begin{equation}\label{PCA2}
A_{n \times m}=U_{n \times m} \Sigma_{m \times m}
\end{equation}
where $\Sigma$ is the matrix obtained by product of $S$ and $V^T$ matrices. Each column of $\Sigma$ consists of weights for the $u_i$'s $\in U$ to produce the original earthquake suite $A$. For example, $i^{th}$ column of $A$ can be produced by multiplying the basis matrix $U$ with $i^{th}$ column of $\Sigma$.  This gives a $m$-dimensional encoding for every earthquake (columns of $\Sigma$), and can be used as a feature representation to be fed into a machine learning model. Now, to create a training set one can generate $m$-dimensional random vectors and use it to produce corresponding time histories needed for training. Moreover, any new earthquake $P_{n \times 1}$ that the machine learning model would be used to predict EDPs for, can be projected onto the $U_{n \times m}$ basis:

\begin{equation}
\sigma_{m \times 1} = U_{m \times n}^TP_{n \times 1}
\end{equation}

\noindent where $\sigma_{m \times 1}$ is the weight vector of the earthquake $P$ when projected onto the $U$ basis. Please note that, conventionality of the input to the machine learning model increase with $m$, the number of earthquakes. However, this can be limited by intelligently selecting only a few large columns of $U$ matrix  that are relevant for an optimal representation of $A$ based on the relative magnitude of the corresponding singular values.  Figure \ref{SVD}, illustrates this point more clearly. Consider 1000 random samples in 2-dimensions with a clear `directionality' or correlation. After performing SVD of this $1000 \times 2$ matrix, it can be noted that the direction of first column of $U$ (represented as $u_1$) captures almost the entire variation in the dataset (91\%). The direction for the second column of U ($u_2$) captures the remaining variance. For many practical applications concerning this dataset, one can just consider the first column of $U$ as a sufficient univariate representation. Similar to this example,  based on the linear dependence of columns of $A$, the last few singular values in $S$ in Equation(\ref{PCA1}) might have very small magnitude and hence can be ignored. Assuming first $p$ singular values to be sufficiently large, one can only consider first $p$ rows of $\Sigma$ in (\ref{PCA2}), along with the corresponding first $p$ columns of $U$. Hence Equation (\ref{PCA2}) can be approximately written as 
\begin{equation}\label{PCA3}
A_{n \times m} \approx U_{n \times p}\Sigma_{p \times m}
\end{equation}

\noindent to obtain the $p$-dimensional encoding for every earthquake.  This $p-$ dimensional vector might now be treated as input to the machine learning model thus putting a limit to the increase of input vector size with the increase in number of earthquakes in the initial suite.

Note that SVD is a linear transformation. There are other methods like Kernel PCA, \cite{Geron} along with other highly non-linear methods such as greedy selection \cite{elad2010sparse} or autoencoders \cite{goodfellow2016deep} etc. which project the data to a set of non-linear basis. These transformations and their impact on machine learning prediction would be a part of the research teams future endeavors.  However, despite its limitations, SVD as shall be shown later is very effective in extracting features for training machine learning models. In the next section, the procedure to select a suitable machine learning algorithm is discussed. 

\section{Selection of suitable machine learning model among a competing set}\label{section4}
As mentioned earlier, selection of a machine learning model and tuning its hyper-parameters is a complex problem affecting the surrogate model's accuracy and efficiency. Therefore, a systematic approach needs to be carried out for selection of a suitable machine learning surrogate model. After going through extensive literature, four algorithms were selected in this study to be candidate surrogate models. However, it must be noted that if one deems another machine learning tool to be appropriate, that can be easily integrated with the proposed framework. A brief description of the models used here is provided in the following, while a detailed description can be found in Refs.\cite{Geron,goodfellow2016deep,hastie_09,padgett_ml}.

\subsection{Decision trees (DT) and Random forest regression (RF)} Decision trees recursively partition the input space based on local regression models.  A greedy optimization algorithm with the objective of minimizing the loss between the prediction of the tree and the actual measurement is utilized to find the optimal partition of the input space.  The number of recursive partitions can be tuned by adjusting the maximum depth upto which a tree can go before stopping.  If left unconstrained, the decision tree would closely fit the training data which might lead to over-fitting.  However, adjusting the hyper-parameter to a lesser depth affects model accuracy. 

To overcome potential over-fitting with a single tree without affecting accuracy of prediction one can use random forests (RF). RF creates a large cluster of DTs with different branches and depths and averages the output as its prediction.  Such a method is called \textit{bagging} where the average of a number of models result in an unbiased model. Number of DTs in the RF model and the maximum depth of each DT are two important tuning parameters that affect the efficiency and accuracy of the prediction. For ease of implementation, the trees are generally allowed to fit the data closely while over-fitting is avoided by tuning the number of trees in the cluster. DT and RF can be readily implemented using the python library Scikit-Learn \cite{scikit-learn} where the hyper-parameters, maximum depth for DT and number of trees for RF,  can be easily tuned.

\subsection{Support vector regression(SVR)} SVR uses non-linear kernel function to map data into high-dimensional feature space. In this feature space, it is expected that the data attains more ``favorable" features which is then captured by a simple linear regression.  SVR is robust to over-fitting and also provides flexibility by kernel selection.  Tuning parameters are type of kernel function that maps data into the high-dimensional feature space, regularization parameter ($\lambda$ ) and  the allowed error tolerance ($\epsilon$) for the linear regression in high dimensional feature space.  In this study, the radial basis function is selected for its flexibility in capturing high degree of non-linearity in data.  Like DT and RF,  SVR can be easily implemented in Scikit-Learn. 

\subsection{Deep neural networks (DNN)} A deep network typically consists of an input layer, an output layer and a stack of hidden layers connecting the input layer to the output layer.  The hidden layers project the inputs into spaces of different dimensions via various linear and non-linear projections, successively, to produce the model output.  One hidden layer feeds into another via a linear transformation and within a hidden layer, the output of the previous layer is undergone a non-parametric non-linear transformation. The process is repeated for all hidden layers up until the output layers.  This stage is called forward propagation. Weights of the linear transformations can be obtained by minimizing the mean square error (MSE) loss between the model predicted output and the actual data.  This is done by back propagating the loss gradient via chain rule.  In this study, the rectified linear unit (ReLU) is used as non-linear transformation within a hidden layer \cite{dahl2013improving}. This is because ReLU activation functions are more efficient and are generally shown to be best for deep neural networks because of fewer vanishing gradients problems when it is back-propagated through the hidden layers as compared to other activation functions like sigmoid \cite{dahl2013improving,hochreiter1998vanishing}.  The deep neural network library PyTorch \cite{PyTorch} is used in this study for implementation of the DNN model. 

For DT, RF,  and SVR a cross validation approach is used to find optimal set of hyper-parameters that best predict the output of the FE model \cite{Geron}. Broadly, cross validation is based on the idea of holding out a portion of data and training only on a subset. The performance of the trained model is then analyzed on the held out cross-validation  set which is also a subset of the training set. This is repeated multiple times in a structured way and set of hyper-parameters that perform the best on average are selected finally \cite{mohri2018foundations}. This enables us to find the most generalizable model. In the case of DNN, the optimal values of number of neurons and layers were decided heuristically by running several different number of neurons and layer combinations as suggested in \cite{Papadrakakis}.  Among the tuned machine learning models, the best performing one is selected on the basis of some performance metric.

\begin{figure}[!htb]
\begin{center}
\includegraphics[width=\textwidth]{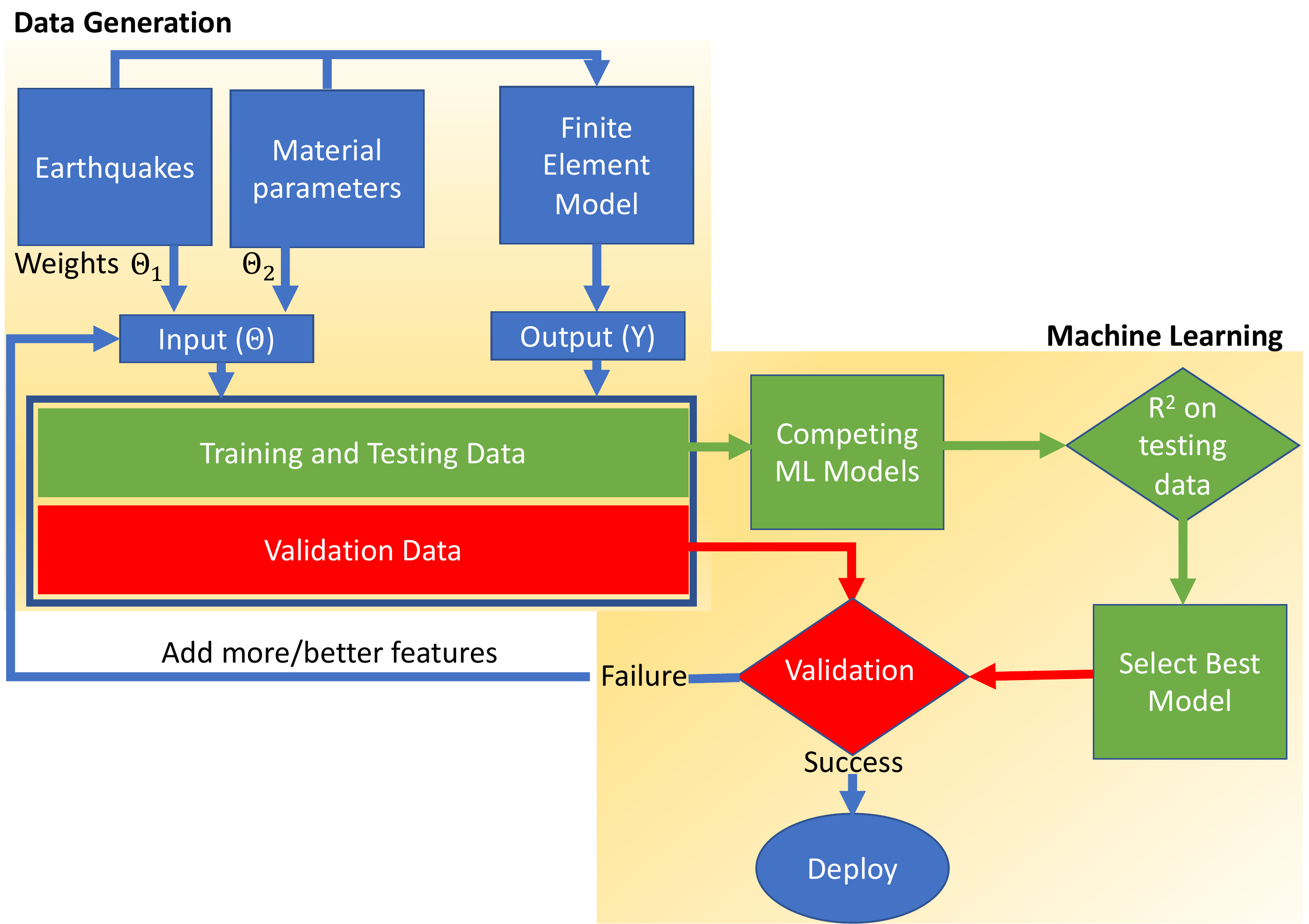}
\end{center}

\caption{Flowchart summarizing the proposed framework.}
\label{flowchart}
\end{figure}

Once the best machine learning model is chosen, its performance should be tested on a carefully designed validation data set. The data set should comprise of input output order pairs that has not been previously seen while training the model. If the machine learning model predict well for such an unseen validation data set it can be deployed for use. If it does not perform well, then one needs to either add more number of features or higher quality features in the input set, $\Teta$ and retrain the model. The reader should not confuse the validation set with the cross-validation set used for hyper-parameter selection. The proposed framework has been summarized in Figure \ref{flowchart}. The application of the framework will be illustrated in the following sections. 

\section{Results and Discussion}\label{section5}

\subsection{1Dof and 3Dof FEM models}
To illustrate the proposed framework,  one-story and three-story buildings are considered,  which are modeled as spring-mass-damper system in OpenSEES \cite{opensees}.  Chopra \cite{chopra2000dynamics} indicates that in most cases,  the first mode,  or the first few modes of a multi-degree of freedom stick model of the structure can provide useful approximation of the total response of the building,  and therefore, considered here. The springs are assumed to be uniaxial bilinear steel material with kinematic hardening but no isotropic hardening, and are modeled using the Steel01 \cite{Menegotto1973MethodOA}  material in OpenSEES.  Elastic material element with specified damping stiffness was used in parallel with the non-linear spring and a damper to capture the damping behavior of the material.   For the one-story building, only one set of parallel springs are used, while for the three-story building, three set of parallel springs are used in series. The objective of the problem is to accurately predict the response of these FE models for an earthquake,  using machine learning  given a set of values for spring constitutive model parameters.

\subsection{Choice of initial suite of ground motion}
\begin{figure}[H]
\begin{center}
\includegraphics[valign=c,width=0.19\textwidth]{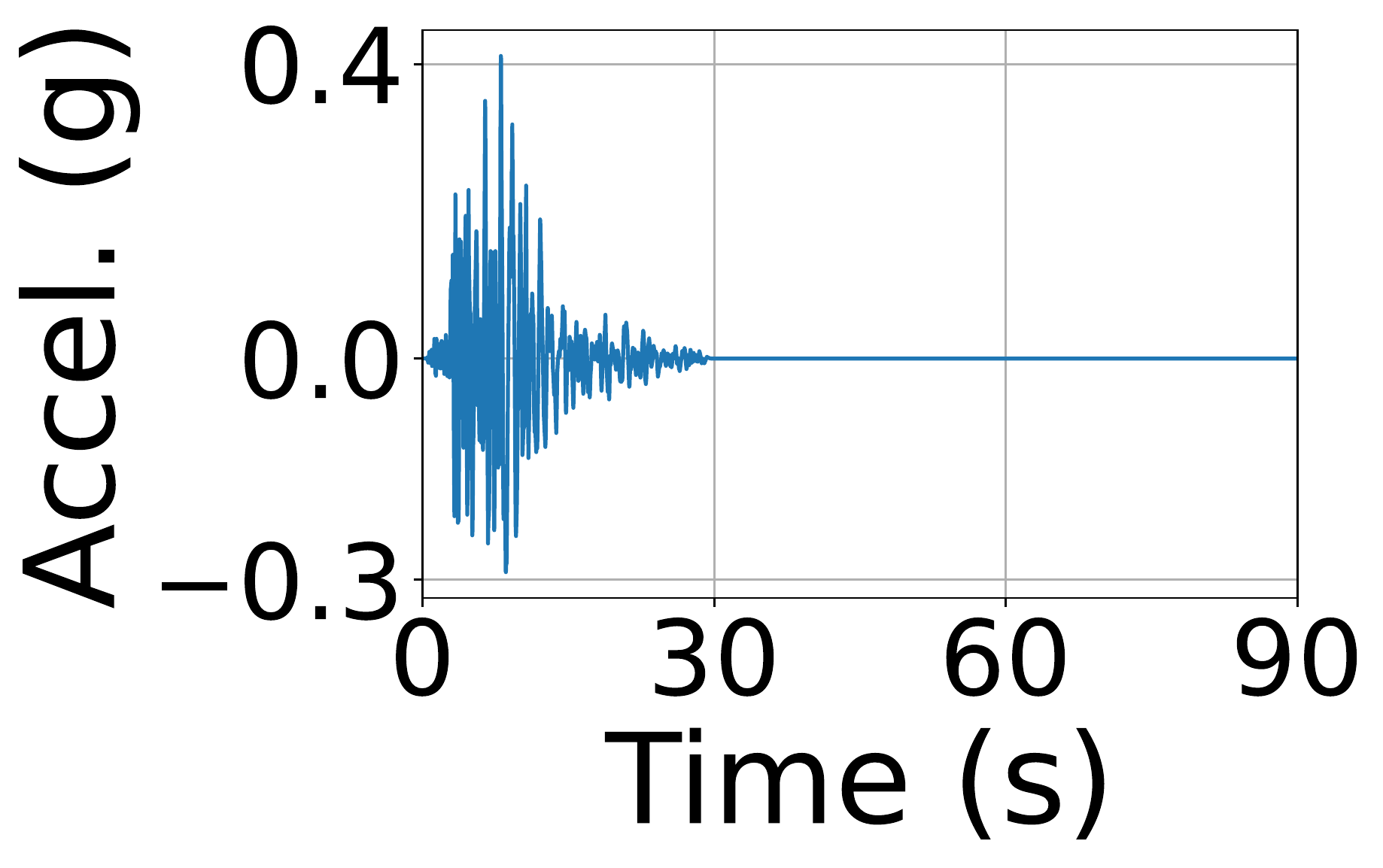} 
\includegraphics[valign=c,width=0.19\textwidth]{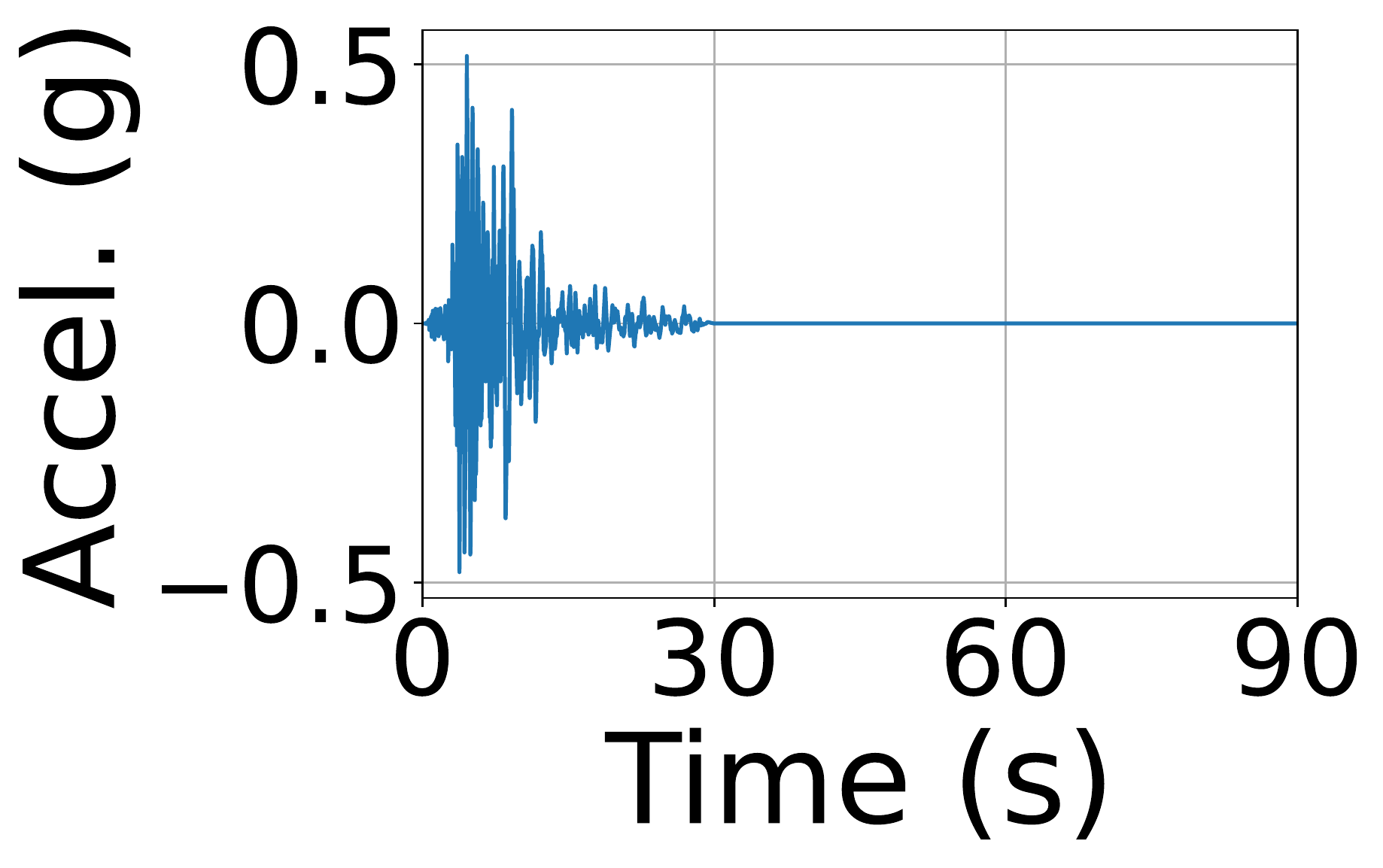} 
\includegraphics[valign=c,width=0.19\textwidth]{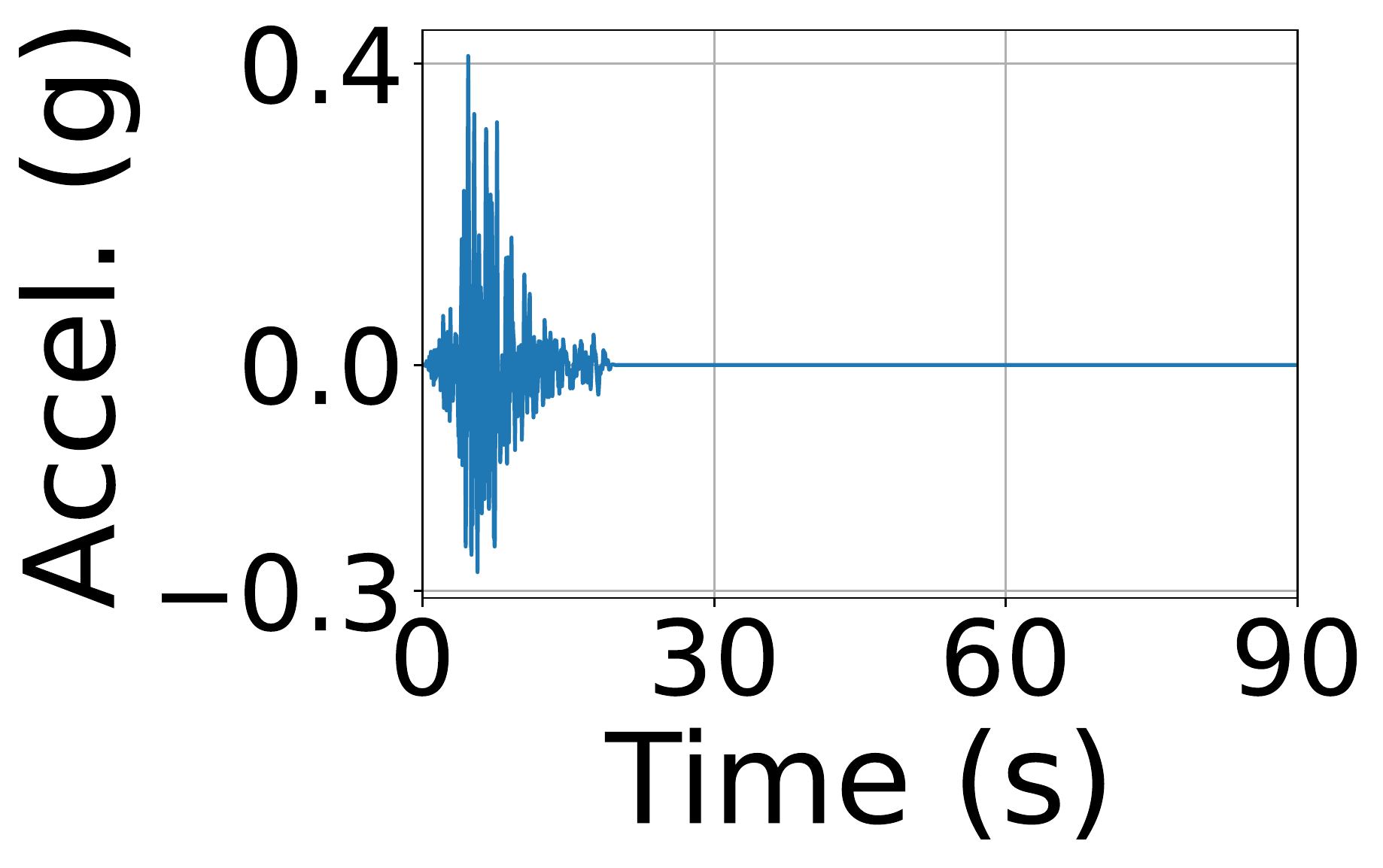}
\includegraphics[valign=c,width=0.19\textwidth]{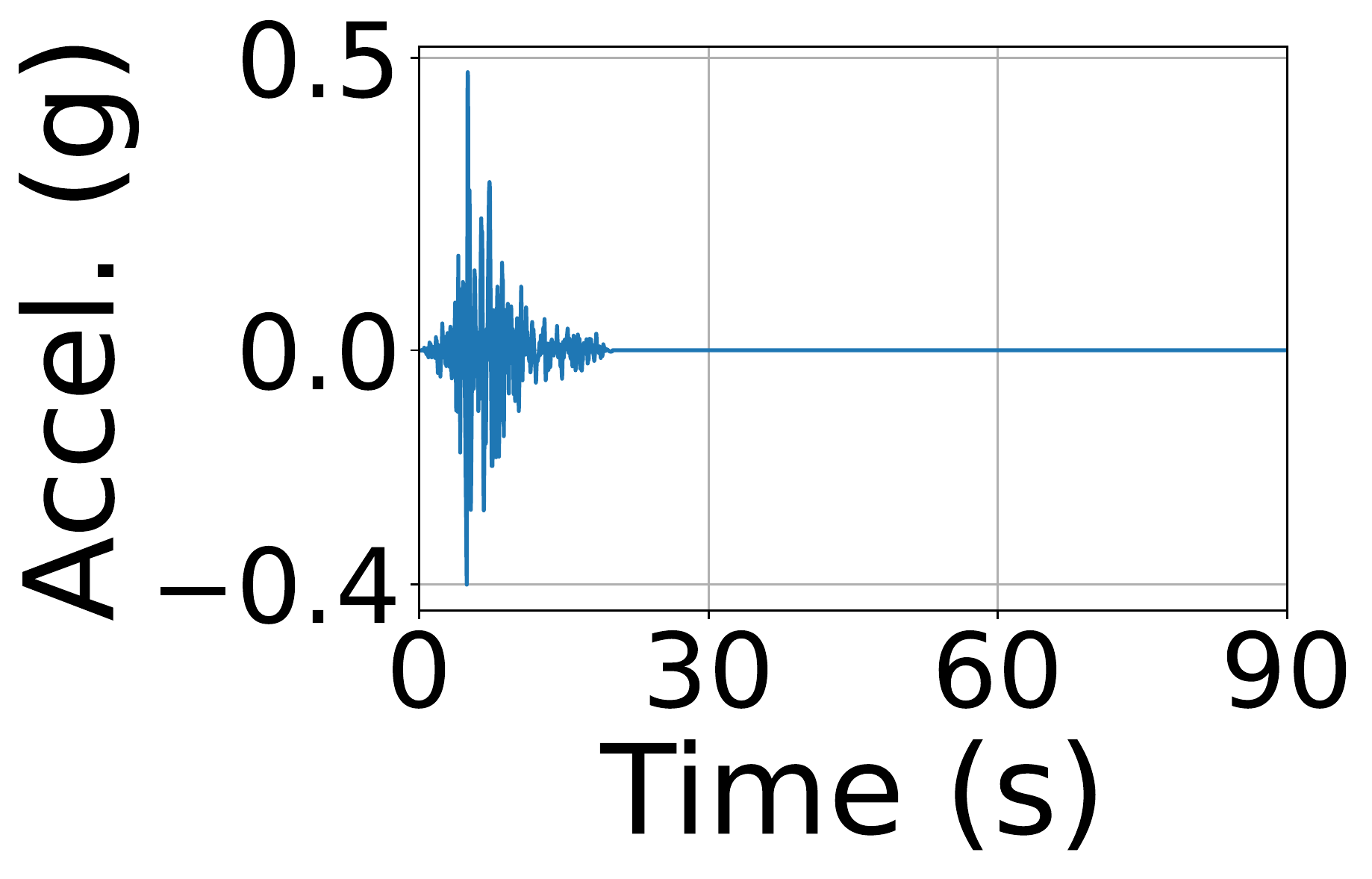}
\includegraphics[valign=c,width=0.19\textwidth]{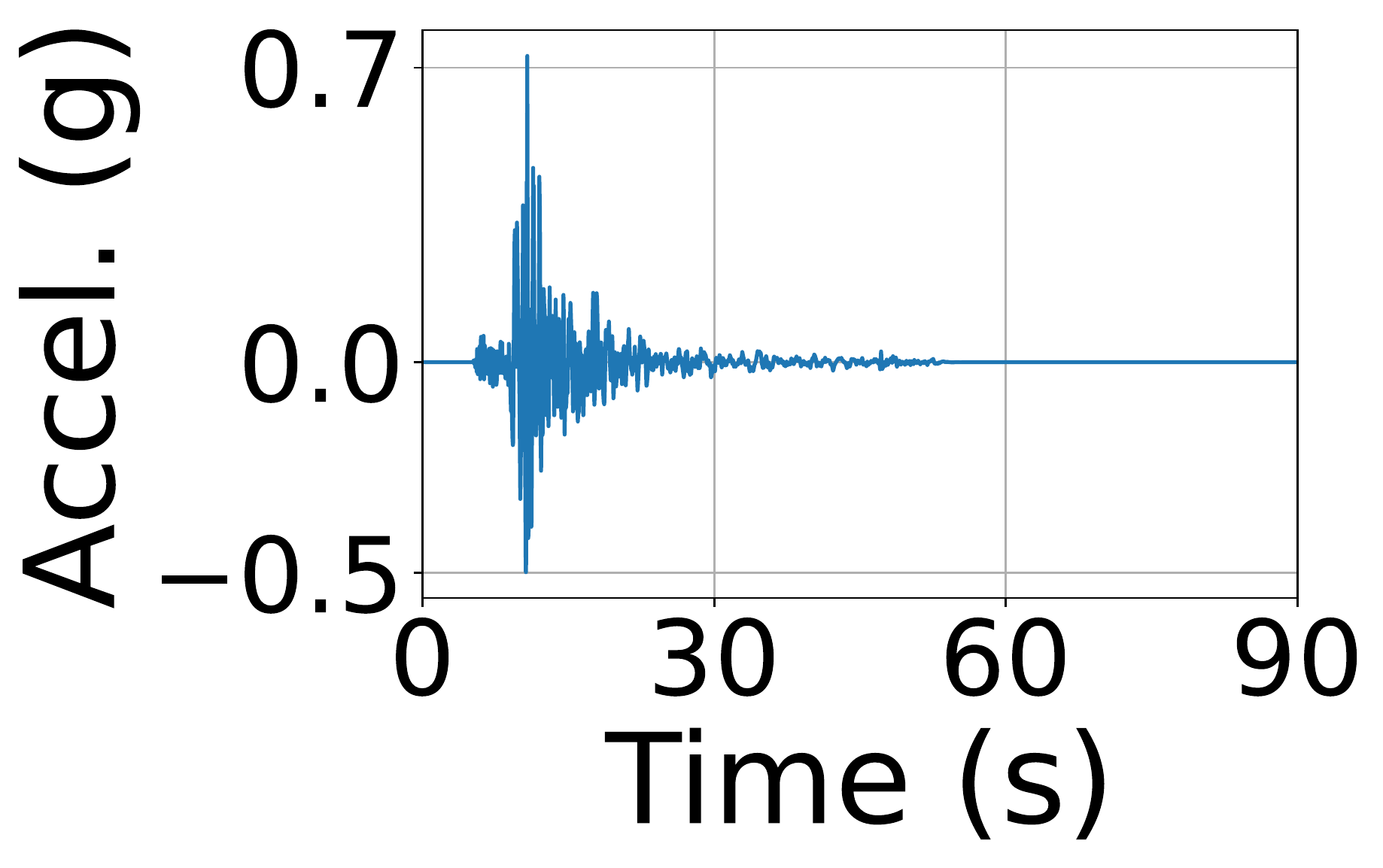} \\
\vspace*{0.35truecm}
\includegraphics[valign=c,width=0.19\textwidth]{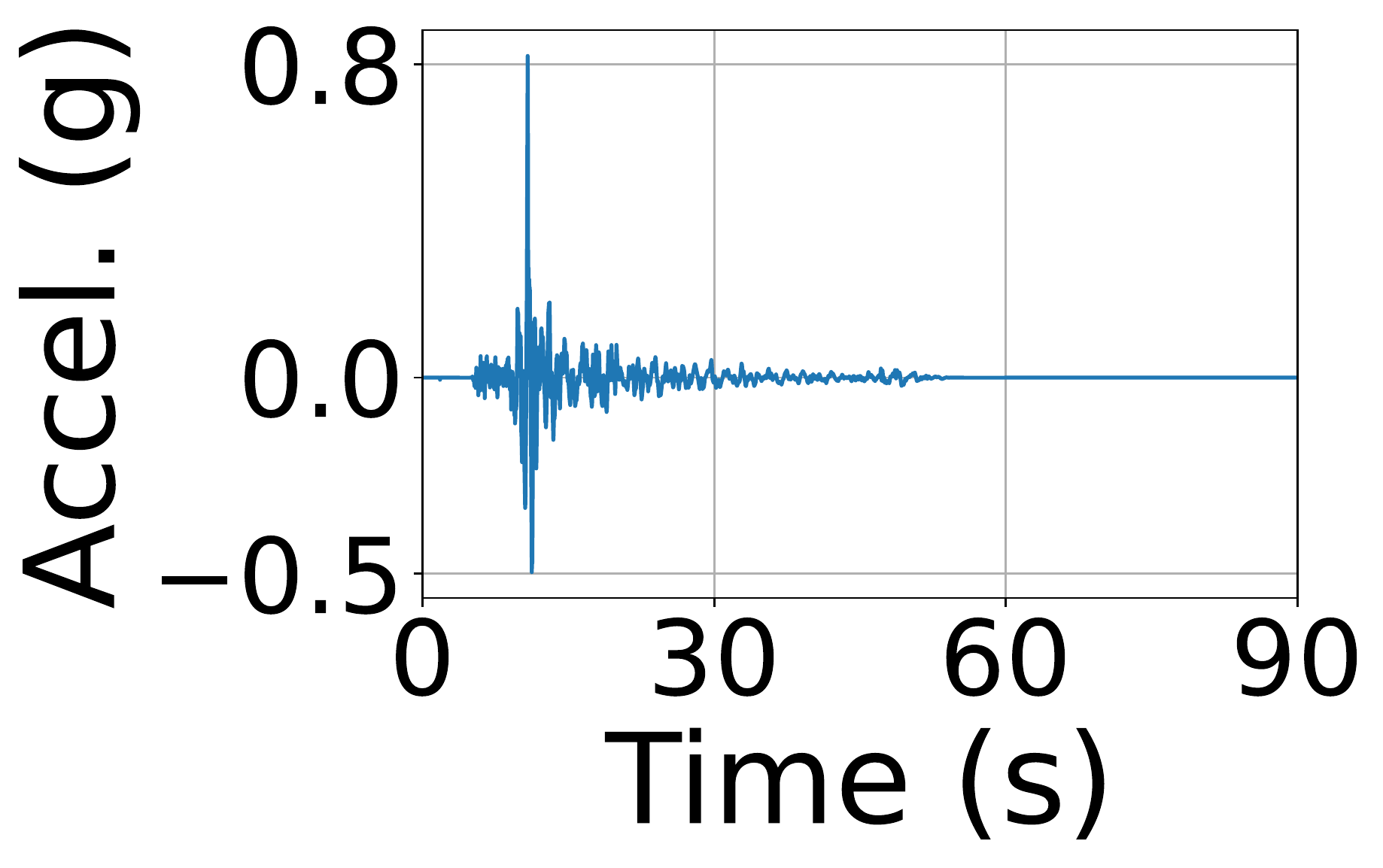} 
\includegraphics[valign=c,width=0.19\textwidth]{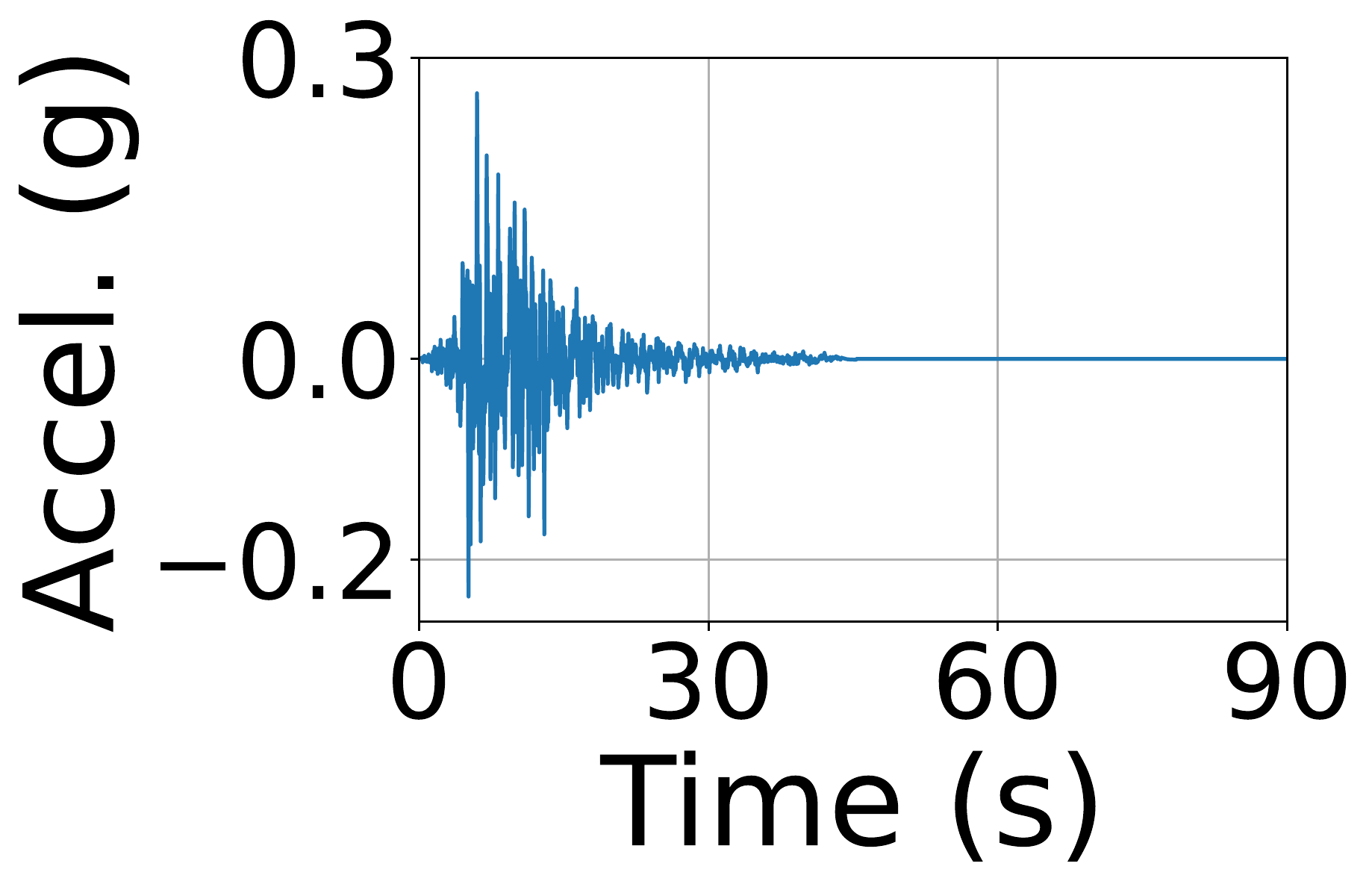} 
\includegraphics[valign=c,width=0.19\textwidth]{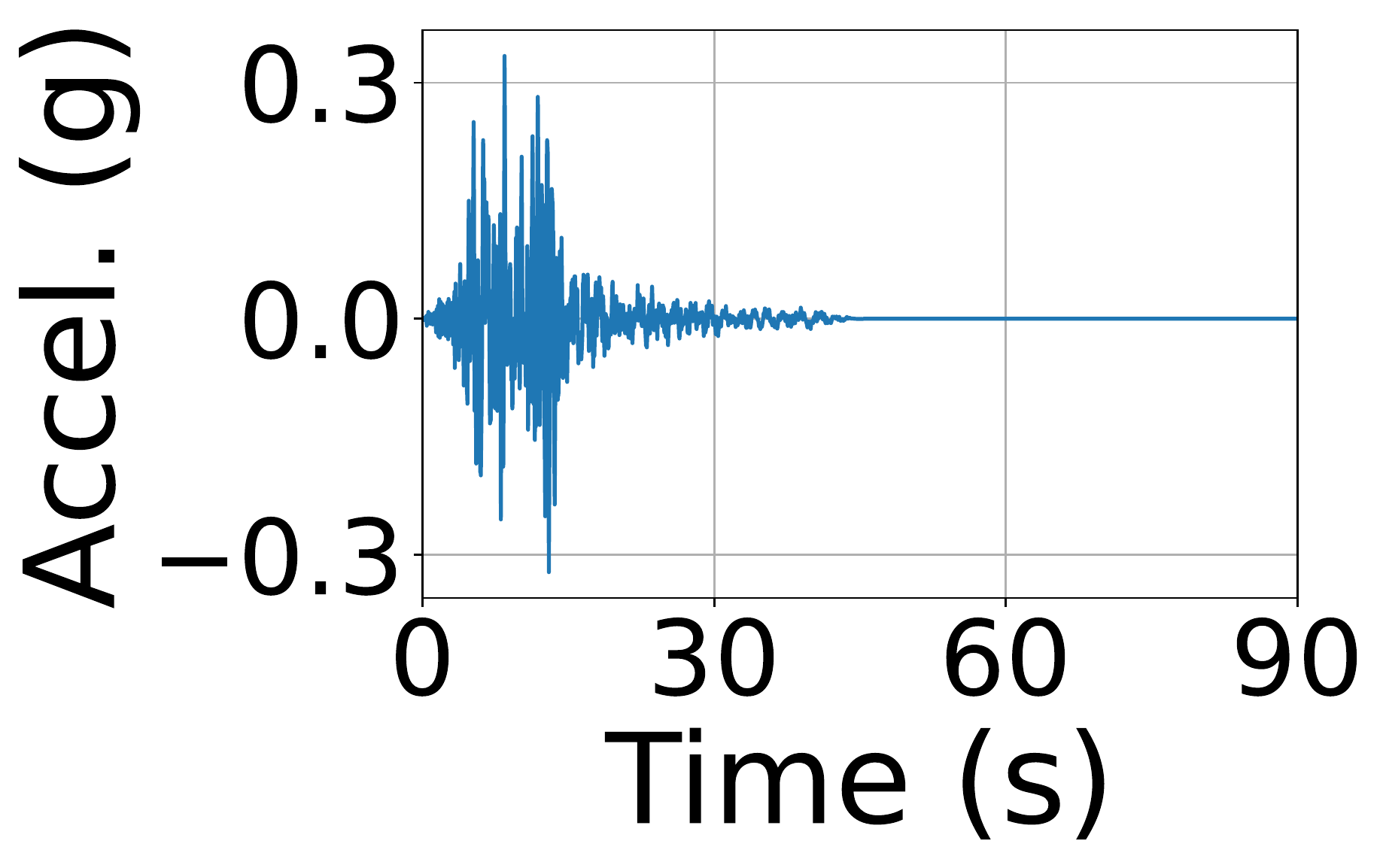}
\includegraphics[valign=c,width=0.19\textwidth]{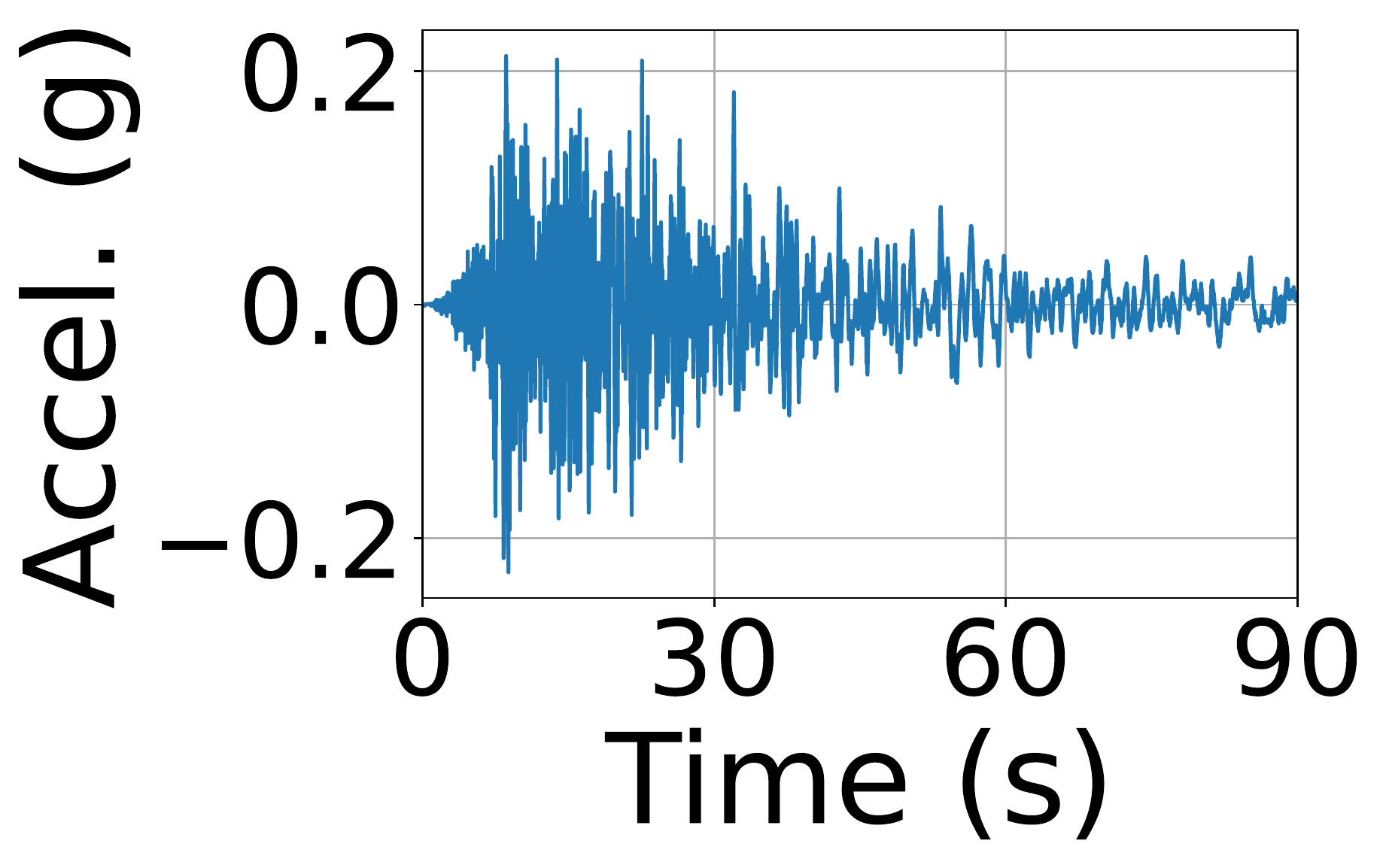}
\includegraphics[valign=c,width=0.19\textwidth]{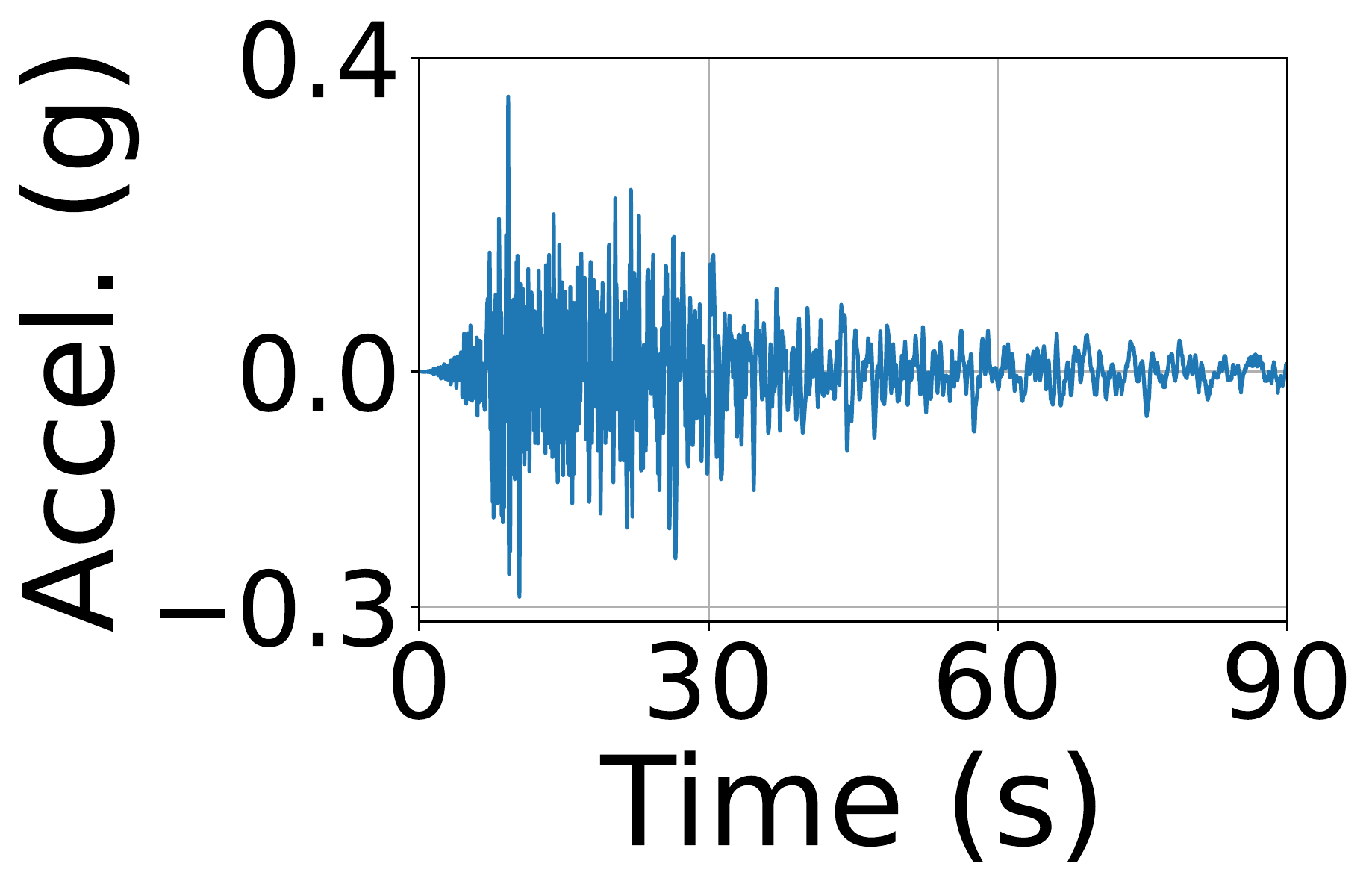} \\
\vspace*{0.35truecm}
\includegraphics[valign=c,width=0.19\textwidth]{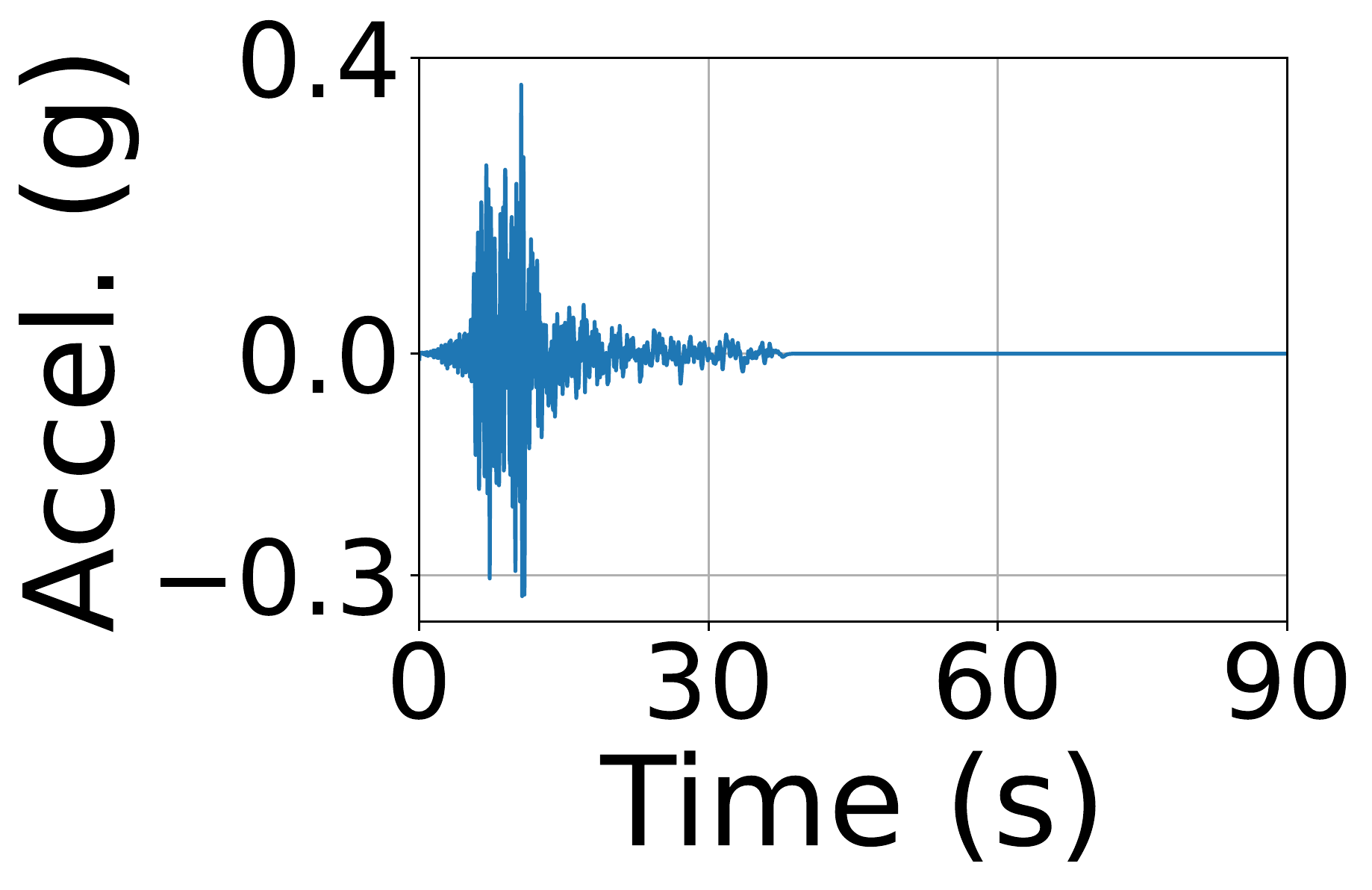} 
\includegraphics[valign=c,width=0.19\textwidth]{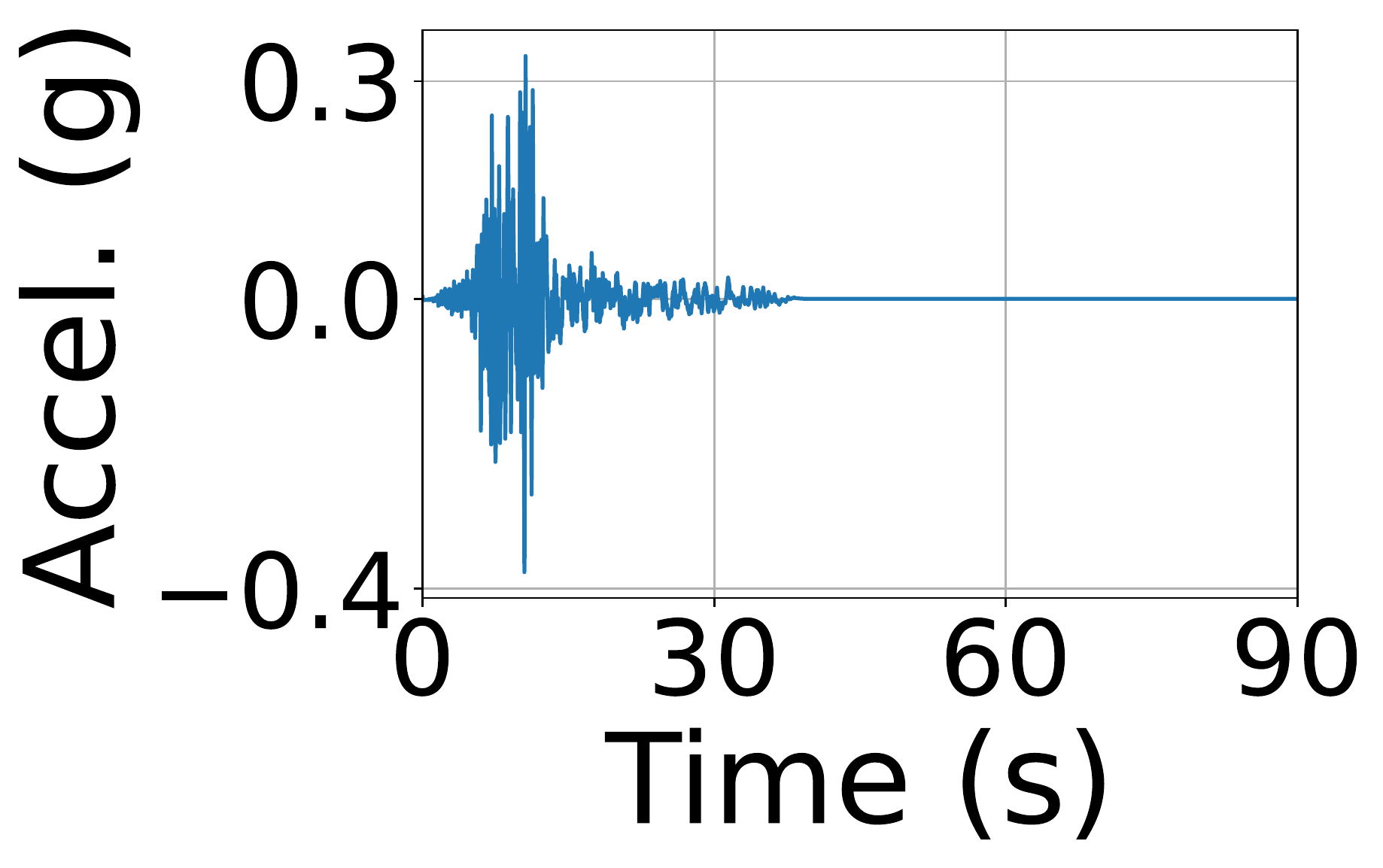} 
\includegraphics[valign=c,width=0.19\textwidth]{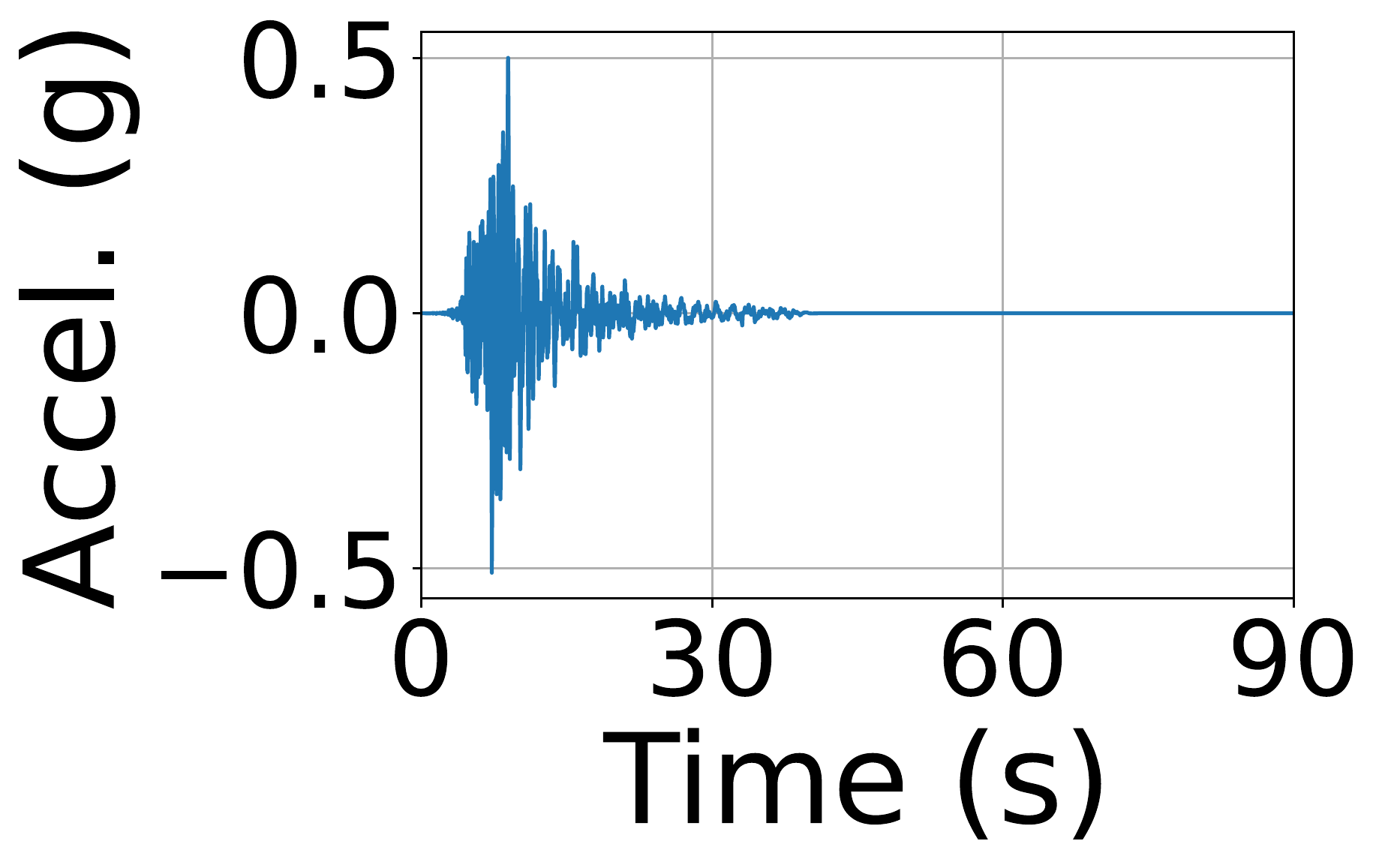}
\includegraphics[valign=c,width=0.19\textwidth]{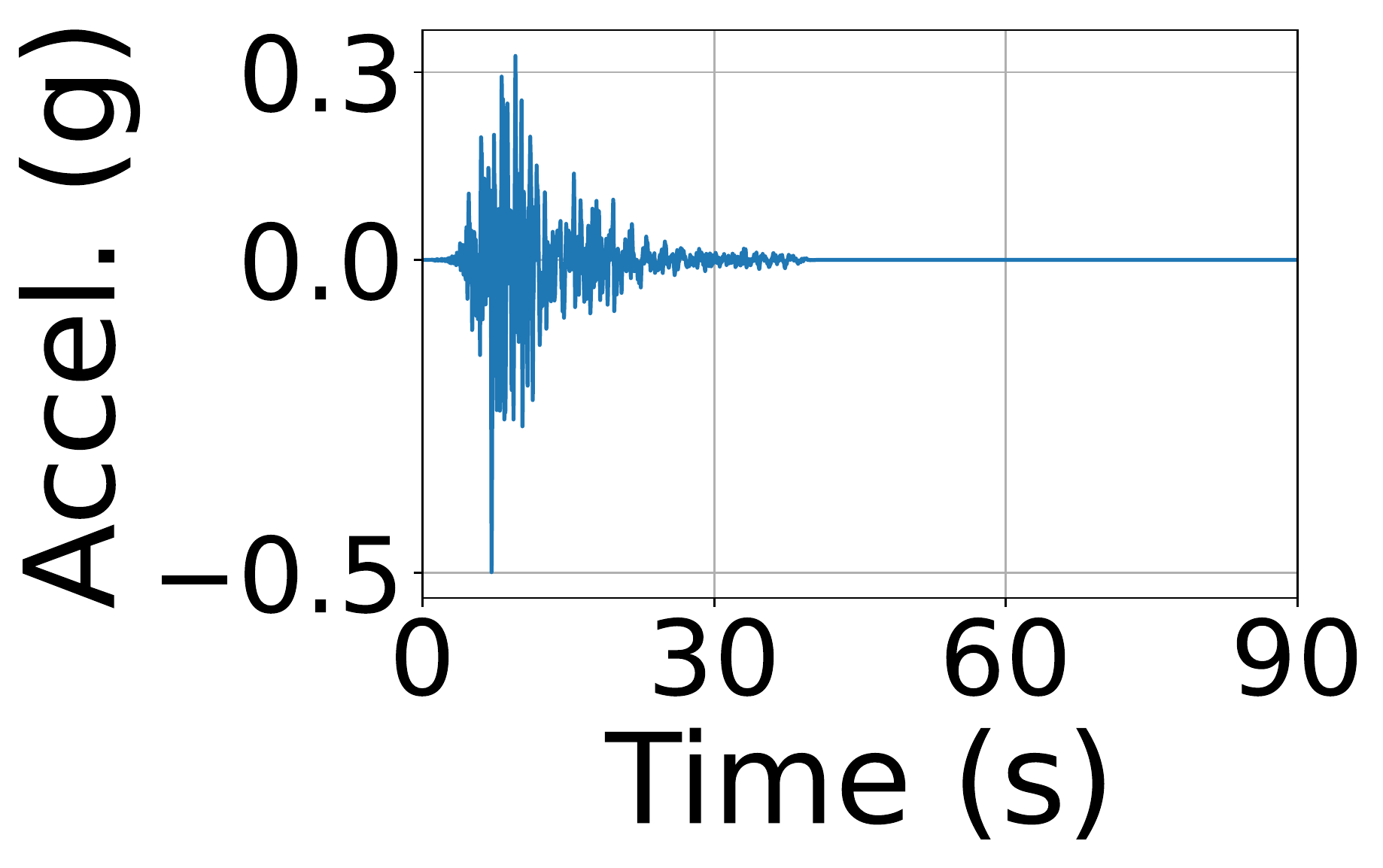}
\includegraphics[valign=c,width=0.19\textwidth]{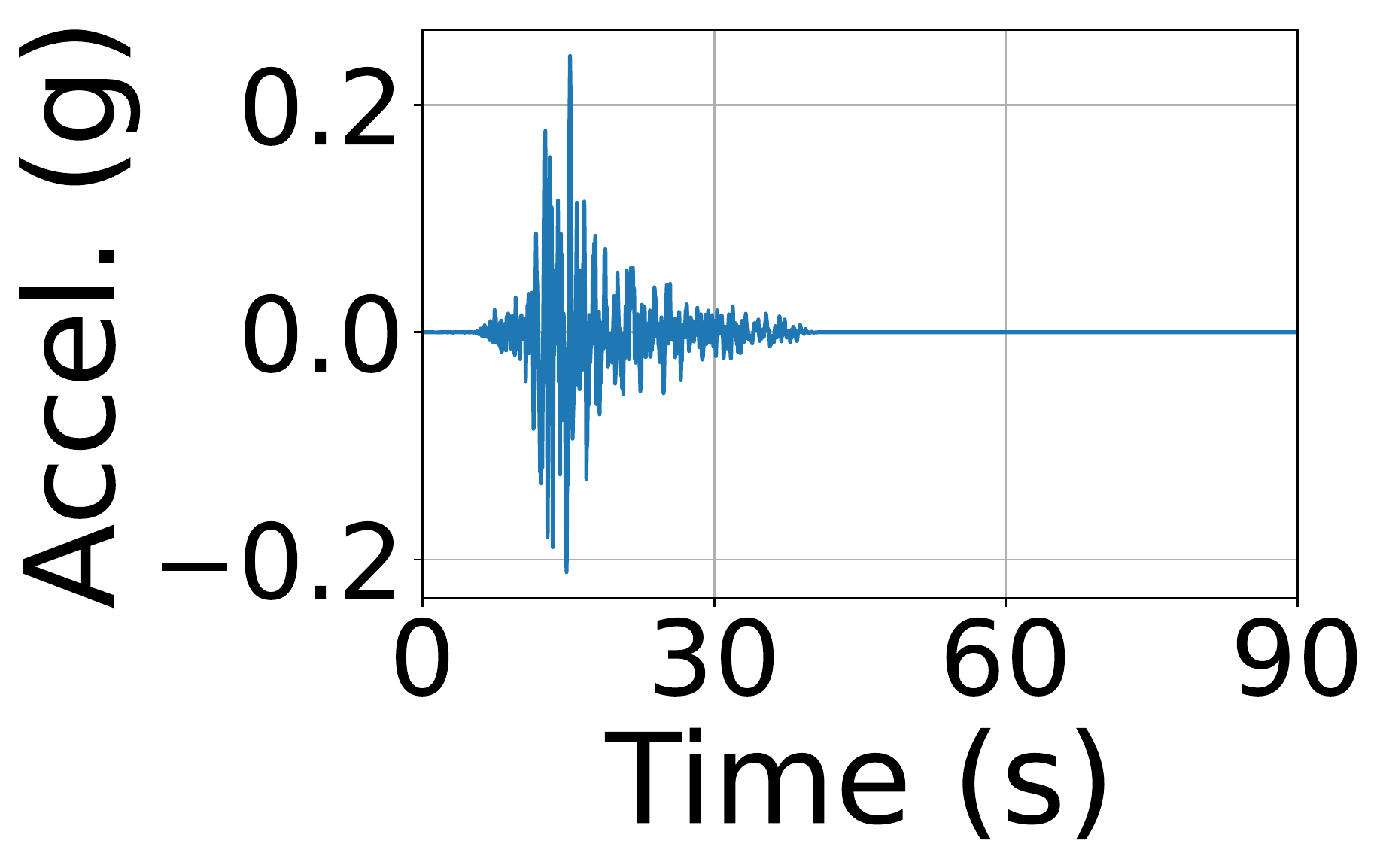} \\
\vspace*{0.35truecm}
\includegraphics[valign=c,width=0.19\textwidth]{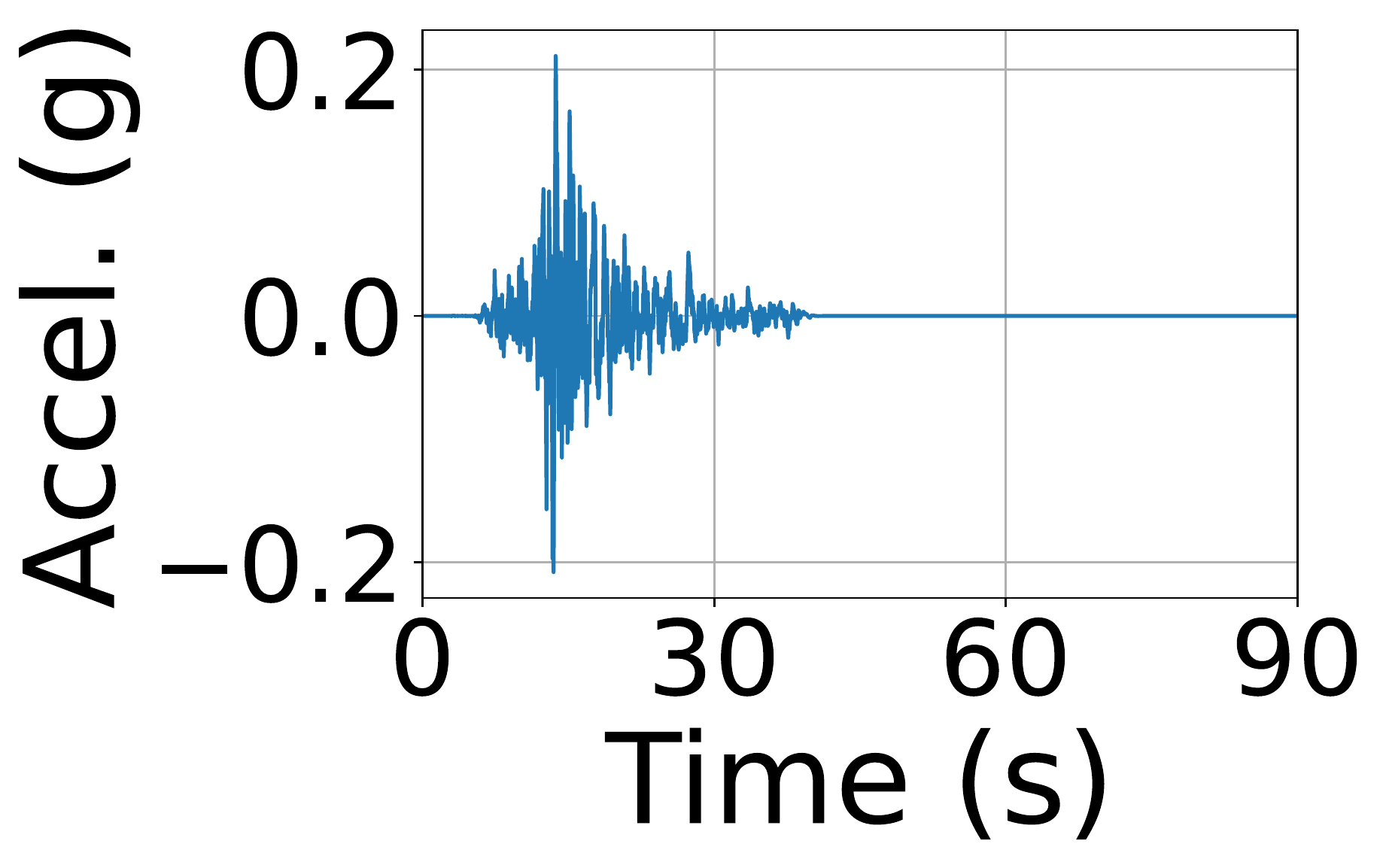} 
\includegraphics[valign=c,width=0.19\textwidth]{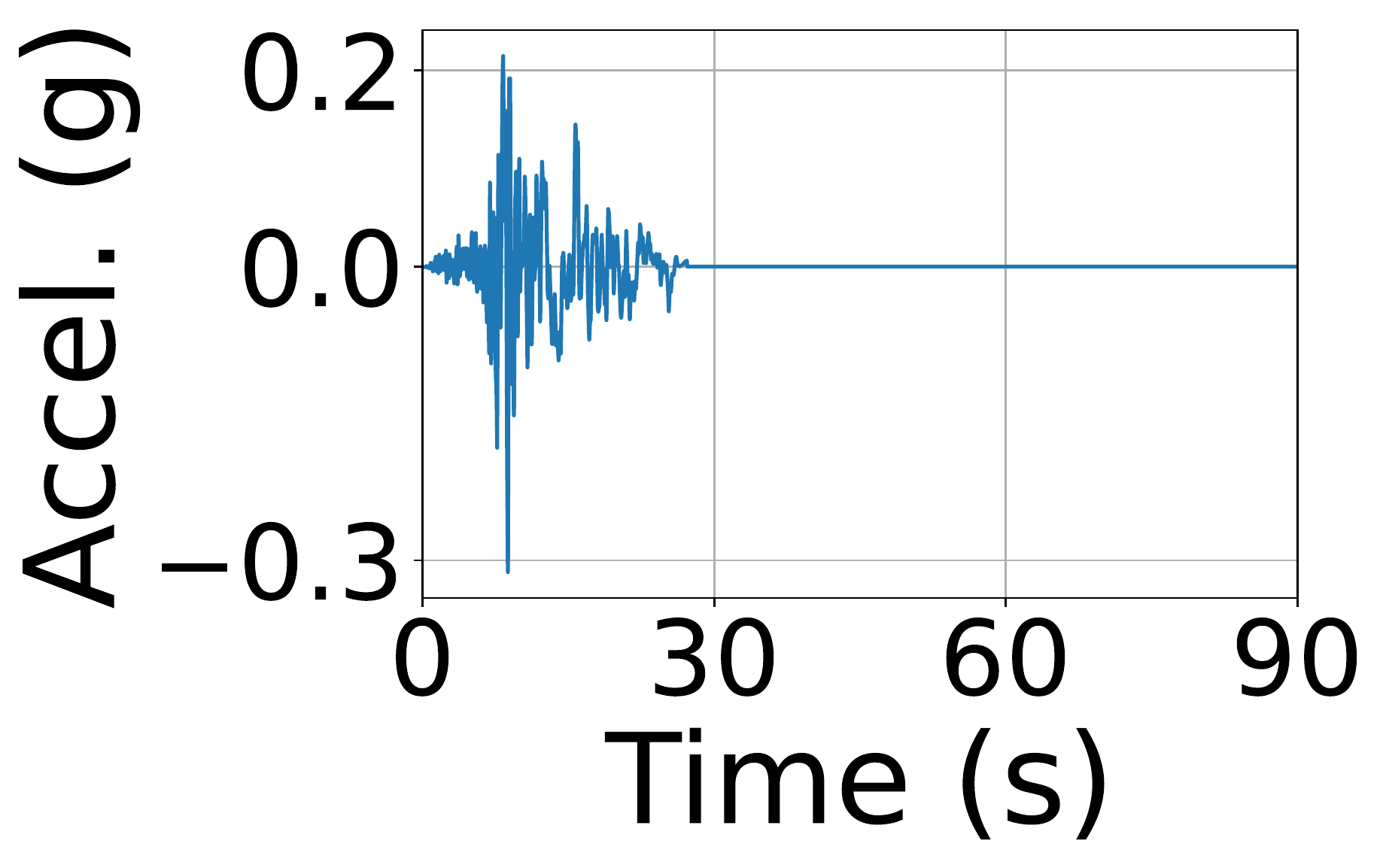} 
\includegraphics[valign=c,width=0.19\textwidth]{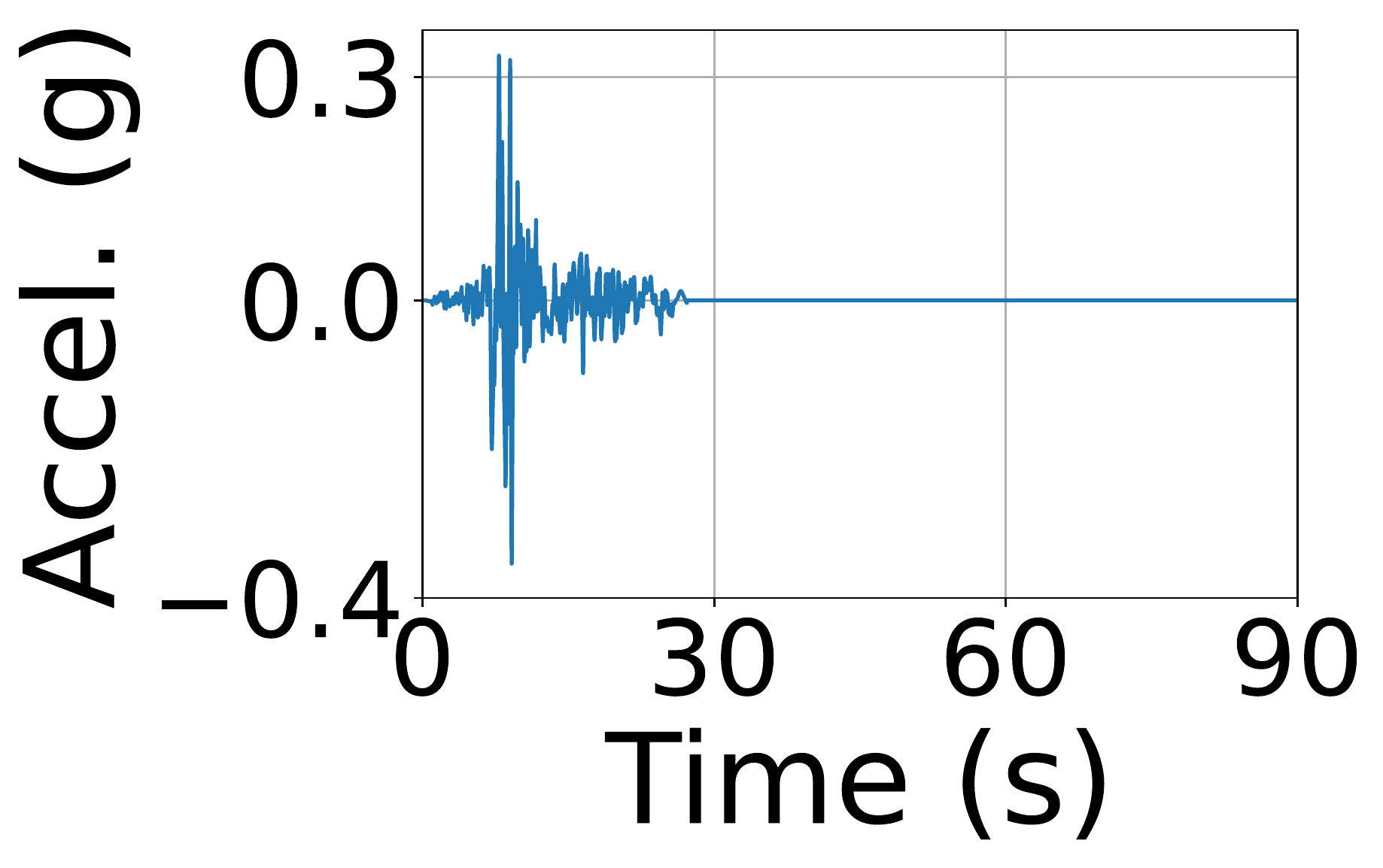}
\includegraphics[valign=c,width=0.19\textwidth]{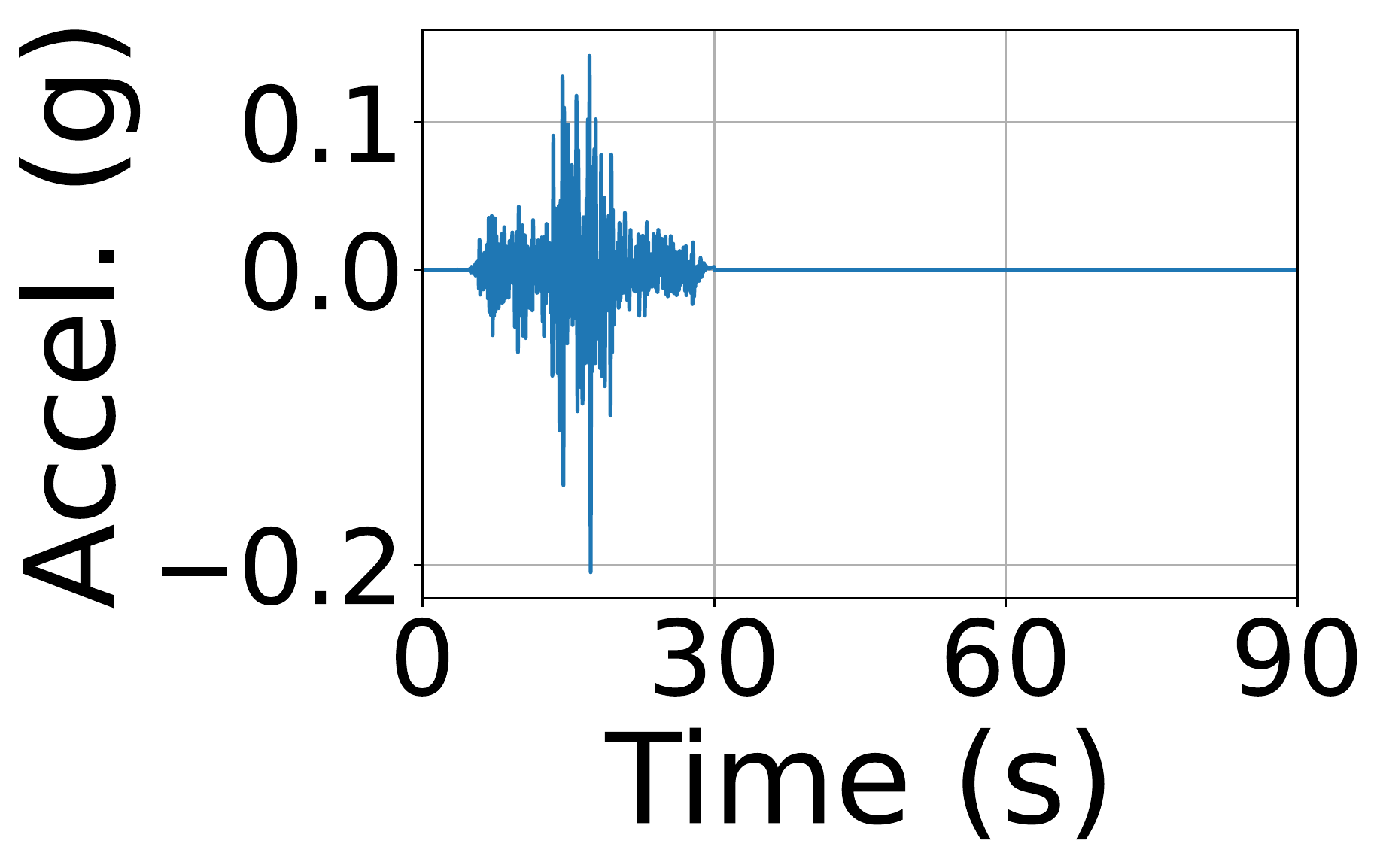}
\includegraphics[valign=c,width=0.19\textwidth]{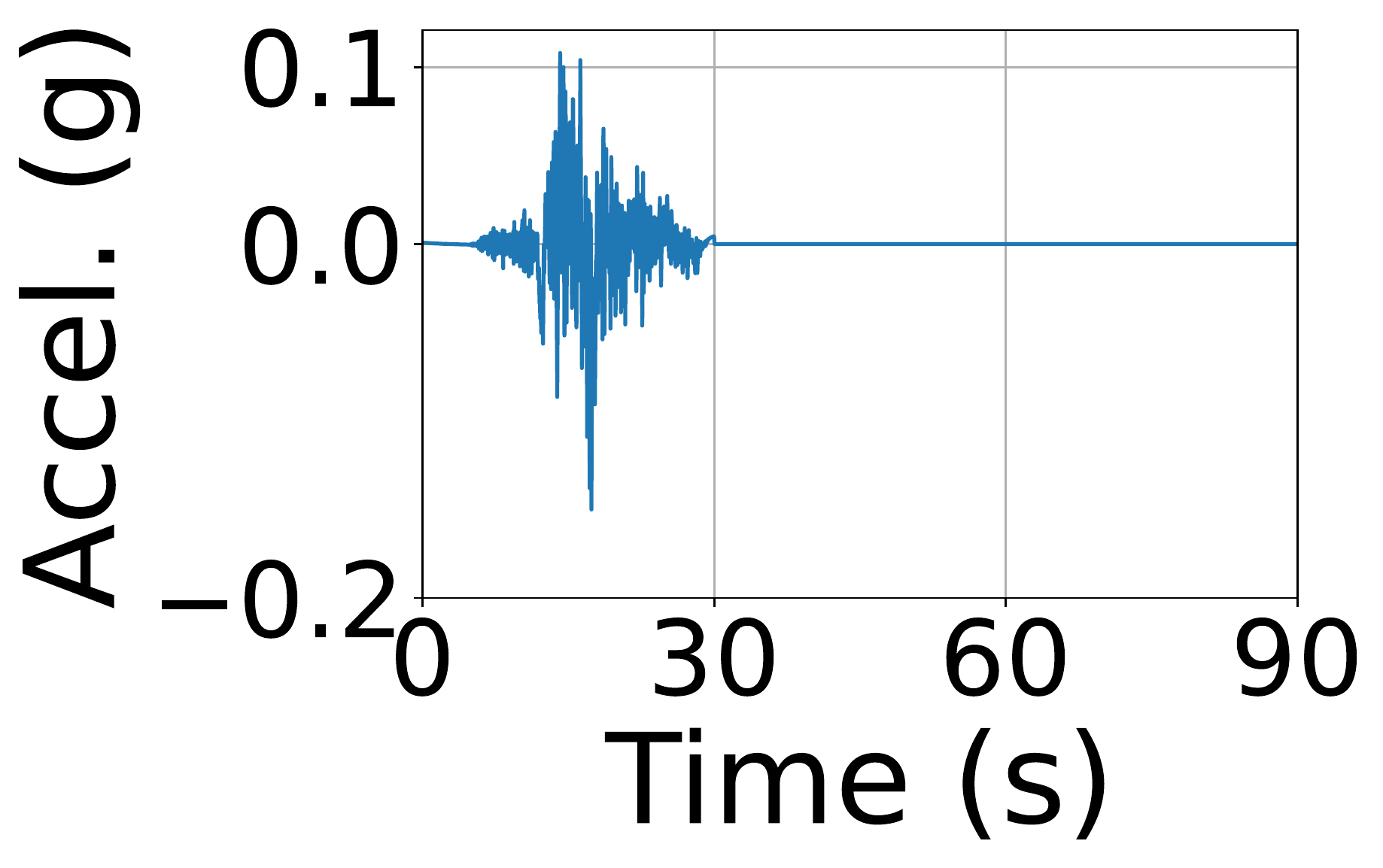} \\
\vspace*{0.35truecm}
\includegraphics[valign=c,width=0.19\textwidth]{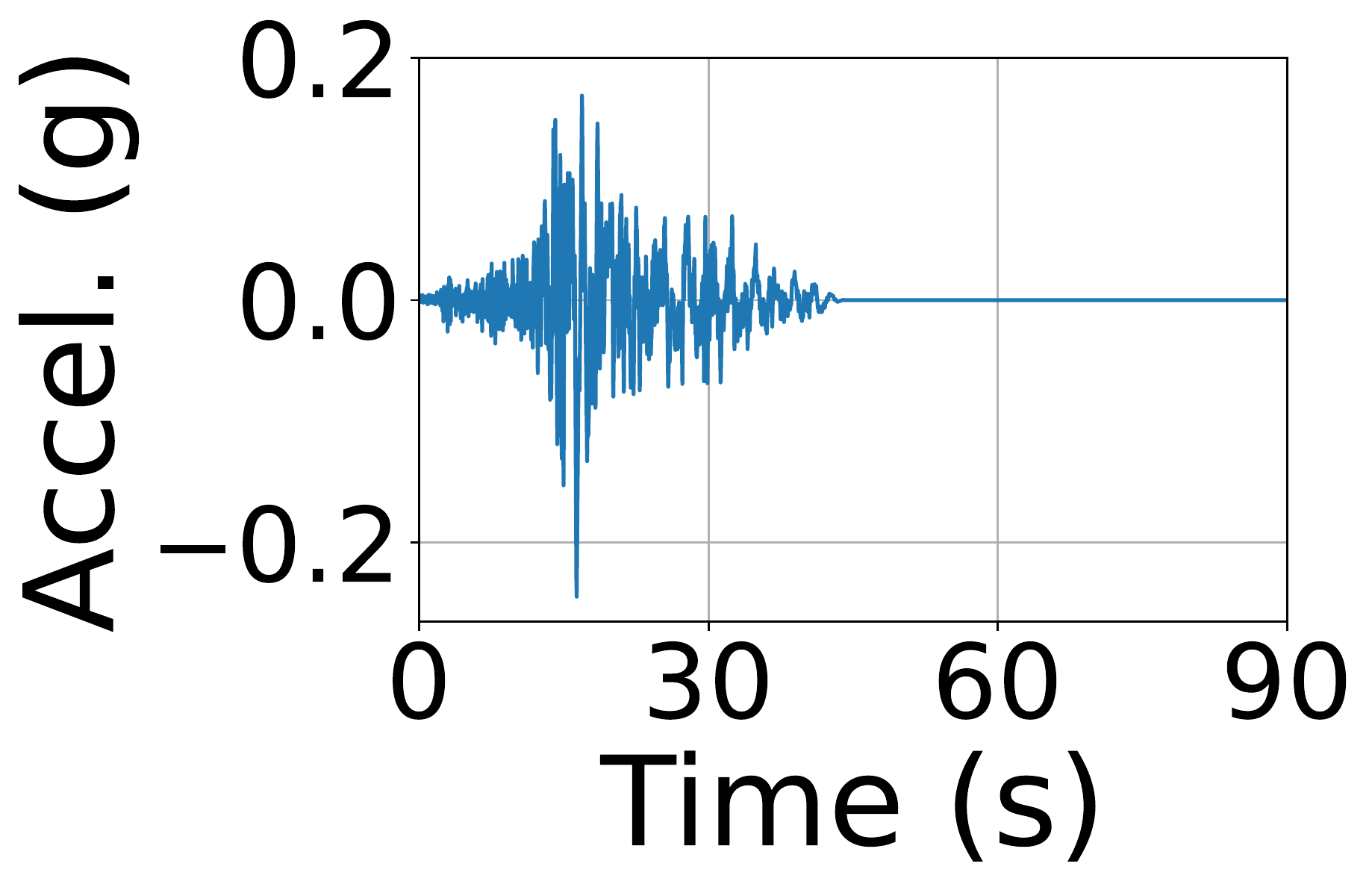} 
\includegraphics[valign=c,width=0.19\textwidth]{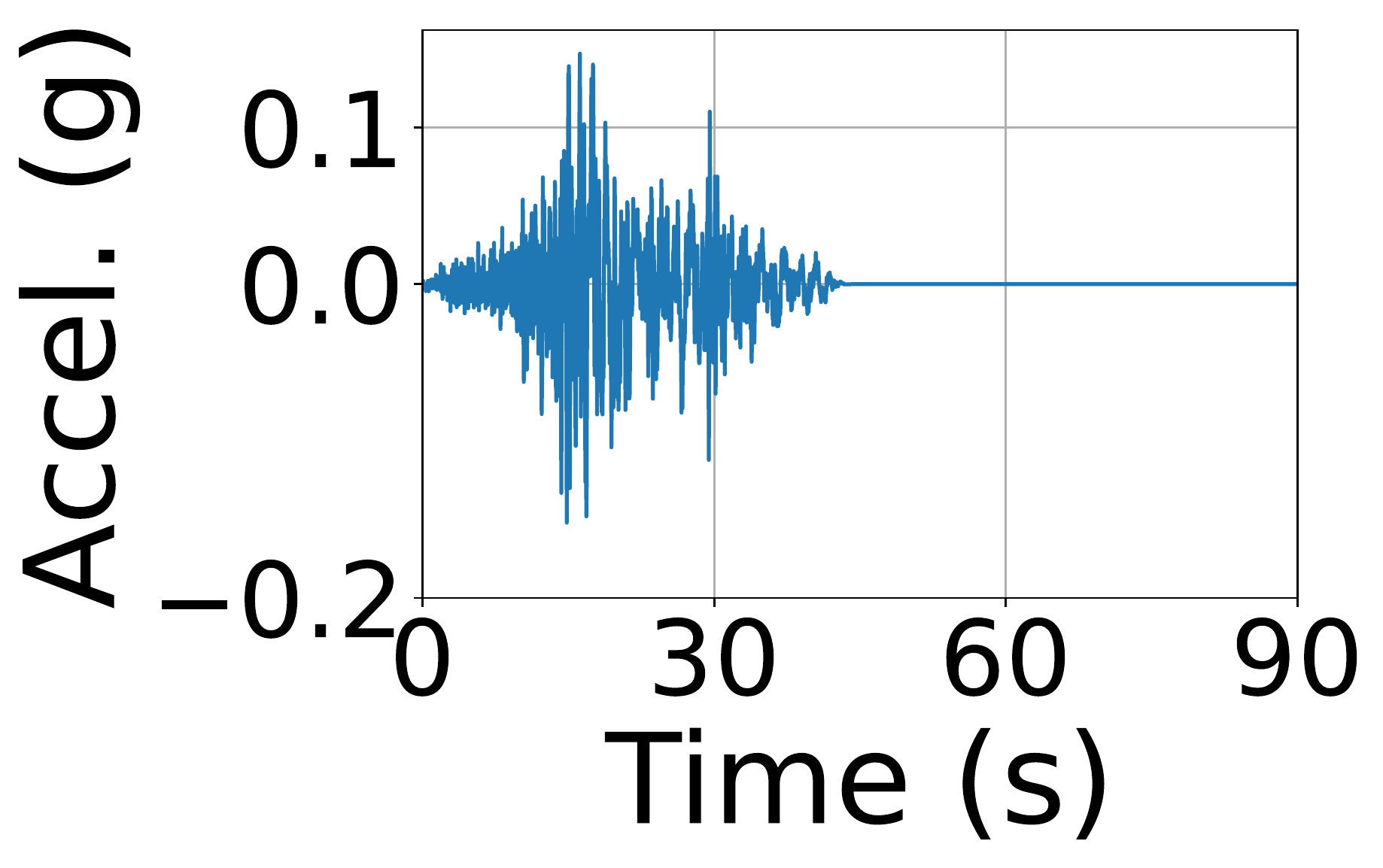} 
\includegraphics[valign=c,width=0.19\textwidth]{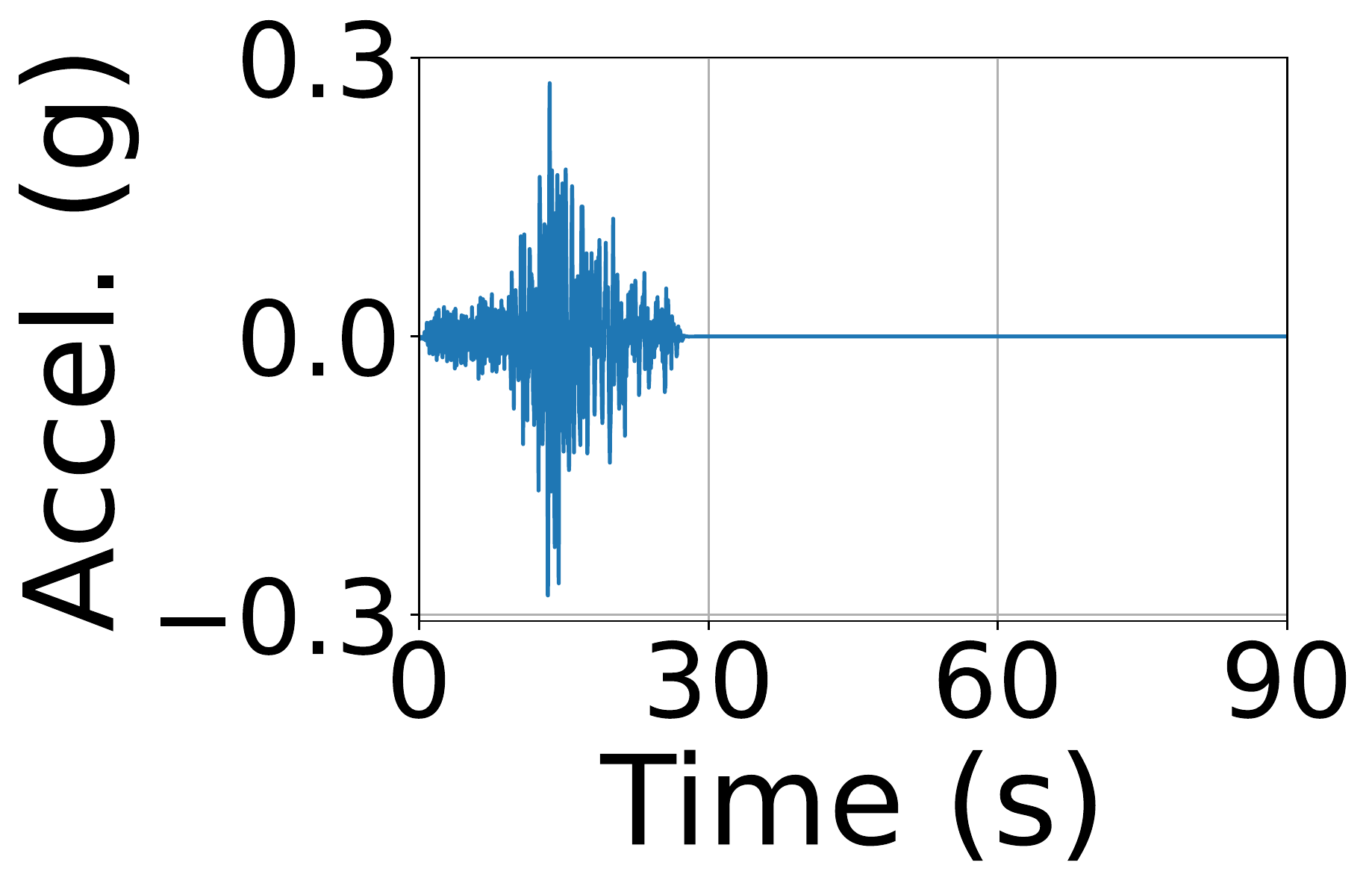}
\includegraphics[valign=c,width=0.19\textwidth]{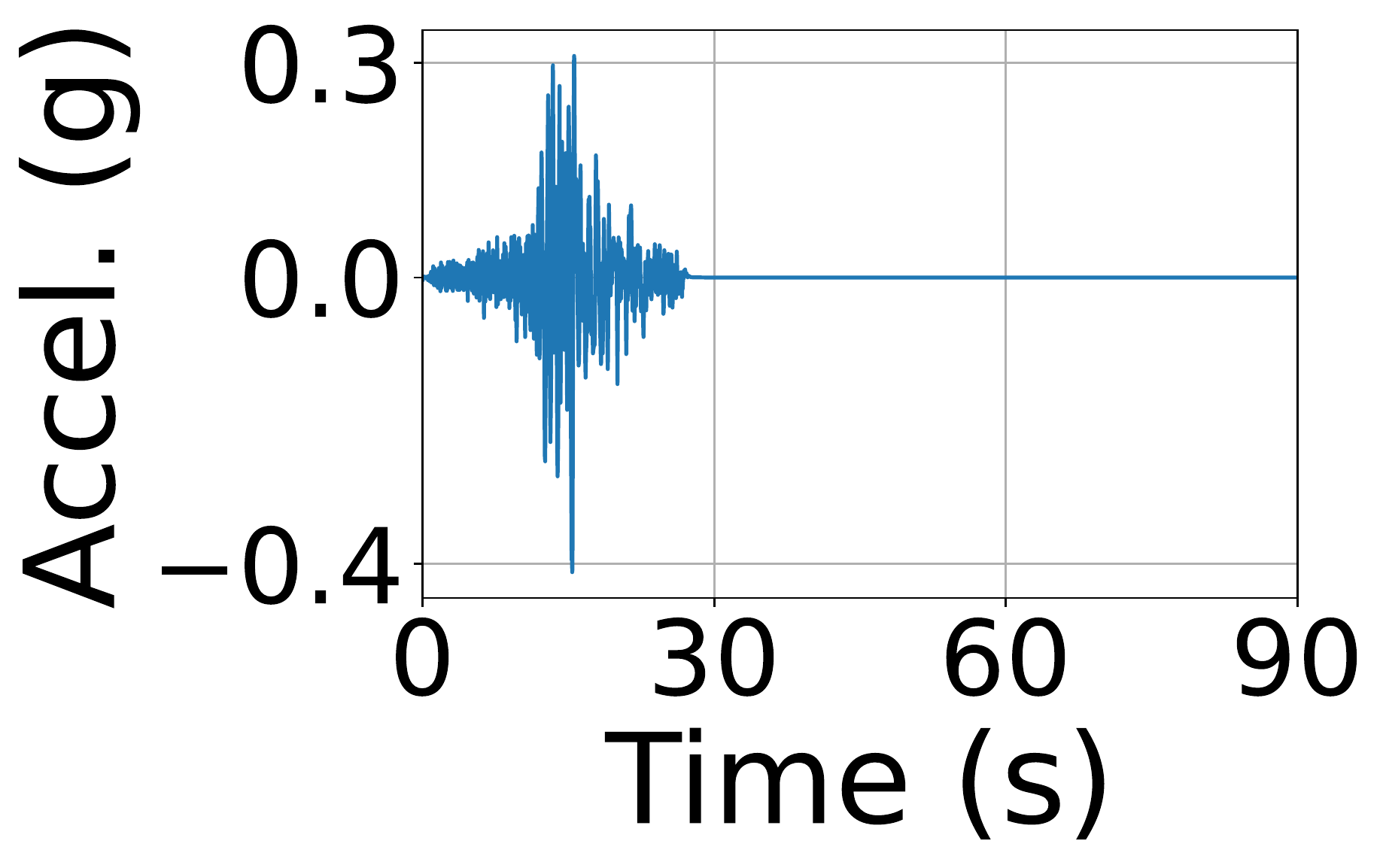}
\includegraphics[valign=c,width=0.19\textwidth]{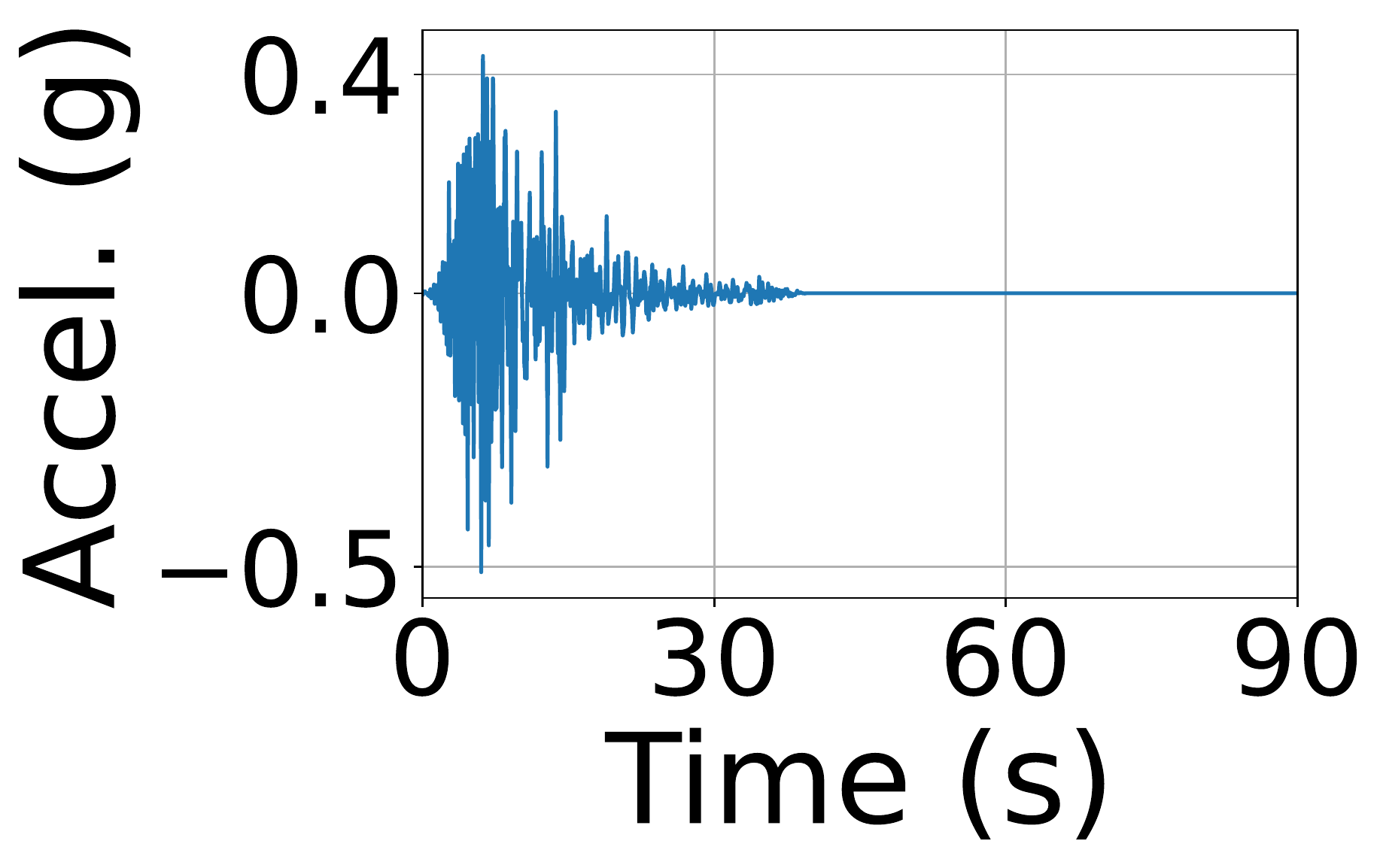} \\
\vspace*{0.35truecm}
\includegraphics[valign=c,width=0.19\textwidth]{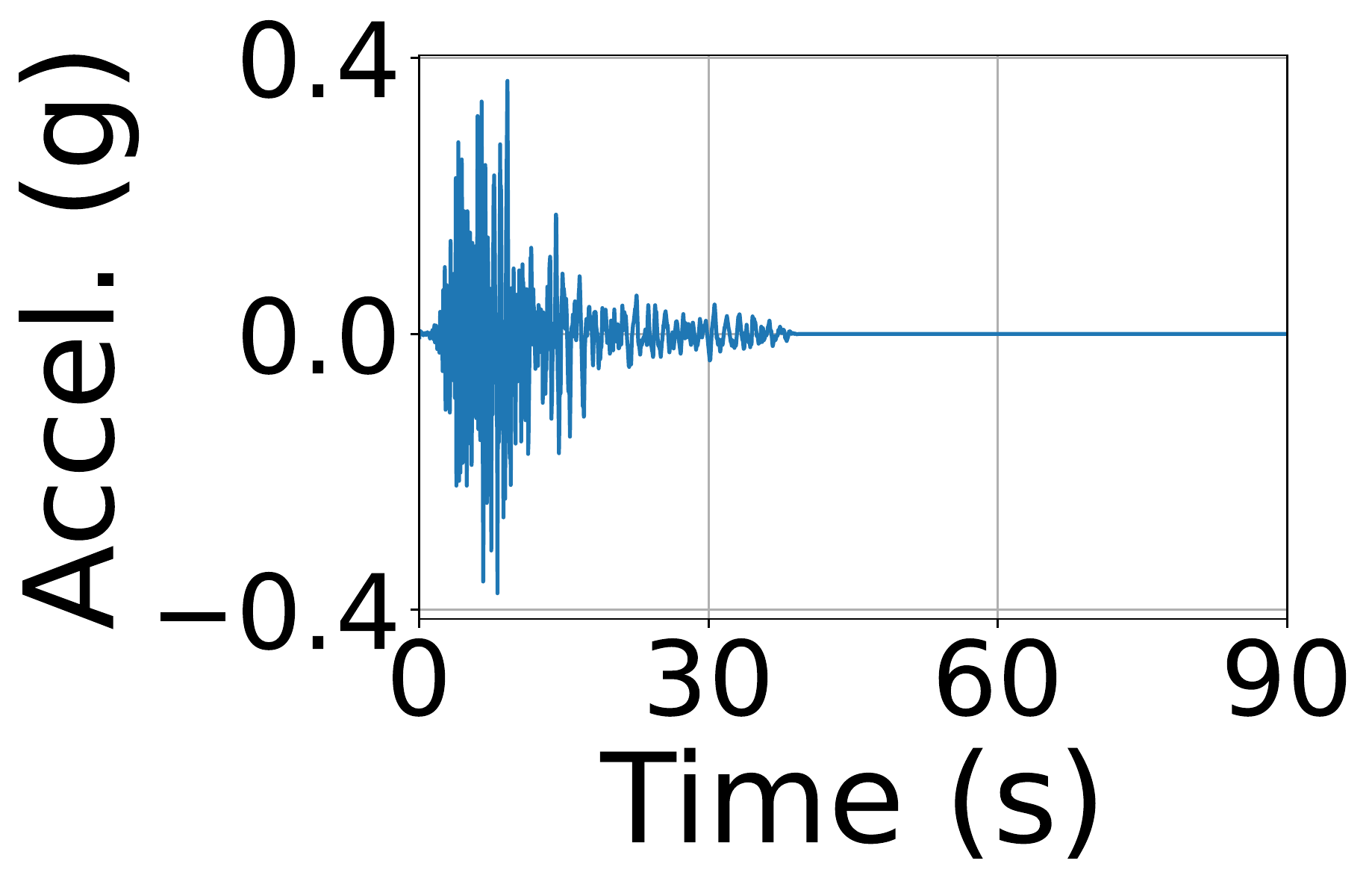} 
\includegraphics[valign=c,width=0.19\textwidth]{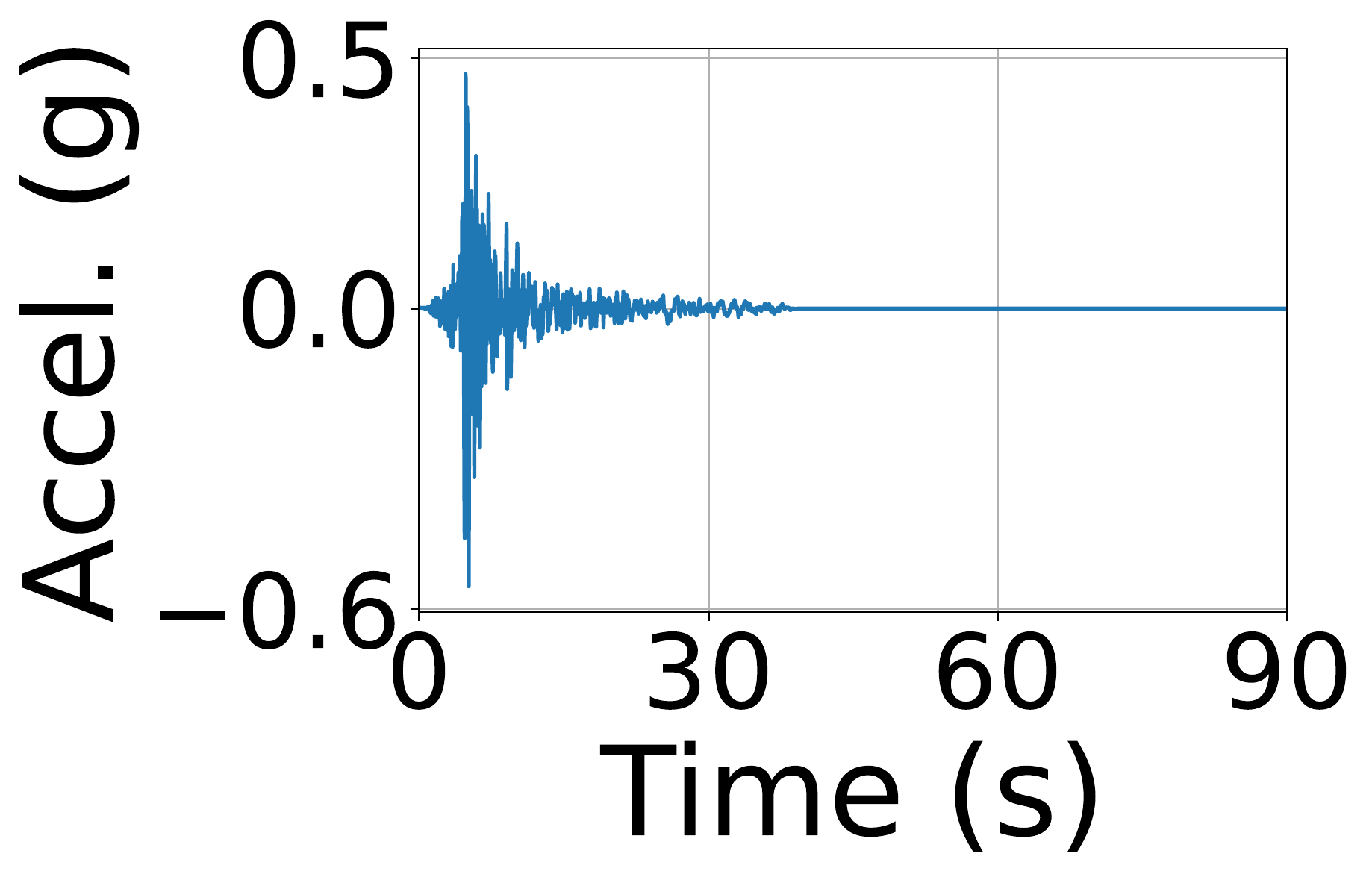} 
\includegraphics[valign=c,width=0.19\textwidth]{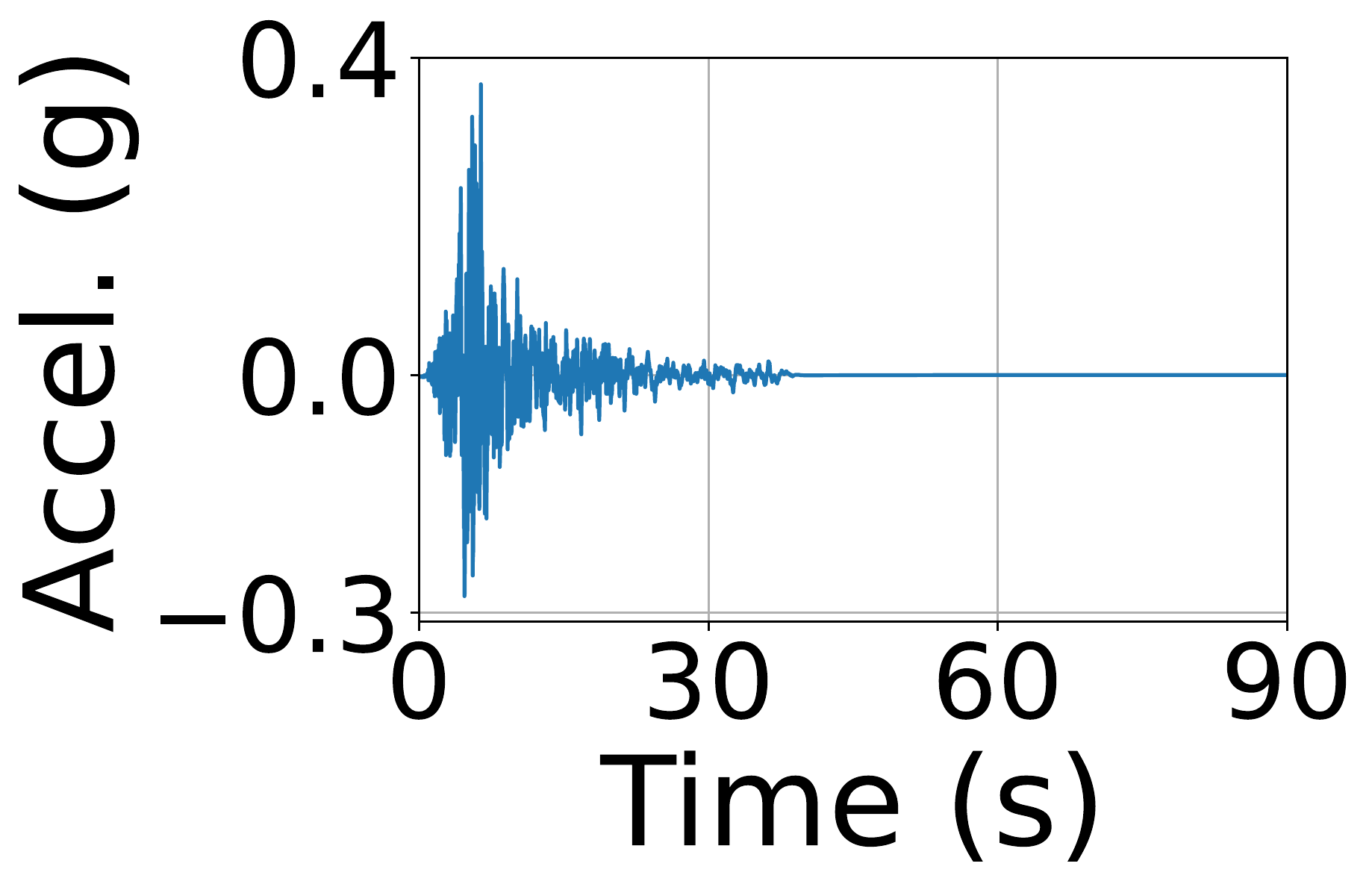}
\includegraphics[valign=c,width=0.19\textwidth]{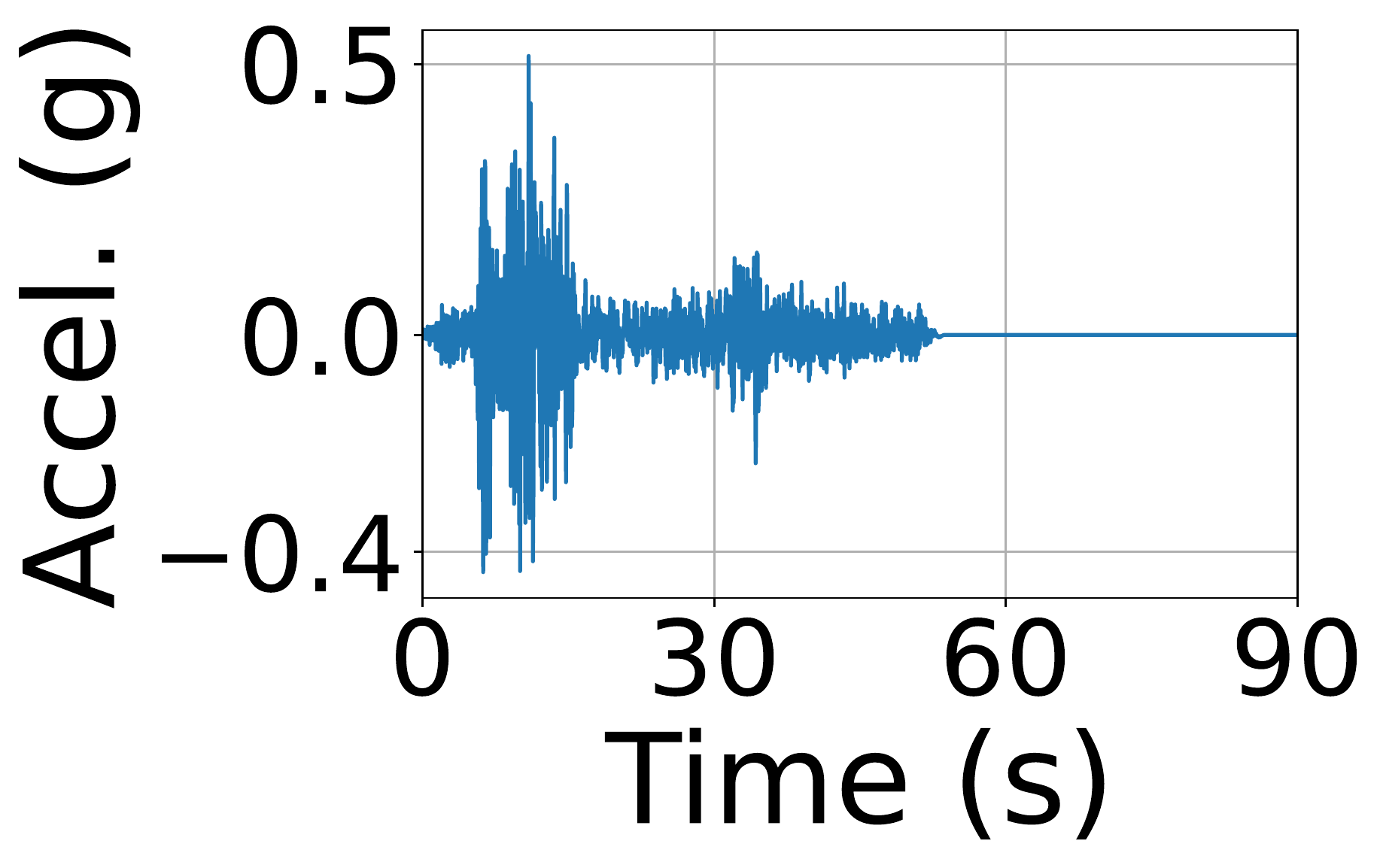}
\includegraphics[valign=c,width=0.19\textwidth]{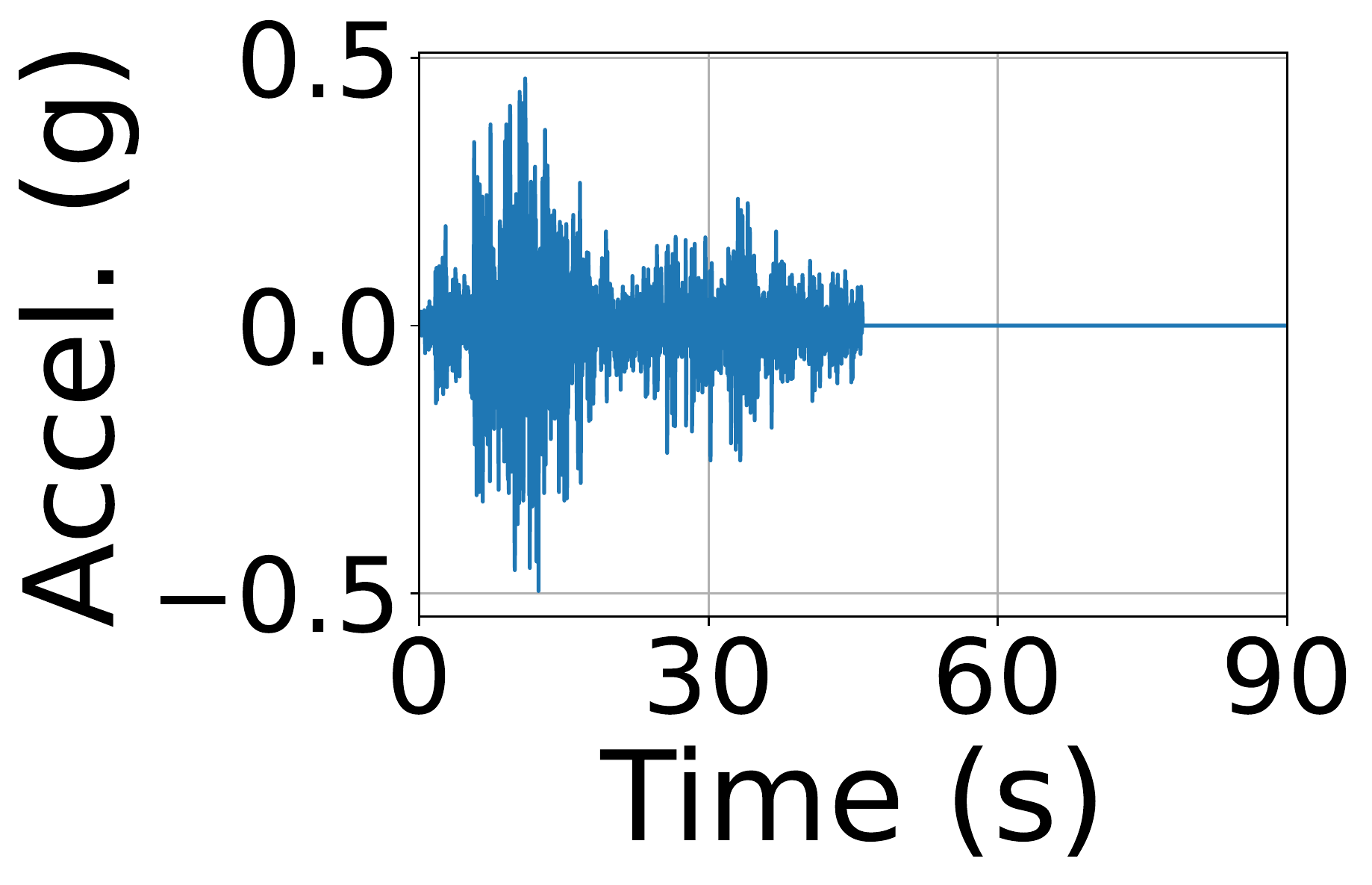} \\
\vspace*{0.35truecm}
\includegraphics[valign=c,width=0.19\textwidth]{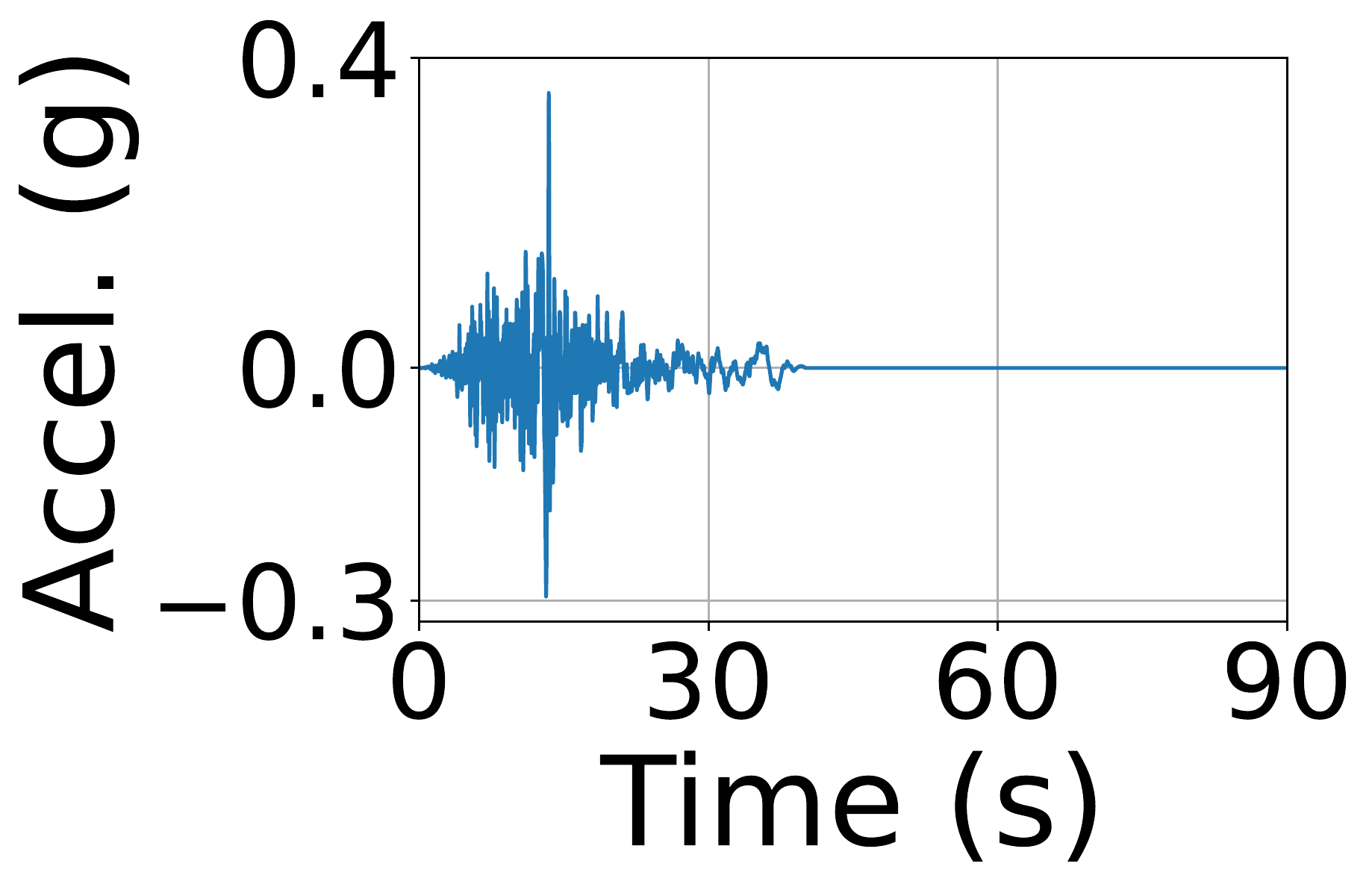} 
\includegraphics[valign=c,width=0.19\textwidth]{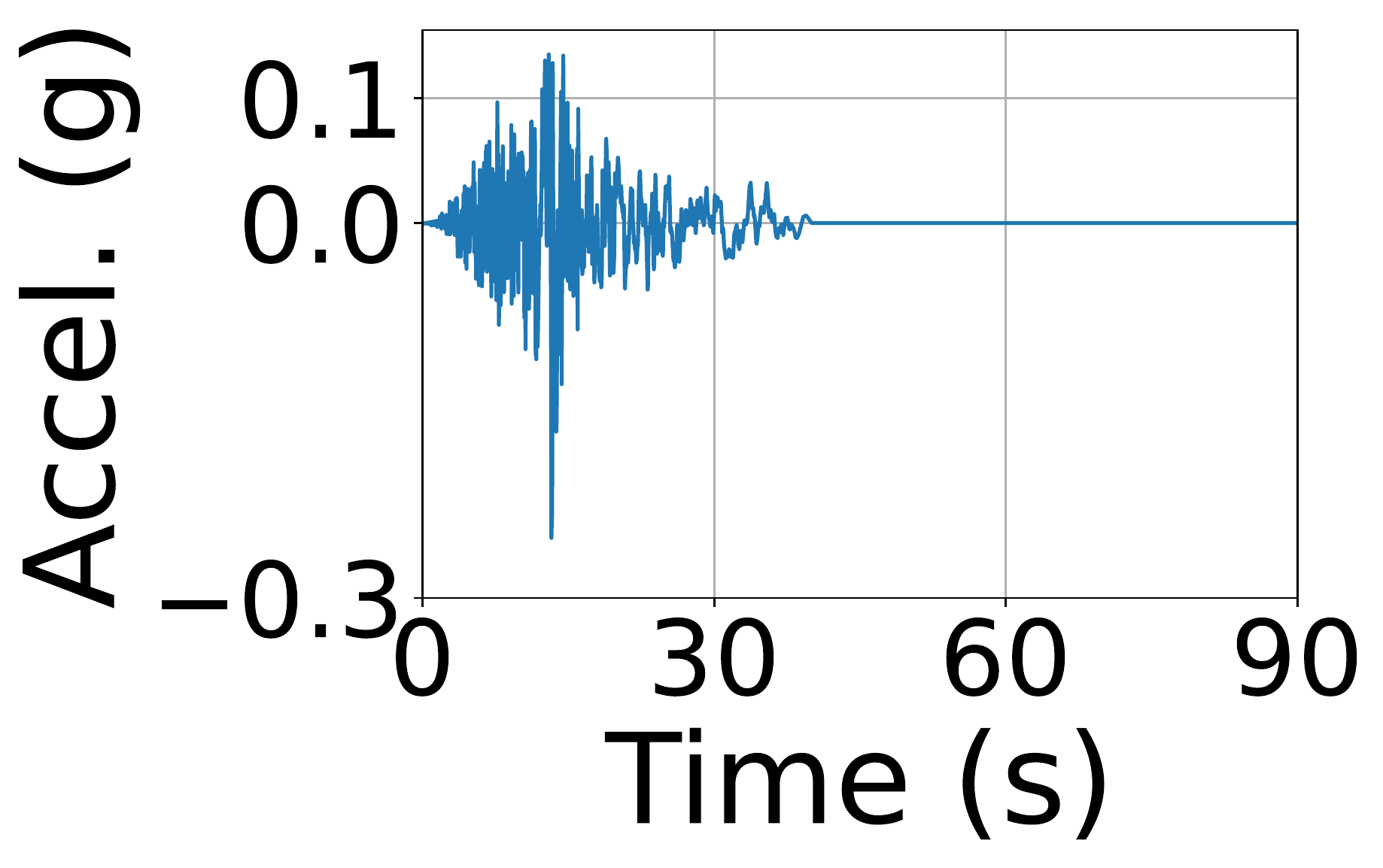} 
\includegraphics[valign=c,width=0.19\textwidth]{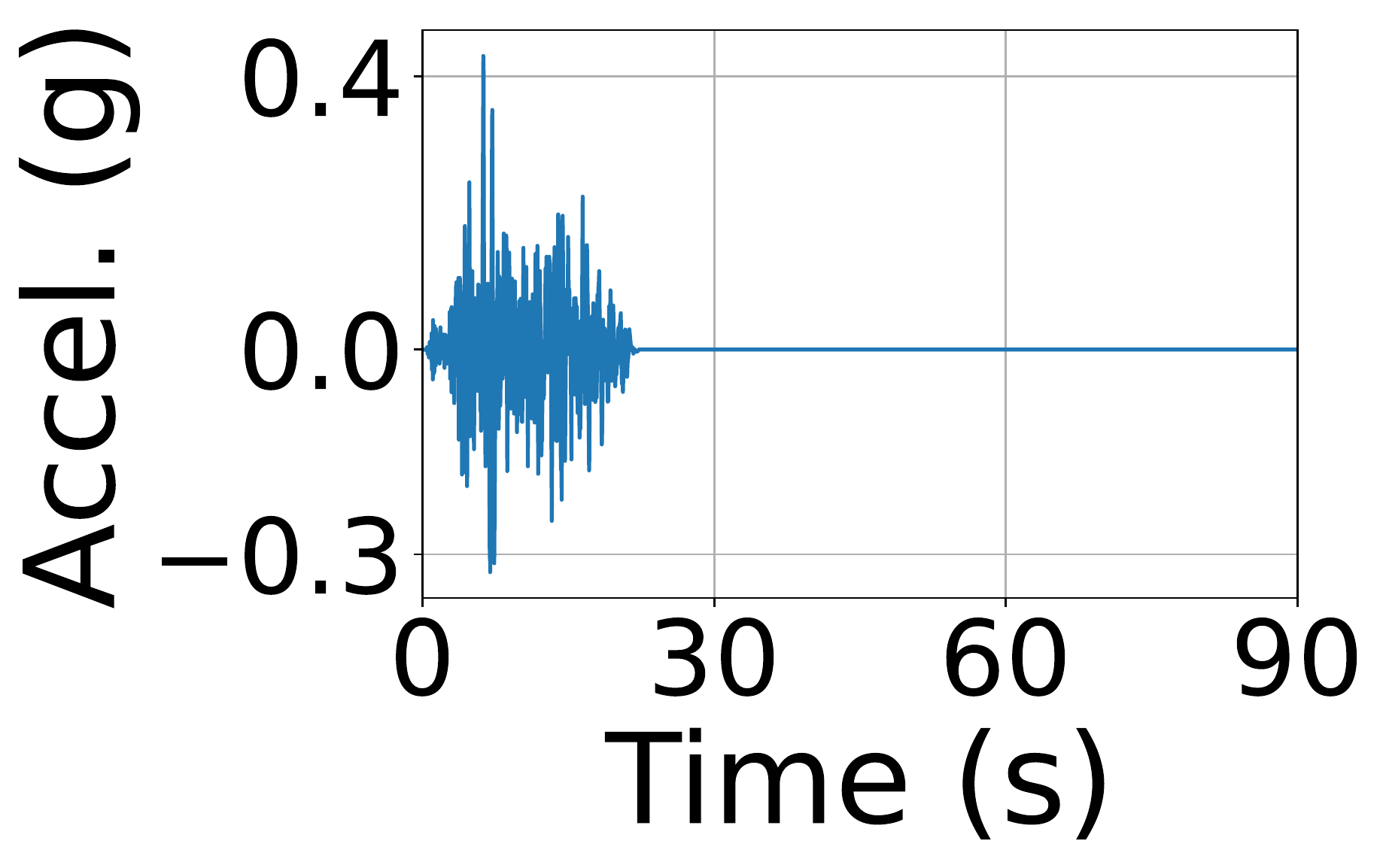}
\includegraphics[valign=c,width=0.19\textwidth]{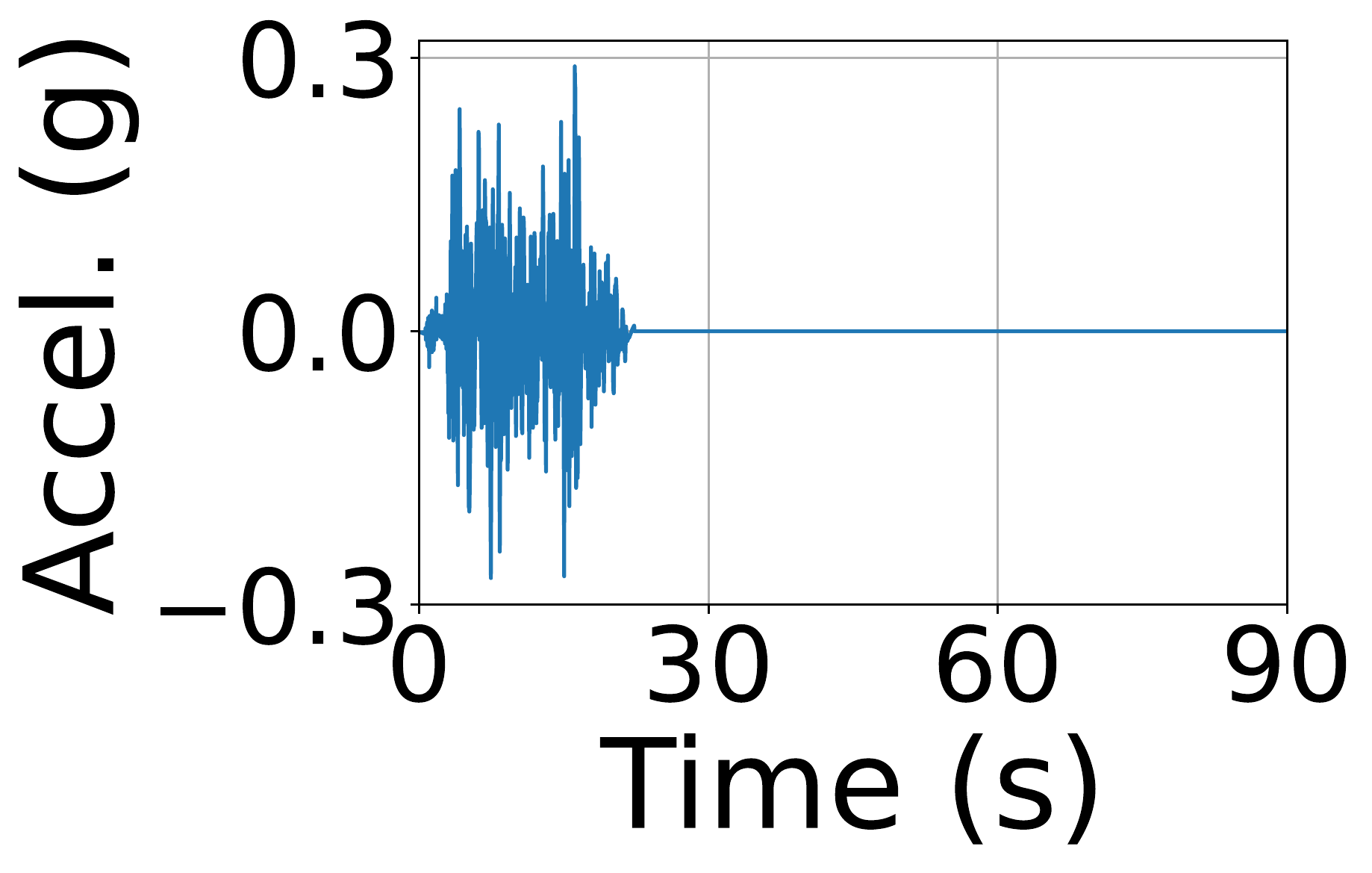}
\includegraphics[valign=c,width=0.19\textwidth]{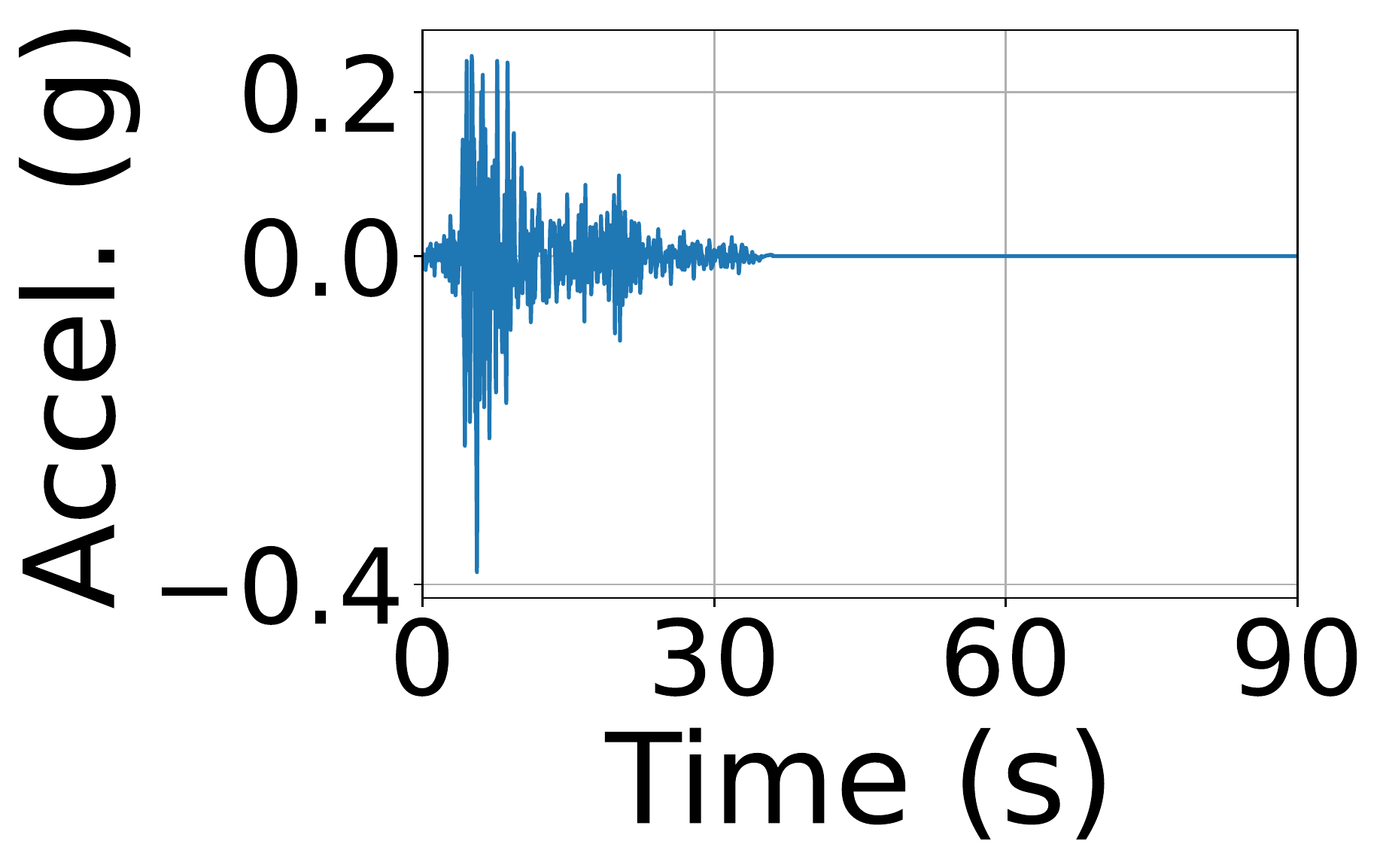} \\
\vspace*{0.35truecm}
\includegraphics[valign=c,width=0.19\textwidth]{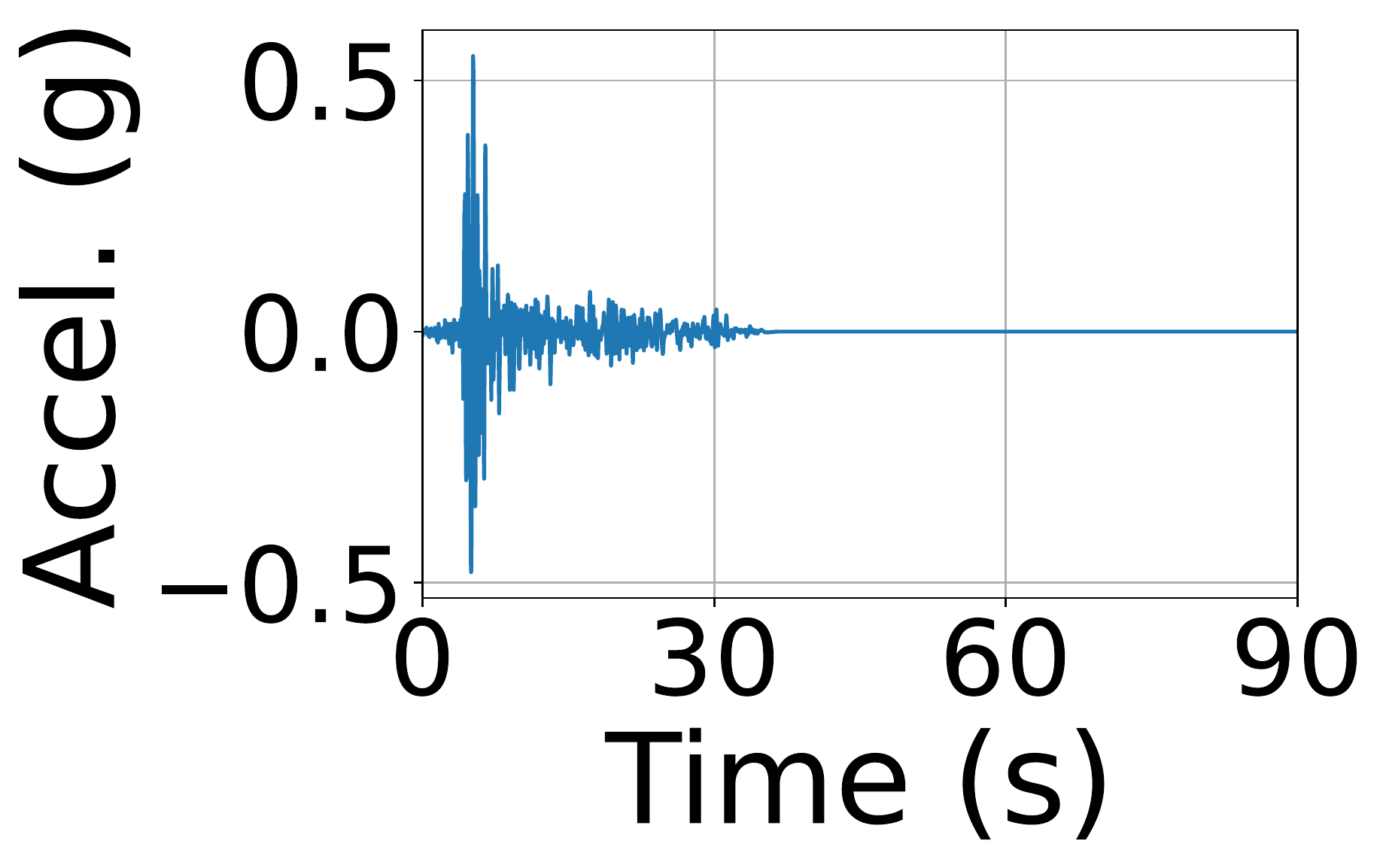} 
\includegraphics[valign=c,width=0.19\textwidth]{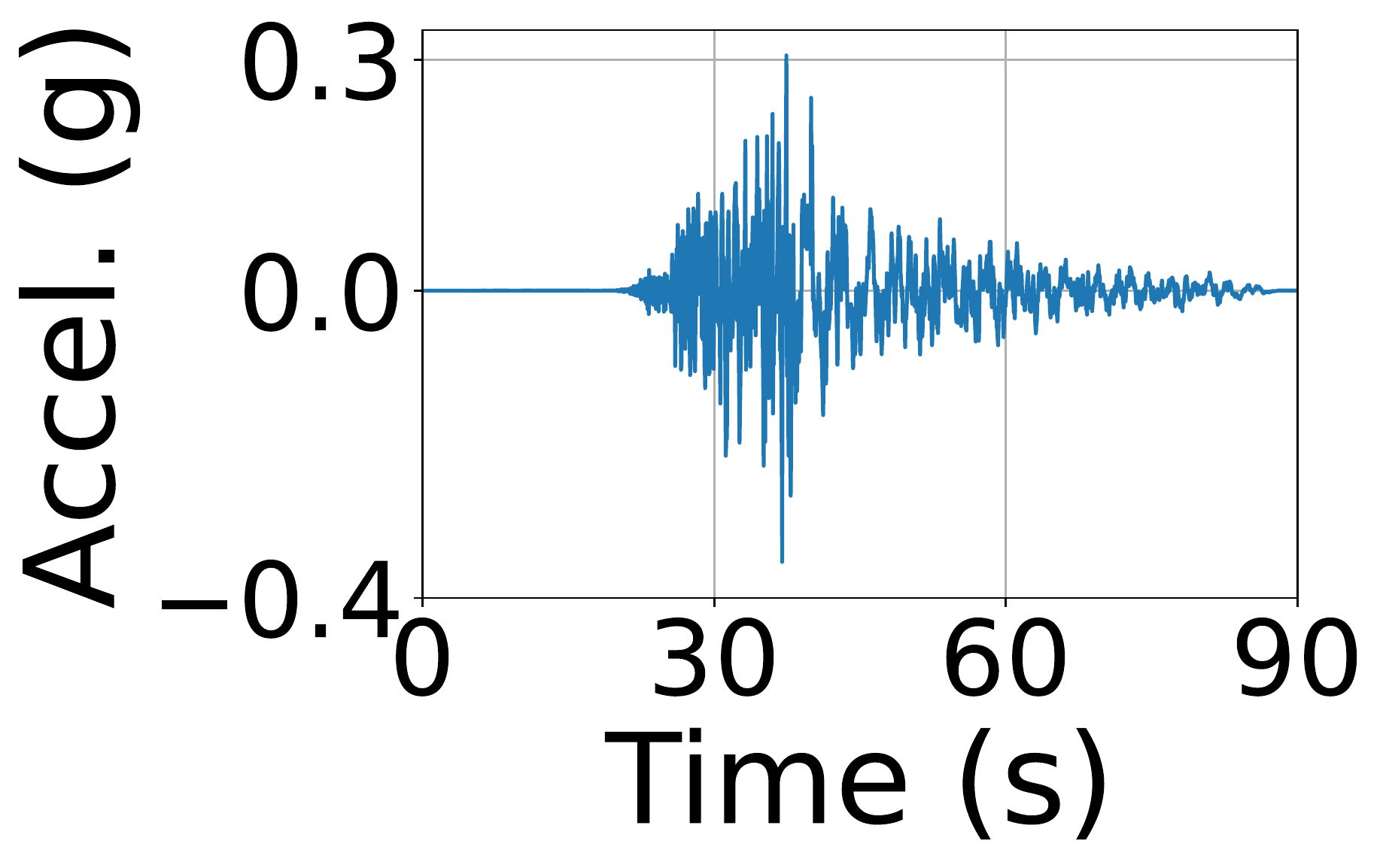} 
\includegraphics[valign=c,width=0.19\textwidth]{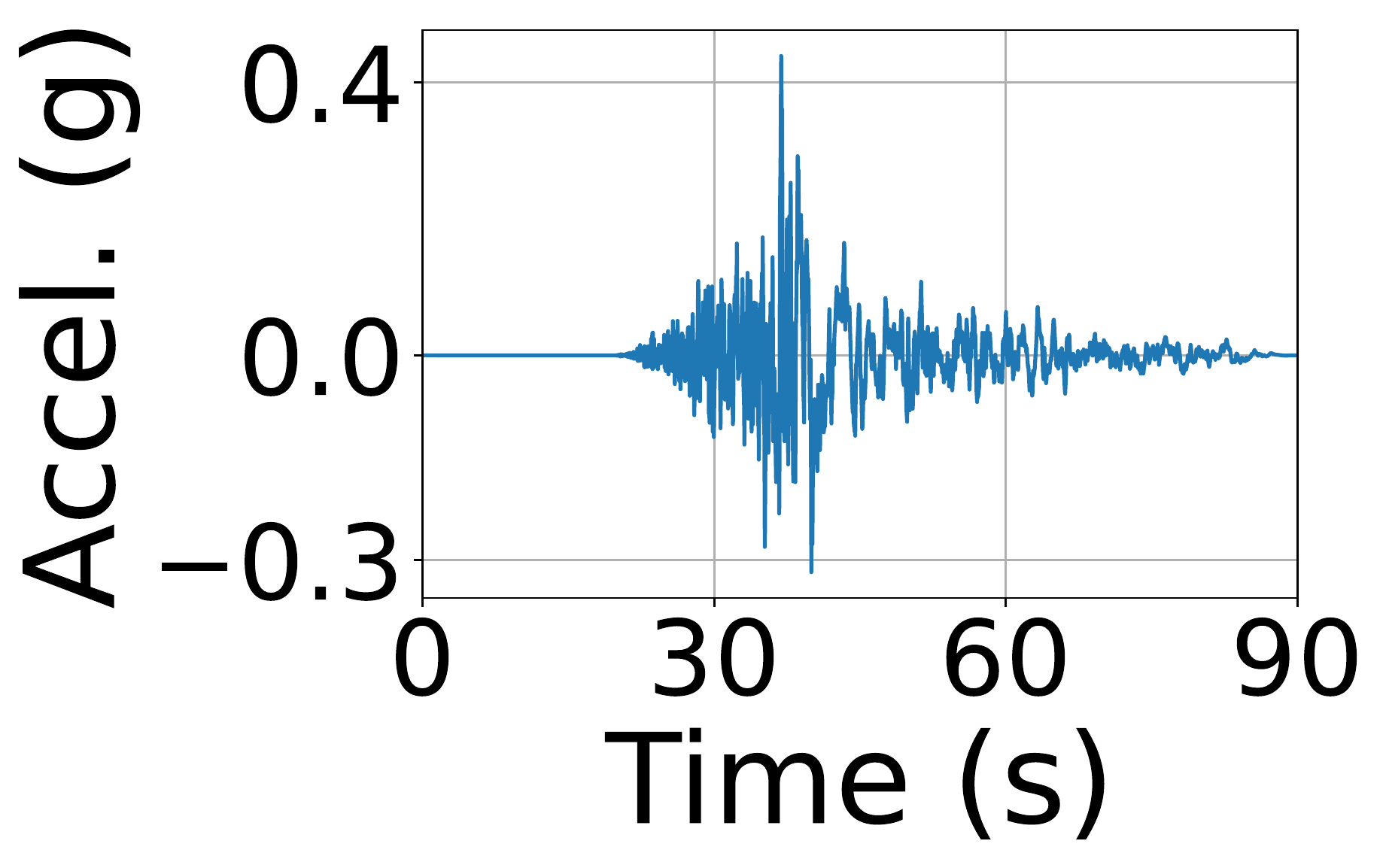}
\includegraphics[valign=c,width=0.19\textwidth]{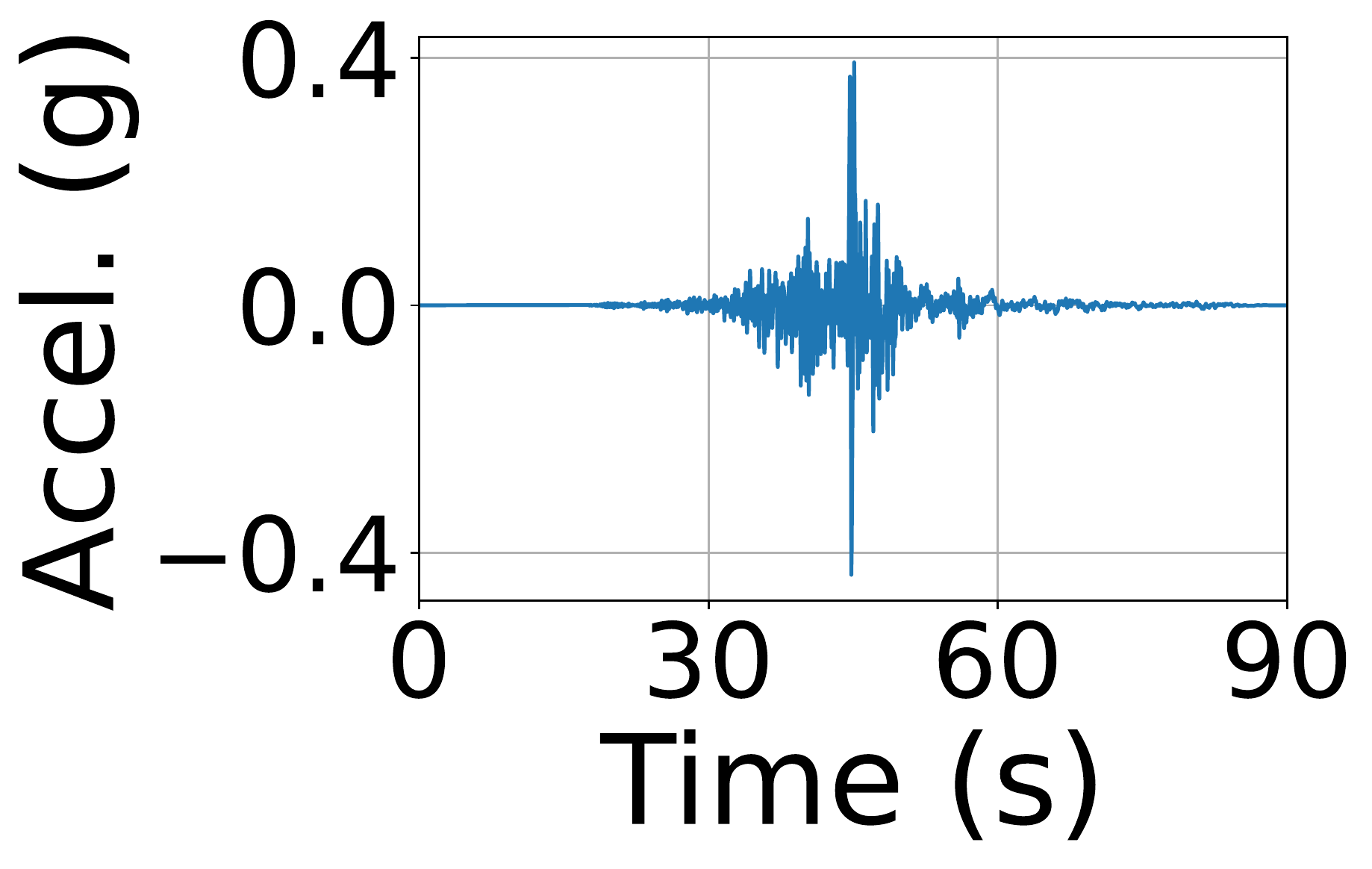}
\includegraphics[valign=c,width=0.19\textwidth]{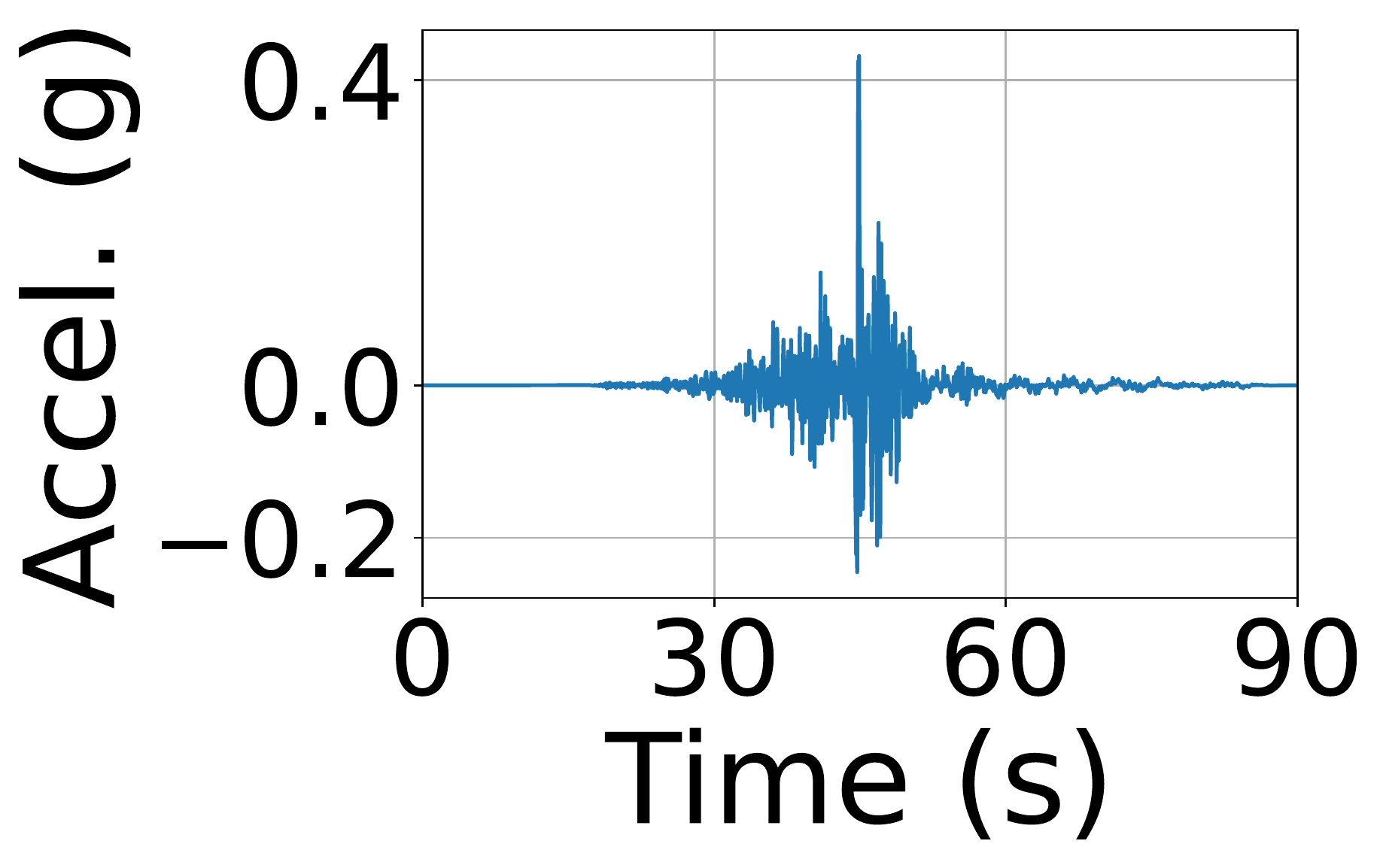} \\
\vspace*{0.35truecm}
\includegraphics[valign=c,width=0.19\textwidth]{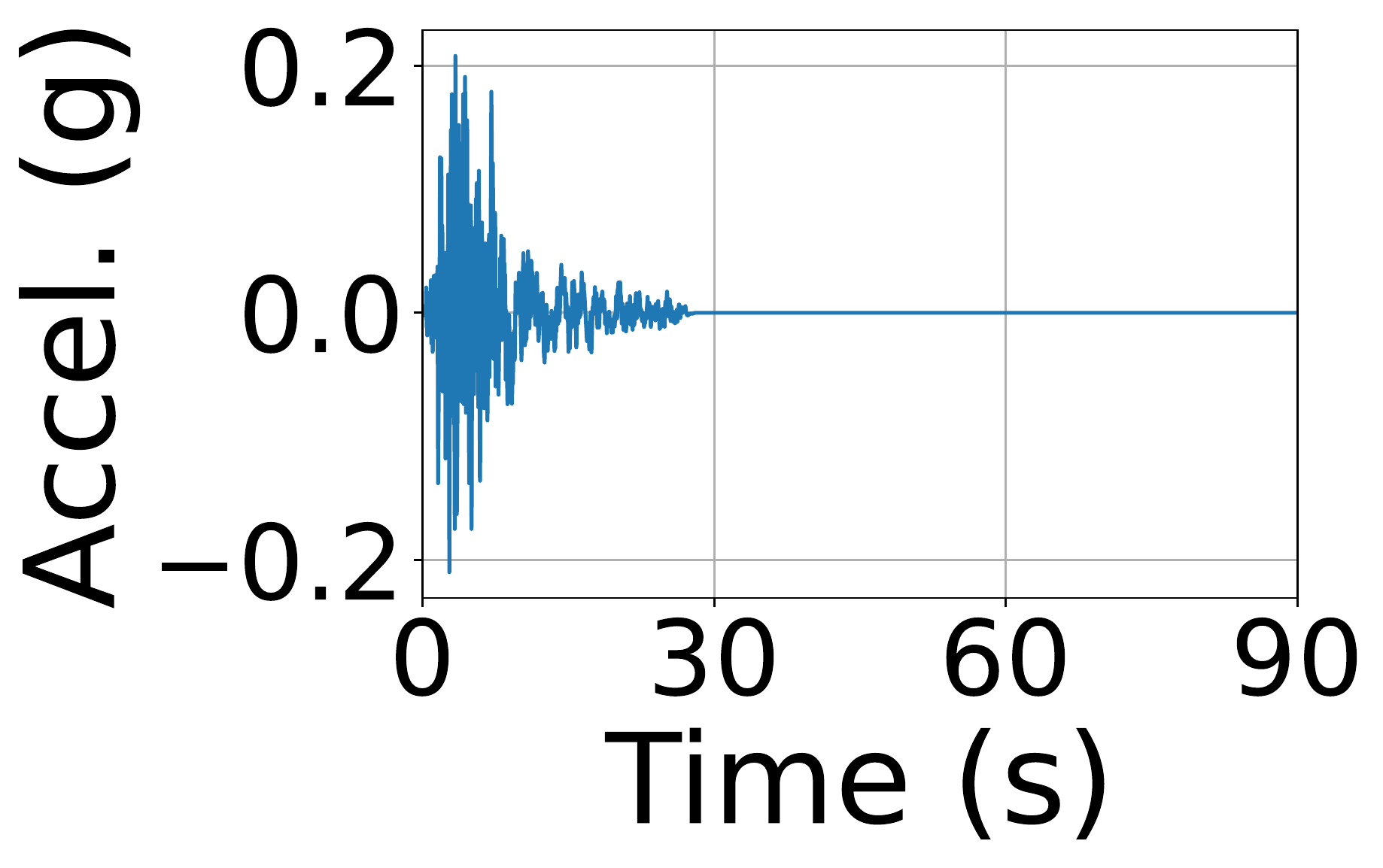} 
\includegraphics[valign=c,width=0.19\textwidth]{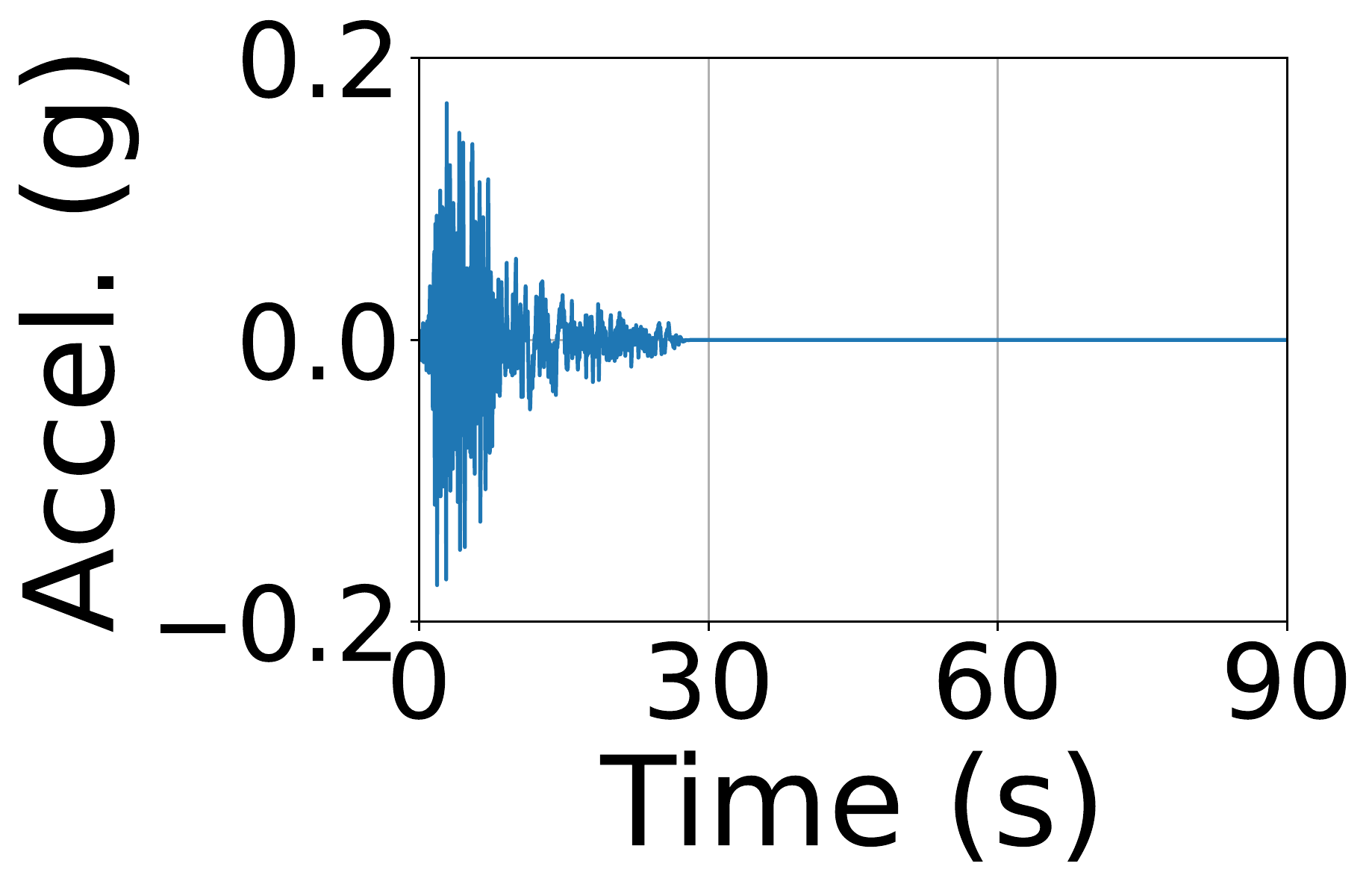} 
\includegraphics[valign=c,width=0.19\textwidth]{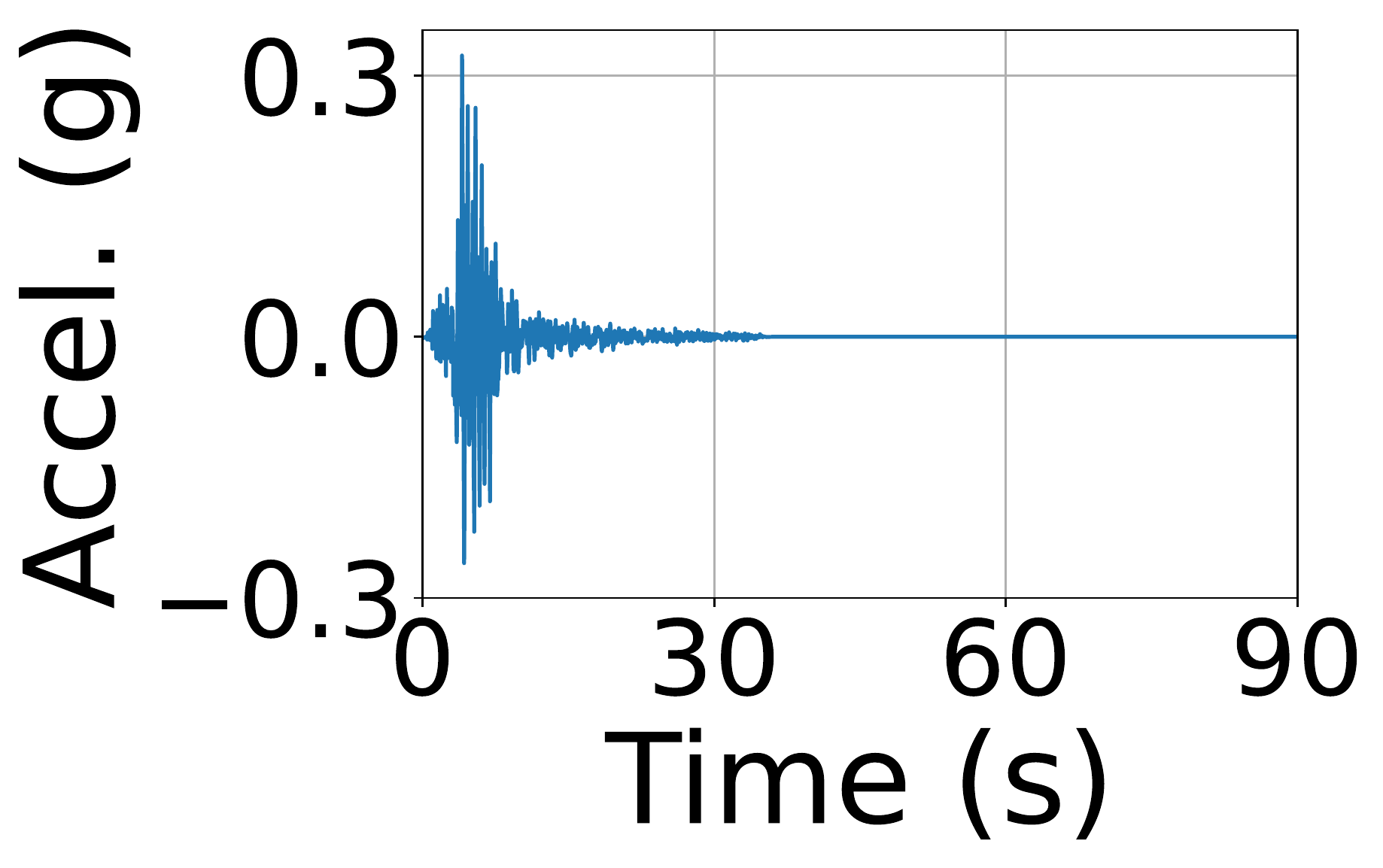}
\includegraphics[valign=c,width=0.19\textwidth]{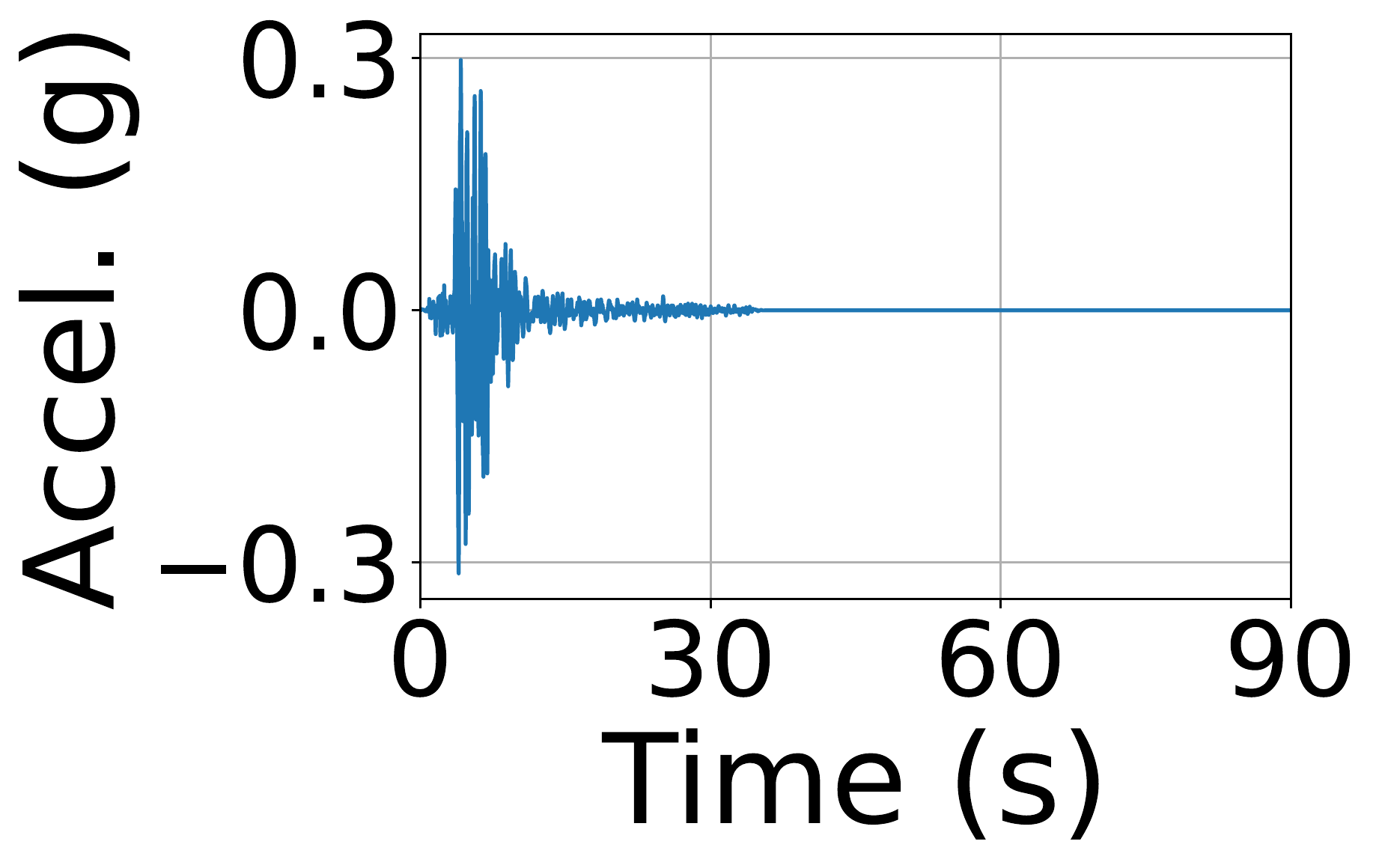}\\

\caption{The FEMA P695 earthquake suite.}
\label{eq_suite}
\end{center}
\end{figure}

The ground motion selection is one of the most important factors in training and developing accurate surrogate models, as the suite needs to be a good representation of the historic earthquakes.  In this study, a set of 22 pair of ground motions representing magnitudes of range 6.5-7.6 recorded on firm soil (rock or stiff, $V_s >$ 180 m/s) were selected from the FEMA P695 \cite{femap695} far-field suite. The list of the selected ground motions and their characteristics are presented in Table \ref{table_eq}. All the ground motions are recorded at sites located greater than or equal to 10 km from fault rupture (strike slip and reverse thrust), referred to as “far-field” motions. To ensure broad representation of different recorded earthquakes, this set contains far-field records selected from most large-magnitude events in the PEER NGA database \cite{doi:10.1193/1.2894831}. The inherent differences in event magnitude, distance to source, source type and site conditions of these ground motions considers the record-to-record variability that is essential to capture the ground motion uncertainties. More information on the record selection criteria can be found in \cite{femap695}. Figure \ref{eq_suite} shows the acceleration time histories of the earthquakes in the  suite. 

\begin{table}[H]
\caption{FEMA P695 far-field ground motions.}

\begin{tabular}{llp{0.2\textwidth}llll}
\hline
       &                          &                        &      &     &    \multicolumn{2}{c}{PGA (g)}\\
ID & Event                    & Station                & Soil & M   & Major & Minor \\
\hline 
\hline 
1      & Northridge, 1994         & Beverly Hills - Mulhol & D    & 6.7 & 0.516 & 0.416 \\
2      & Northridge, 1994         & Canyon County - WLC    & D    & 6.7 & 0.482 & 0.410 \\
3      & Duzce, 1999              & Bolu                   & D    & 7.1 & 0.822 & 0.728 \\
4      & Hector Mine, 1999        & Hector                 & C    & 7.1 & 0.337 & 0.266 \\
5      & Imperial Valley, 1979    & Delta                  & D    & 6.5 & 0.351 & 0.238 \\
6      & Imperial Valley, 1979    & El Centro Array \#11   & D    & 6.5 & 0.380 & 0.364 \\
7      & Kobe, Japan, 1995        & Nishi-Akashi           & C    & 6.9 & 0.509 & 0.503 \\
8      & Kobe, Japan, 1995        & Shin-Osaka             & D    & 6.9 & 0.243 & 0.212 \\
9      & Kocaeli, Turkey, 1999    & Duzce                  & D    & 7.5 & 0.358 & 0.312 \\
10     & Kocaeli, Turkey, 1999    & Arcelik                & C    & 7.5 & 0.219 & 0.150 \\
11     & Landers, 1992            & Yermo Fire Station     & D    & 7.3 & 0.245 & 0.152 \\
12     & Landers, 1992            & Coolwater              & D    & 7.3 & 0.417 & 0.283 \\
13     & Loma Prieta, 1989        & Capitola               & D    & 6.9 & 0.529 & 0.443 \\
14     & Loma Prieta, 1989        & Gilroy Array \#3       & D    & 6.9 & 0.555 & 0.367 \\
15     & Manji, Iran, 1990        & Abbar                  & C    & 7.4 & 0.515 & 0.496 \\
16     & Superstition Hills, 1987 & El Centro Imp. Co.     & D    & 6.5 & 0.358 & 0.258 \\
17     & Superstition Hills, 1987 & Poe Road (temp)        & D    & 6.5 & 0.446 & 0.300 \\
18     & Cape Mendocino, 1992     & Rio Dell Overpass      & D    & 7.0 & 0.549 & 0.385 \\
19     & Chi-Chi, Taiwan, 1999    & CHY101                 & D    & 7.6 & 0.440 & 0.353 \\
20     & Chi-Chi, Taiwan, 1999    & TCU045                 & C    & 7.6 & 0.512 & 0.474 \\
21     & San Fernando, 1971       & LA - Hollywood Stor    & D    & 6.6 & 0.210 & 0.174 \\
22     & Friuli, Italy, 1976      & Tolmezzo               & C    & 6.5 & 0.351 & 0.315\\
\hline 
\end{tabular}
\label{table_eq}
\end{table}

\subsection{Characterization of ground motion using SVD}
\begin{figure}[h!]
\begin{center}
\subfigure[\label{exp_variance_cum} ]{\includegraphics[width=0.47\textwidth]{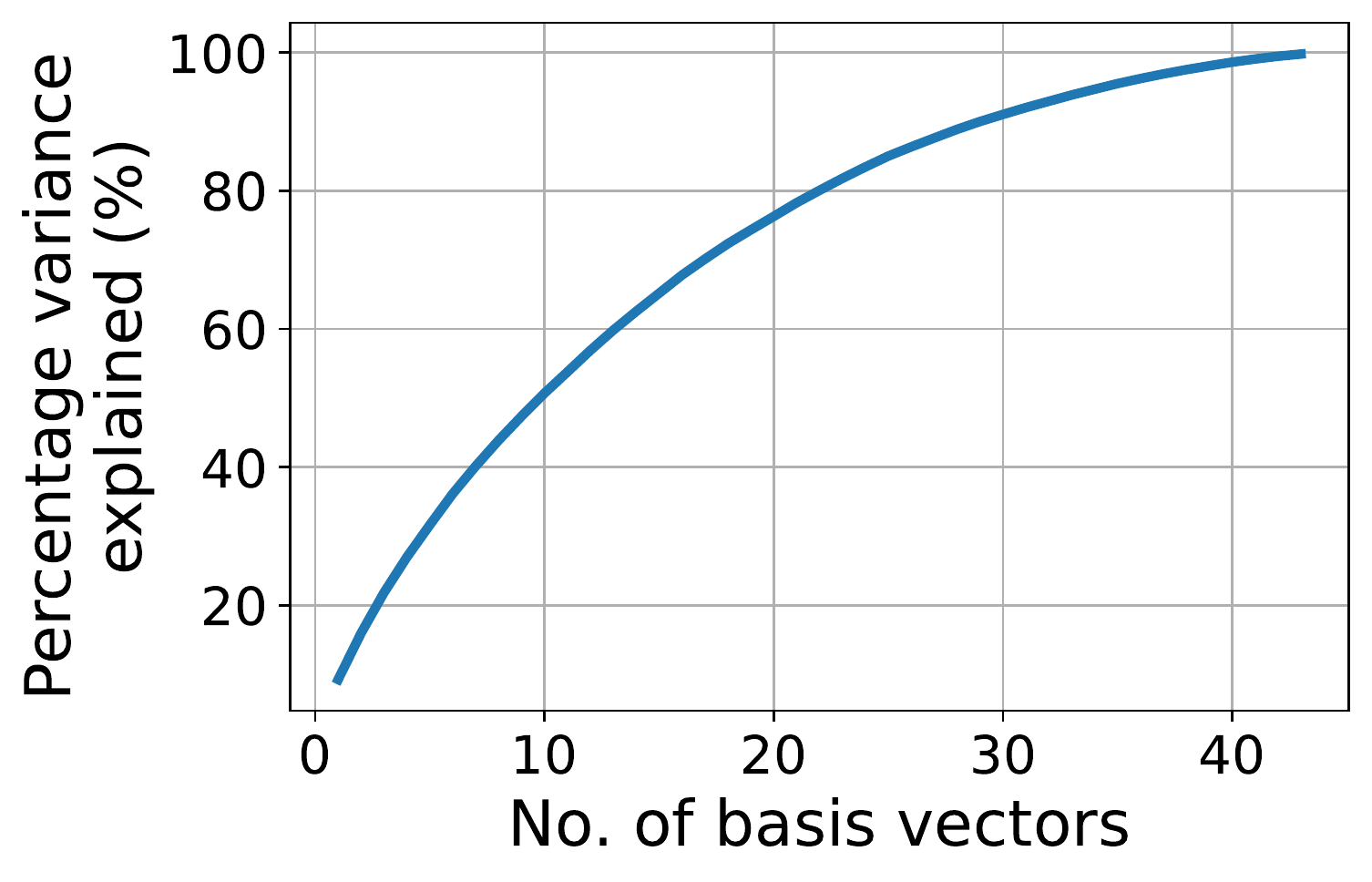}} 
\hspace*{0.2truecm}
\subfigure[\label{exp_variance} ]{\includegraphics[width=0.47\textwidth]{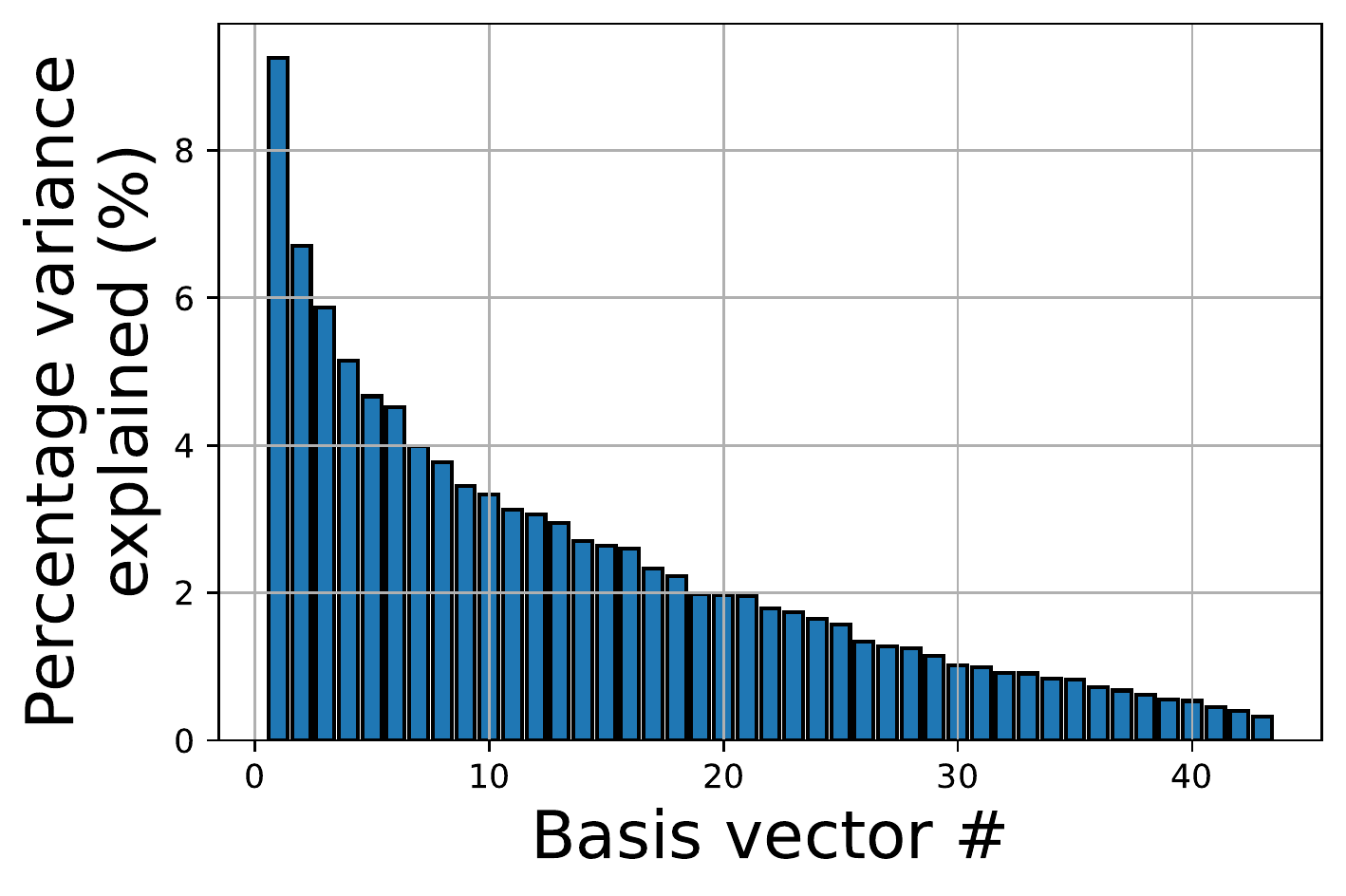}} 
\caption{(a)  Cumulative percentage of variance explained by increasing the number of basis vectors used. (b) Contribution of each basis vector to the  total explained variance.}
\label{svd_efficiency}
\end{center}
\end{figure}

\begin{figure}
\begin{center}
\includegraphics[valign=c,width=0.19\textwidth]{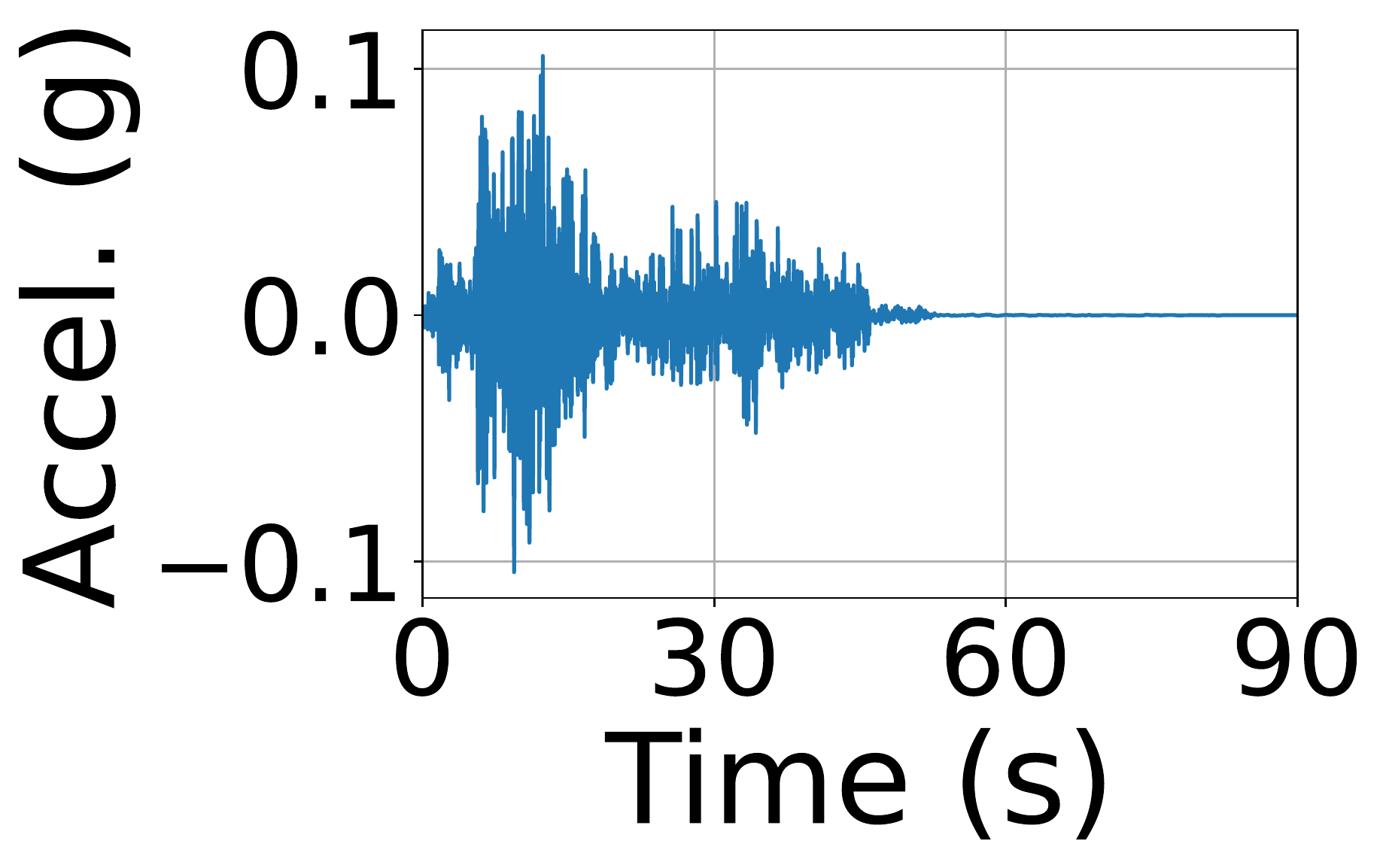} 
\includegraphics[valign=c,width=0.19\textwidth]{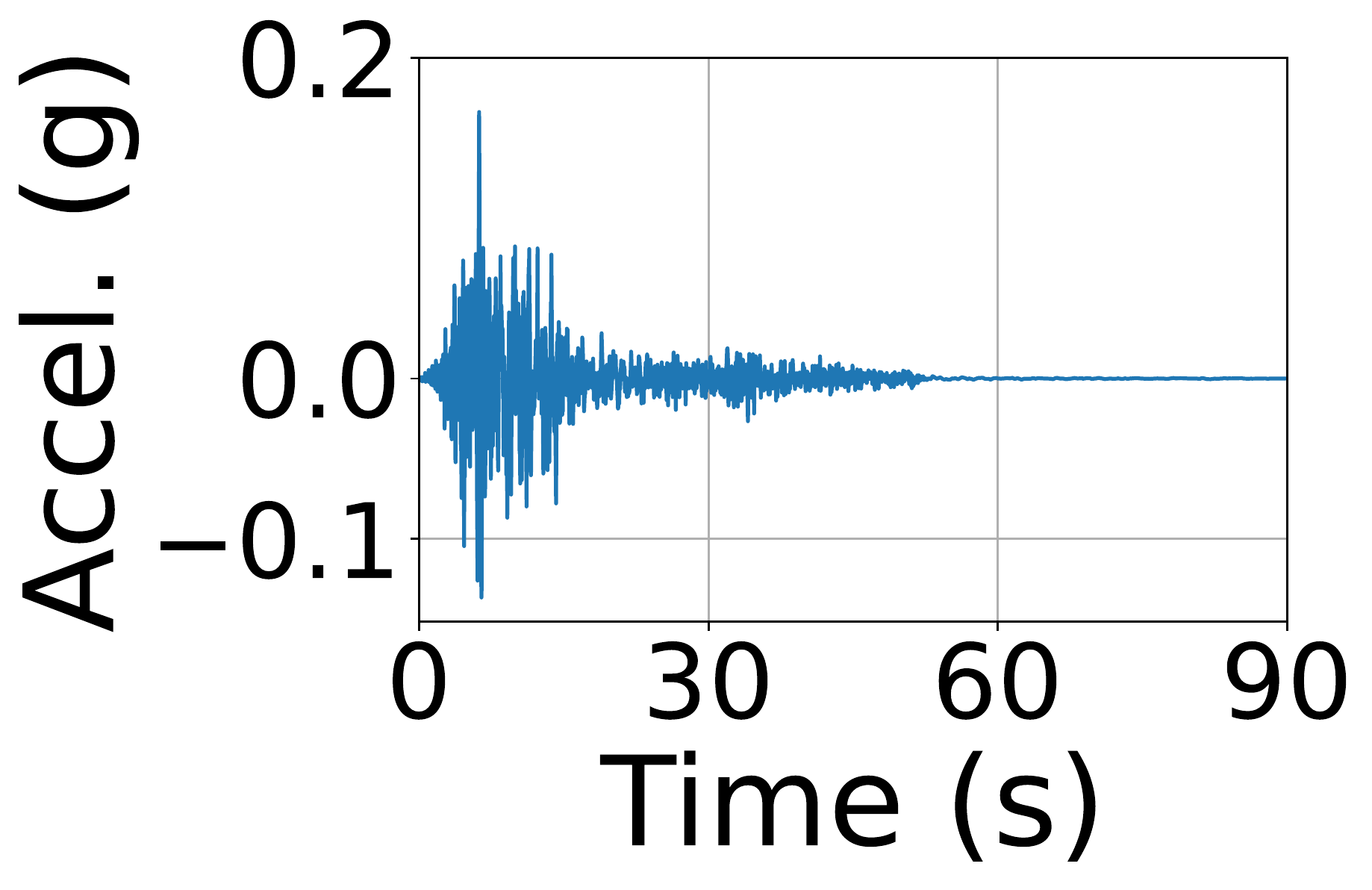} 
\includegraphics[valign=c,width=0.19\textwidth]{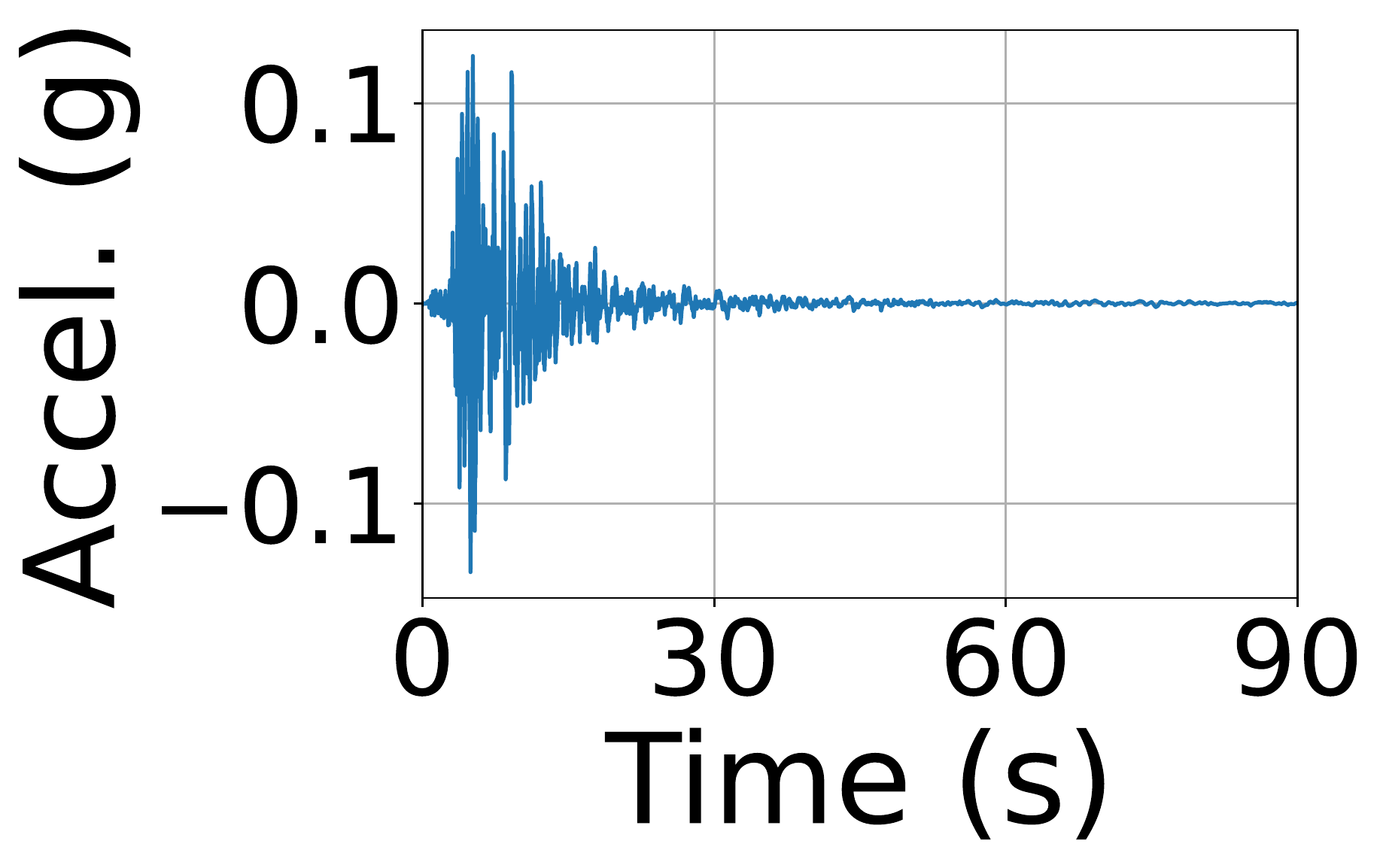}
\includegraphics[valign=c,width=0.19\textwidth]{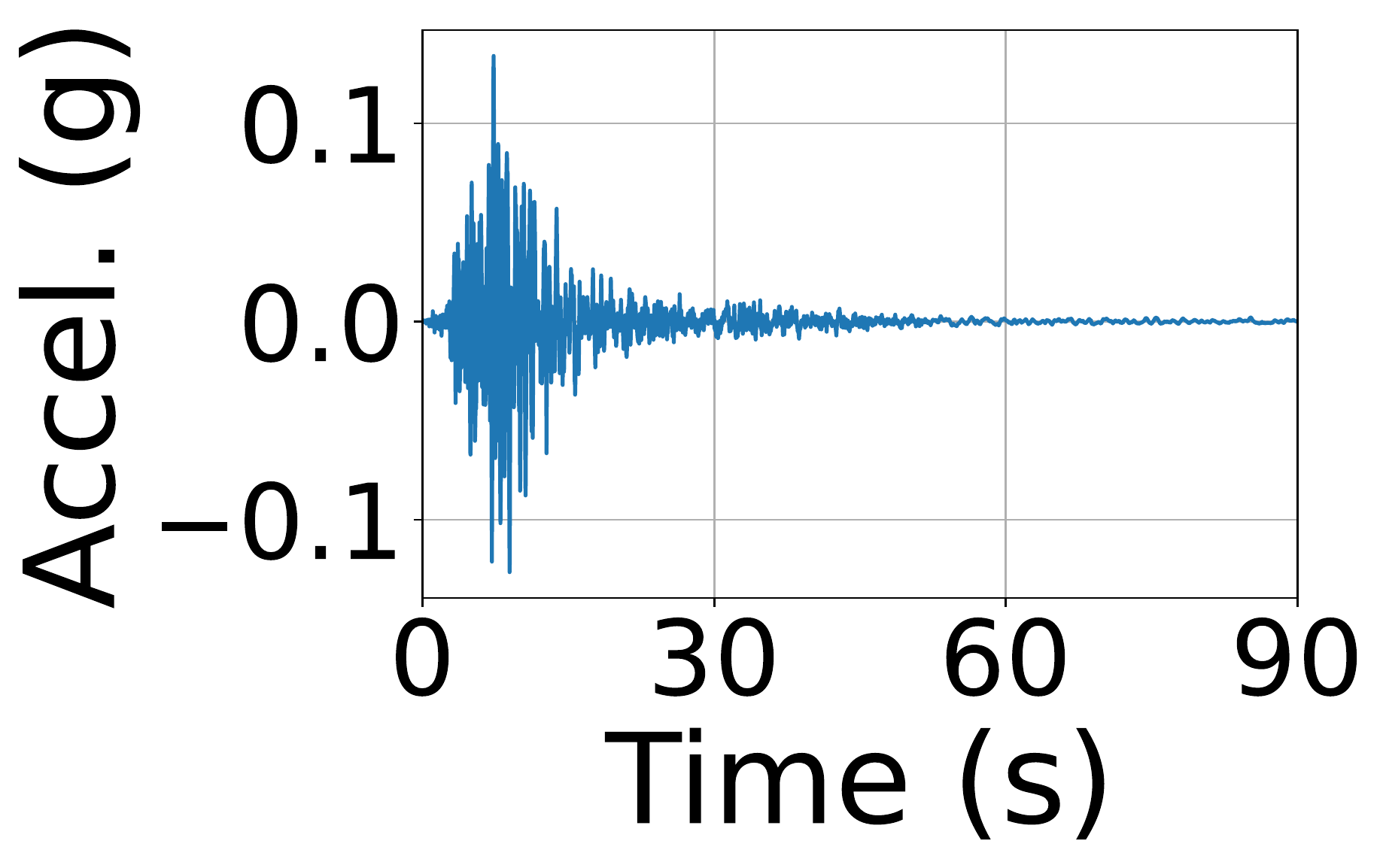}
\includegraphics[valign=c,width=0.19\textwidth]{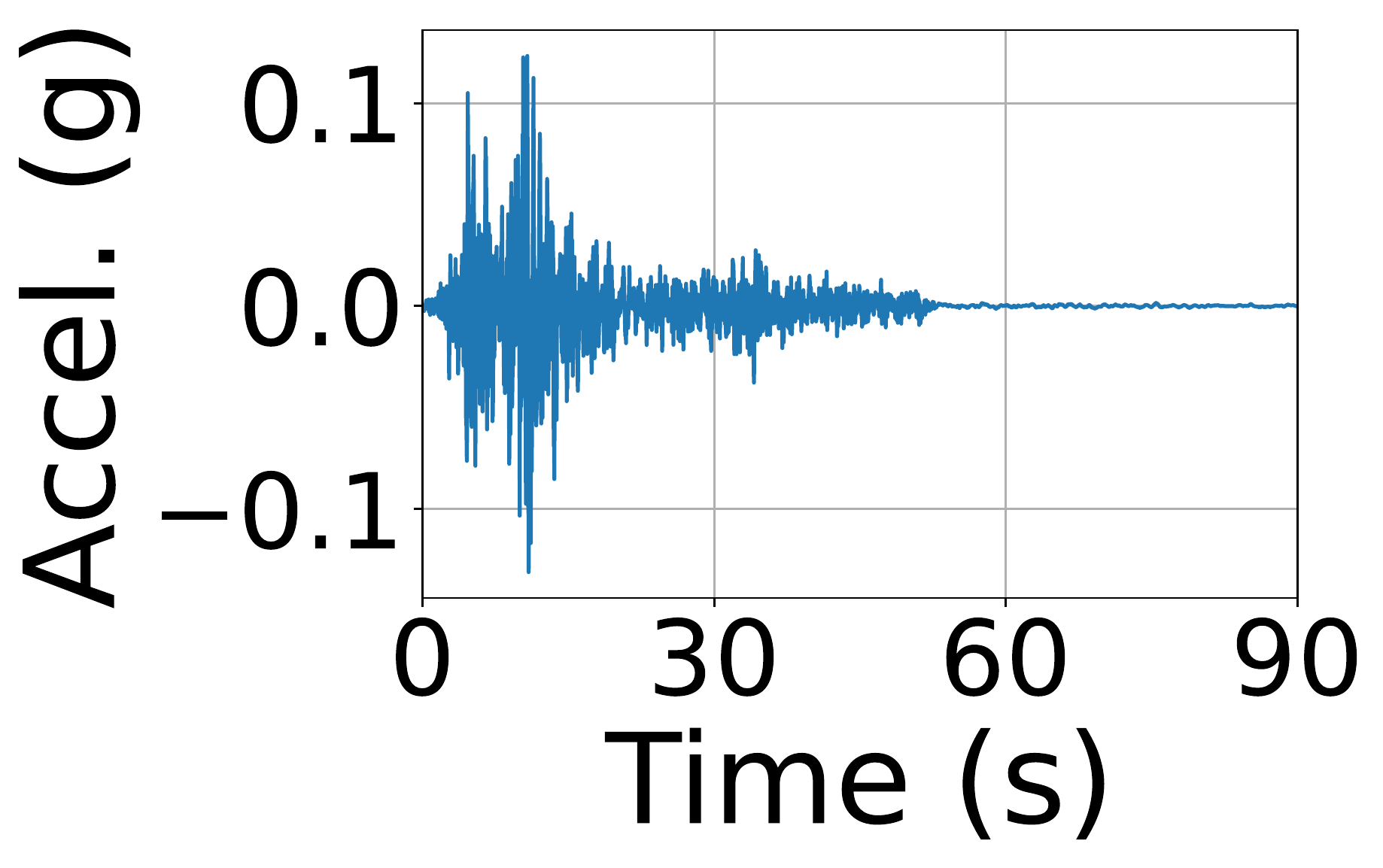} \\
\vspace*{0.35truecm}
\includegraphics[valign=c,width=0.19\textwidth]{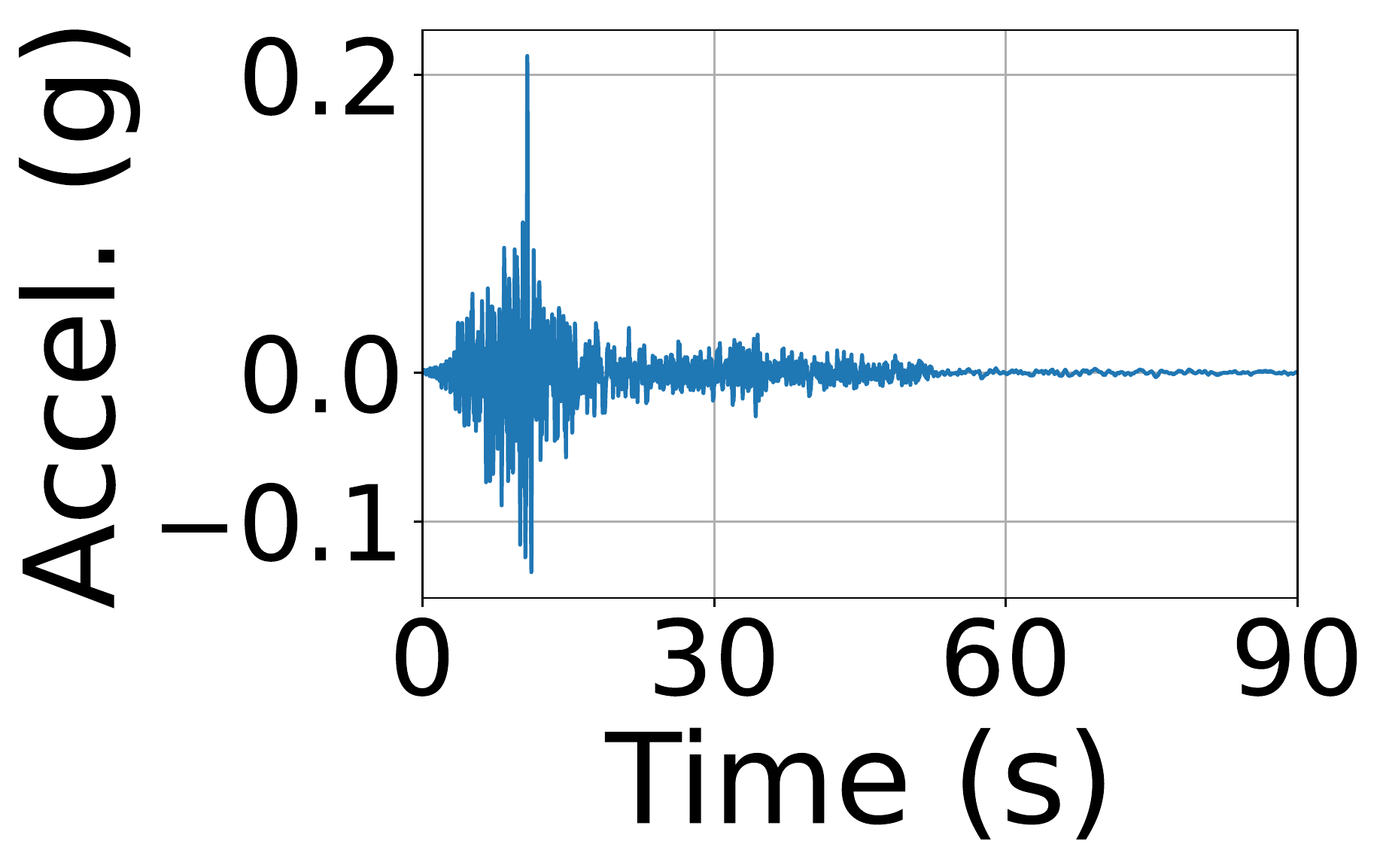} 
\includegraphics[valign=c,width=0.19\textwidth]{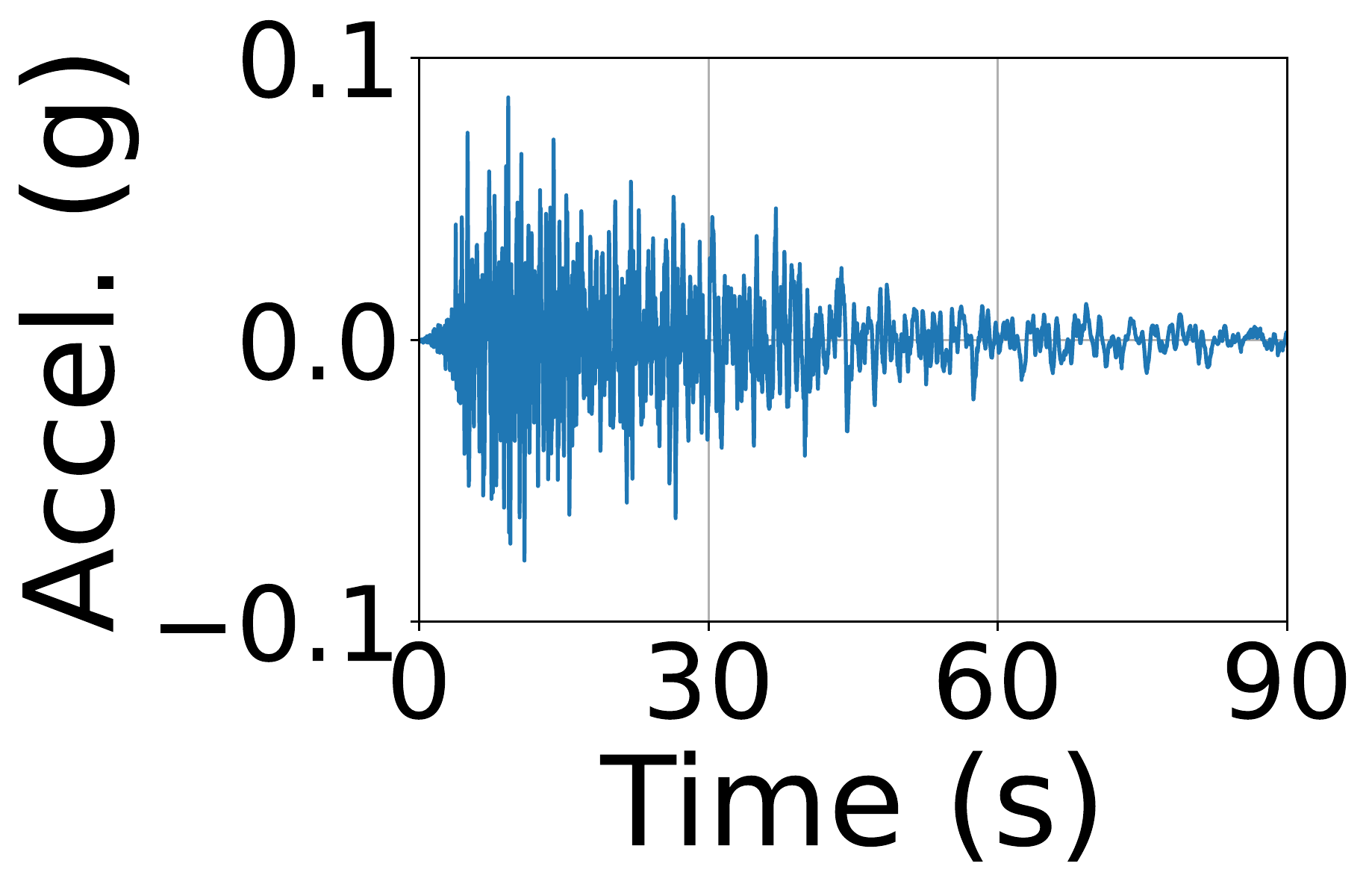} 
\includegraphics[valign=c,width=0.19\textwidth]{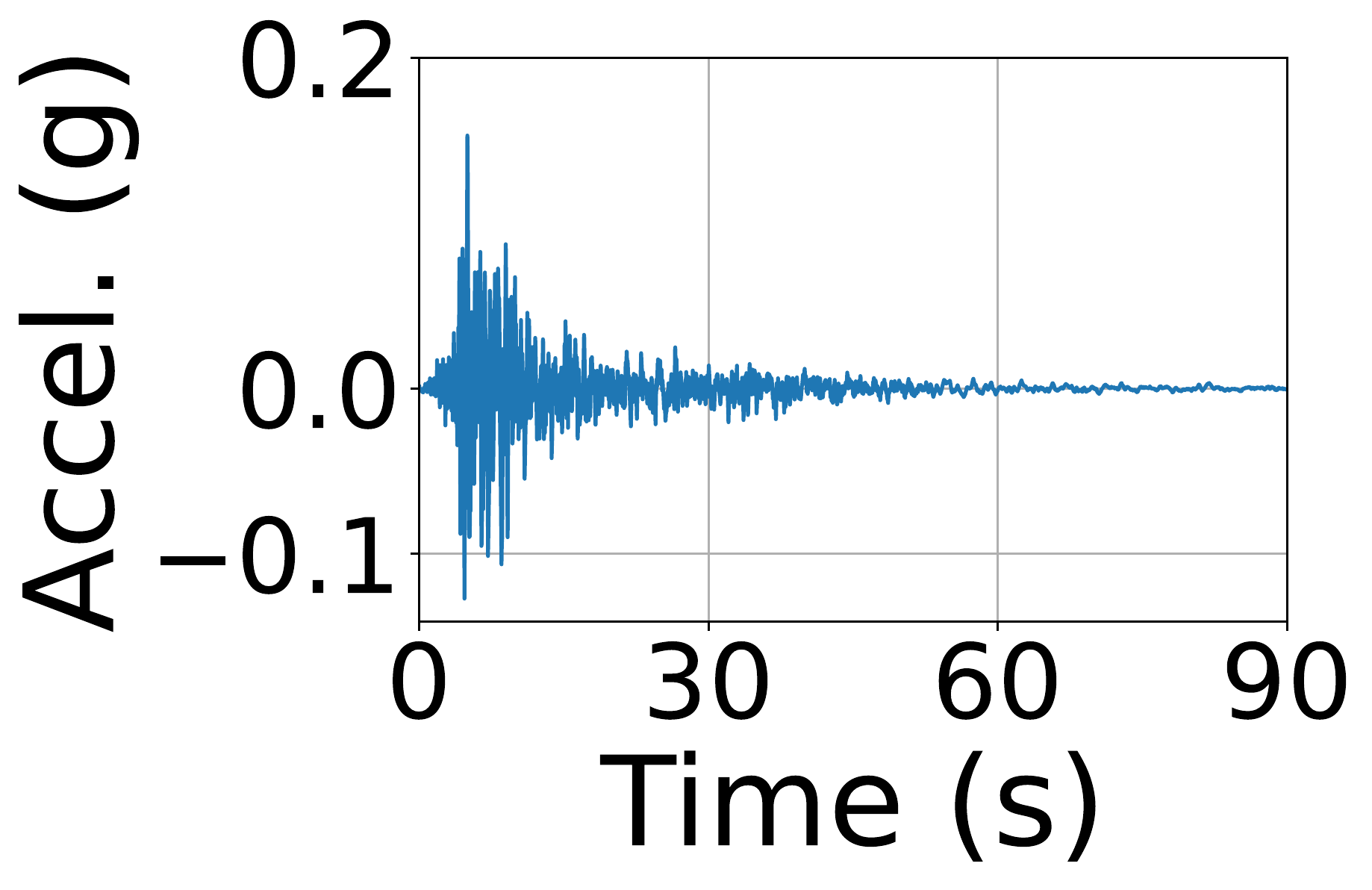}
\includegraphics[valign=c,width=0.19\textwidth]{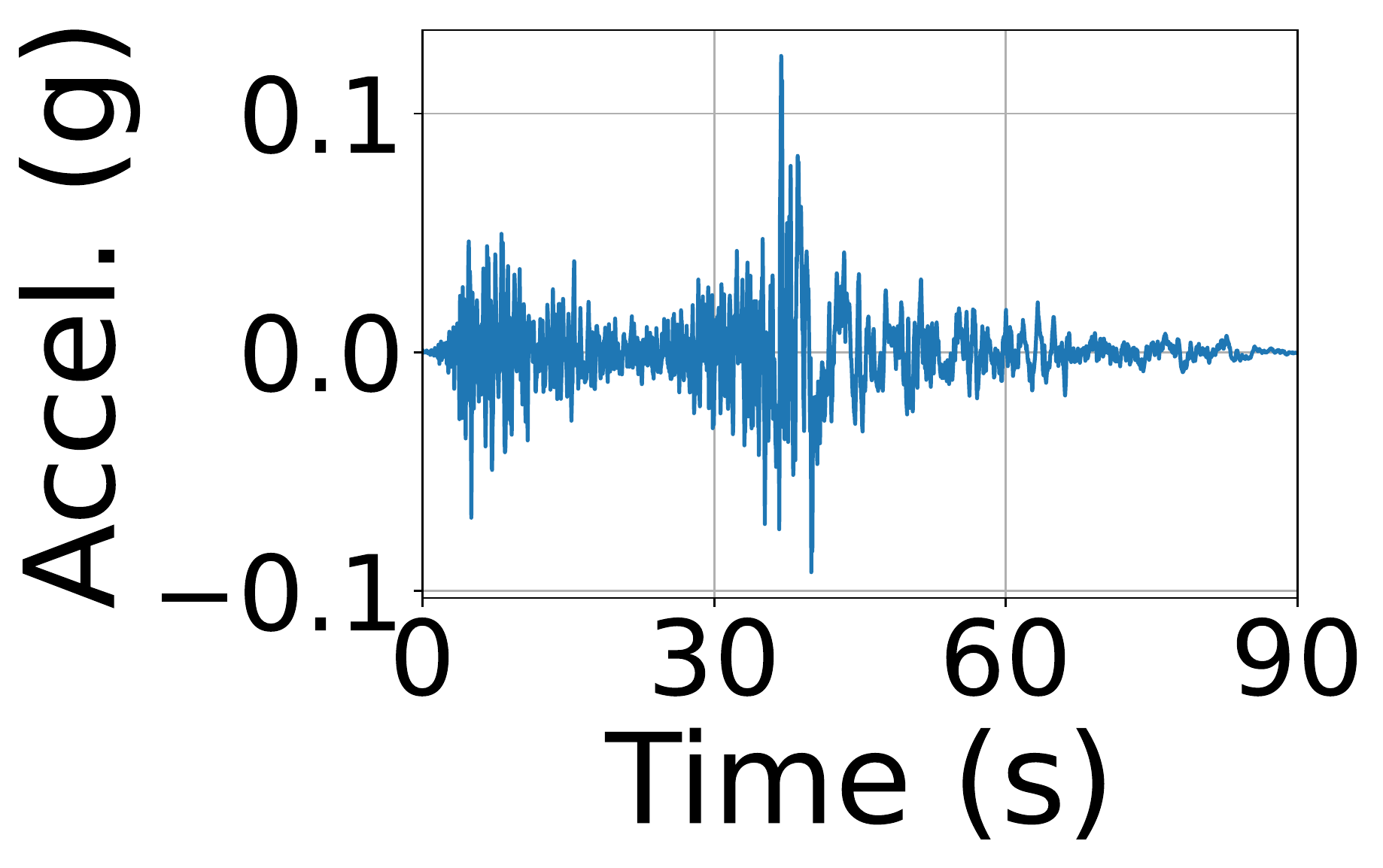}
\includegraphics[valign=c,width=0.19\textwidth]{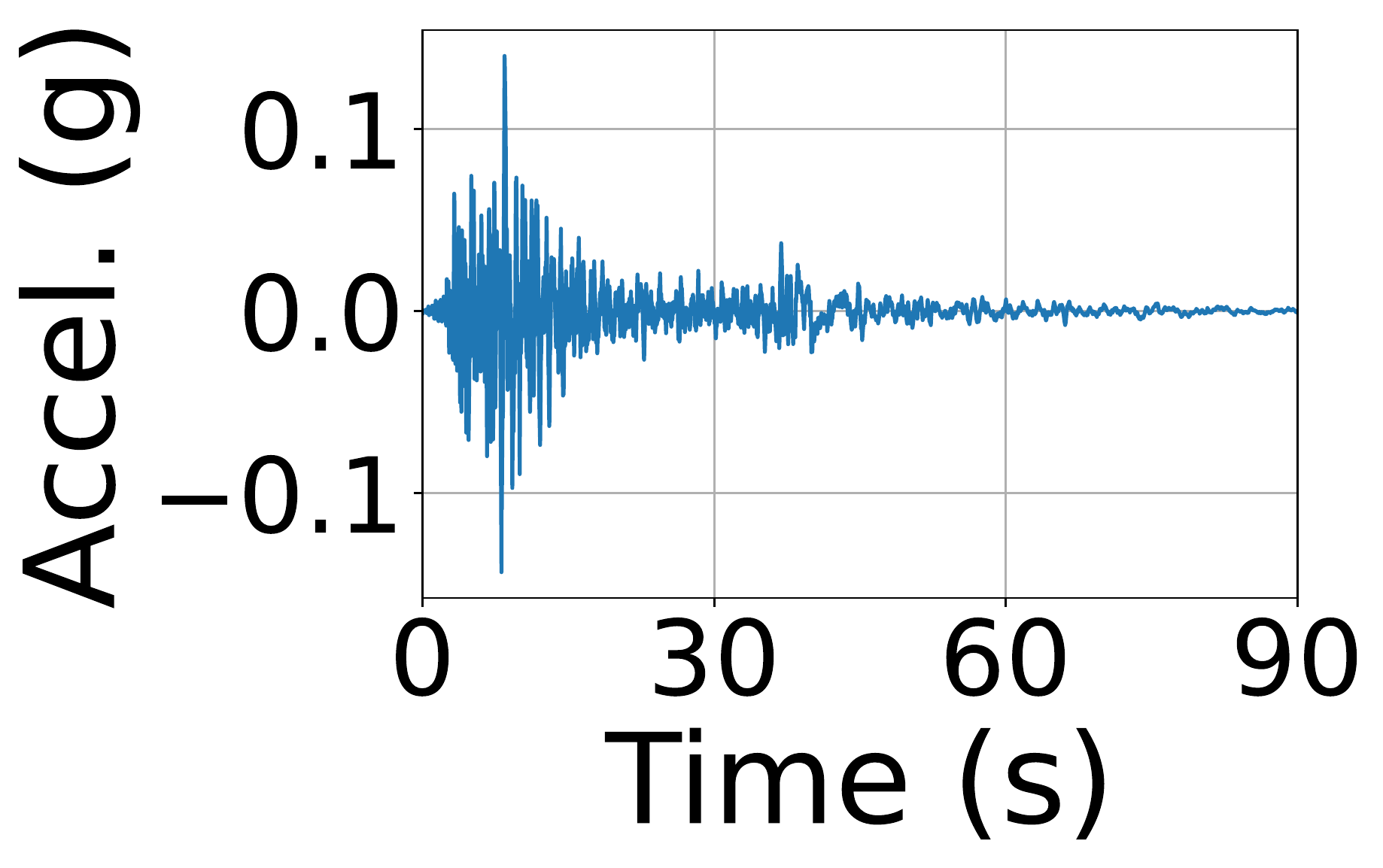} \\
\vspace*{0.35truecm}
\includegraphics[valign=c,width=0.19\textwidth]{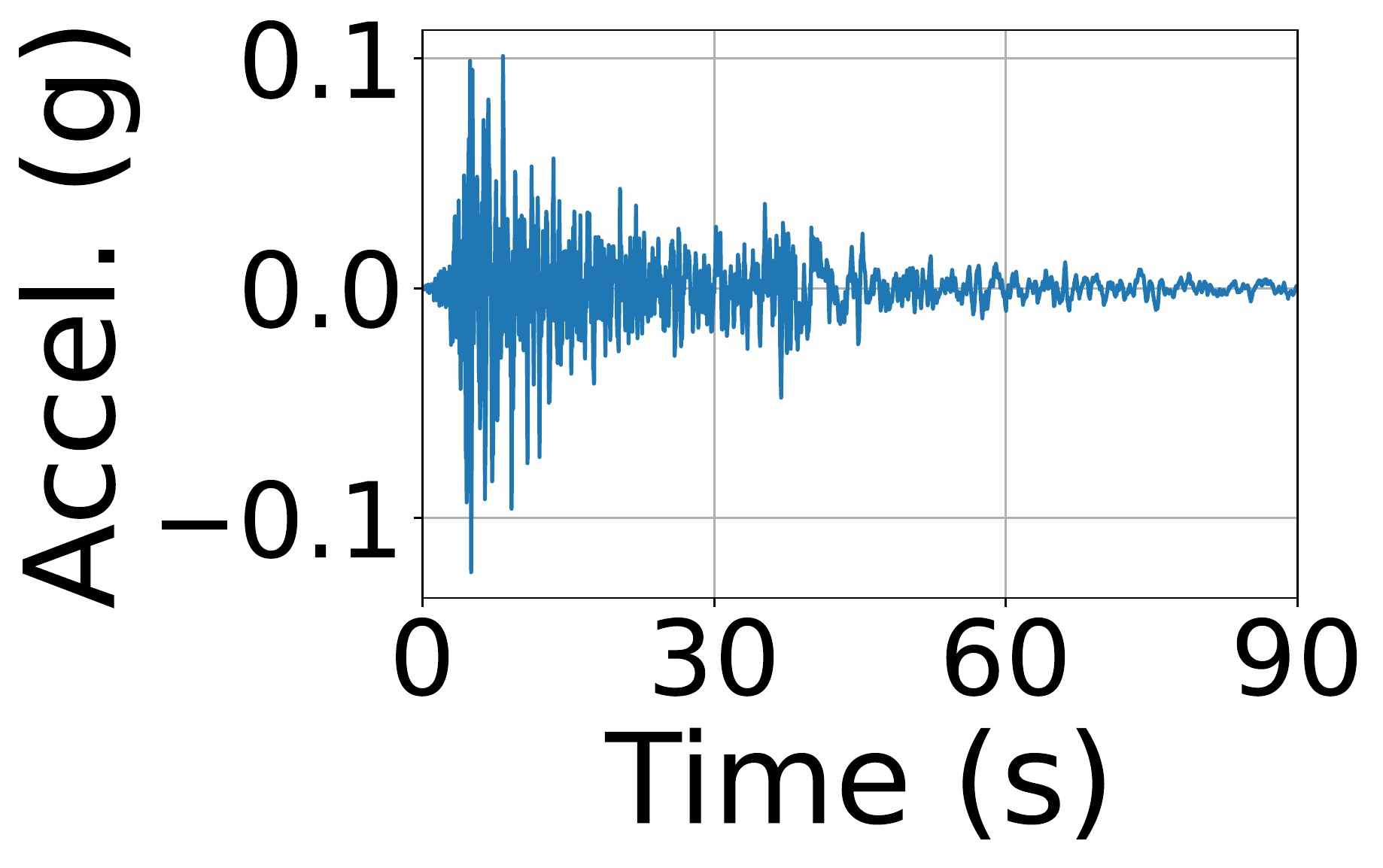} 
\includegraphics[valign=c,width=0.19\textwidth]{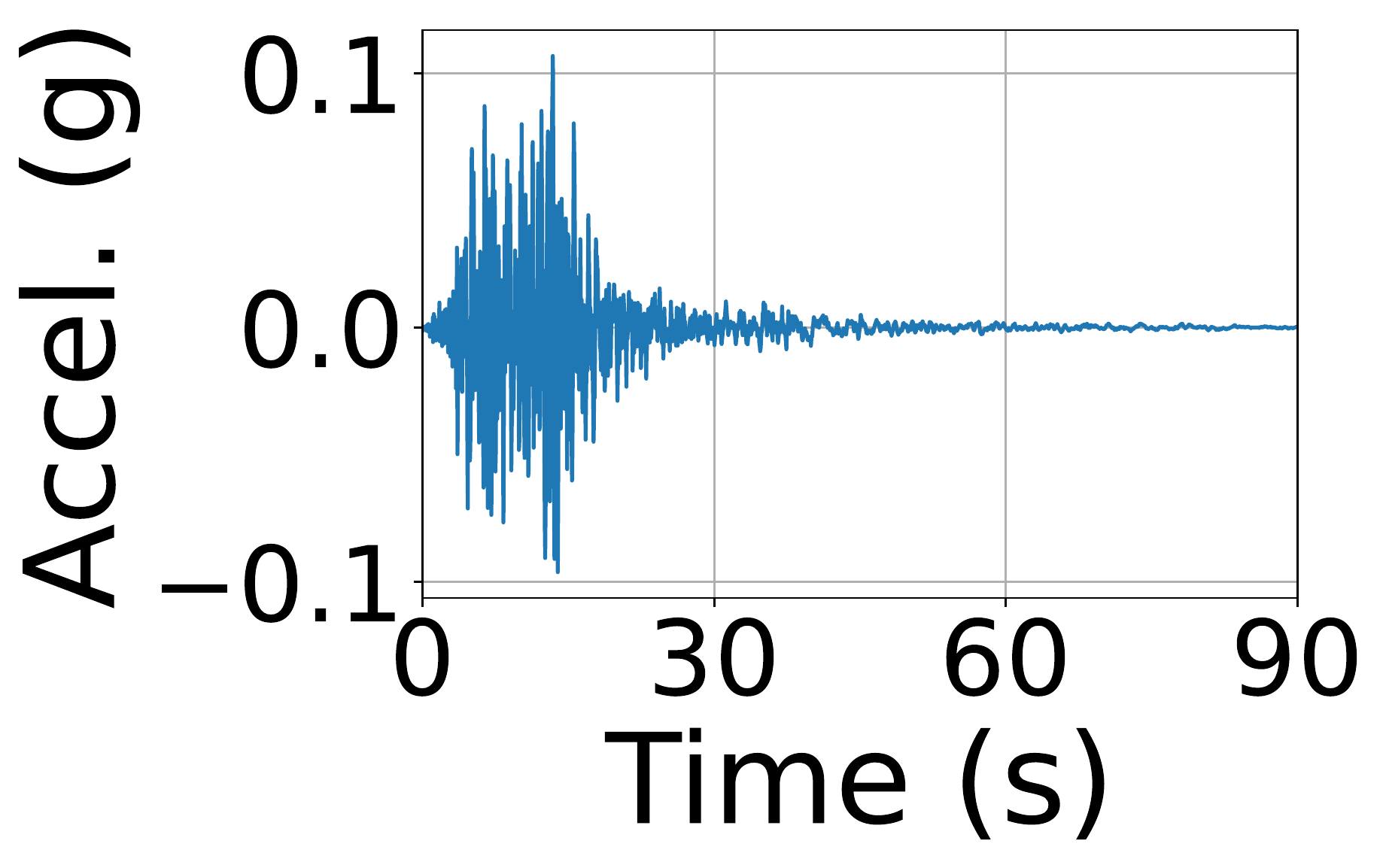} 
\includegraphics[valign=c,width=0.19\textwidth]{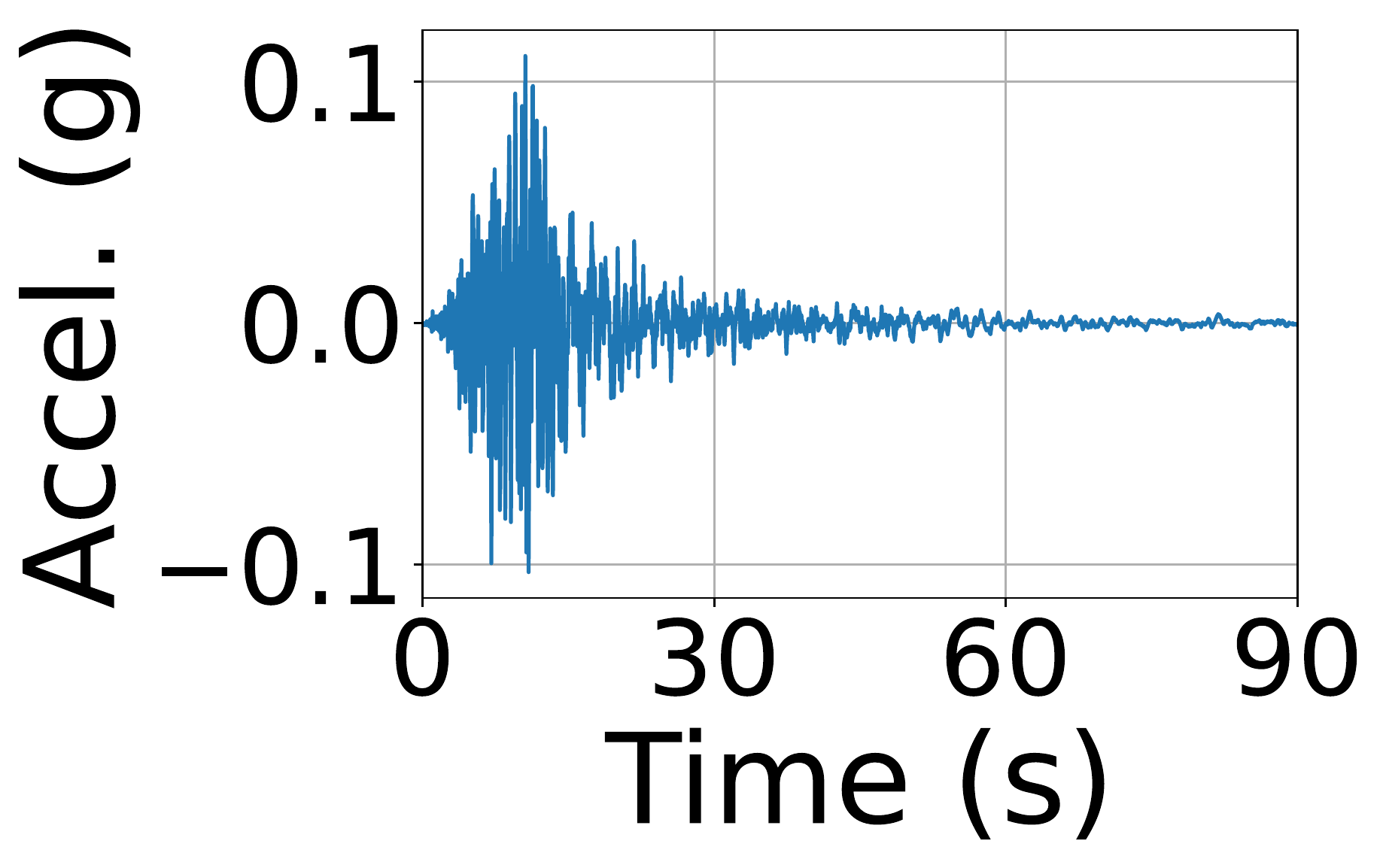}
\includegraphics[valign=c,width=0.19\textwidth]{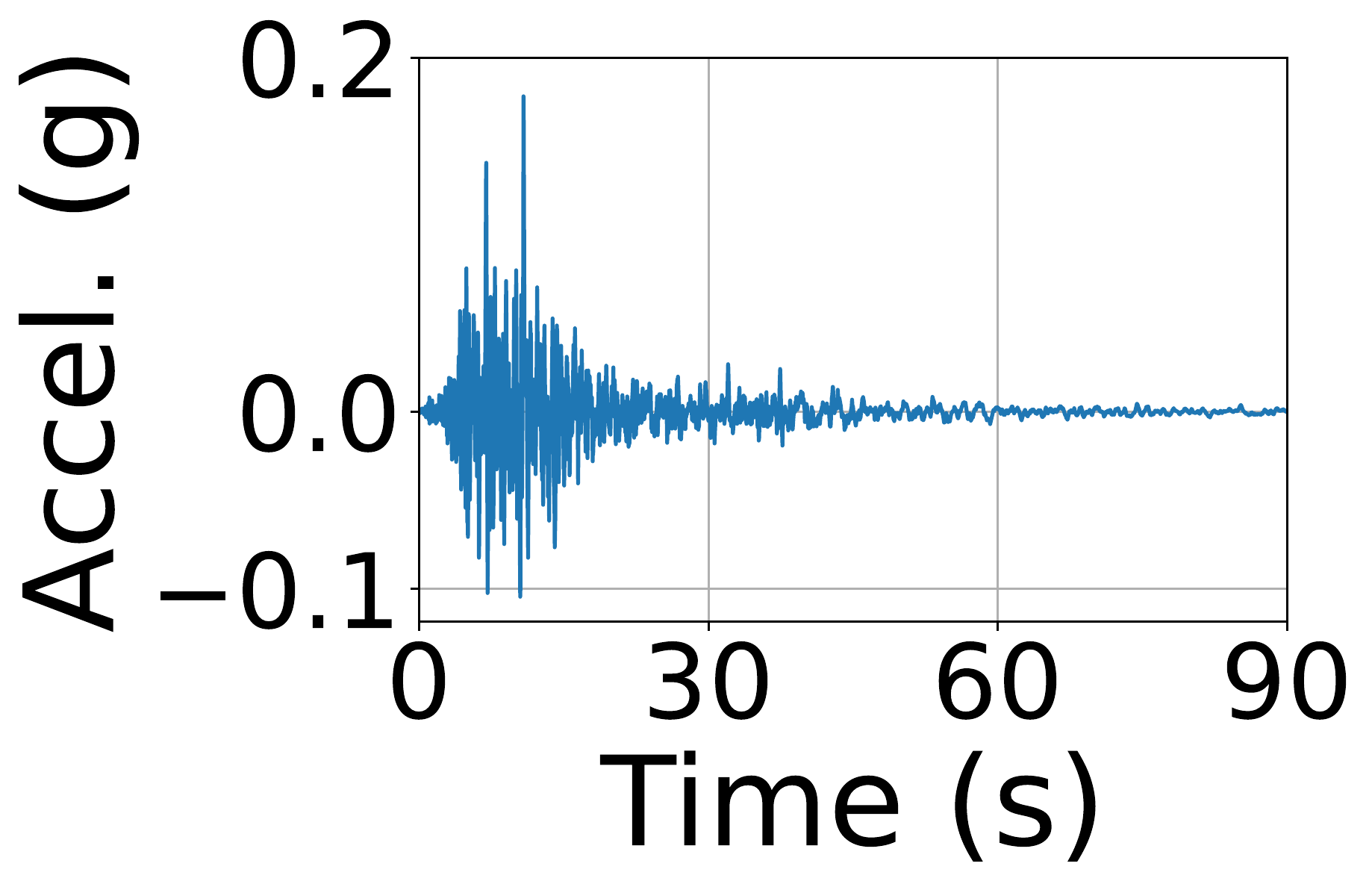}
\includegraphics[valign=c,width=0.19\textwidth]{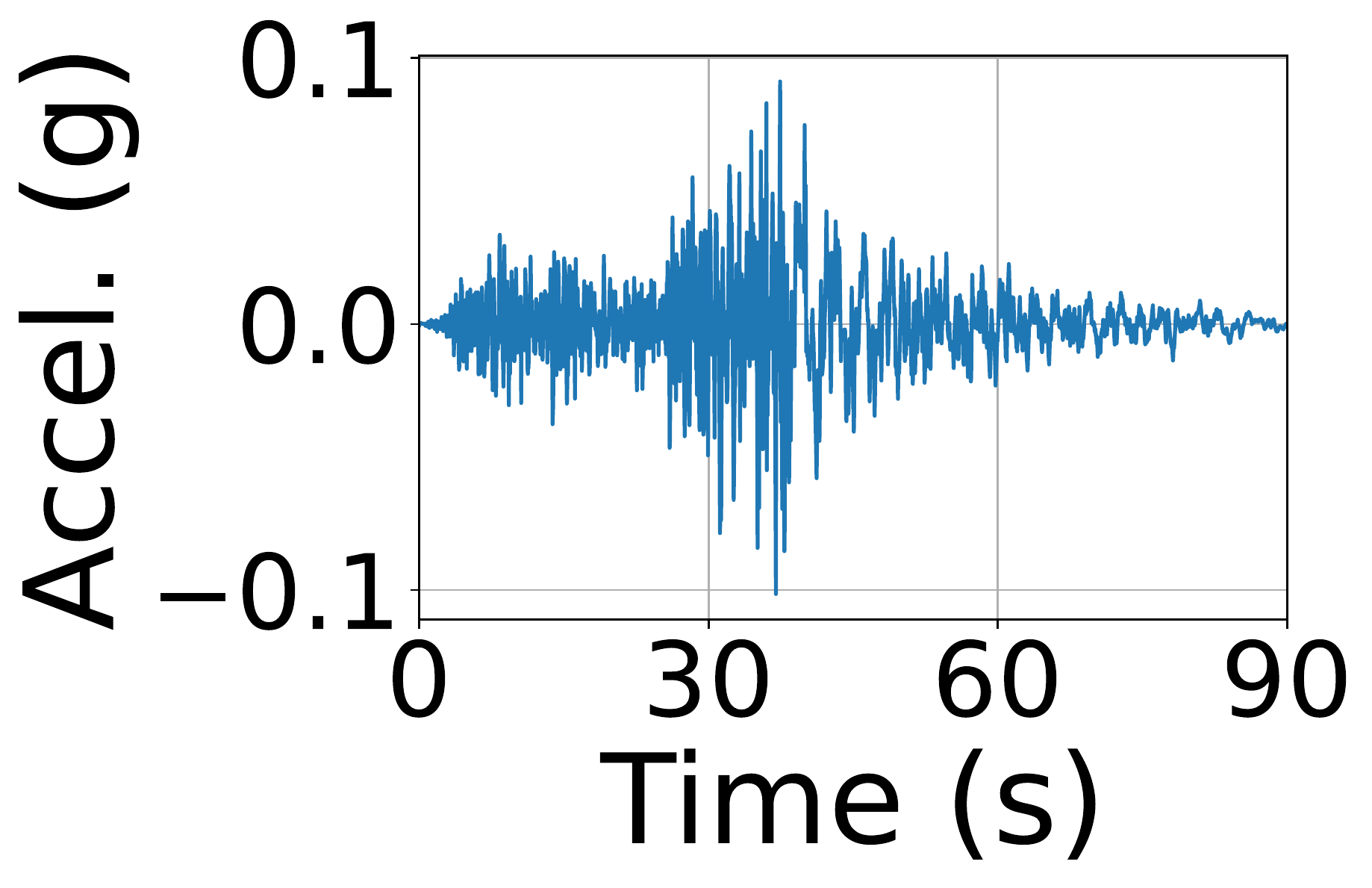} \\
\vspace*{0.35truecm}
\includegraphics[valign=c,width=0.19\textwidth]{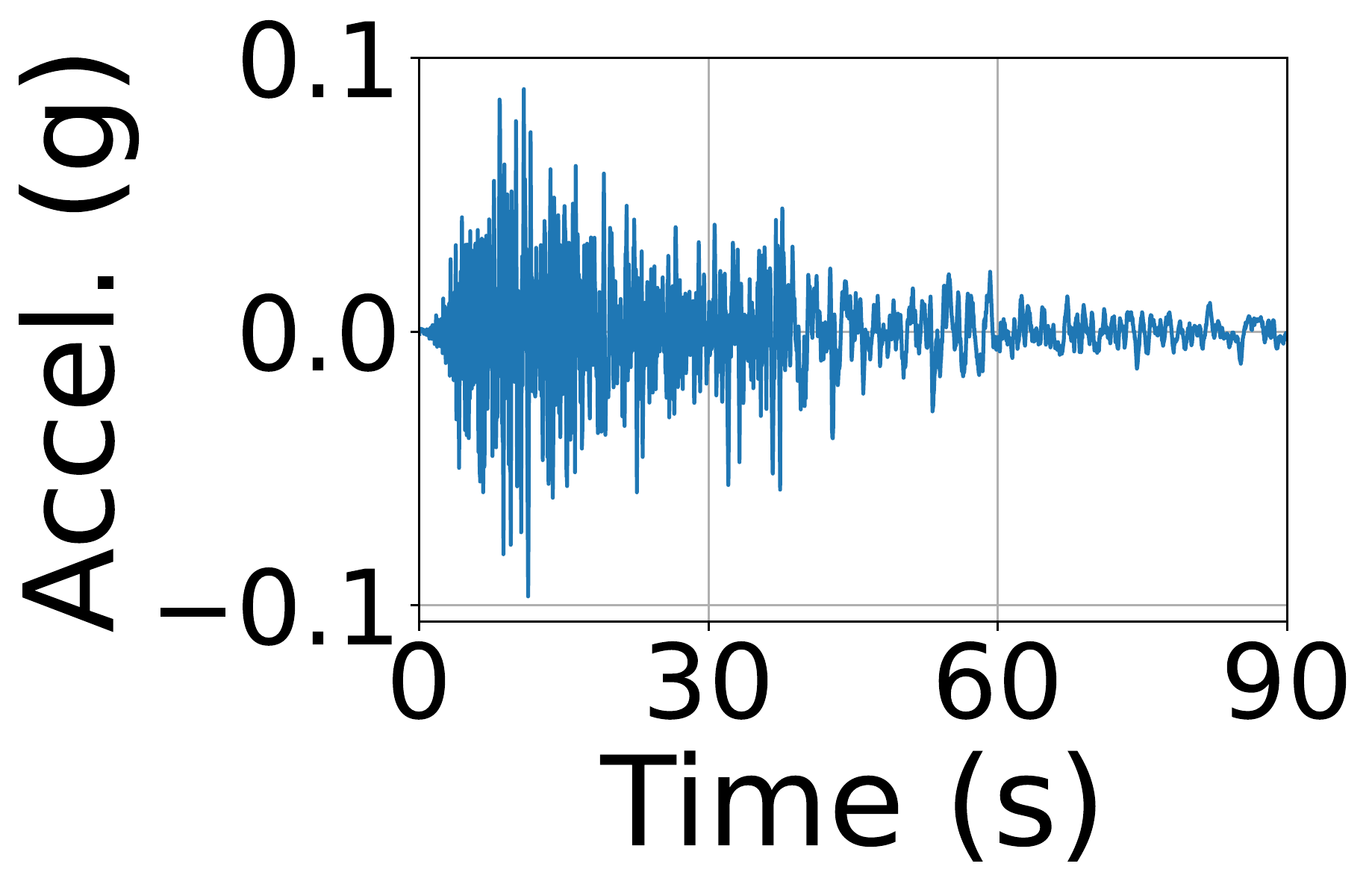} 
\includegraphics[valign=c,width=0.19\textwidth]{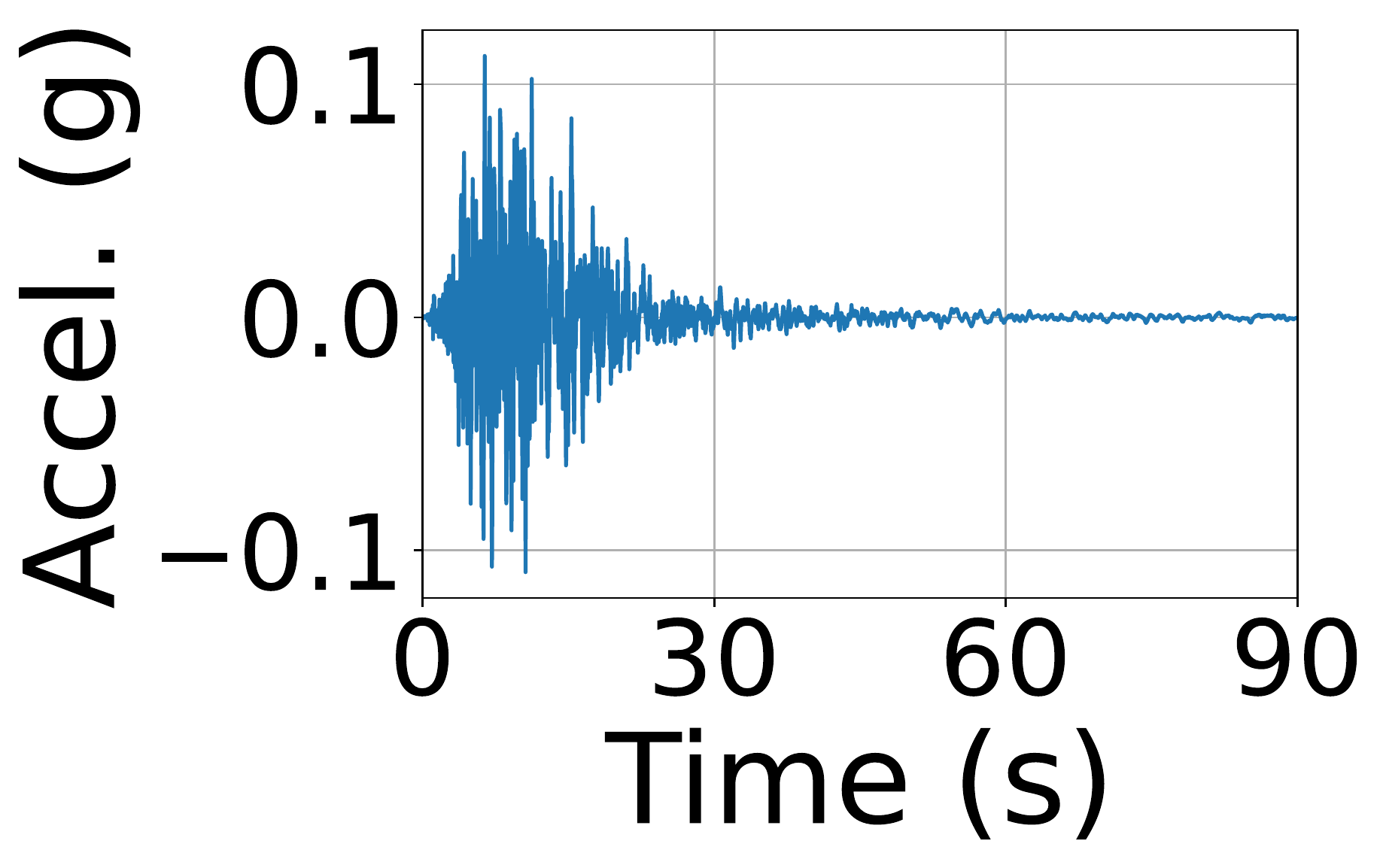} 
\includegraphics[valign=c,width=0.19\textwidth]{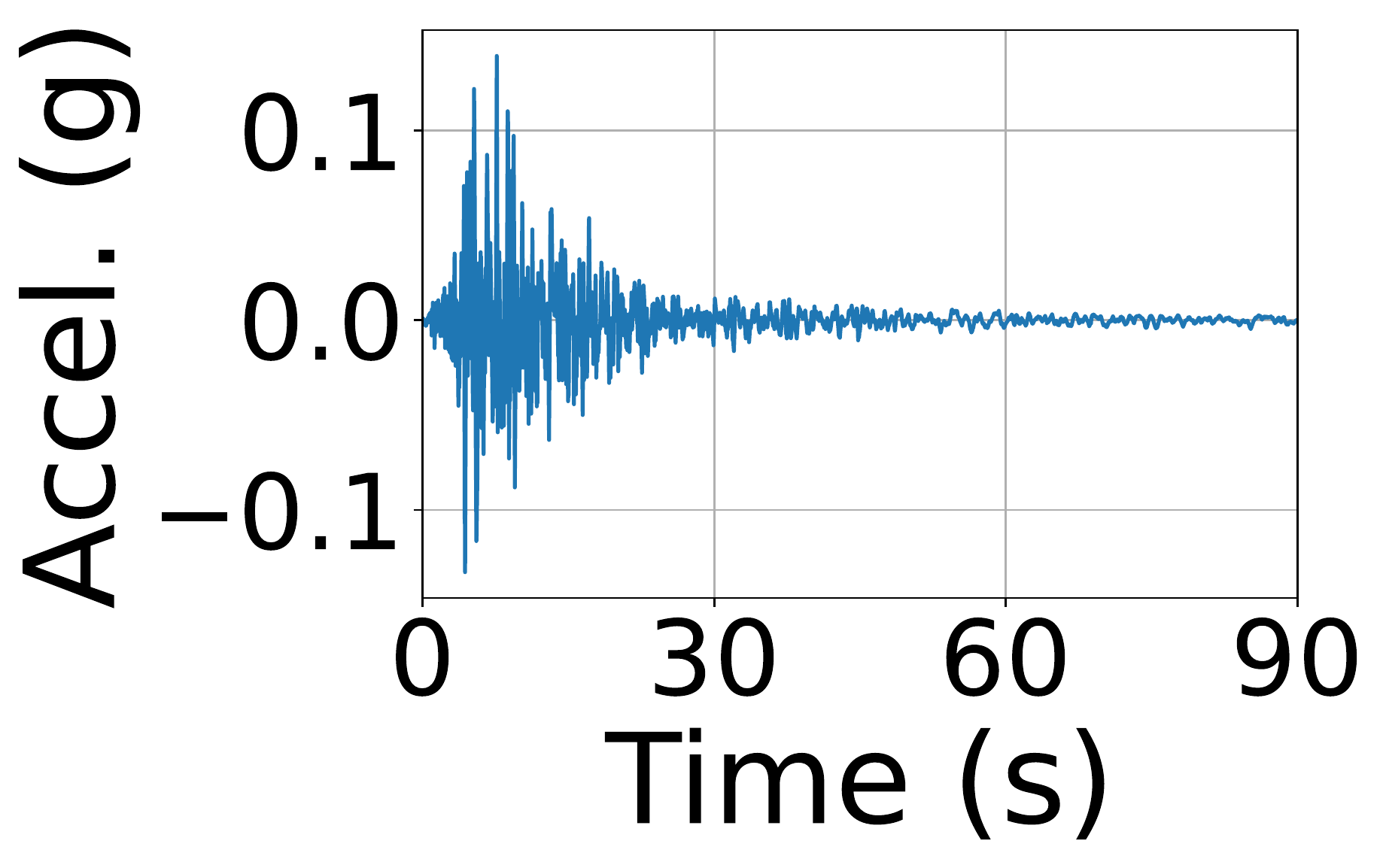}
\includegraphics[valign=c,width=0.19\textwidth]{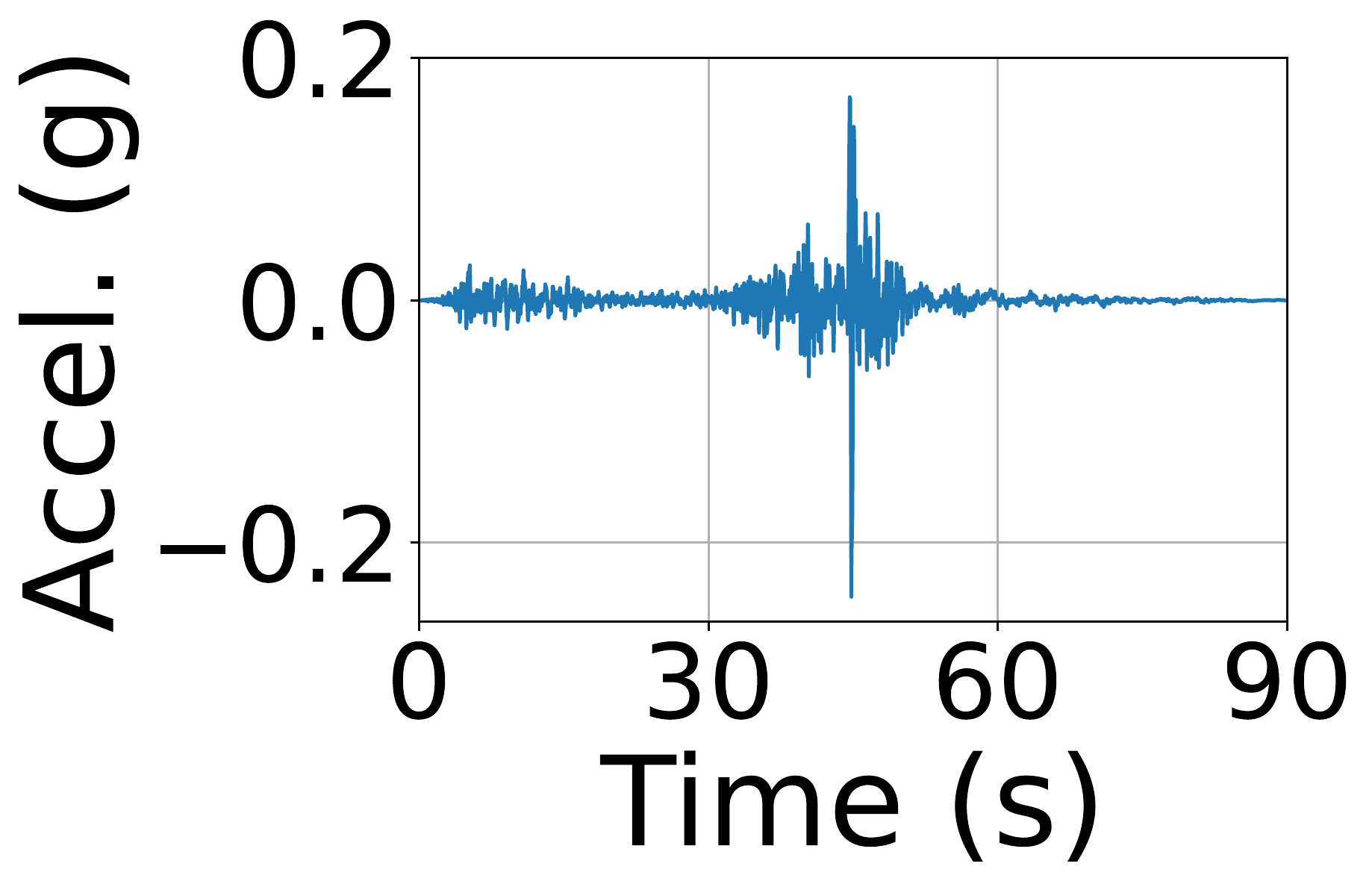}
\includegraphics[valign=c,width=0.19\textwidth]{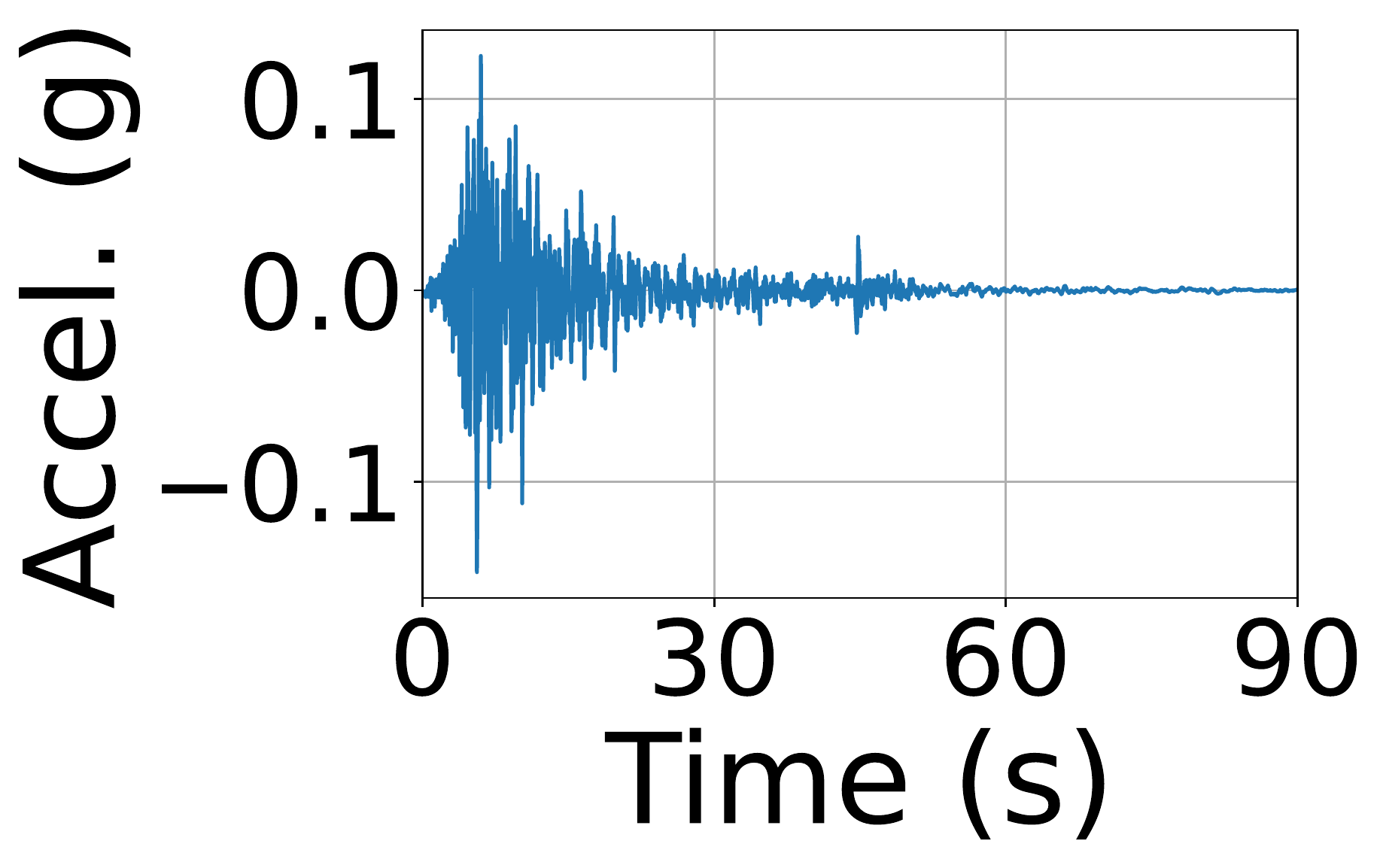} \\
\vspace*{0.35truecm}
\includegraphics[valign=c,width=0.19\textwidth]{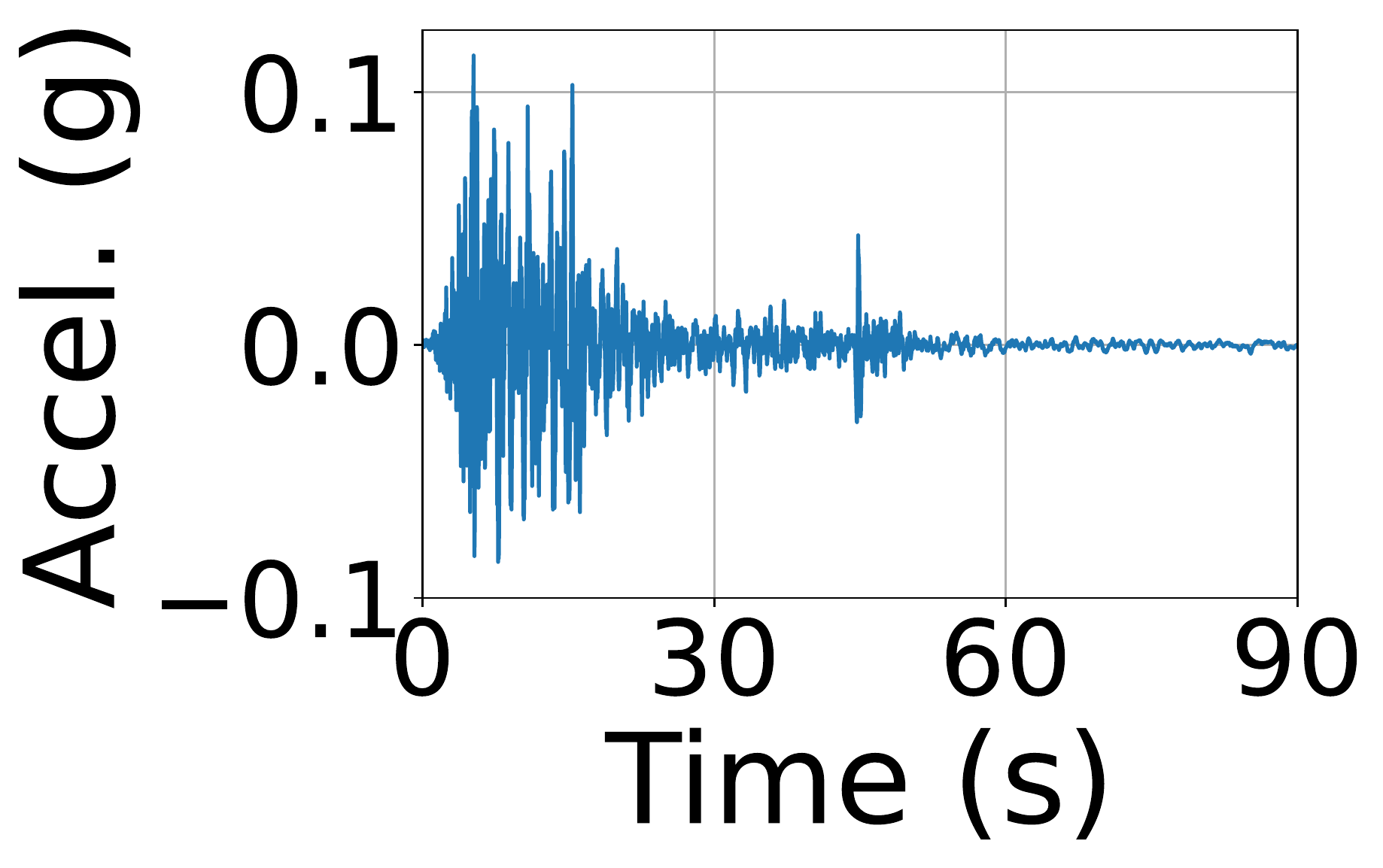} 
\includegraphics[valign=c,width=0.19\textwidth]{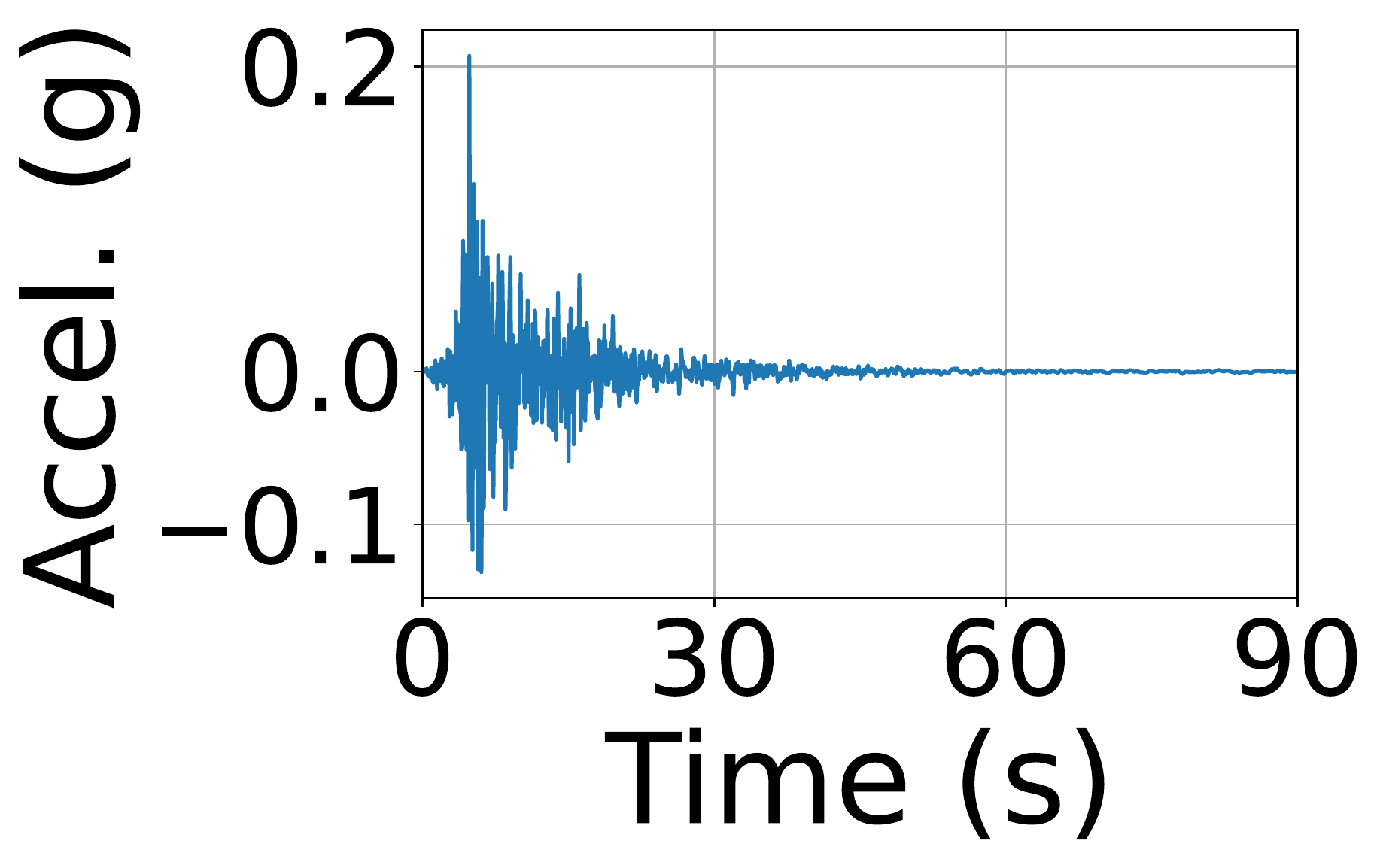} 
\includegraphics[valign=c,width=0.19\textwidth]{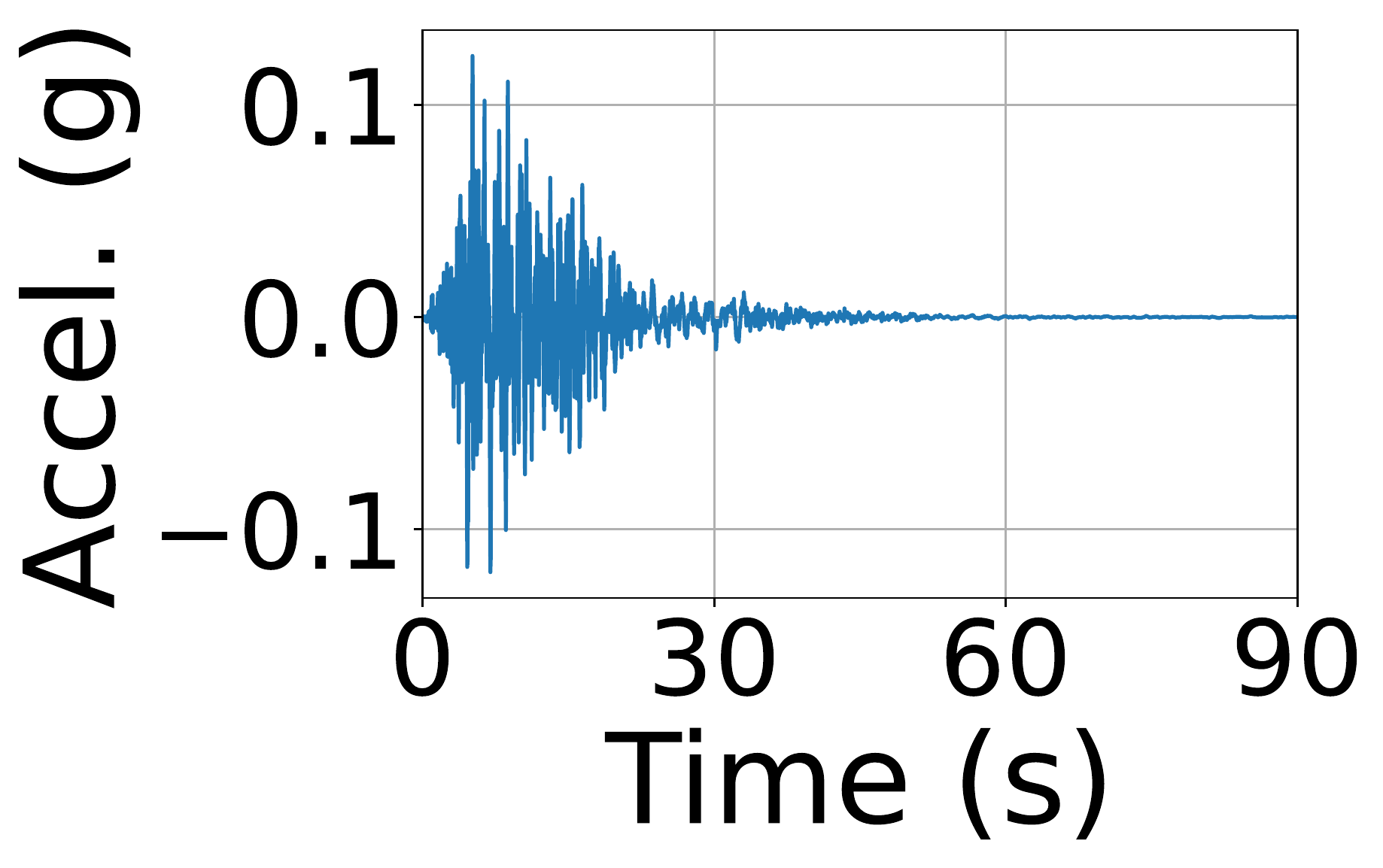}
\includegraphics[valign=c,width=0.19\textwidth]{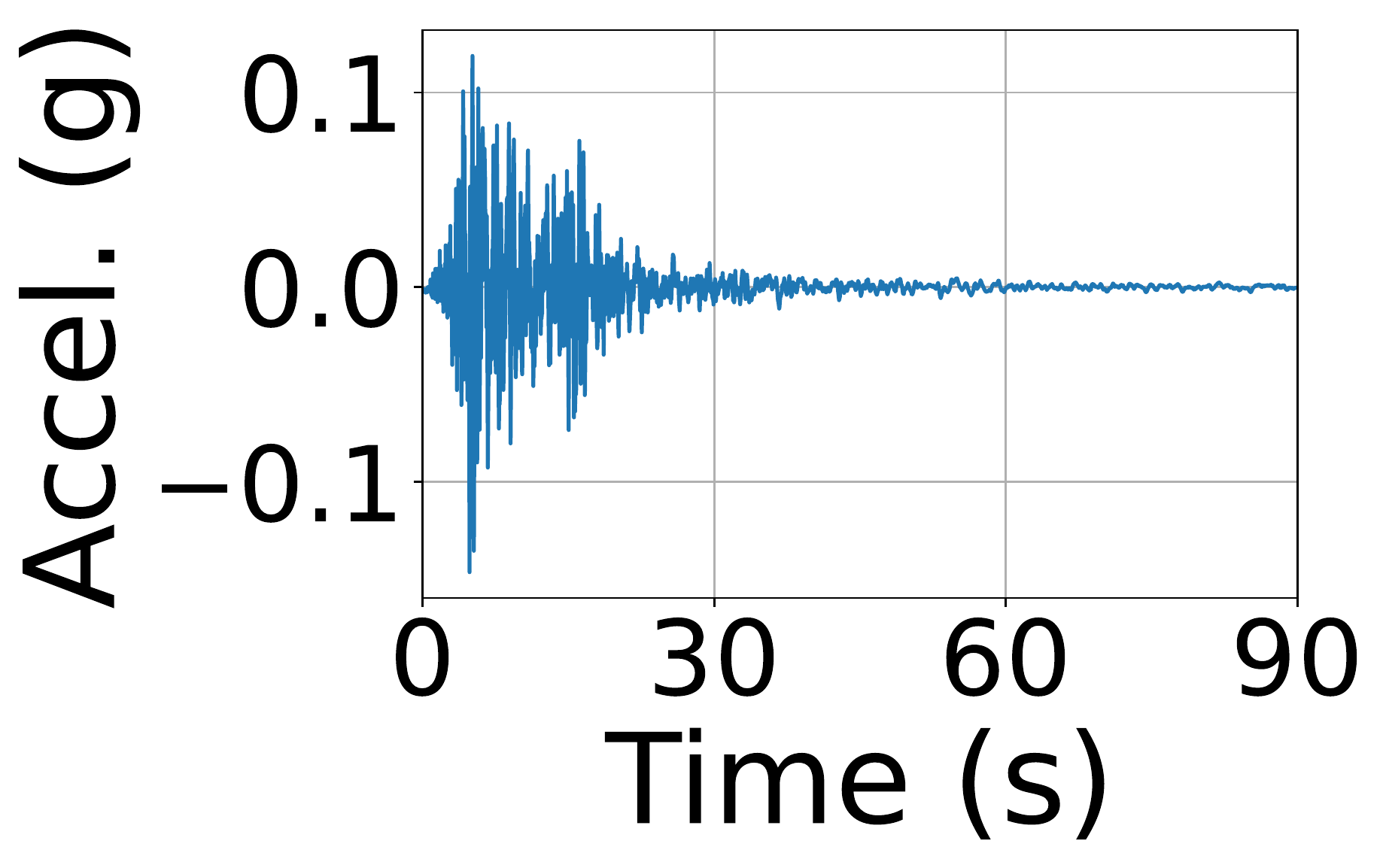}
\includegraphics[valign=c,width=0.19\textwidth]{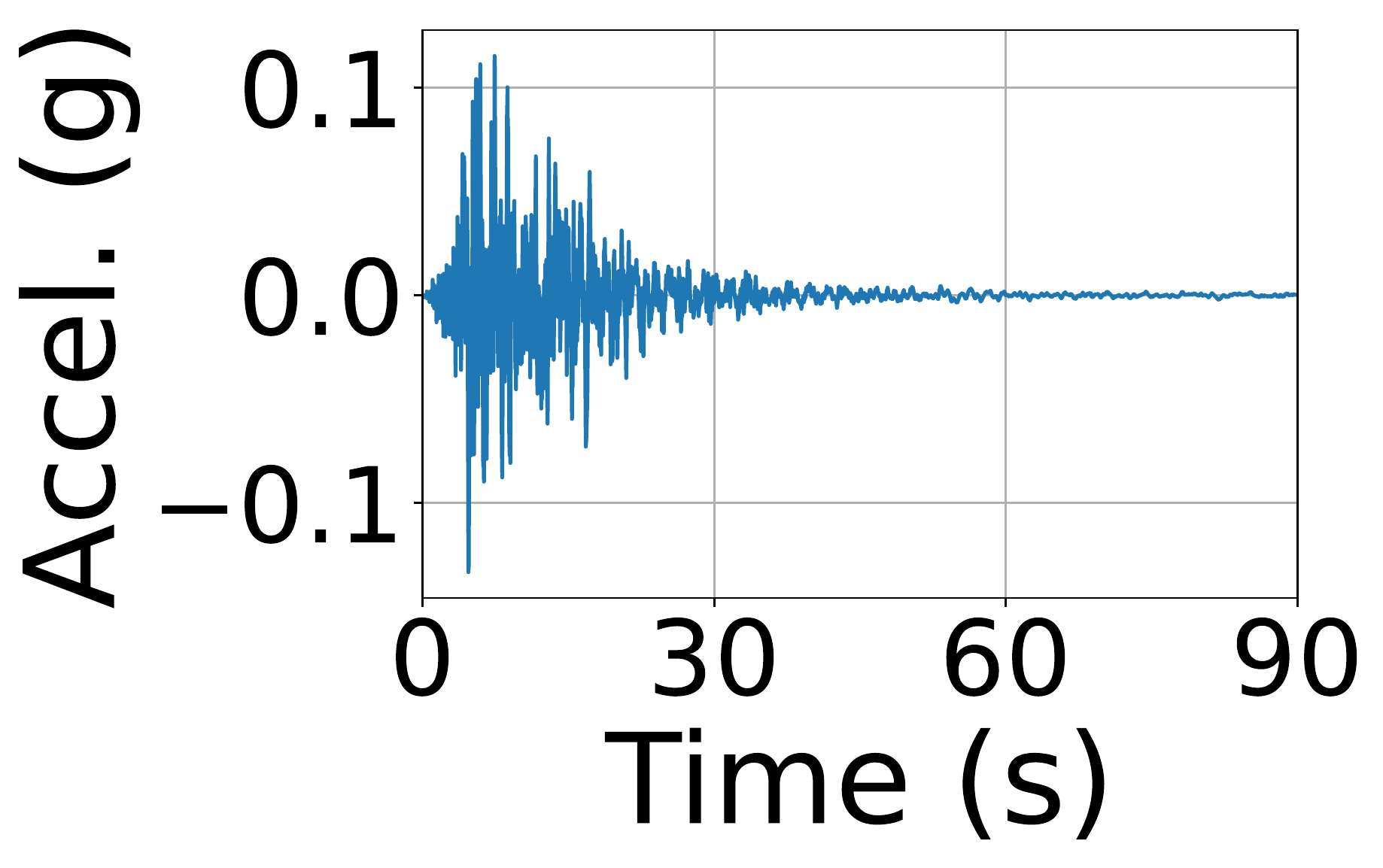} \\
\vspace*{0.35truecm}
\includegraphics[valign=c,width=0.19\textwidth]{figures/eq_basis_20} 
\includegraphics[valign=c,width=0.19\textwidth]{figures/eq_basis_21} 
\includegraphics[valign=c,width=0.19\textwidth]{figures/eq_basis_22}
\includegraphics[valign=c,width=0.19\textwidth]{figures/eq_basis_23}
\includegraphics[valign=c,width=0.19\textwidth]{figures/eq_basis_24} \\
\vspace*{0.35truecm}
\includegraphics[valign=c,width=0.19\textwidth]{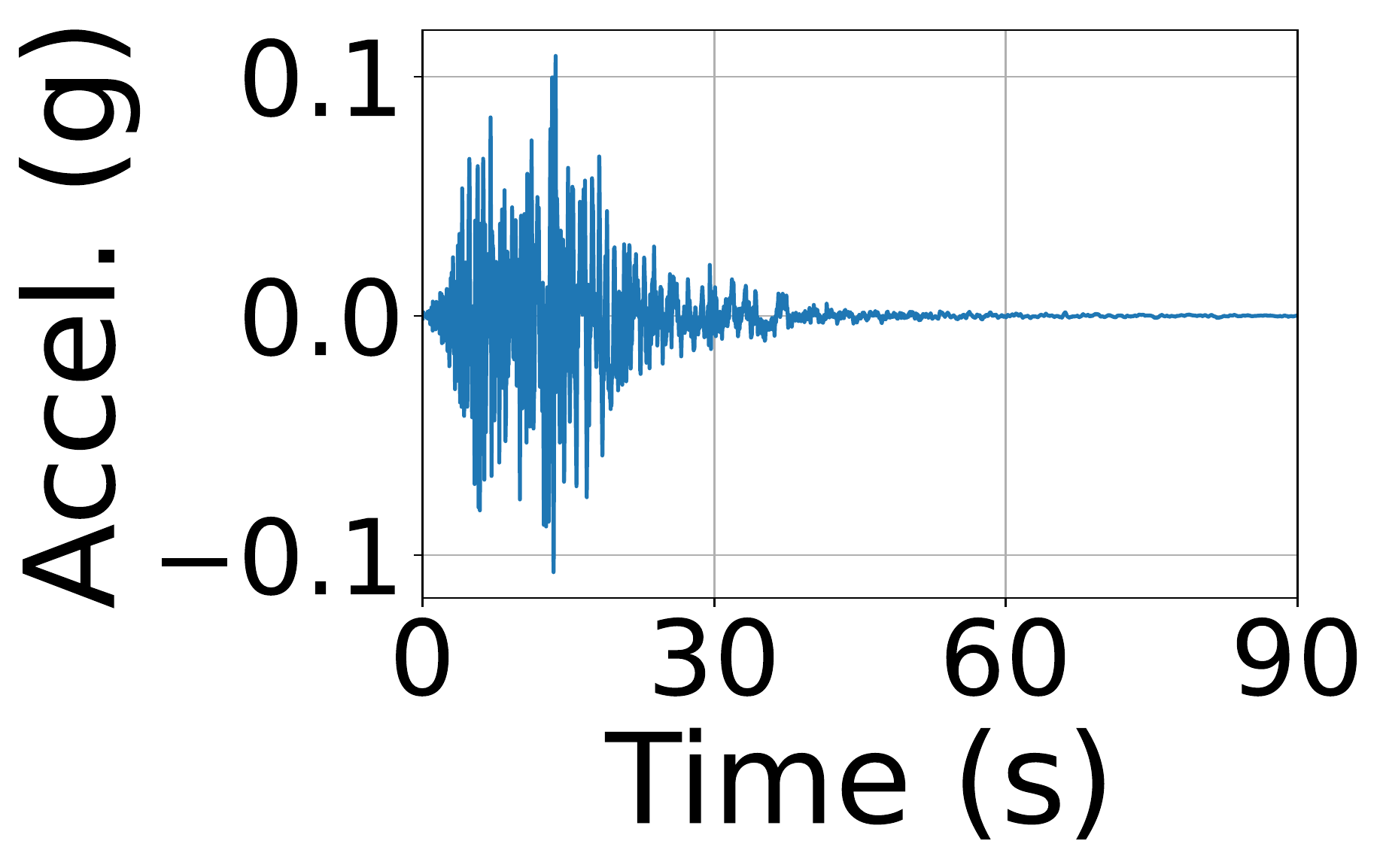} 
\includegraphics[valign=c,width=0.19\textwidth]{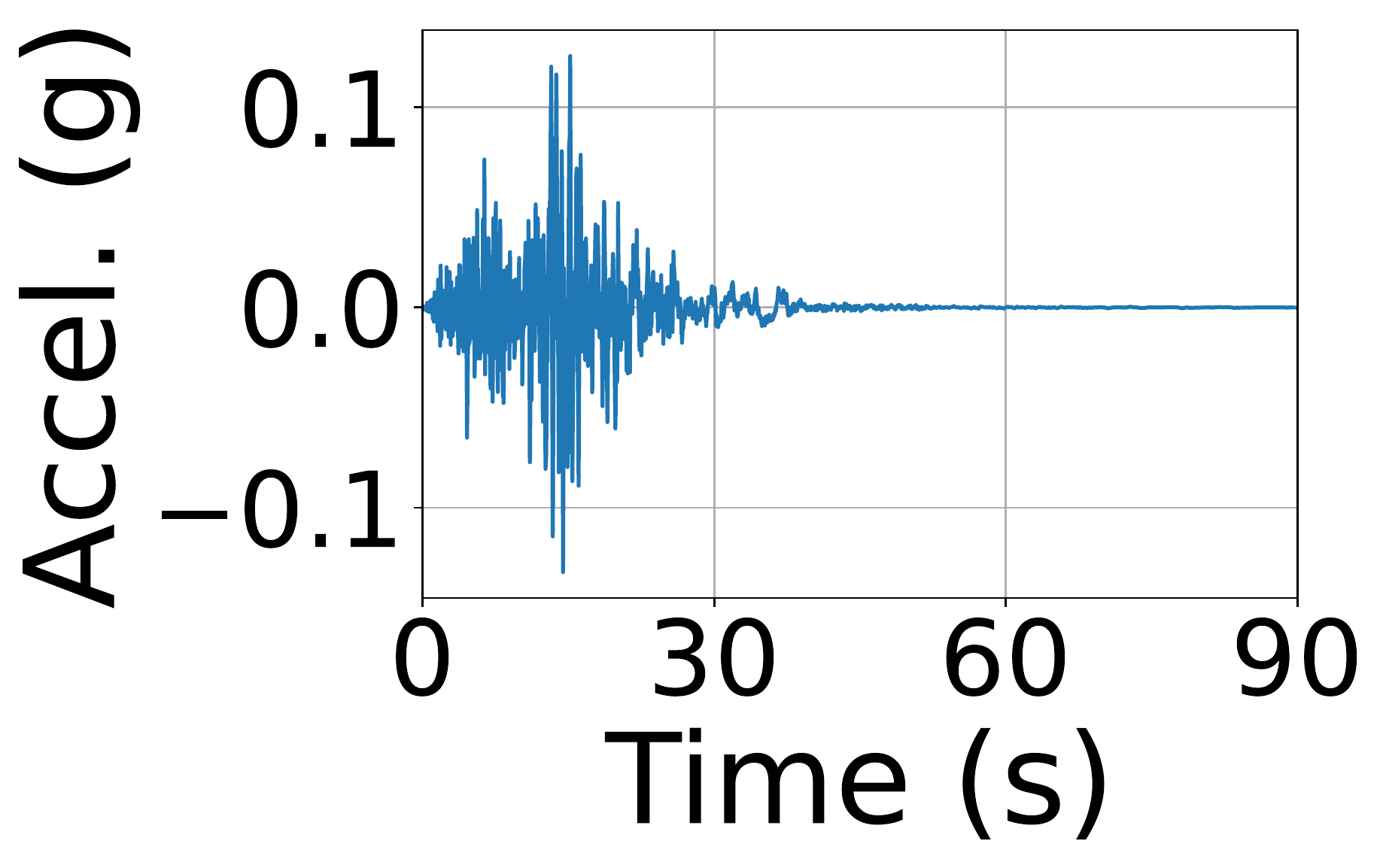} 
\includegraphics[valign=c,width=0.19\textwidth]{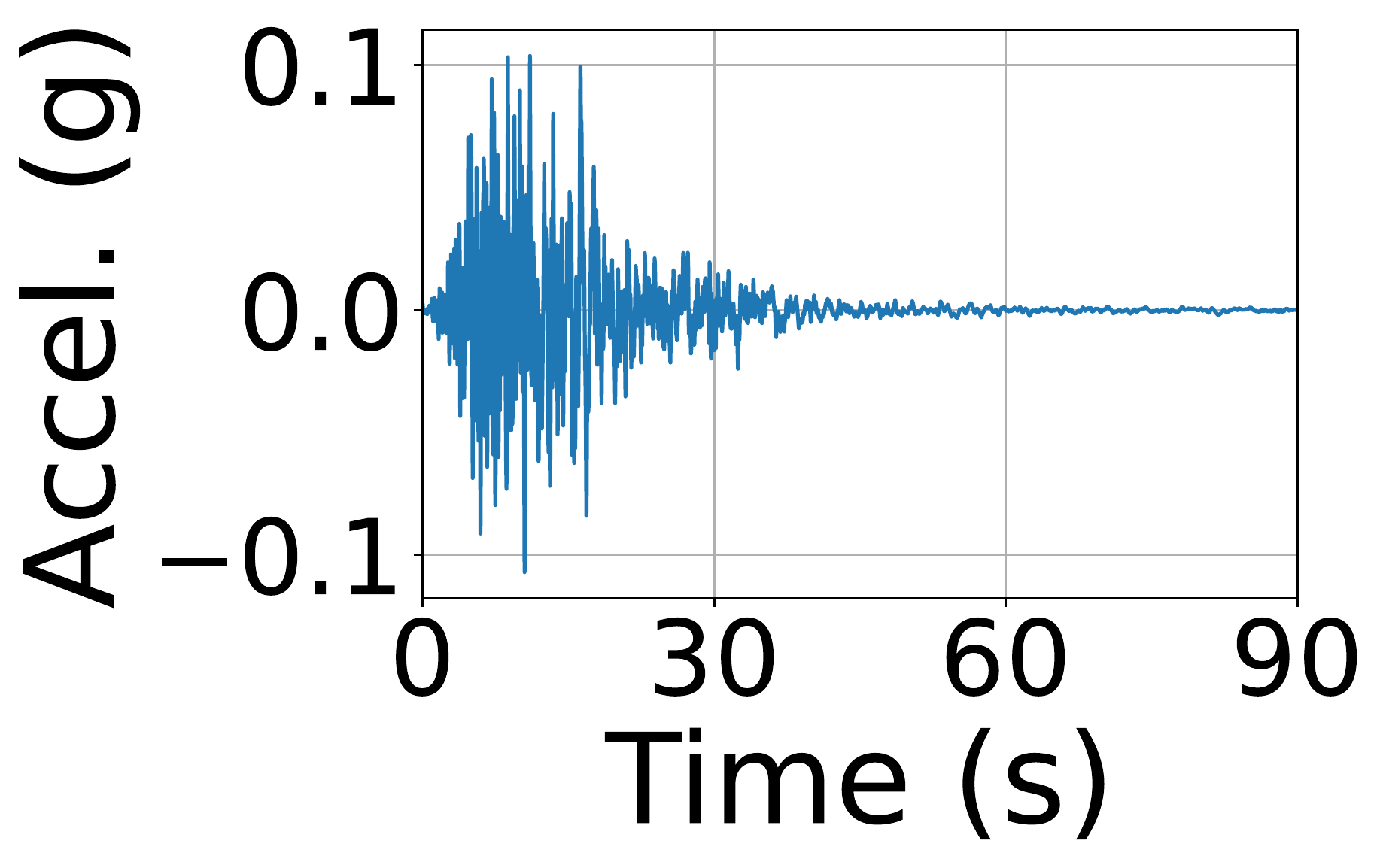}
\includegraphics[valign=c,width=0.19\textwidth]{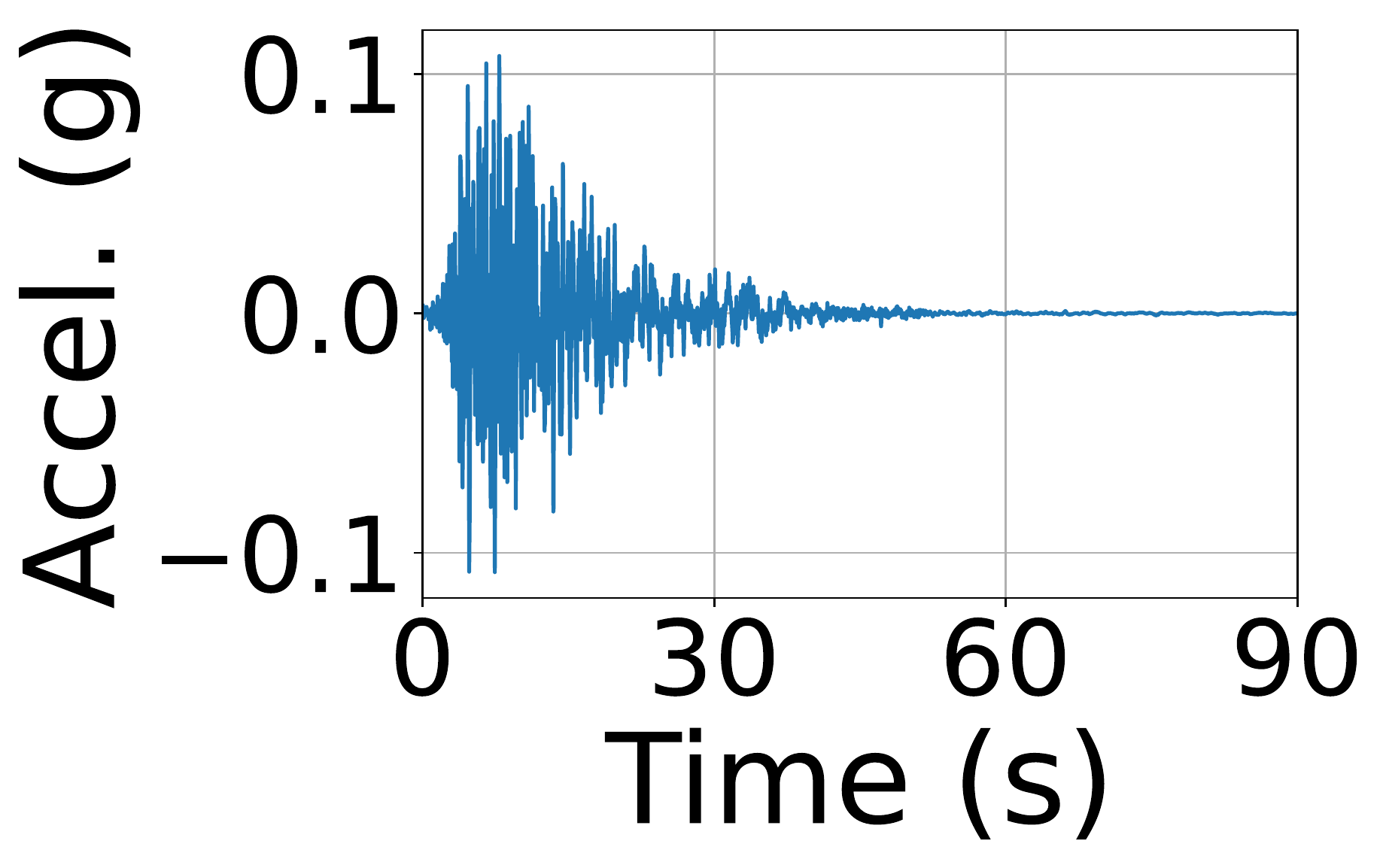}
\includegraphics[valign=c,width=0.19\textwidth]{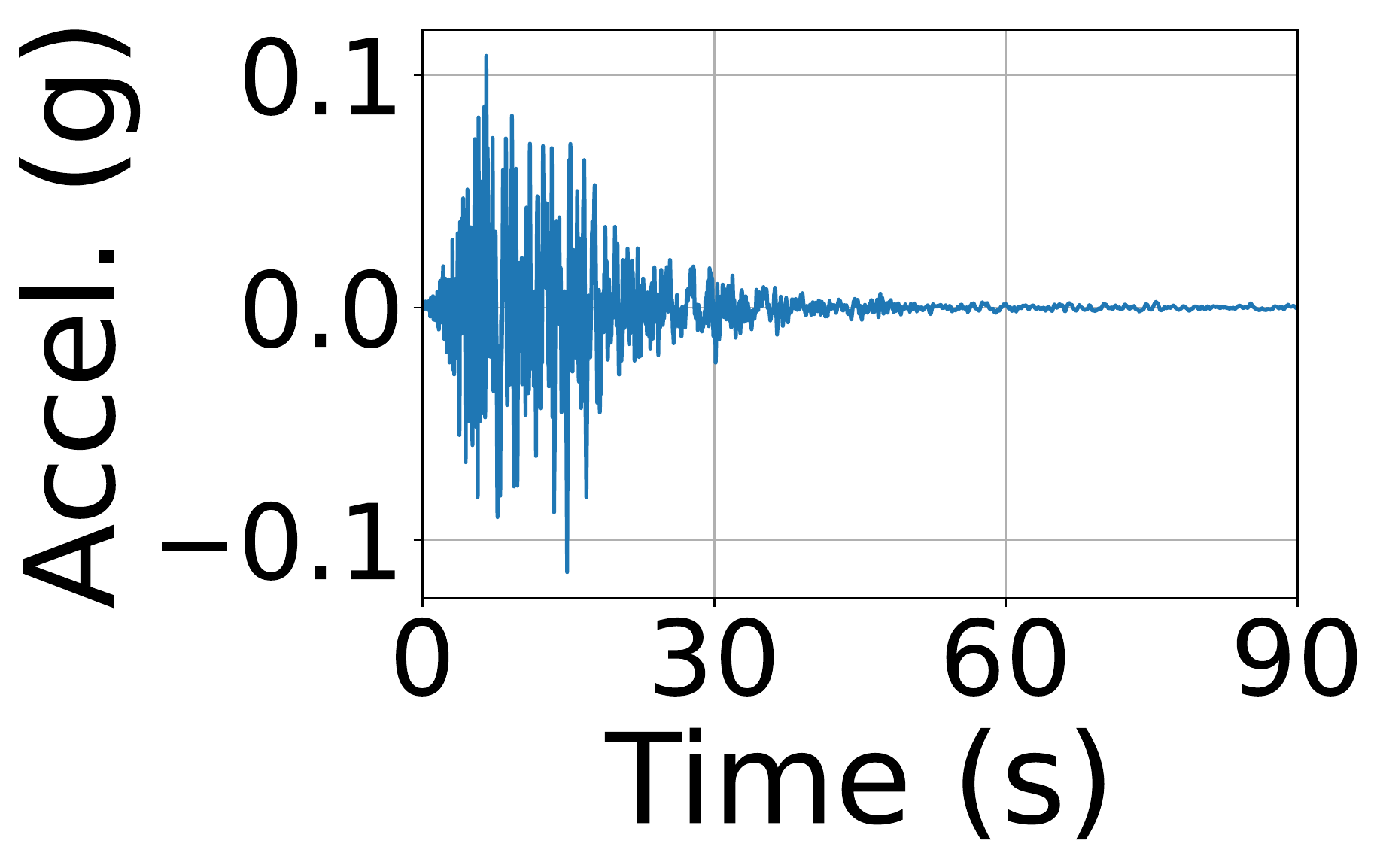} \\
\vspace*{0.35truecm}
\includegraphics[valign=c,width=0.19\textwidth]{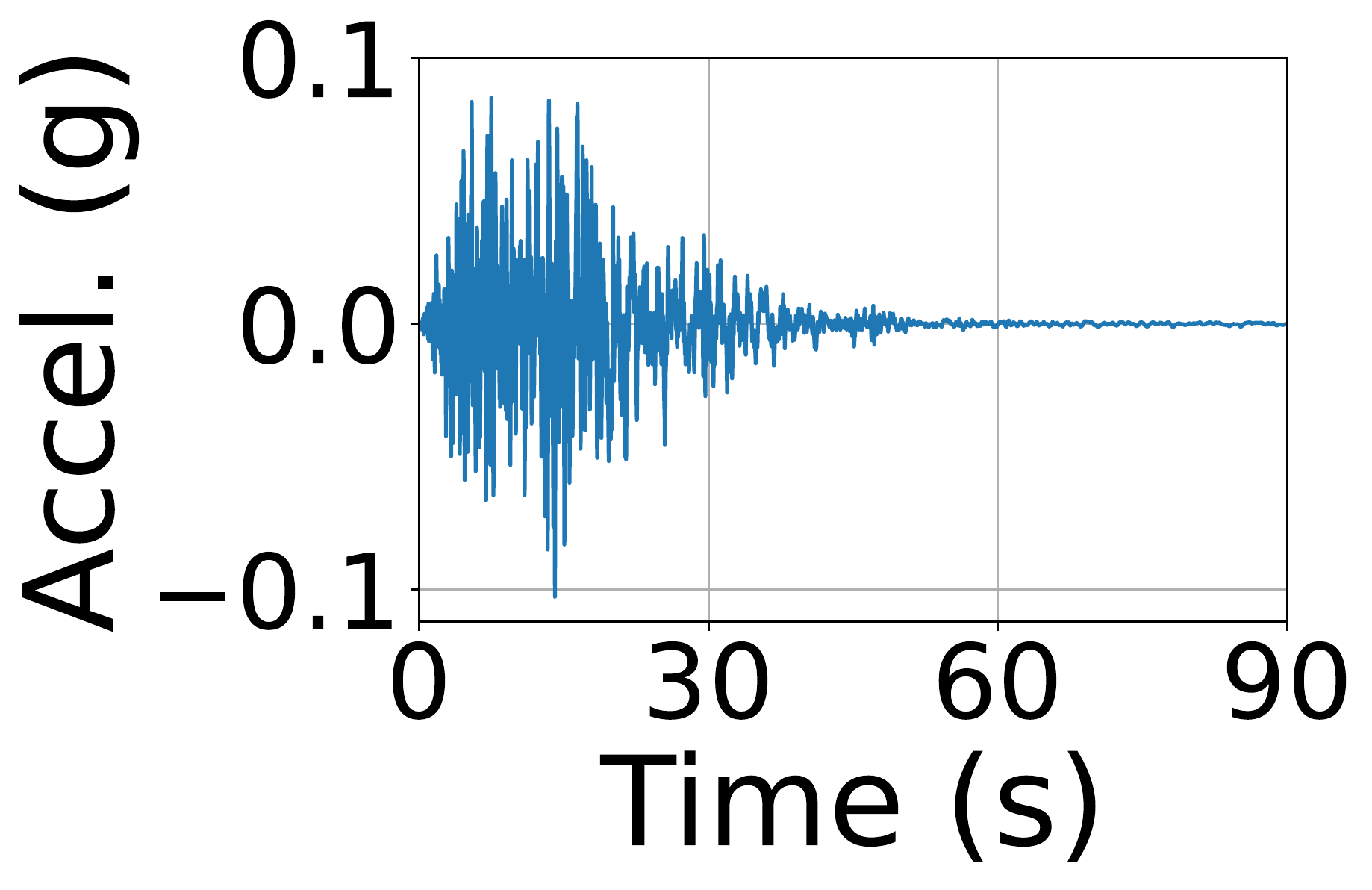} 
\includegraphics[valign=c,width=0.19\textwidth]{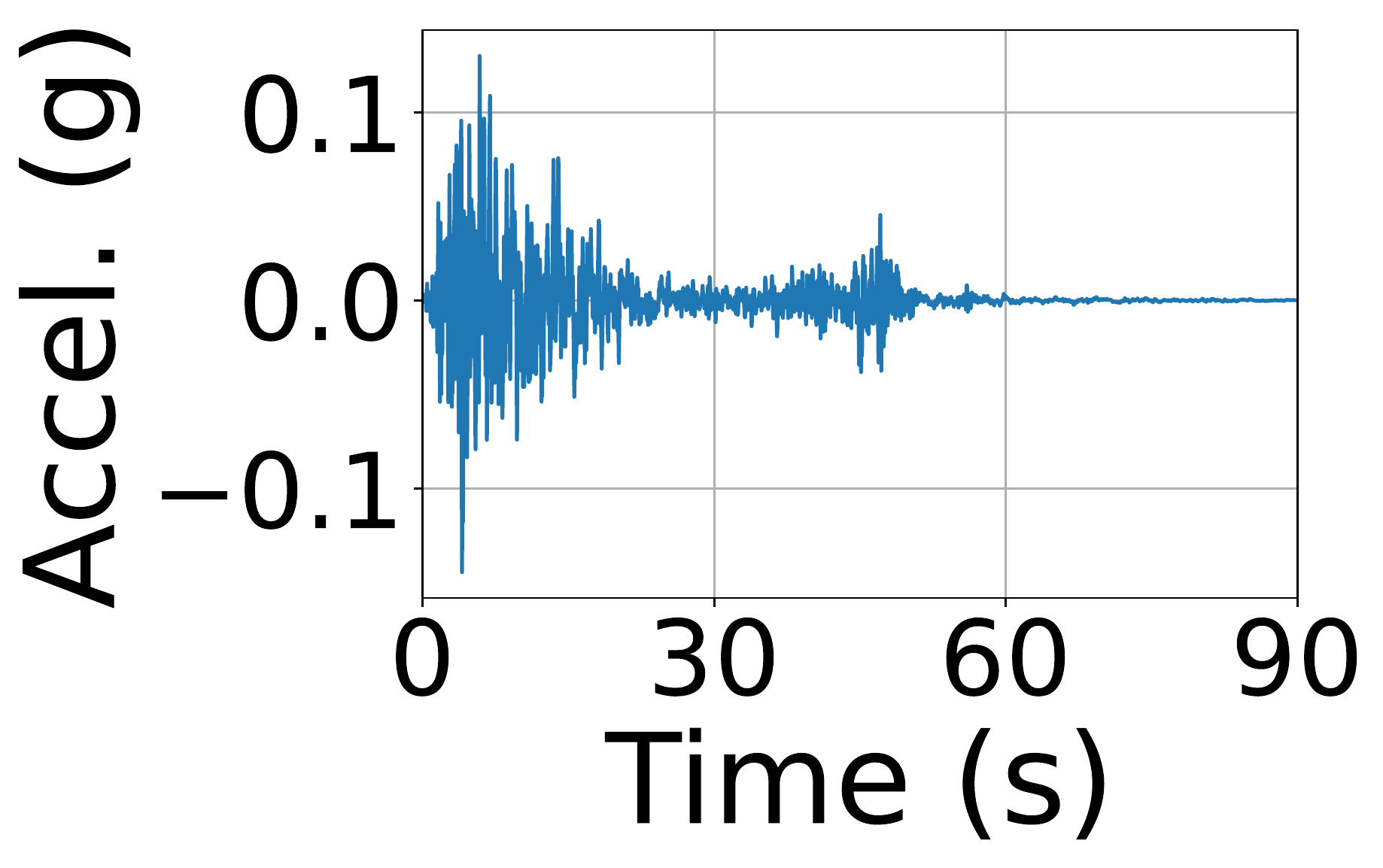} 
\includegraphics[valign=c,width=0.19\textwidth]{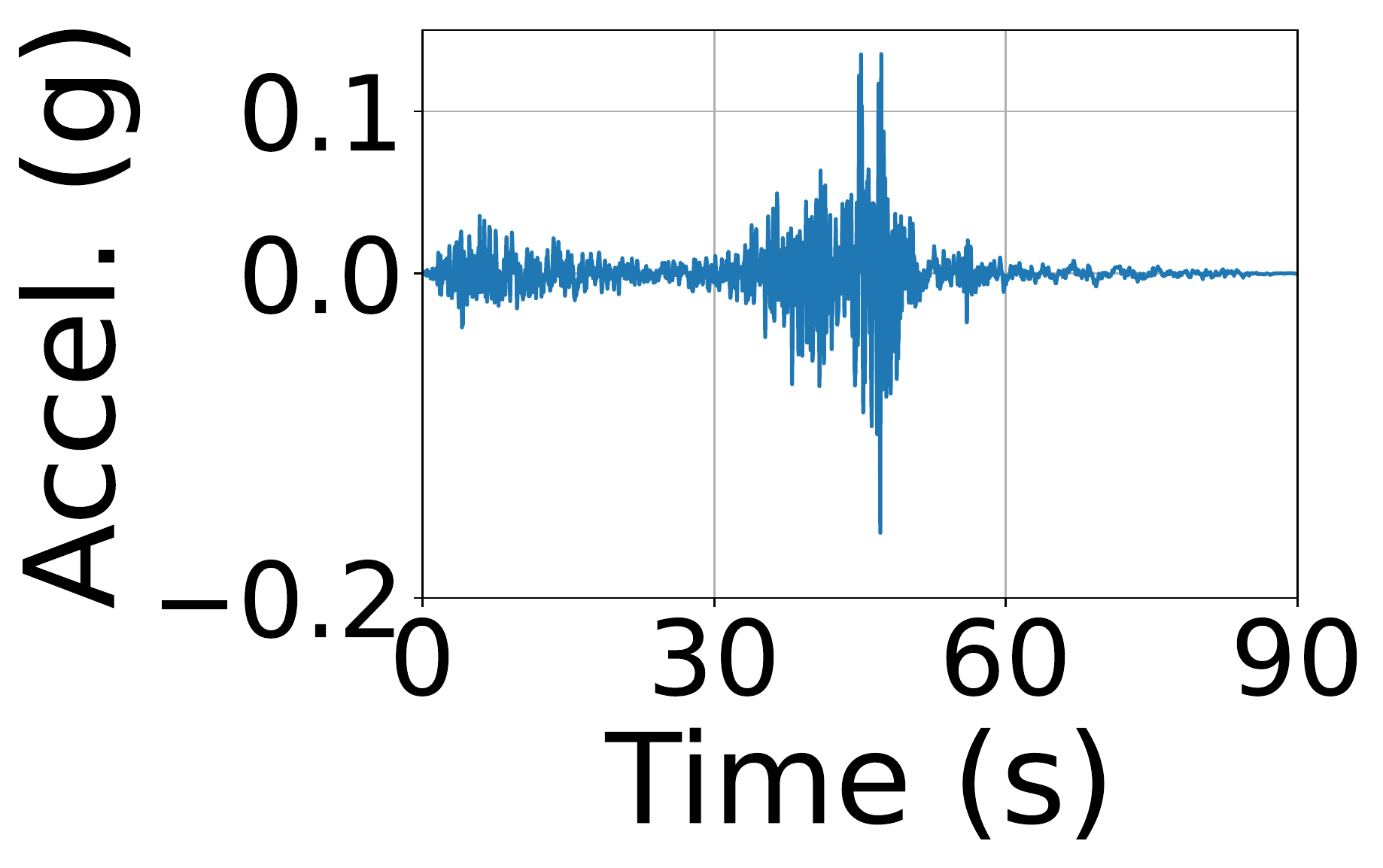}
\includegraphics[valign=c,width=0.19\textwidth]{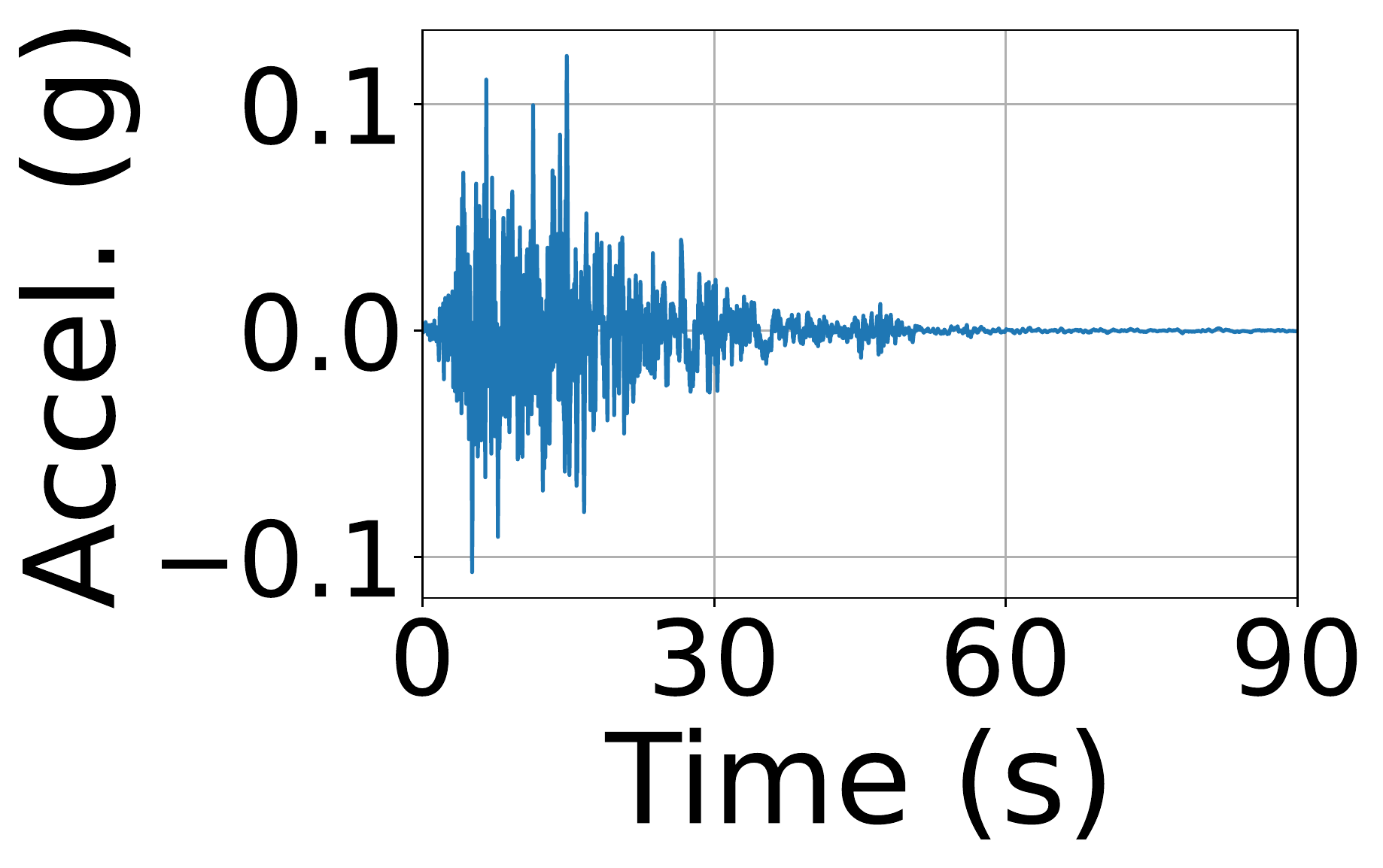}
\includegraphics[valign=c,width=0.19\textwidth]{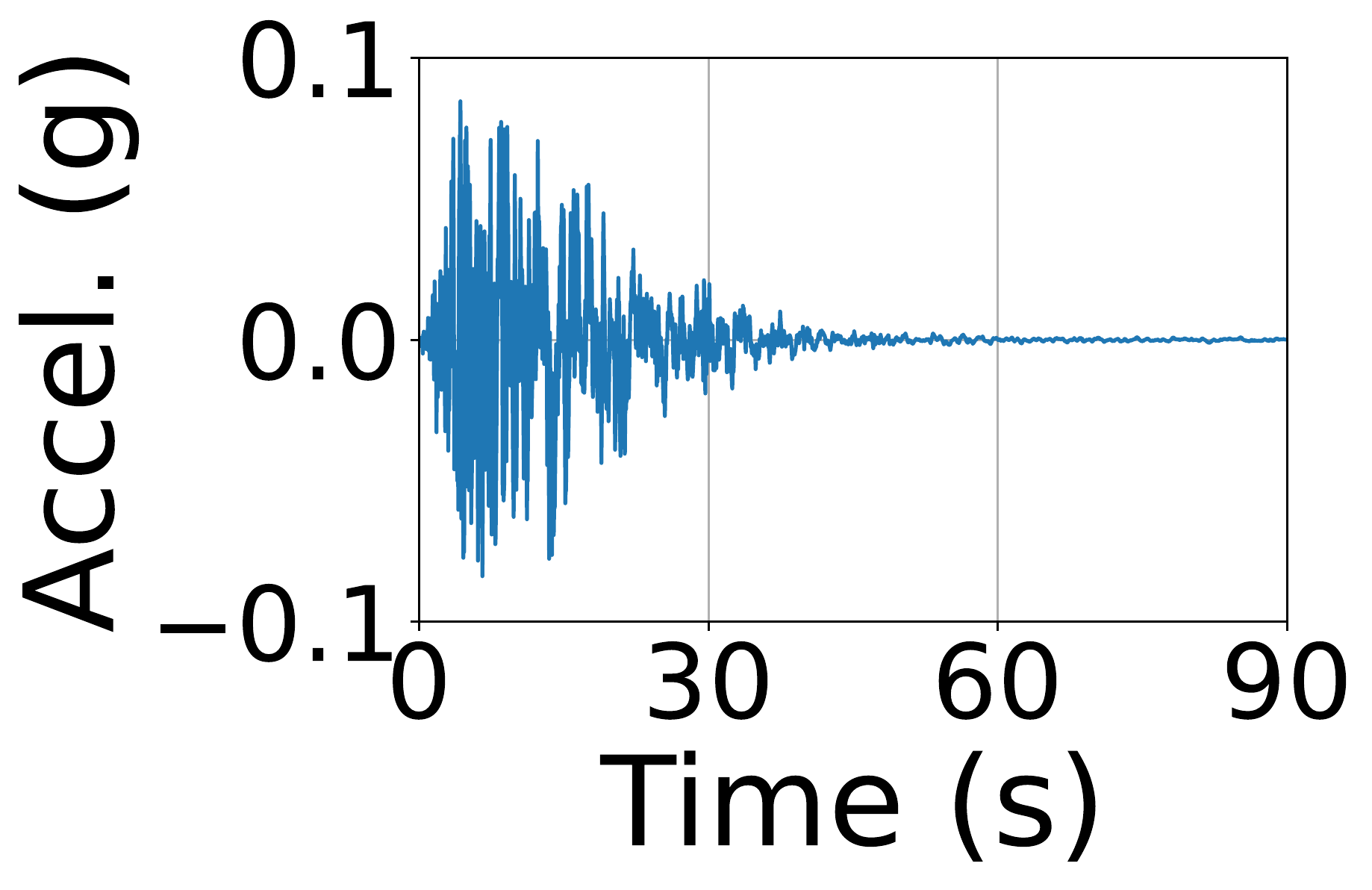} \\
\vspace*{0.35truecm}
\includegraphics[valign=c,width=0.19\textwidth]{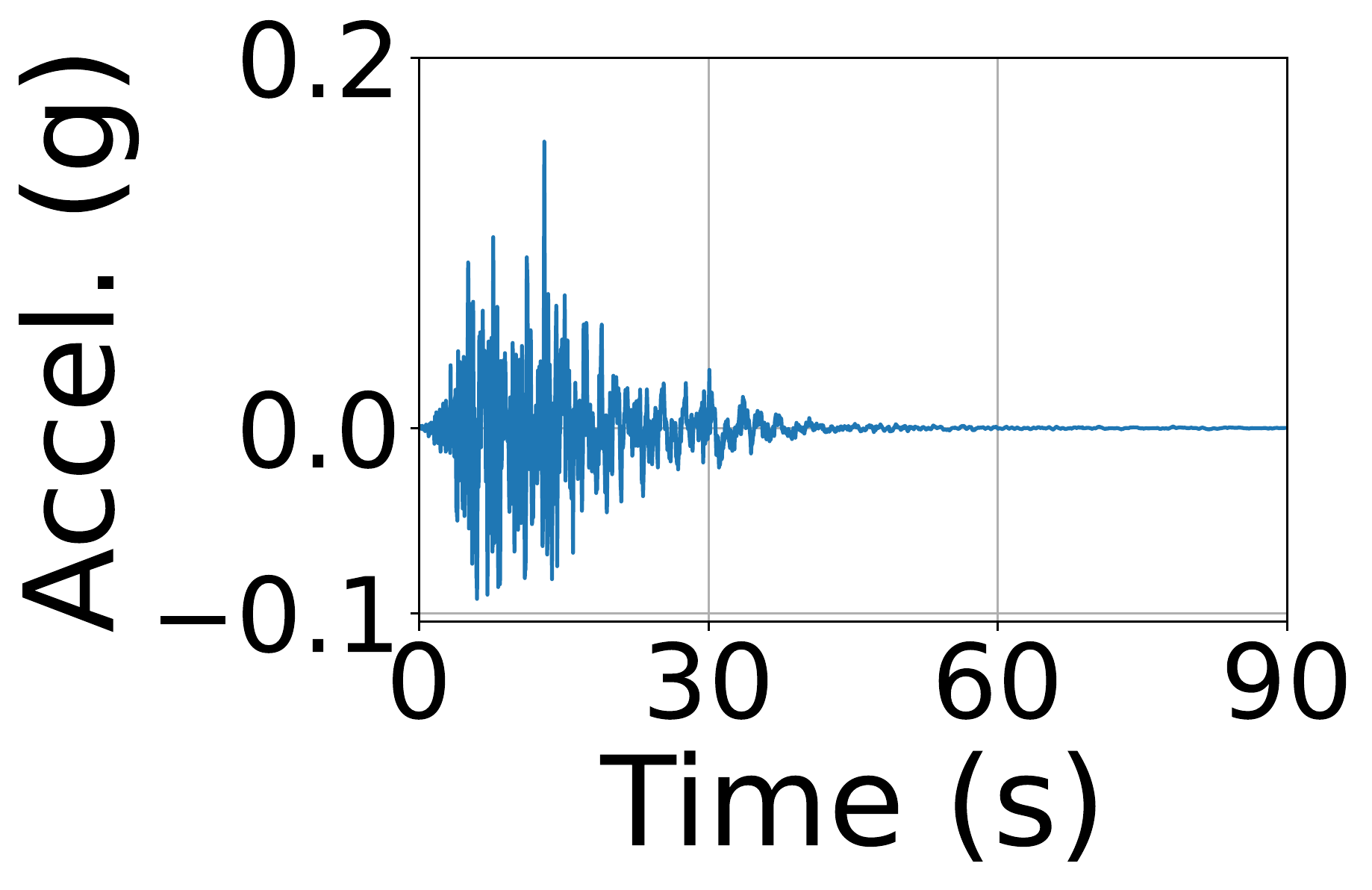} 
\includegraphics[valign=c,width=0.19\textwidth]{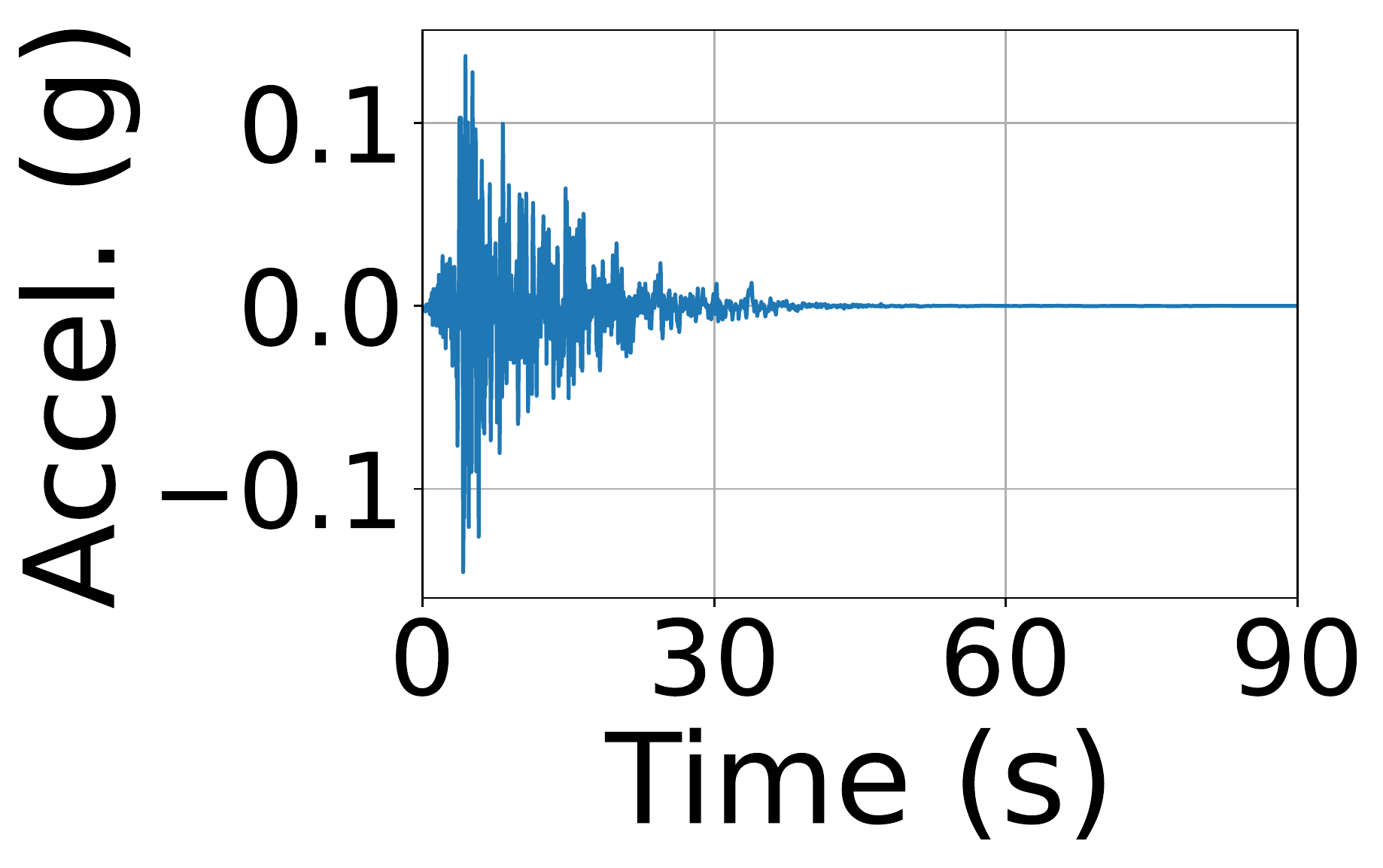} 
\includegraphics[valign=c,width=0.19\textwidth]{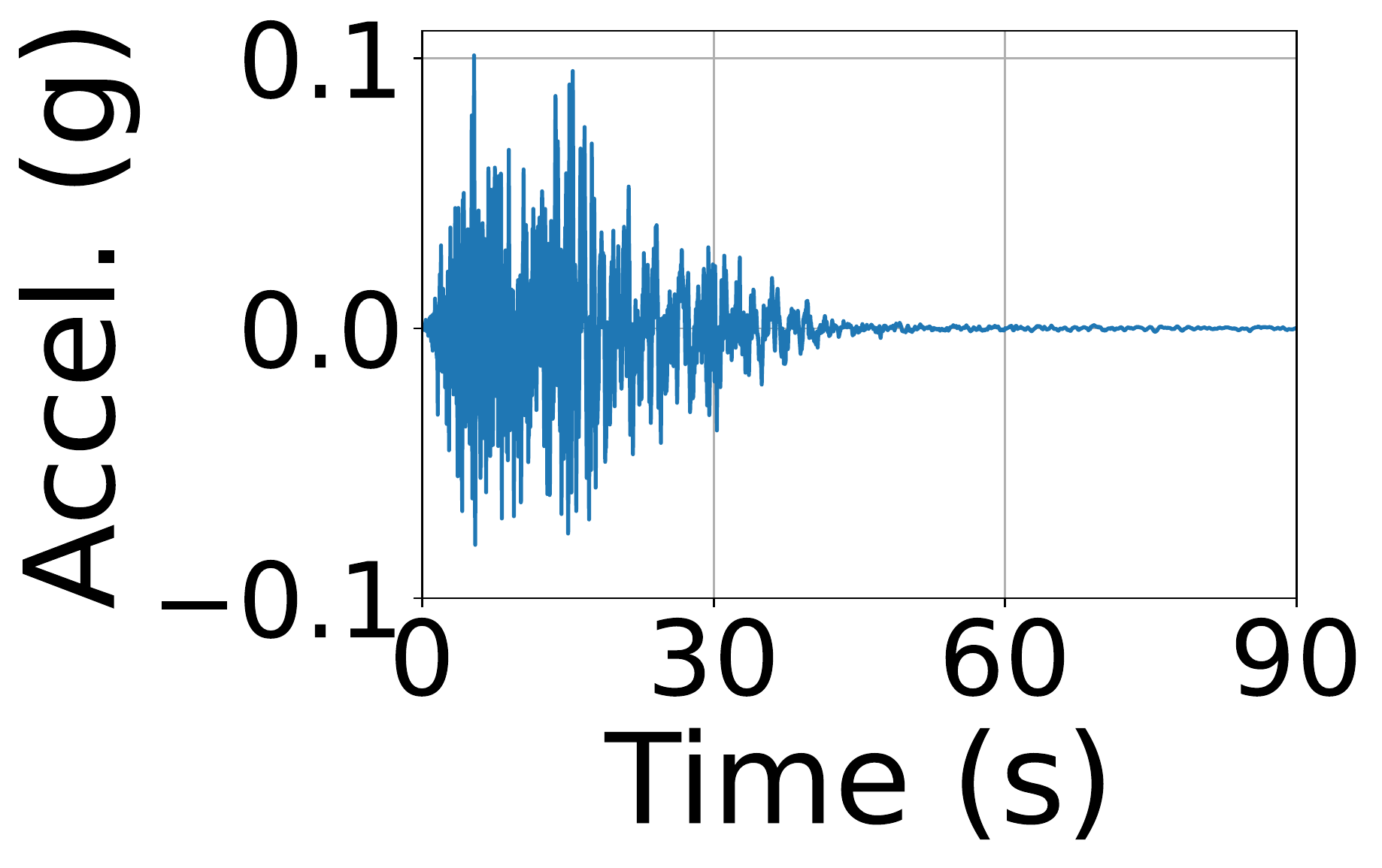}
\includegraphics[valign=c,width=0.19\textwidth]{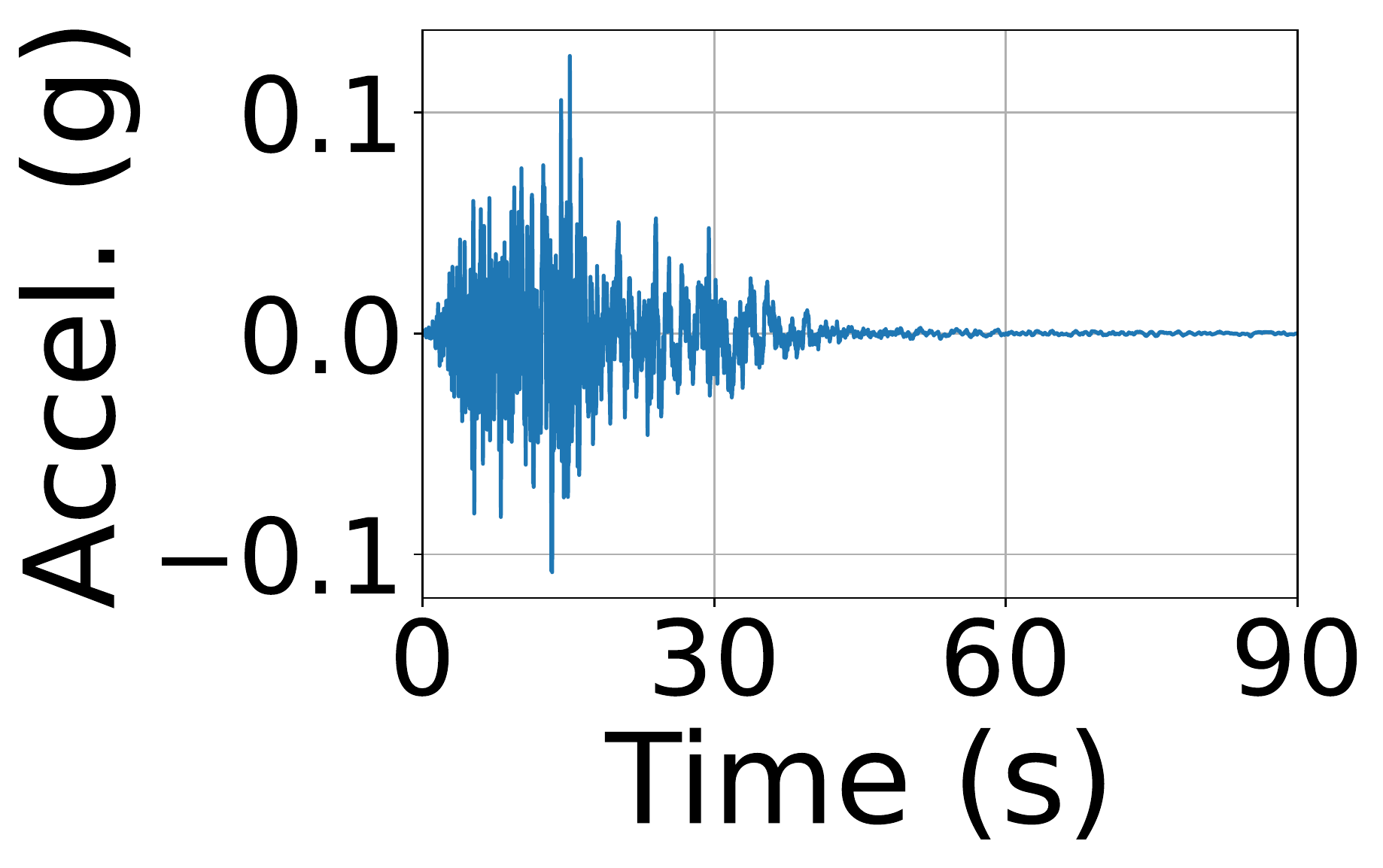}
\includegraphics[valign=c,width=0.19\textwidth]{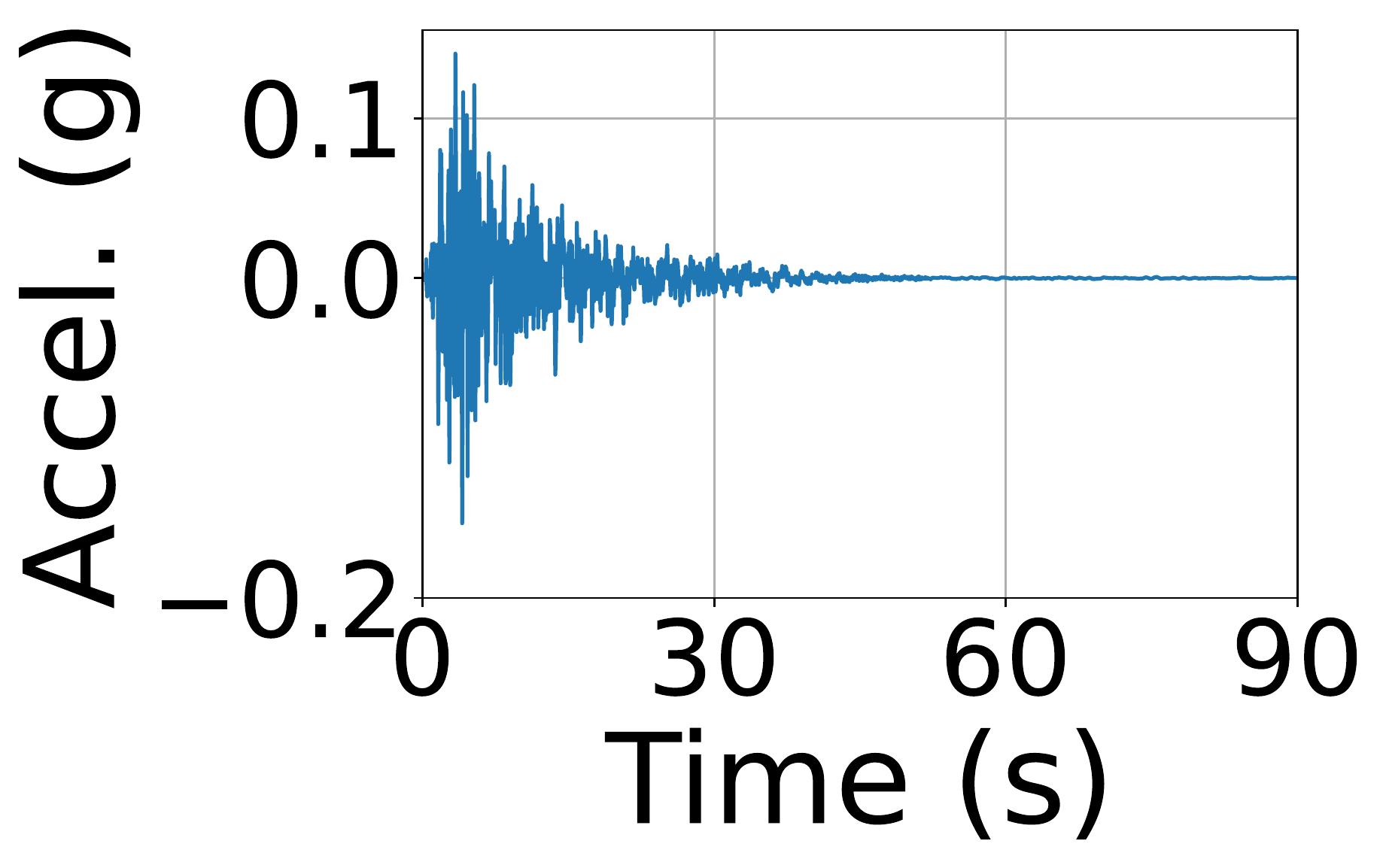} \\

\caption{First forty basis vectors for FEMA P695 earthquake suite which are responsible for 99\% of variability in the suite.Each plot represents a column of the $U_{4500\times40}$ matrix.}
\label{eq_svd}
\end{center}
\end{figure}

Once the ground motion suite is selected,  it needs to be processed for performing SVD according to Equation \ref{PCA2}.  Since different ground motions have different sampling frequencies and record lengths, they were first interpolated to a common sampling frequency and record length. The common sampling frequency was chosen to be the 50Hz or a time step of 0.02s,  which corresponds to the smallest sampling frequency among all the records. The common time history length was chosen to be 90s, the longest record length in the suite.  For motions with duration less than 90s,  zero padding is used. This leads to $A_{4500 \times 44}$ in Equation \ref{PCA2} with $n$, number of time steps being equal to 4500 and $m$, number of earthquakes being equal to 44.  SVD was performed on matrix $A$, which resulted in a 44 columns,  orthonormal basis matrix $U_{4500\times44}$.  First, the basis vectors are arranged in descending order of their singular values,  and then the cumulative explained variance is plotted against the number of basis vectors in Figure \ref{exp_variance_cum}.  It can be observed that the first 40 basis vectors are responsible for explaining 99\% of variation in the ground motion data.  This result is reiterated in Figure \ref{exp_variance}, where the percentage of explained variance are plotted against the basis vector number.  Compared to the first 40 vectors,  the contribution of the the last 4 vectors are insignificant. Therefore, only the first 40 vectors were chosen to be the basis vector for the ground motion suite leading to a $U_{4500 \times 40}$ orthonormal basis matrix.  Each of these 40 basis vectors are plotted in Figure \ref{eq_svd}. Note that one can select a fewer number of basis vectors depending on desired level of accuracy. 

\subsection{Training and testing the machine learning models}\label{training_testing}

To represent the FE models using machine learning models, one needs to train the ML models using a large set of input output pairs.  The output of the machine learning model is chosen to be the response of the building in terms of peak displacement and peak floor acceleration, which are typically used to predict structural and non-structural damages in buildings, respectively,  for risk assessment.  This resulted in a two dimensional output vector $Y$.  In order to train the machine learning model for a range of earthquakes and model parameters, the following steps were adopted.  

\begin{figure}
\begin{center}
\includegraphics[valign=c,width=0.19\textwidth]{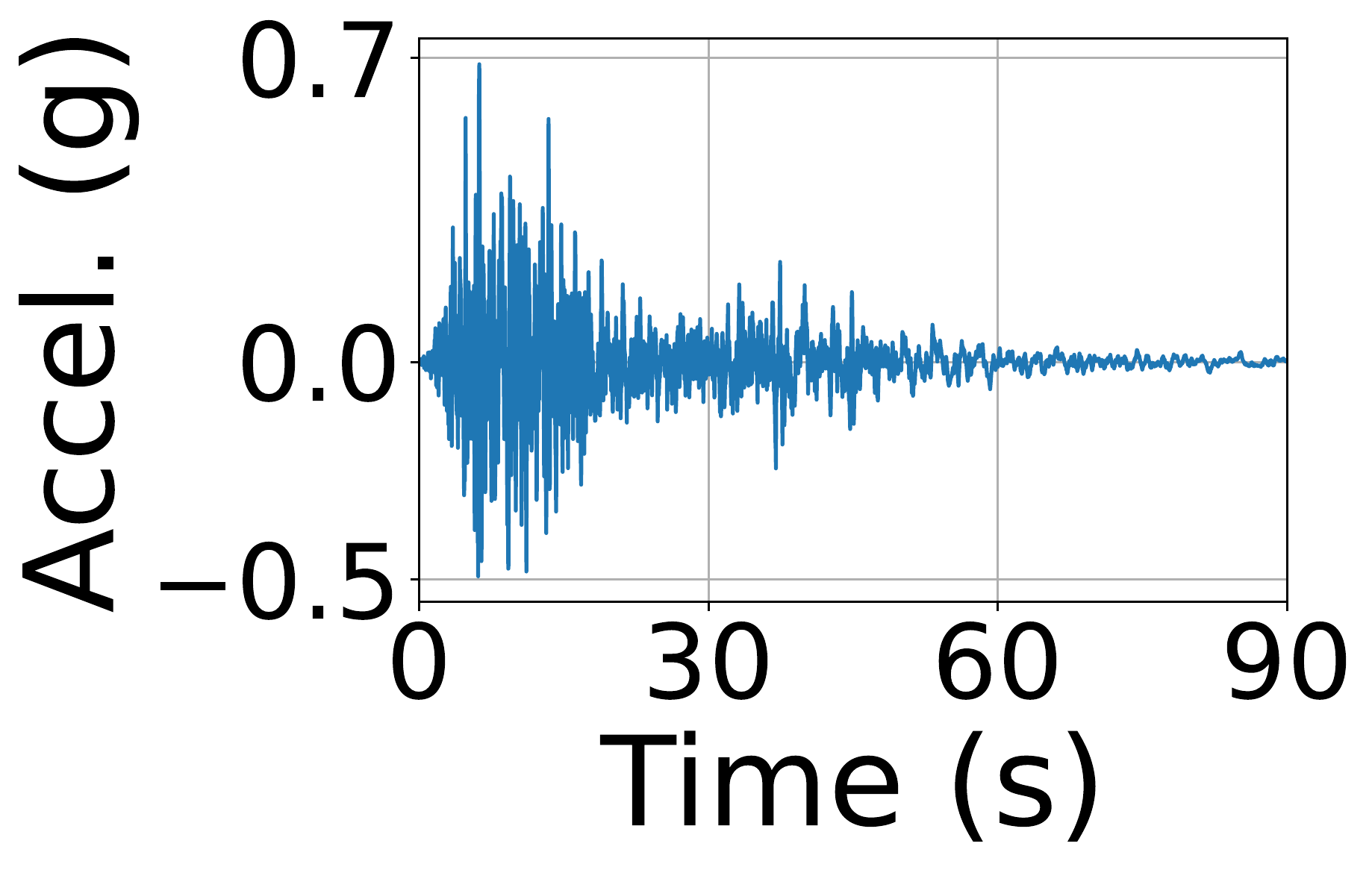} 
\includegraphics[valign=c,width=0.19\textwidth]{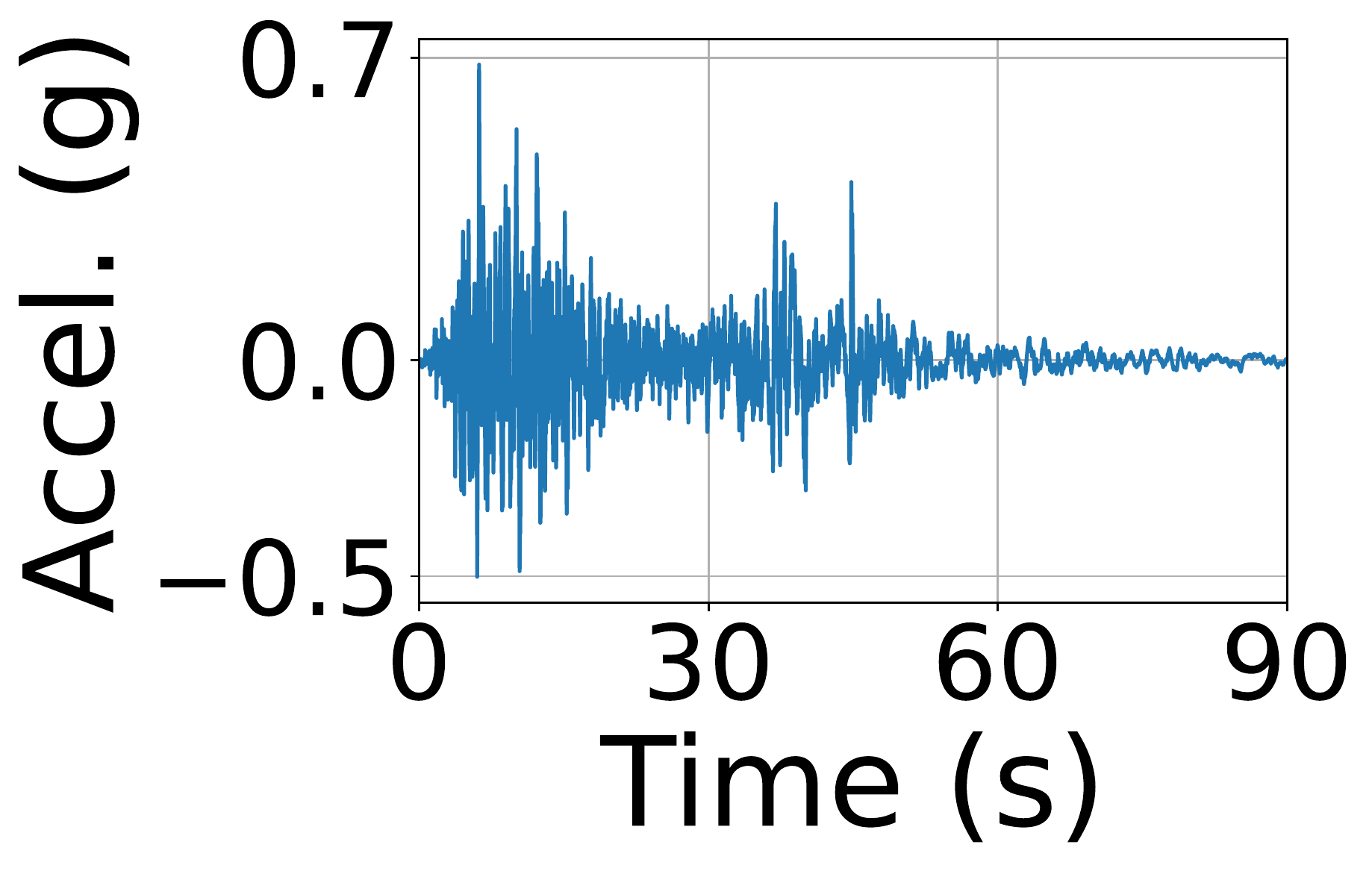} 
\includegraphics[valign=c,width=0.19\textwidth]{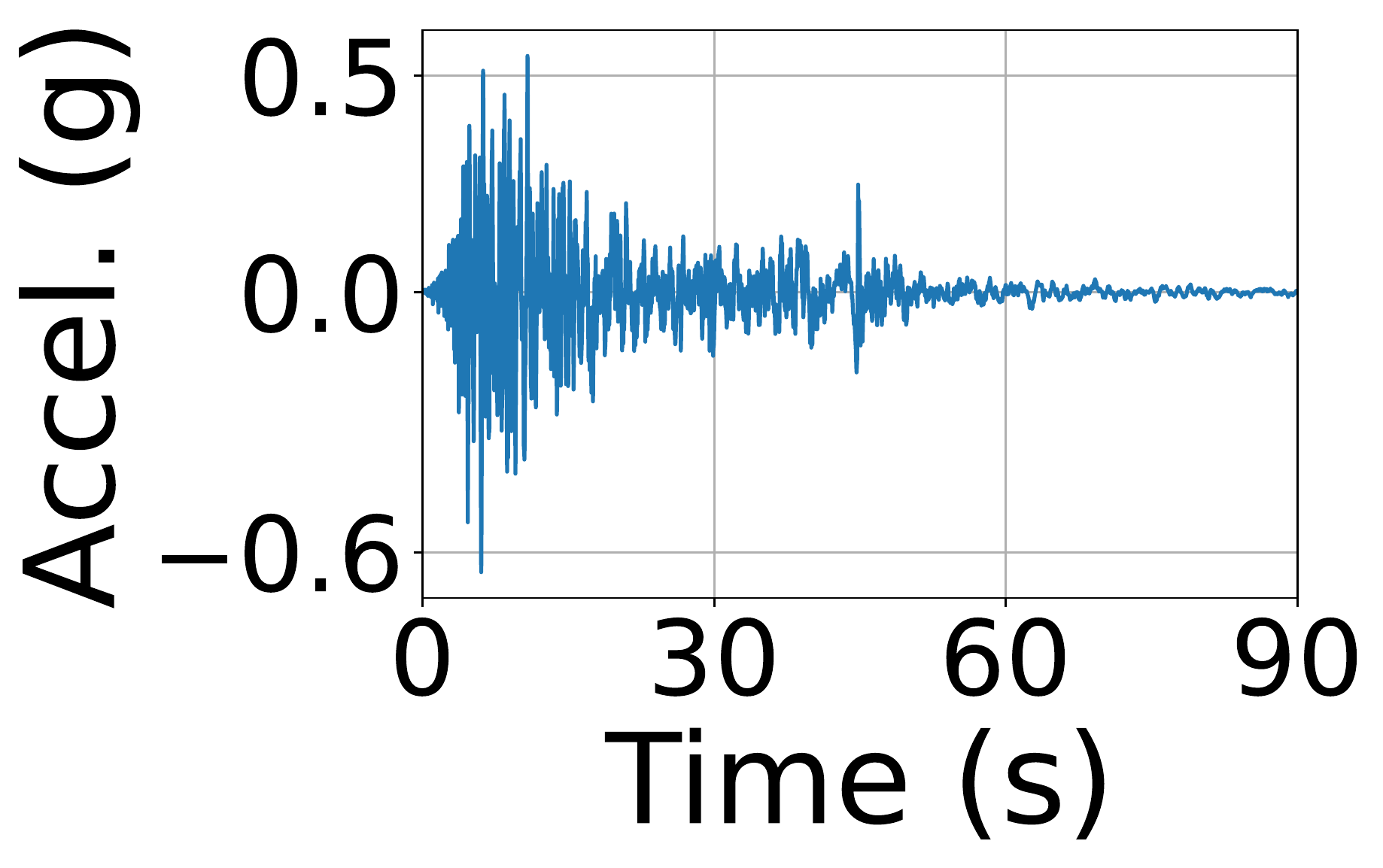}
\includegraphics[valign=c,width=0.19\textwidth]{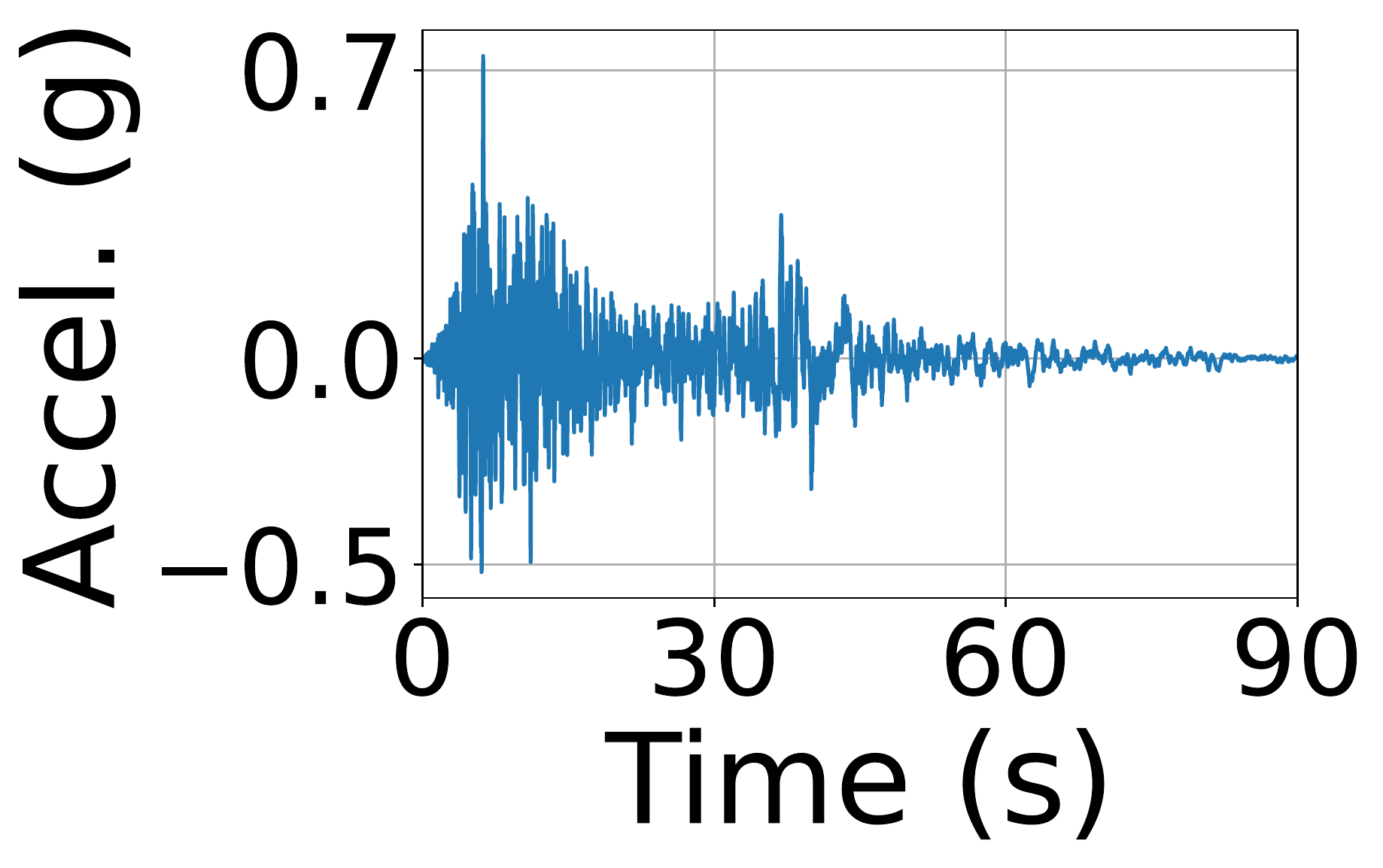}
\includegraphics[valign=c,width=0.19\textwidth]{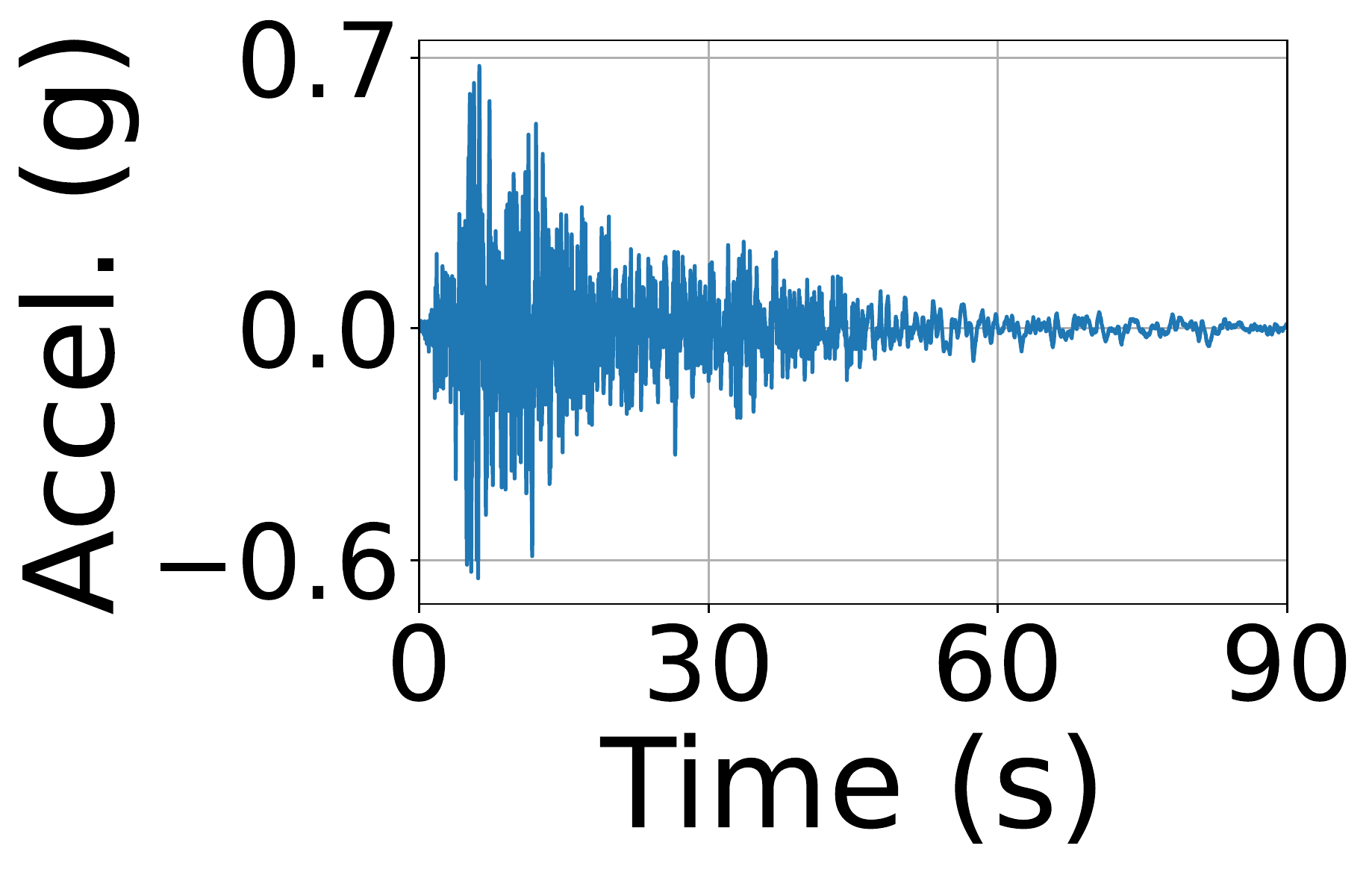} \\
\vspace*{0.35truecm}
\includegraphics[valign=c,width=0.19\textwidth]{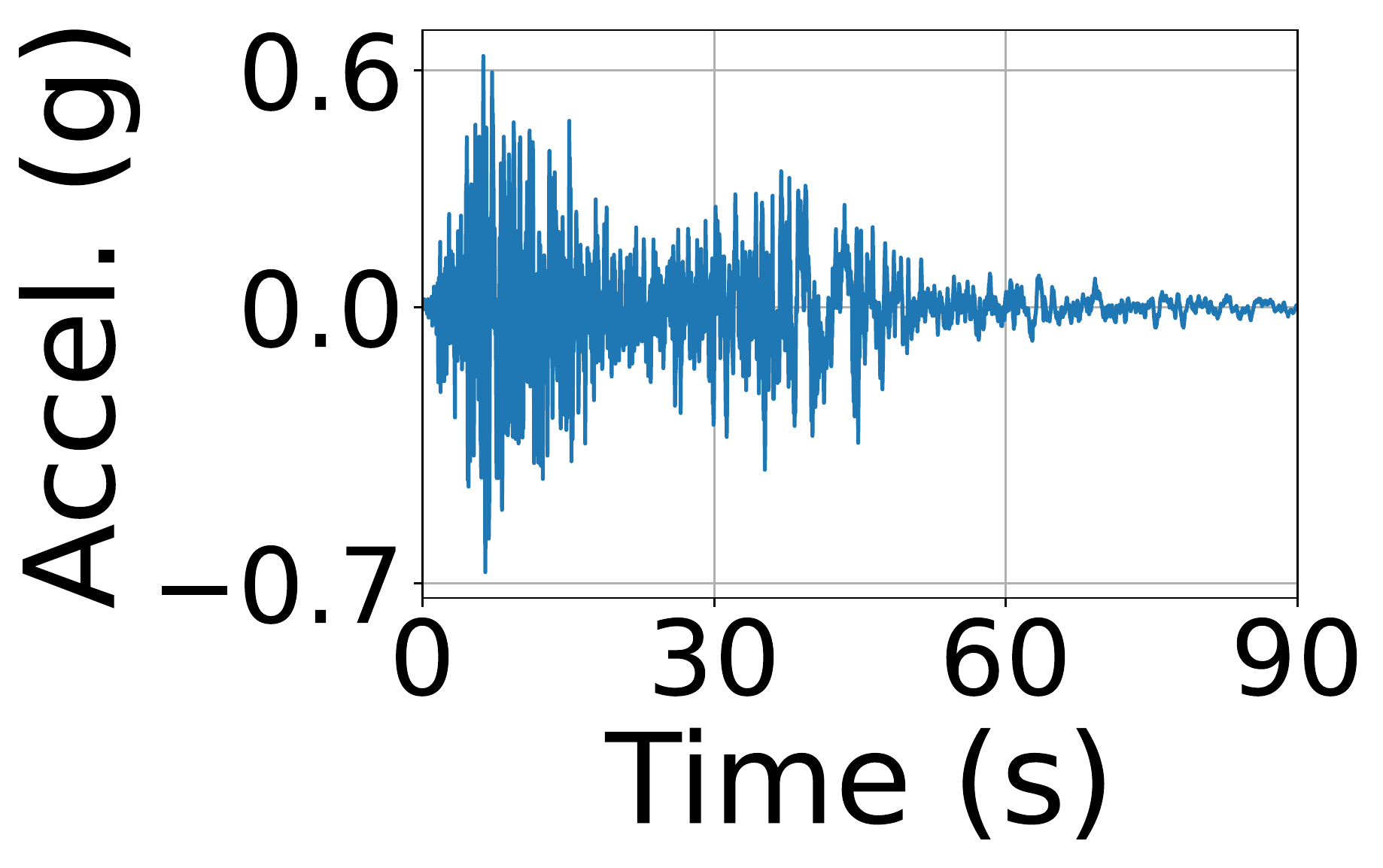} 
\includegraphics[valign=c,width=0.19\textwidth]{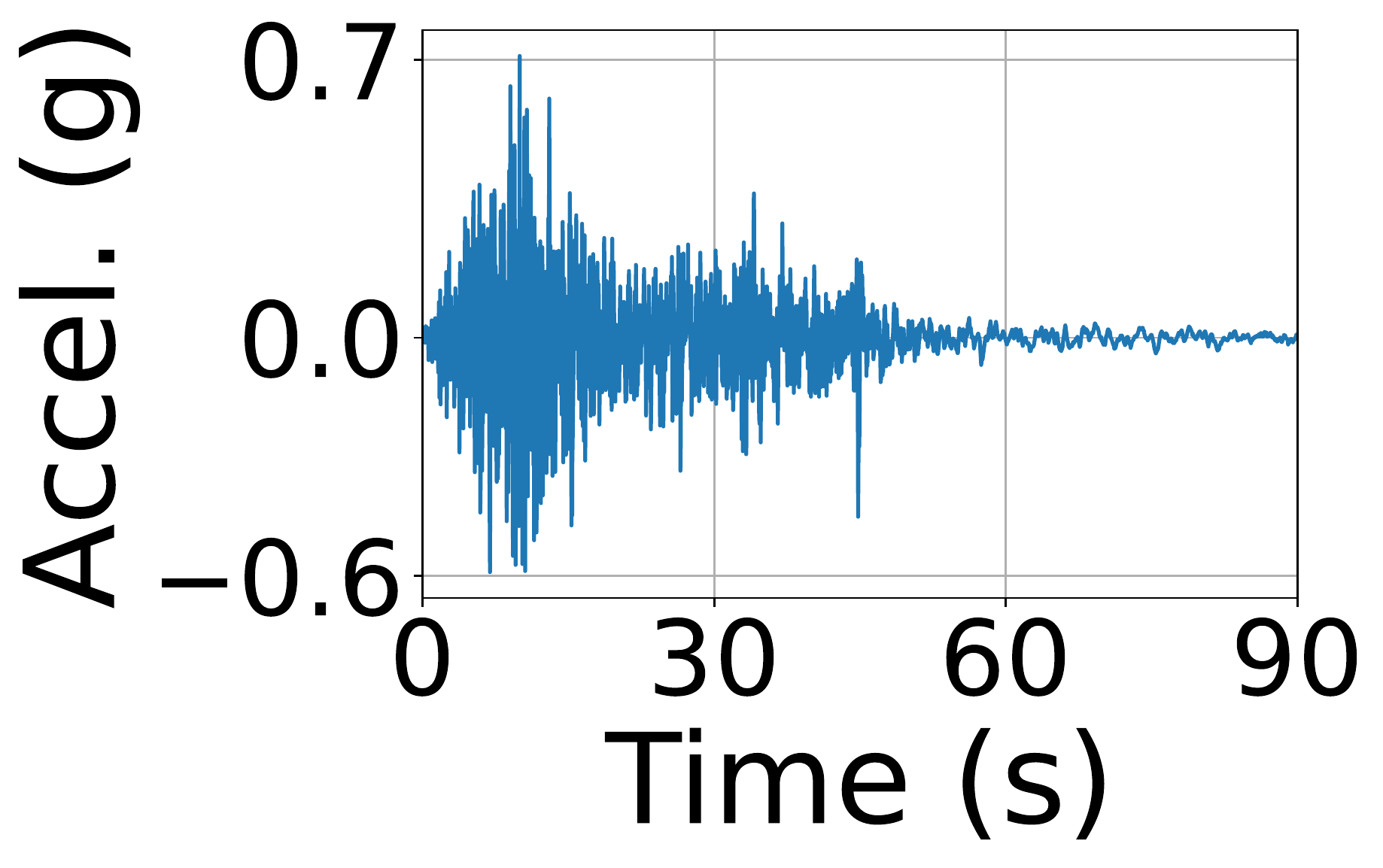} 
\includegraphics[valign=c,width=0.19\textwidth]{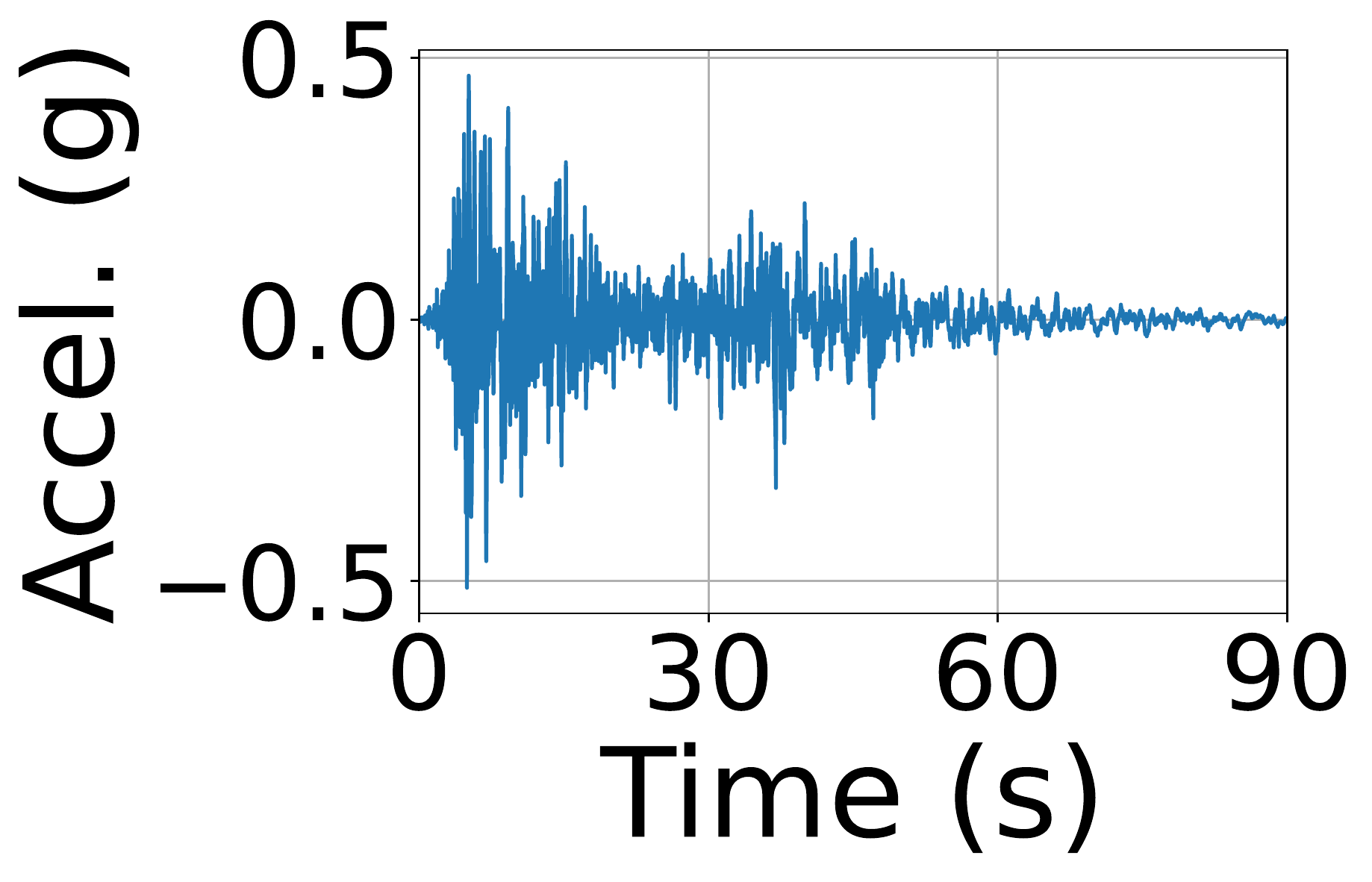}
\includegraphics[valign=c,width=0.19\textwidth]{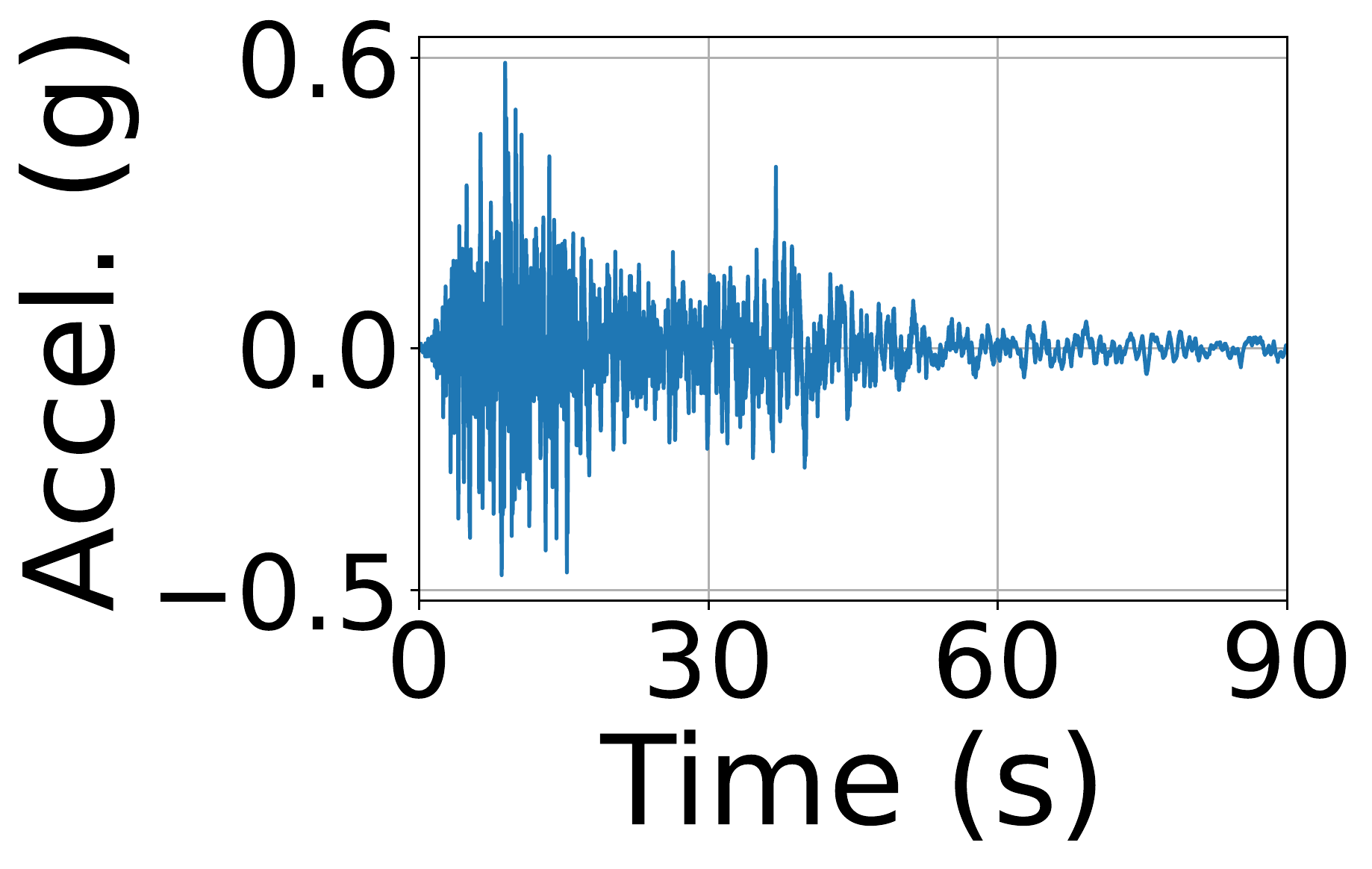}
\includegraphics[valign=c,width=0.19\textwidth]{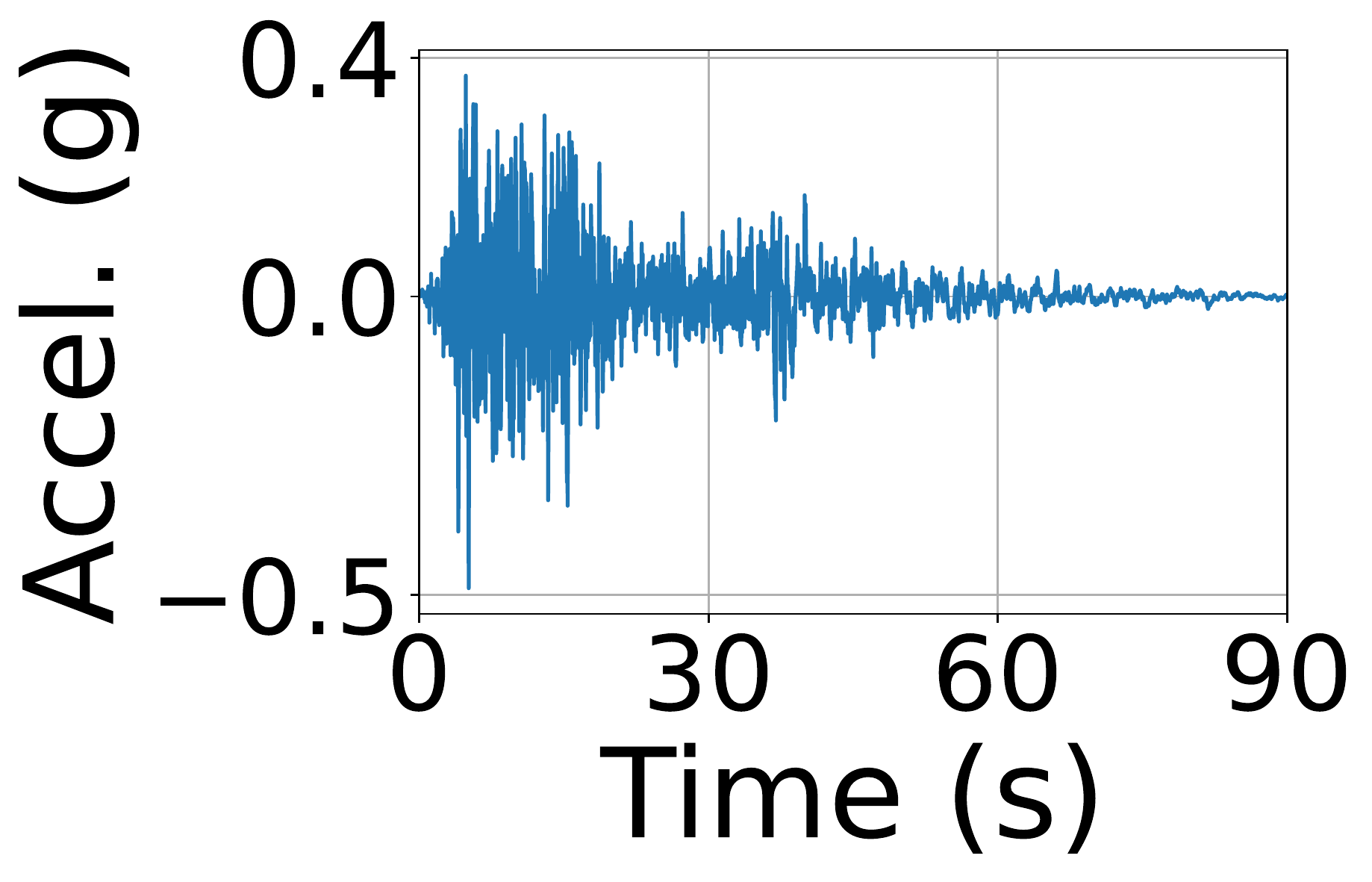} \\

\caption{Earthquake realizations obtained by multiplying random samples of $\Teta_1$ with $U$.}
\label{eq_real}
\end{center}
\end{figure}

First, to produce training ground motions,  the FEMA P695 far-field ground motion suite was projected onto the $U$ basis,  which was obtained from SVD, as described in the previous section.  For each earthquake,  a set of 40 weights were obtained corresponding to each column vector of the $U$ basis, which form the columns of the weight matrix, $\Sigma _{40 \times 44}$ in Equation \ref{PCA2}.  Each column of the weight matrix ,$\Sigma$, was assumed to be a realization of an uniform random vector $\Teta_{1_{40 \times 1}}$. The bounds of $\Teta_1$ were chosen to be maximum and minimum values of each row of $\Sigma$.  Please note that this bound can be extended to capture a wider range of earthquake variation. Now, one can produce new earthquakes by randomly generating values for $\Teta_1$ and multiplying it with $U$.  A few earthquakes corresponding to different realizations of $\Teta_1$ are demonstrated in Figure \ref{eq_real}. 

Secondly,  to account for material variability,  the constitutive model parameters such as elastic modulus ($E$), yield stress ($F_y$),  and damping ratio ($\xi$) were chosen to be uniform random variables and mass of each story was assumed to be 1.0.   A mean value of $E=$ 40 MPa and a variation of $\pm 20\%$ was chosen. Such a choice of $E$ resulted in a mean period of 1 s for the structure.  Mean of $F_y$ was chosen to be 0.7\% of the mean of the elastic modulus with bounds obtained by $\pm 20\%$ of the mean.  This choice of $F_y$ ensures that the response of the FE models reach the nonlinear regime for most of the realizations,  so that the machine learning model can be trained for non-linearity.  Mean of $\xi$ was chosen to be 0.05 with $\pm 20\%$ variation.  The ranges of the constitutive model parameters are shown in Table \ref{structural_param}. Since it was observed that the model response was sensitive to all three spring parameters through out the selected range, all of them were used as inputs to the machine learning models. This lead to a uniform random vector $\Teta_{2_{3 \times 1}}$ for the 1-DoF model, and $\Teta_{2 _{9 \times 1}}$ for the 3-DoF model.  Finally the input vector for the machine learning model $\Teta$ can be written as $\Teta=[\Teta_1, \Teta_2]^T$, which is a 43 dimensional vector for the 1-DoF model and the 49 dimensional vector for the 3-DoF model.  
 \begin{table}[h!]
\caption{Mean and support of the structural parameters.}
\label{structural_param}
\begin{center}
    \begin{tabular}{| c | c | c|}
    \hline
Parameter & Mean & Support\\
\hline \hline   
Elastic modulus($E$)  & 40 MPa  & [32 Mpa,48 Mpa]\\
\hline 
Yield stress ($F_y)$ & 0.28 MPa  & [0.21 Mpa,0.35 Mpa]\\
\hline 
Damping ratio ($\xi$) & 0.05  & [0.04,0.06]\\
\hline 
\end{tabular}
\end{center}
\end{table}

\begin{figure}
\begin{center}
\includegraphics[valign=c,width=0.19\textwidth]{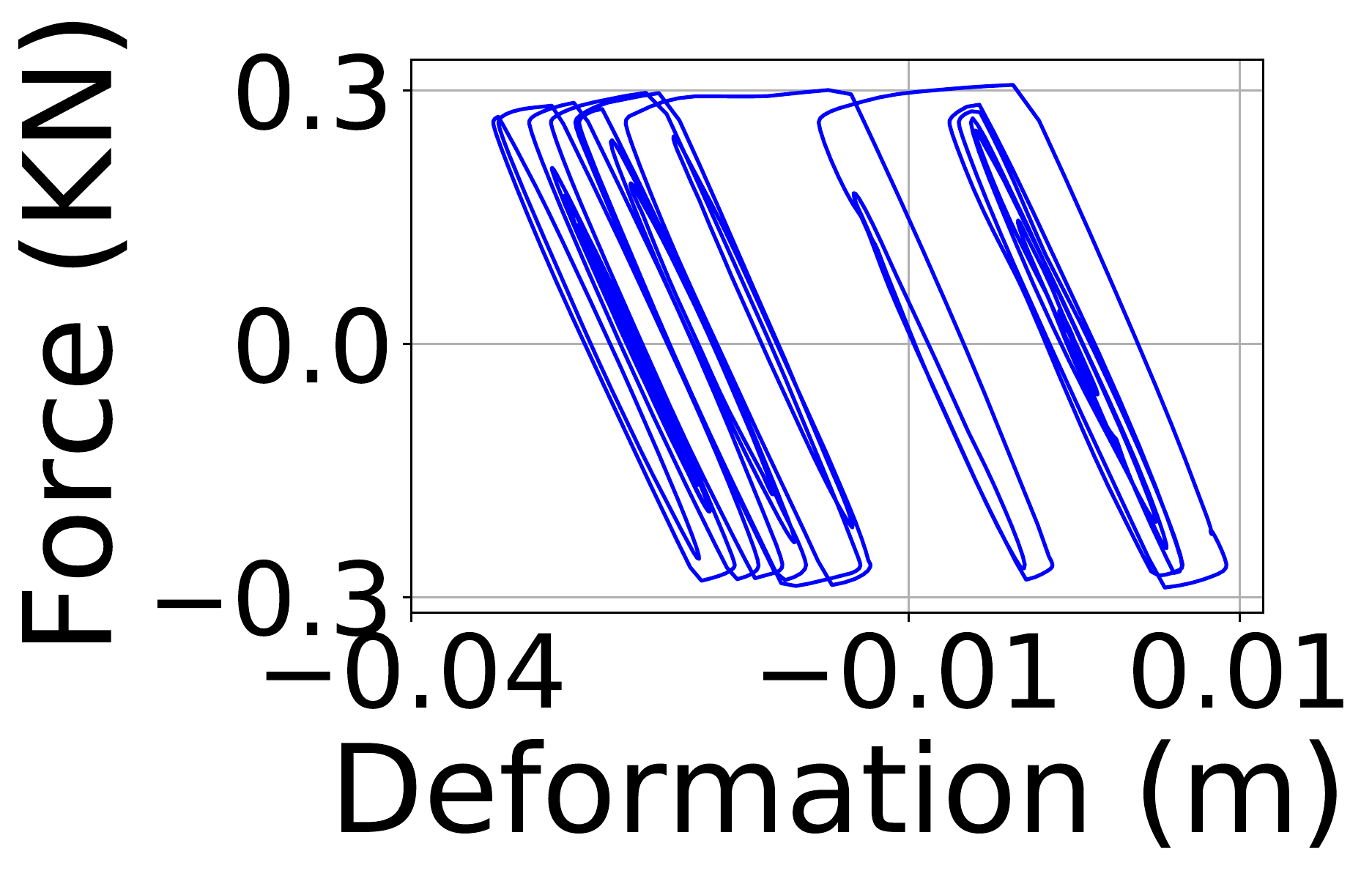} 
\includegraphics[valign=c,width=0.19\textwidth]{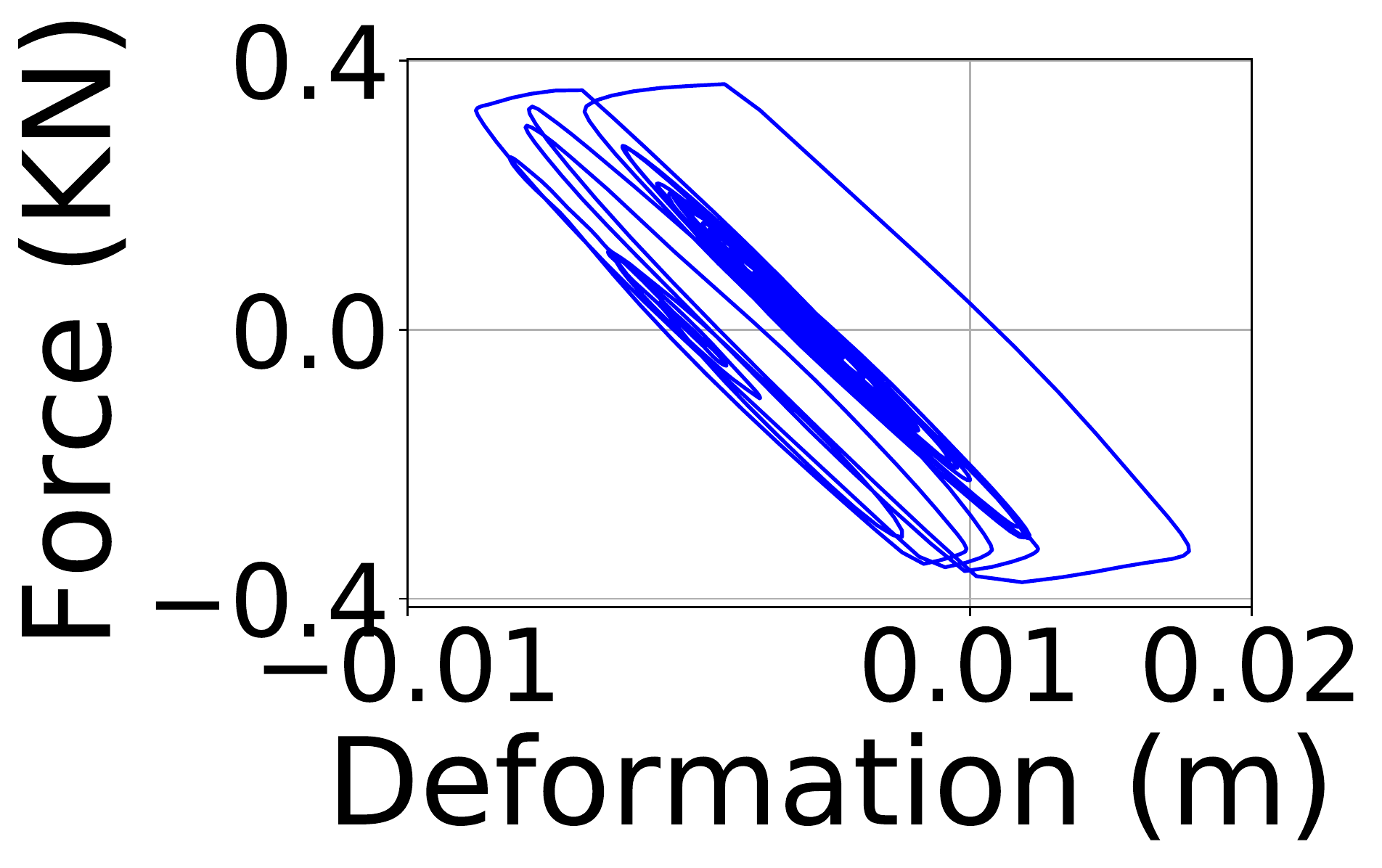} 
\includegraphics[valign=c,width=0.19\textwidth]{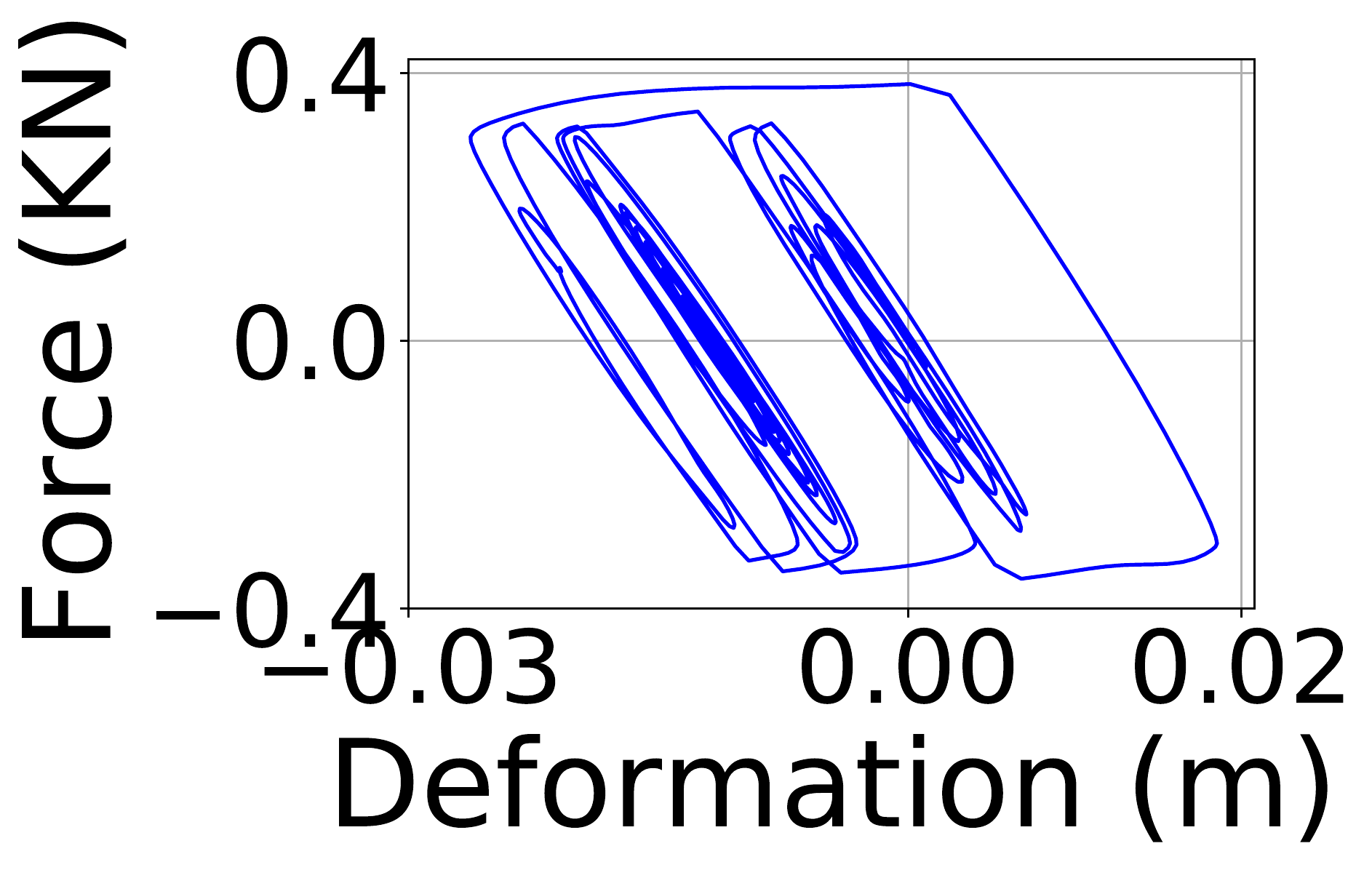}
\includegraphics[valign=c,width=0.19\textwidth]{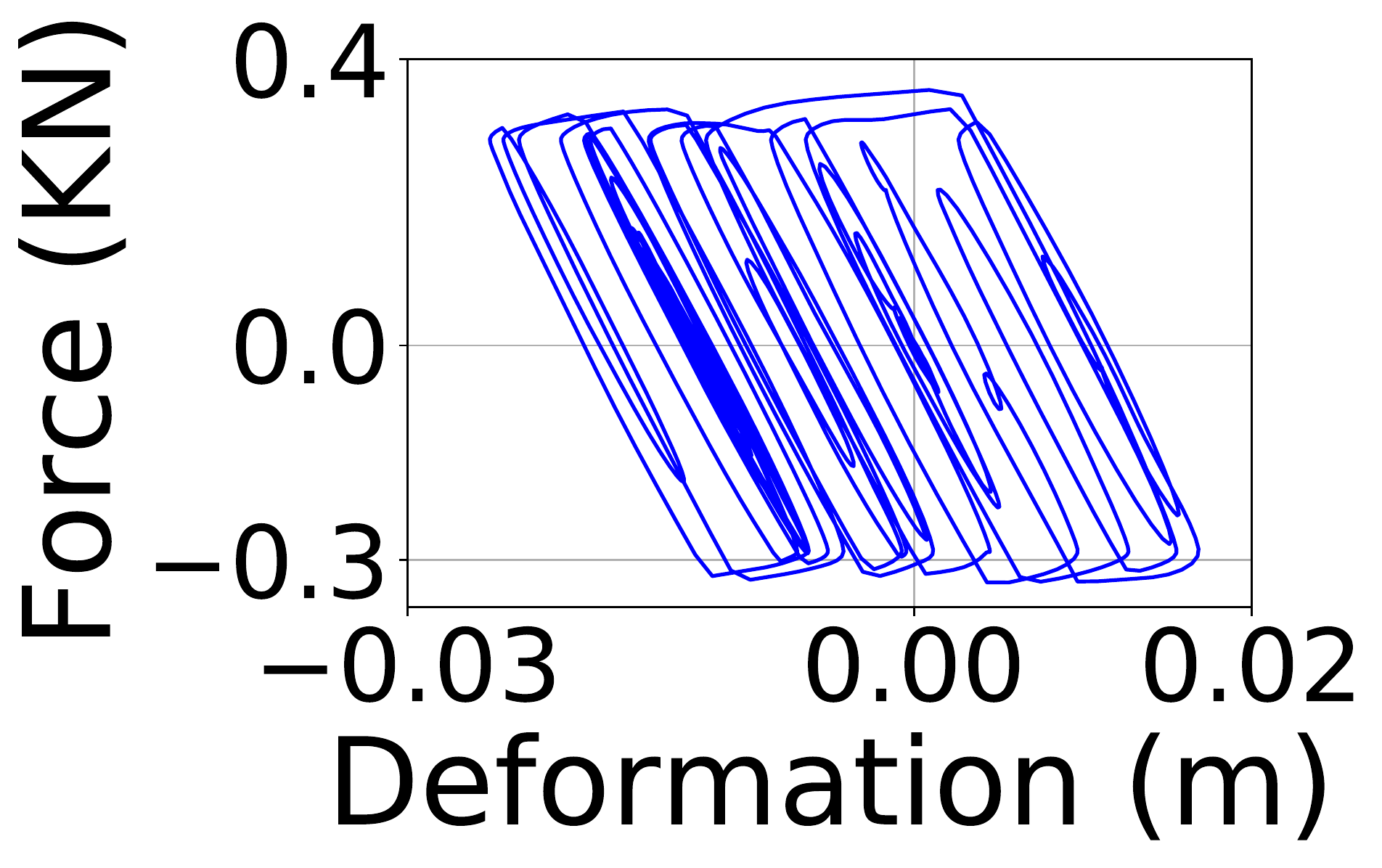}
\includegraphics[valign=c,width=0.19\textwidth]{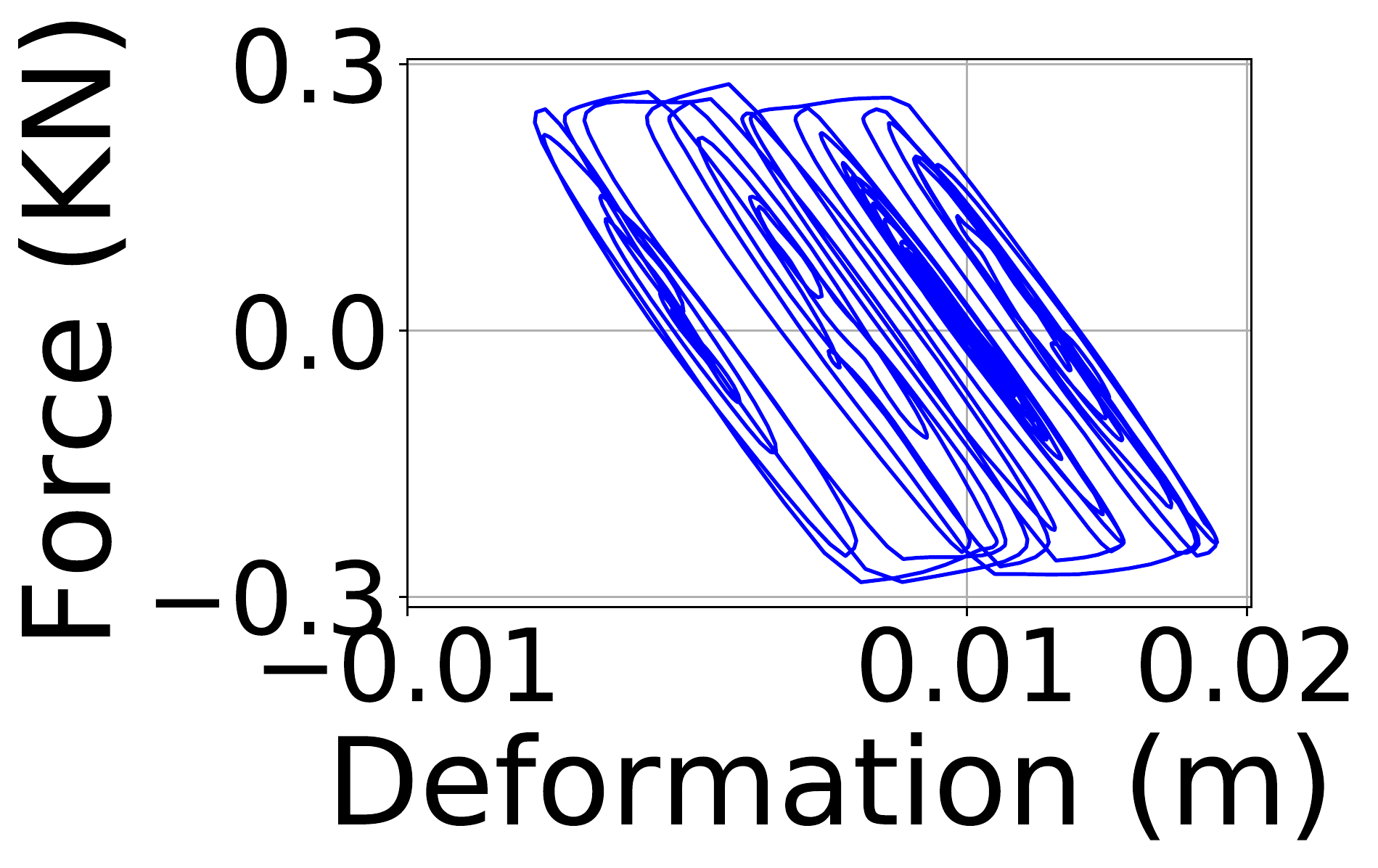} \\
\vspace*{0.35truecm}
\includegraphics[valign=c,width=0.19\textwidth]{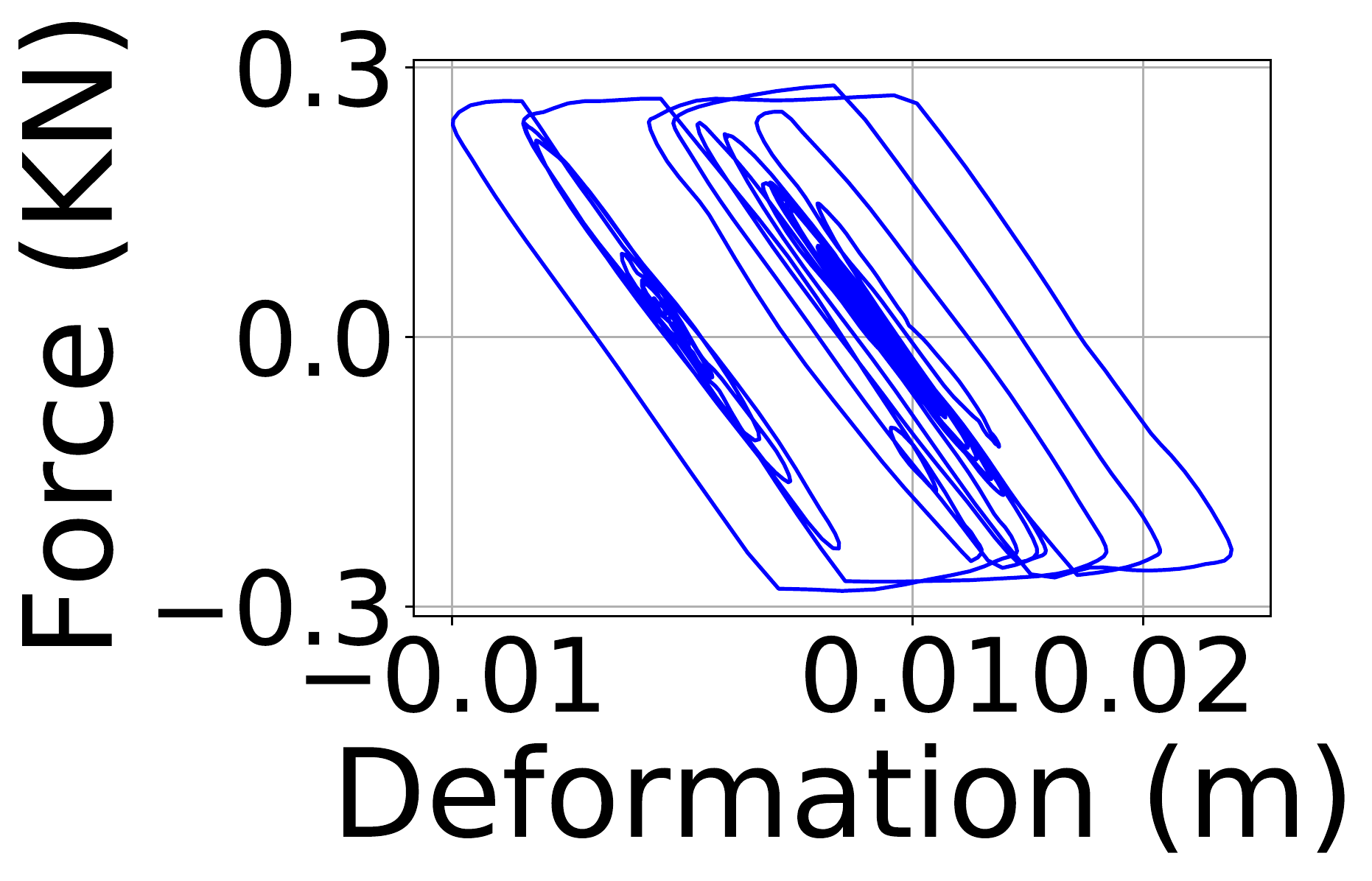} 
\includegraphics[valign=c,width=0.19\textwidth]{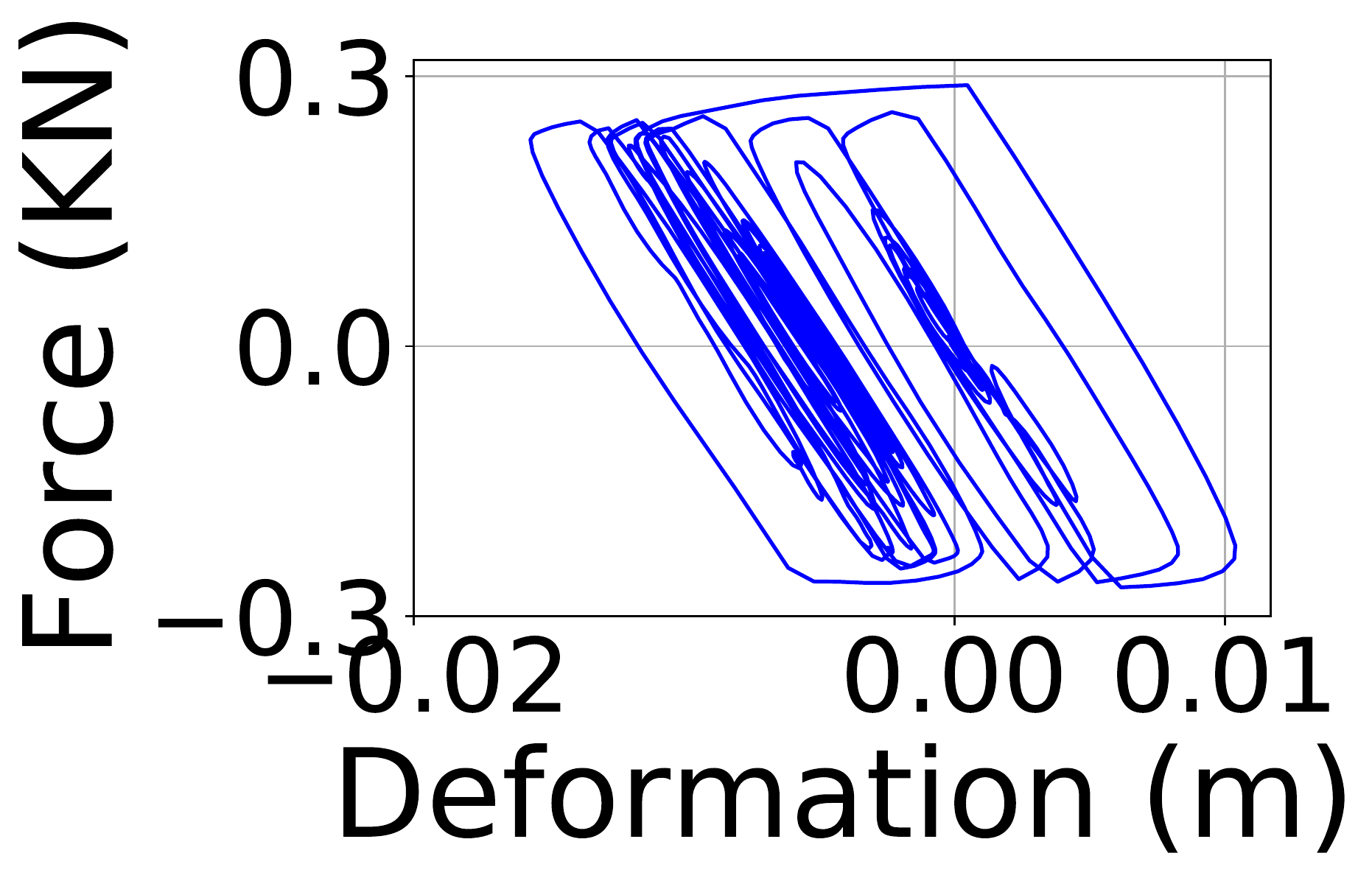} 
\includegraphics[valign=c,width=0.19\textwidth]{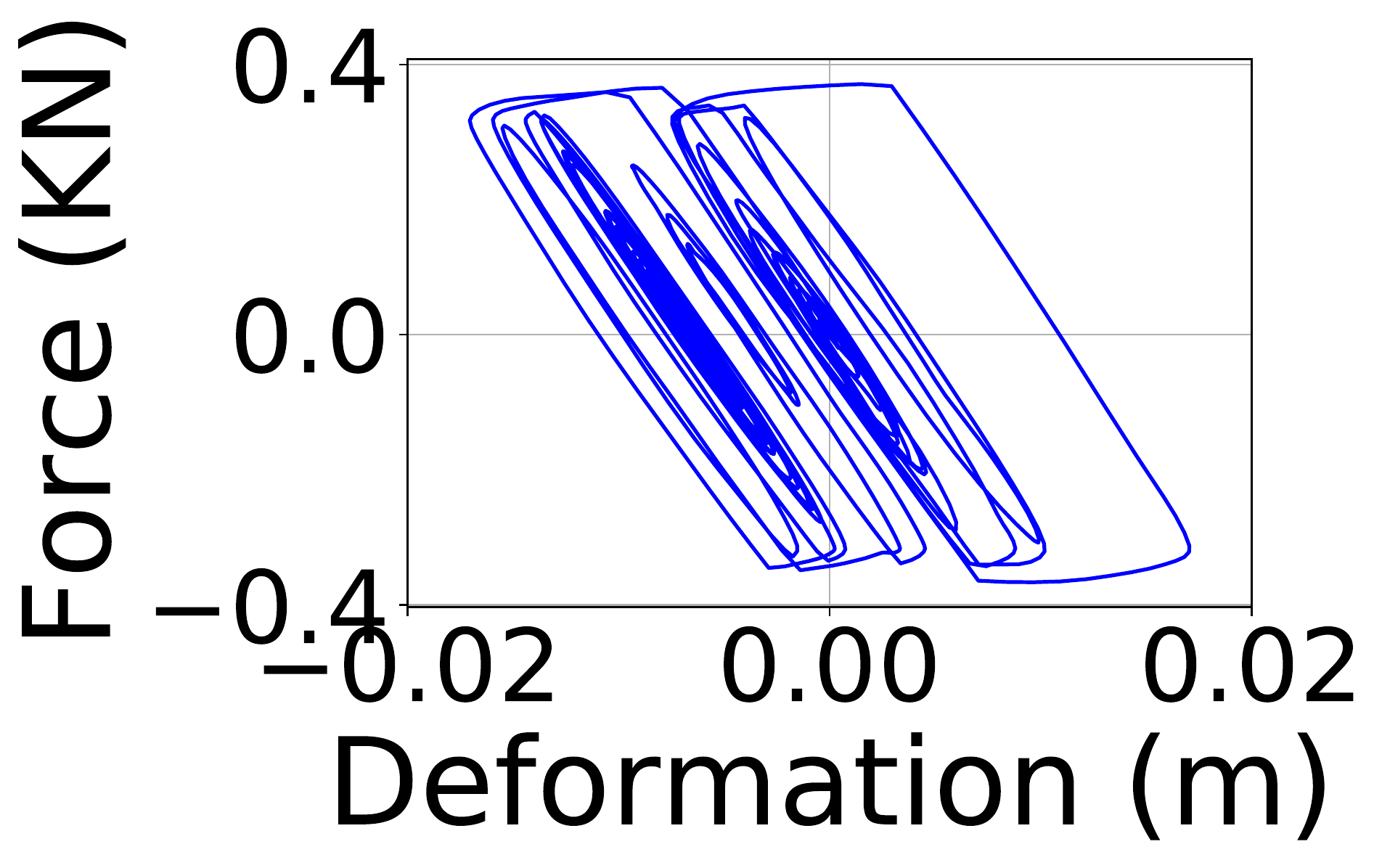}
\includegraphics[valign=c,width=0.19\textwidth]{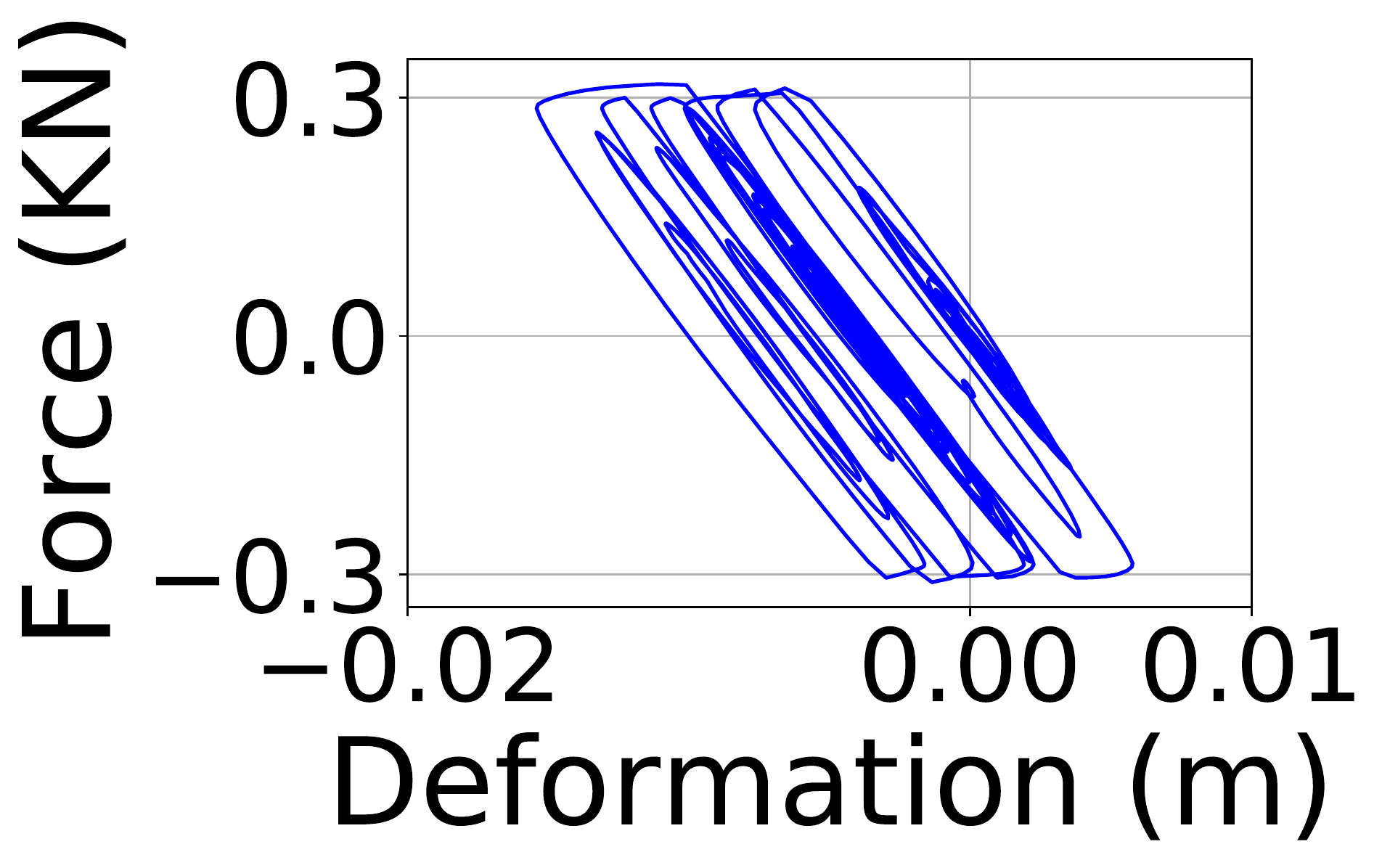}
\includegraphics[valign=c,width=0.19\textwidth]{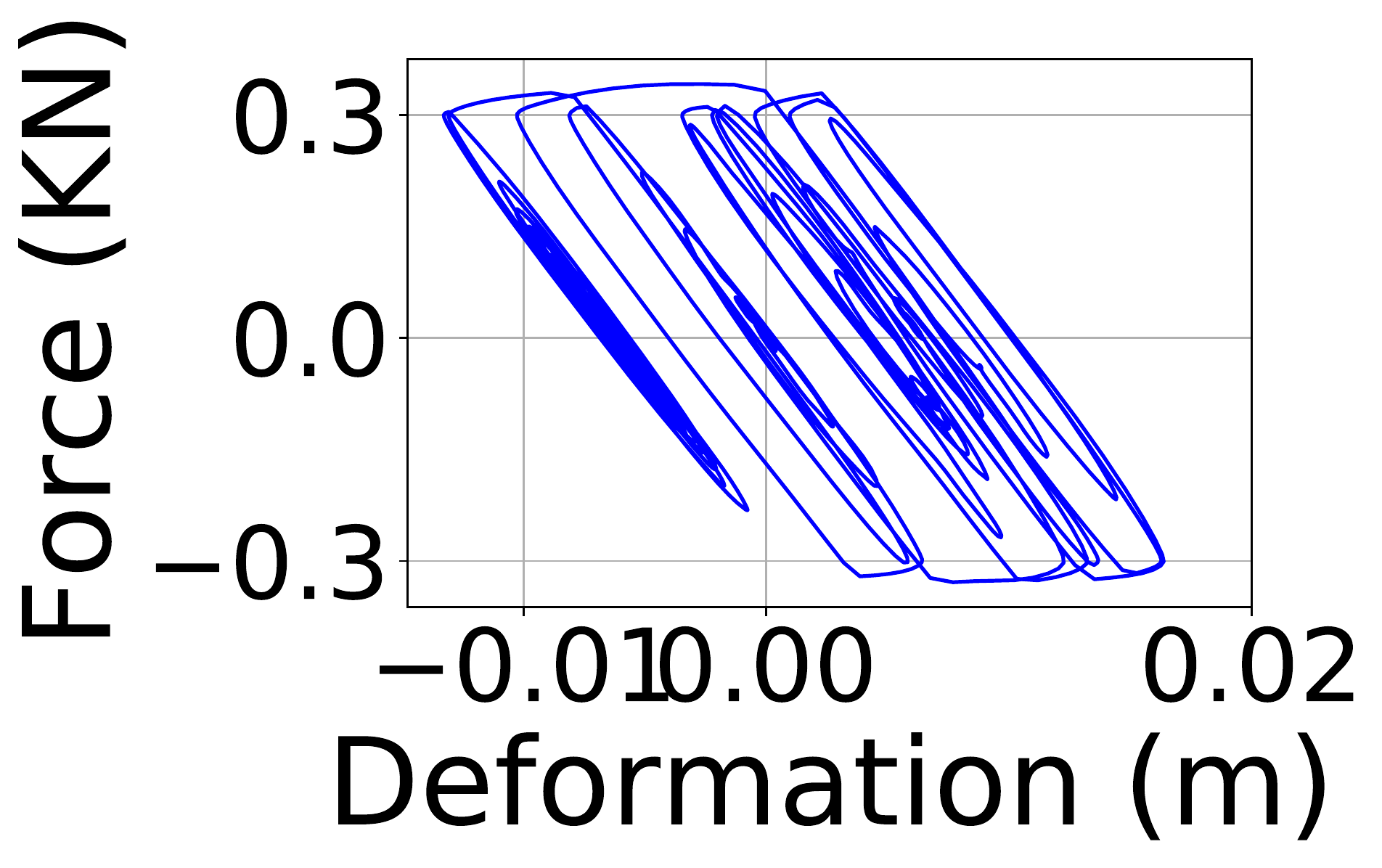} \\
\caption{Force-deformation plots corresponding to realizations of earthquakes shown in Figure \ref{eq_real} and material parameters sampled from Table \ref{structural_param}.}
\label{Force_Def_real}
\end{center}
\end{figure}

Once, the input random vector $\Teta$ and its PDF have been chosen, 500,000 realizations were generated.  At this number of realizations it was observed that all the machine learning models produced their respective best results.  No convergence study was conducted pertaining to the number of points for individual machine learning models, since, along with the hyper-parameter selection it would have resulted in a very complex problem which is beyond the scope of this study.  The FE models were simulated for each of the realizations to obtain the two dimensional output vectors $Y$.  The total simulation time was about 4 hours on 64 Intel i9 computing cores with 64 GB of RAM. Force deformation plots  corresponding to the earthquake time histories demonstrated in Figure \ref{eq_real} are shown in Figure \ref{Force_Def_real}.  It can be noted that the chosen earthquakes and material parameters resulted in highly non-linear behavior and it was observed that about 99\% of all the realizations displayed non-linear behavior. Next, in order to abstain from over-fitting, of the 500,000 input-output data points, 90\% was used for training and 10\% was used for testing.  Now, the competing machine learning models were trained on the training input-output order pair and their performance was tested on the testing input-output order pair.  $R^2$ error  defined in Eq. \ref{r2}, was chosen as the performance metric:
\begin{equation}\label{r2}
R^2 (Y,\hat{Y})= 1-\frac{Variance(Y-(\hat{Y}))}{Variance({Y})}
\end{equation}
where the numerator is the error variance between the FE model output,  $Y$, and machine learning model output, $\hat{Y}$. 

\begin{figure}[h!]
\begin{center}
\subfigure[\label{1d_train} ]{\includegraphics[width=0.45\textwidth]{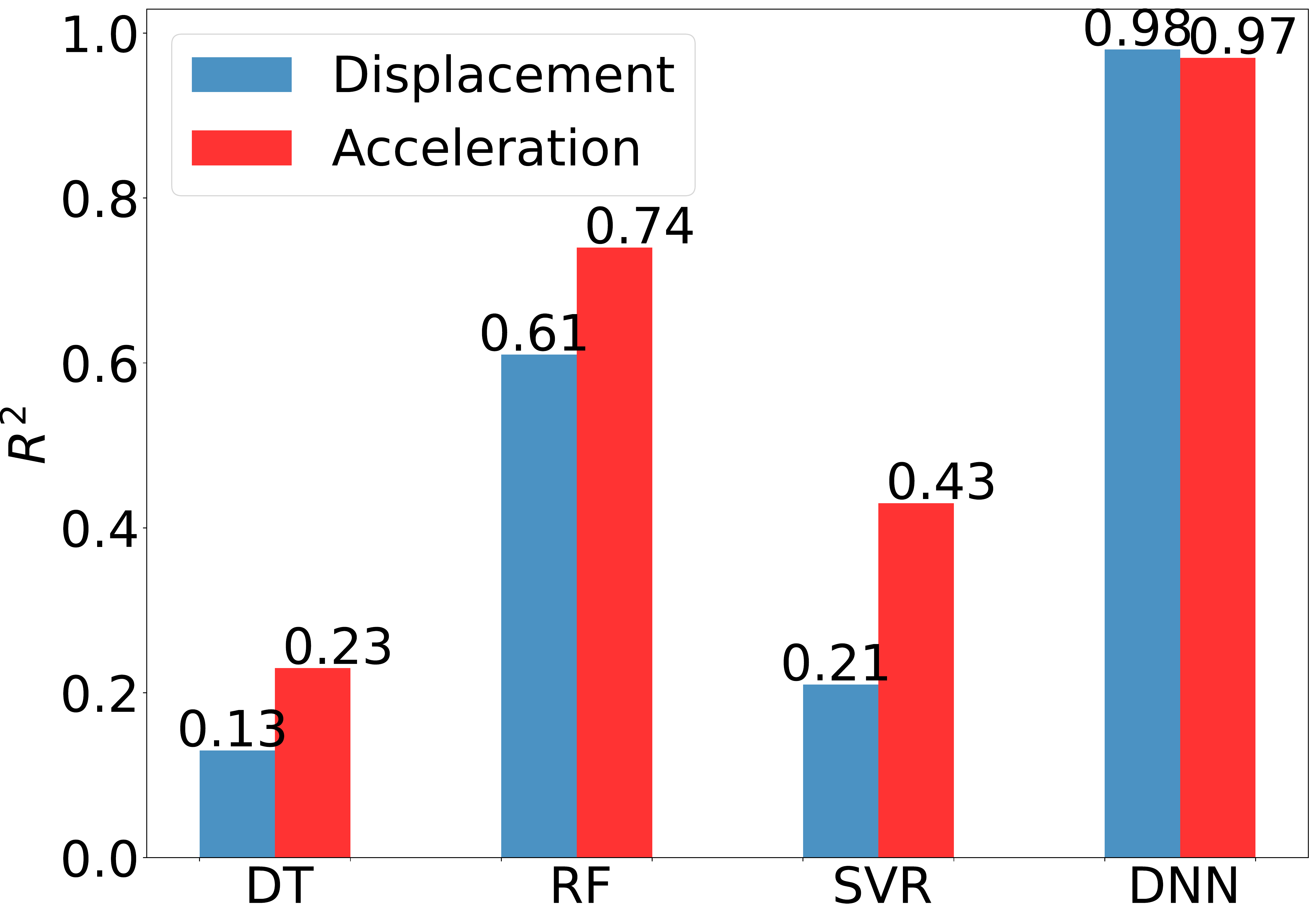}} 
\hspace*{0.1truecm}
\subfigure[\label{1d_test} ]{\includegraphics[width=0.45\textwidth]{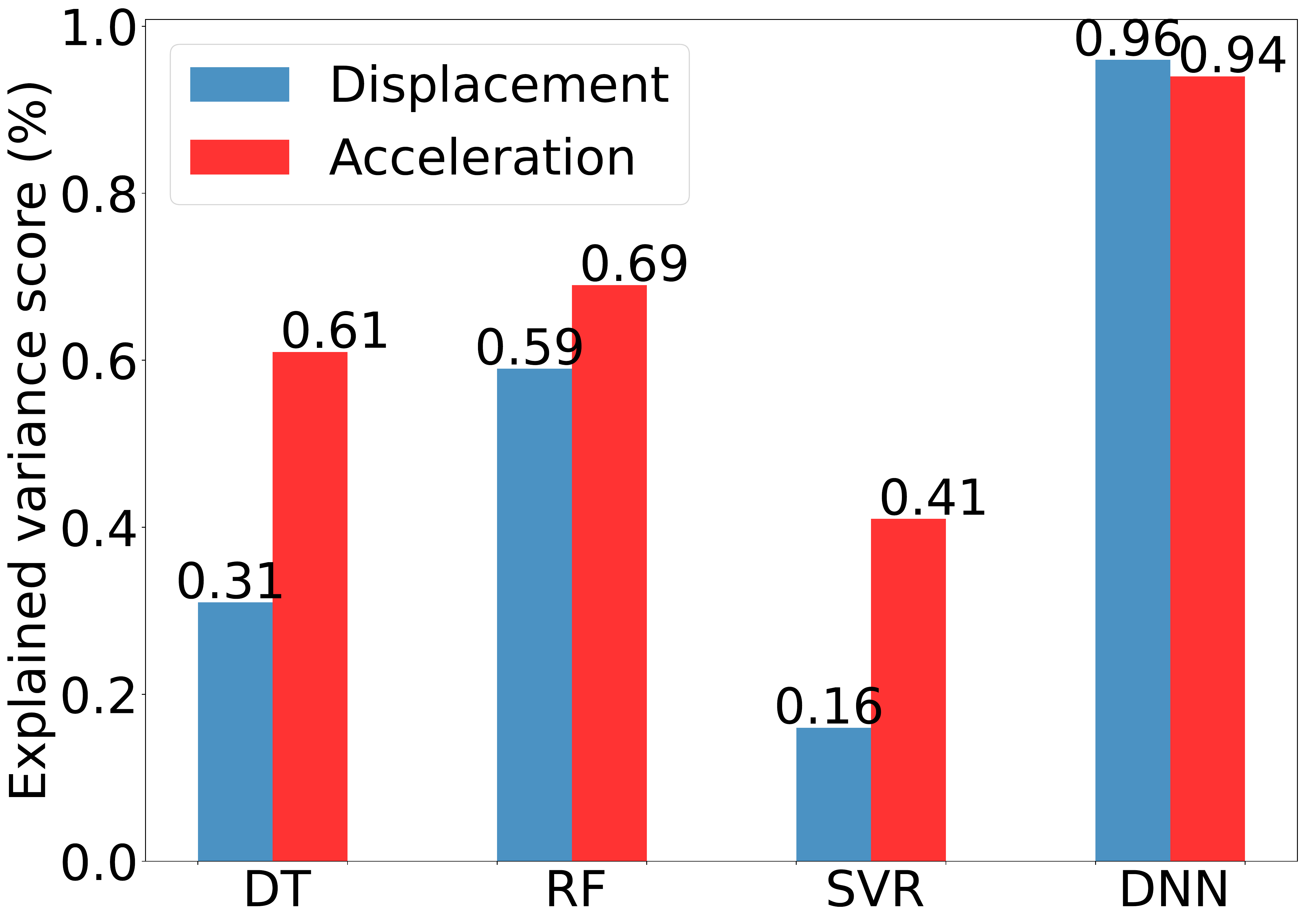}} 
\caption{Performance of various machine learning models in terms of $R^2$ (a) training data  (b) testing data for the one-story building}
\label{test_train_1}
\end{center}
\end{figure}

\begin{figure}[h!]
\begin{center}
\subfigure[\label{3d_train}]{\includegraphics[width=0.45\textwidth]{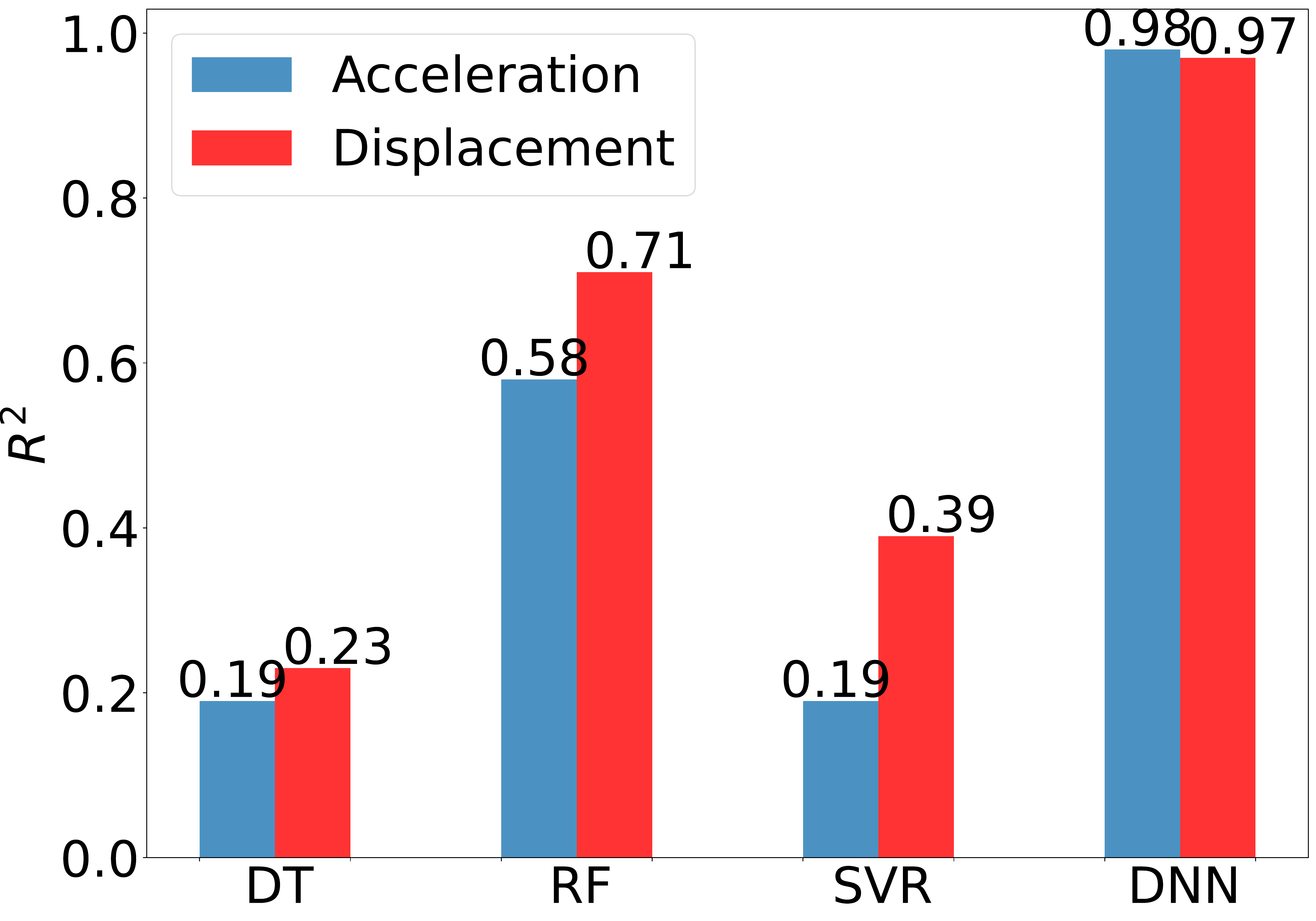}} 
\hspace*{0.1truecm}
\subfigure[\label{3d_test}]{\includegraphics[width=0.45\textwidth]{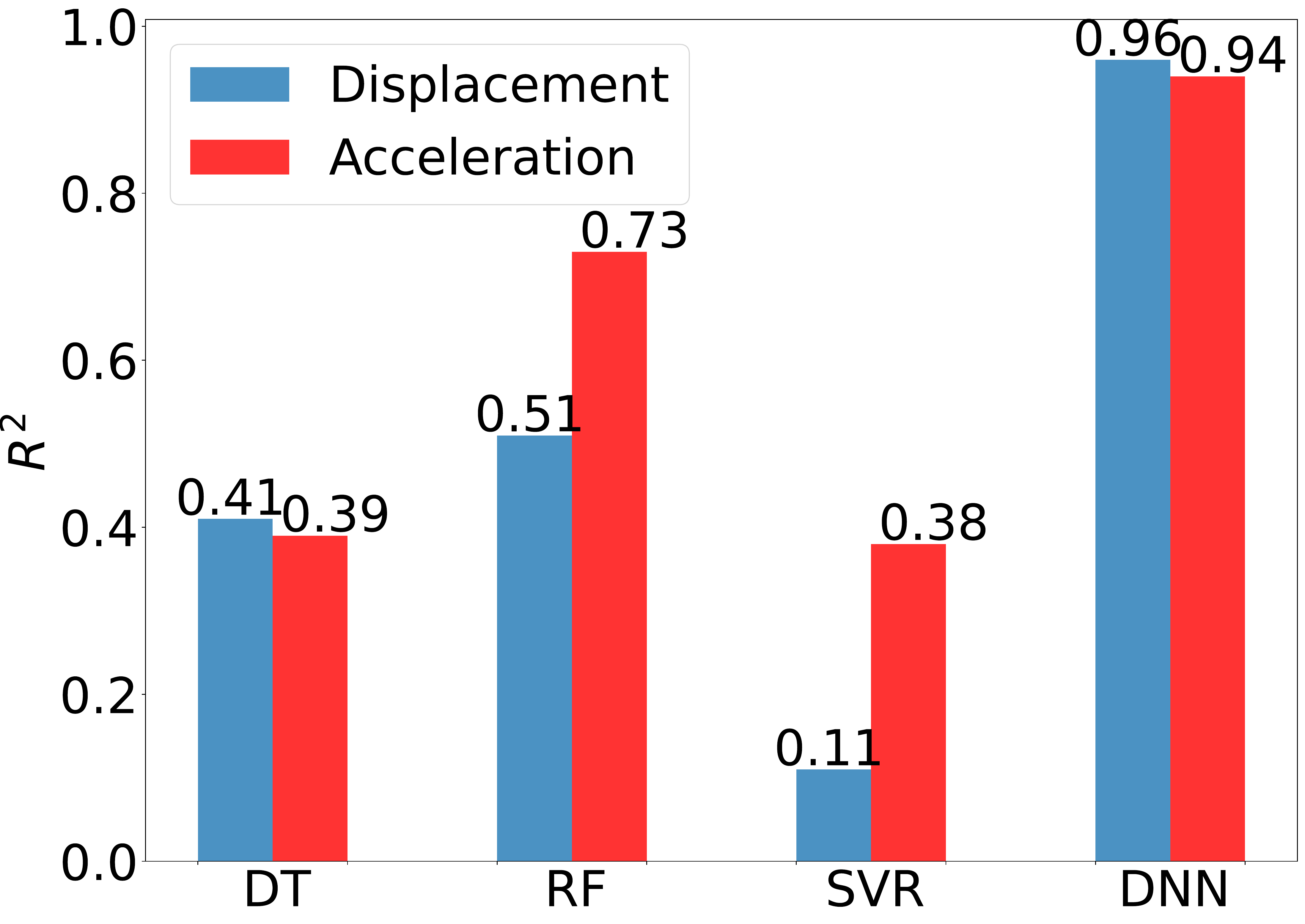}} 
\caption{Performance of various machine learning models in terms of $R^2$on (a) training data  (b) testing data for the three-story building}
\label{test_train_2}
\end{center}
\end{figure}

It must be noted that choosing among the various competing machine learning models is a two step process.  It involves first choosing the best set of hyper parameters for a model type. Once the best set of hyper-parameters are chosen then the best model is selected which produces the least $R^2$ error. As mentioned earlier, a cross validation study was done to chose the best hyper-parmeters for SVR,  DT and RF.   The hyper-parameters for DNN were chosen heuristically by varying the numbers of neurons and layers. The best values of hyper-parameters corresponding to each of the machine learning models mentioned in Section \ref{section4} are shown in Table \ref{hyperparameters}.   $R^2$ error for each of the machine learning model, corresponding to the best hyper parameters,  is shown in Figure \ref{test_train_1}-\ref{test_train_2}.  Figure \ref{test_train_1} (a) corresponds to the training error for the one-story building and \ref{test_train_1} (b) corresponds to the testing error for the one-story building. Similarly,  Figure \ref{test_train_2} (a) corresponds to the training error for the three-story building and \ref{test_train_2} (b) corresponds to the testing error for the three-story building.  It can be observed that the best performance is obtained by DNN in both training and testing, followed by RF, SVR and DT in both 1-DoF and 3-DoF cases.  Intuitively, DNN is expected to perform better for the problem at hand due to the highly non-linear nature of the finite element model. The possibility of adding multiple layers with a variety of non-linearities enables DNN to capture complex structure in data, which is otherwise impossible for machine learning models considering linear or simple non-linear basis functions. Moreover, the stochastic nature of the gradient based training algorithms helps the model to generalize well on unseen samples, thus avoiding overfitting. Therefore,  the proposed model is a DNN with 10 layers and 500 neurons in each layer.  One can appreciate the merit of splitting the data into training and testing sets by checking the result for DT in Figure \ref{test_train_1}. It can be clearly seen that DT overfits the data as demonstrated by the low training error and high testing error.

Now, this DNN model needs to be validated before deploying. Even though the test set was not used for training the models, it was indirectly used for selection of hyper-parameters for SVR, RF and DT.  For DNN,  convergence of error on the testing set was used to stop the forward and backward propagation loops. Therefore, in some sense there was leakage of information from the testing set to the training set. Moreover,  it might so happen that co-incidentally the model might perform well on one test set. Therefore, a thorough validation step is important before deploying the best performing machine learning model,  DNN in our case,  for use. 

 \begin{table}[h!]
\caption{Hyperparameter values that yield the best results for each ML model}
\label{hyperparameters}
\begin{center}
    \begin{tabular}{| p{0.35\textwidth} | p{0.35\textwidth} | c|}
    \hline
Machine Learning Model & Hyperparameters & Values\\
\hline \hline   
Decision trees (DT) & Maximum depth  & [100]\\
\hline 
Random forests (RF) & Number of trees  & [250]\\
\hline 
Support vector regression (SVR) & $\lambda$ and $\epsilon$& [1.5,0.5]\\
\hline 
Deep neural networks (DNN) & Number of layers and neurons in each layer, & [10, 500]\\
\hline 
\end{tabular}
\end{center}
\end{table}

\subsection{Validation of the machine learning surrogate model}

As mentioned earlier validation is critical to effective real world use. The selected DNN model was thoroughly validated as explained in the following.

\subsubsection{Using unseen generated earthquakes and random parameters}
\begin{figure}[h!]
\begin{center}
\subfigure[\label{pdf_fy} ]{\includegraphics[width=0.32\textwidth]{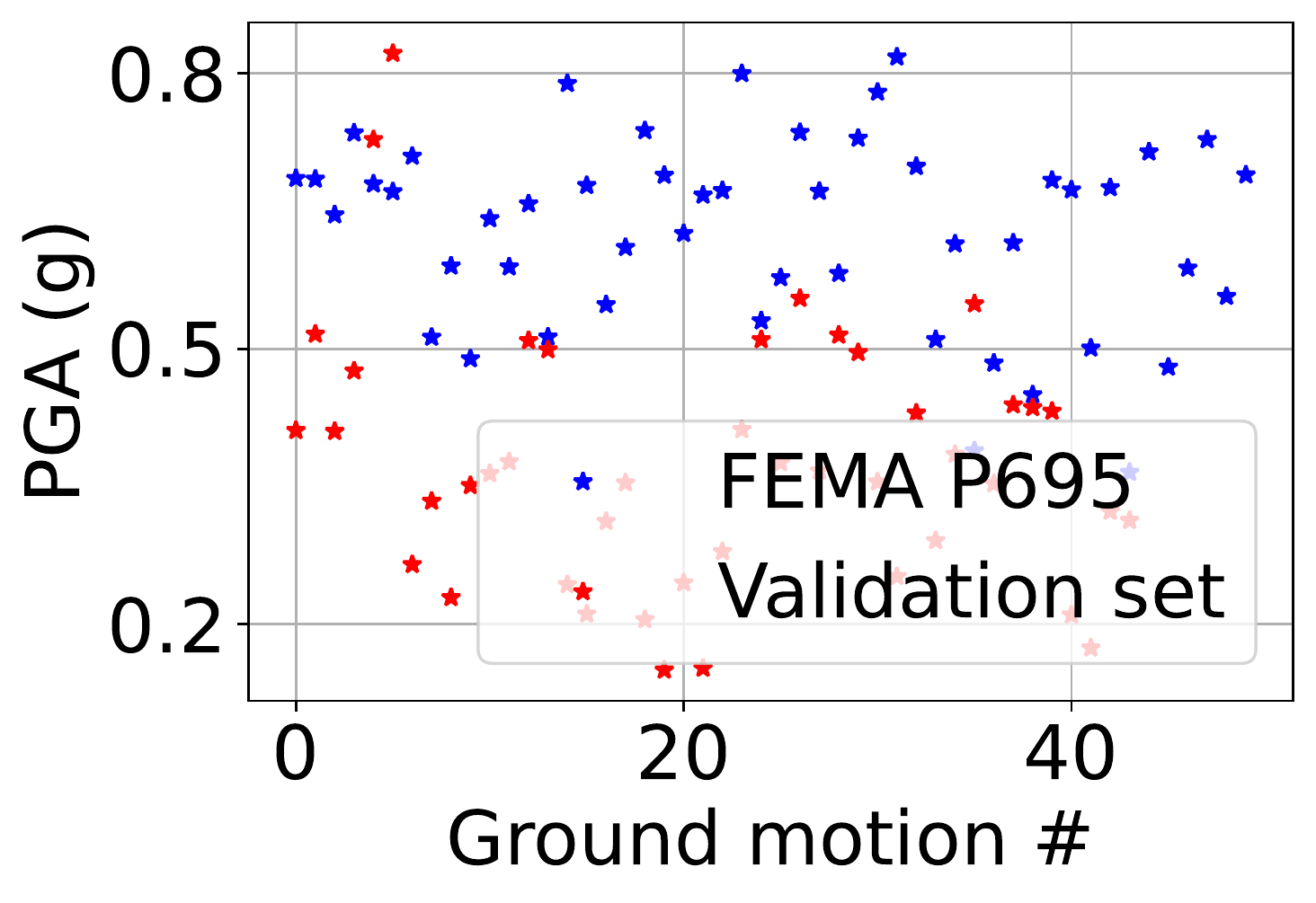}} 
\subfigure[\label{pdf_E} ]{\includegraphics[width=0.32\textwidth]{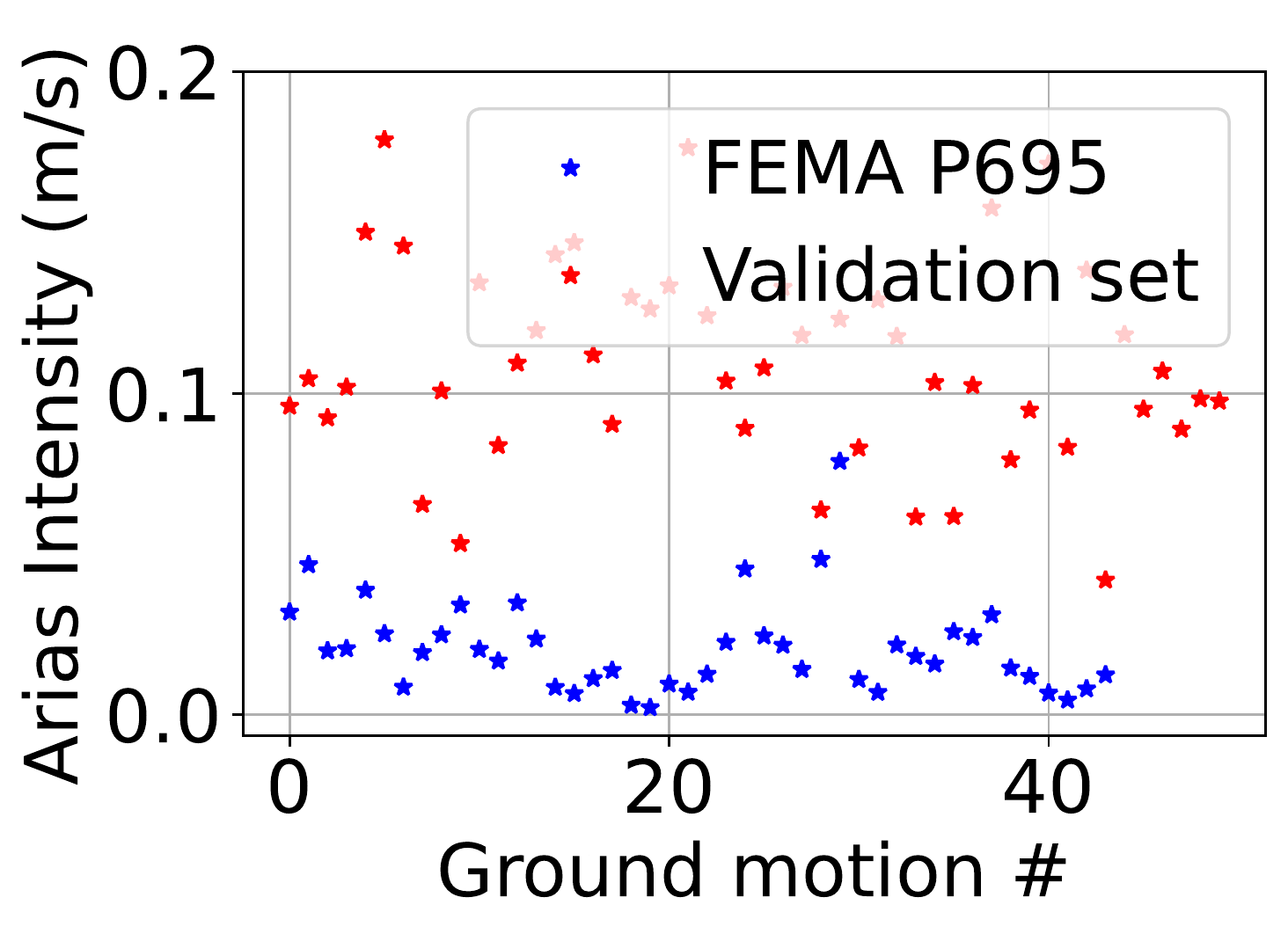}} 
\subfigure[\label{pdf_xi} ]{\includegraphics[width=0.32\textwidth]{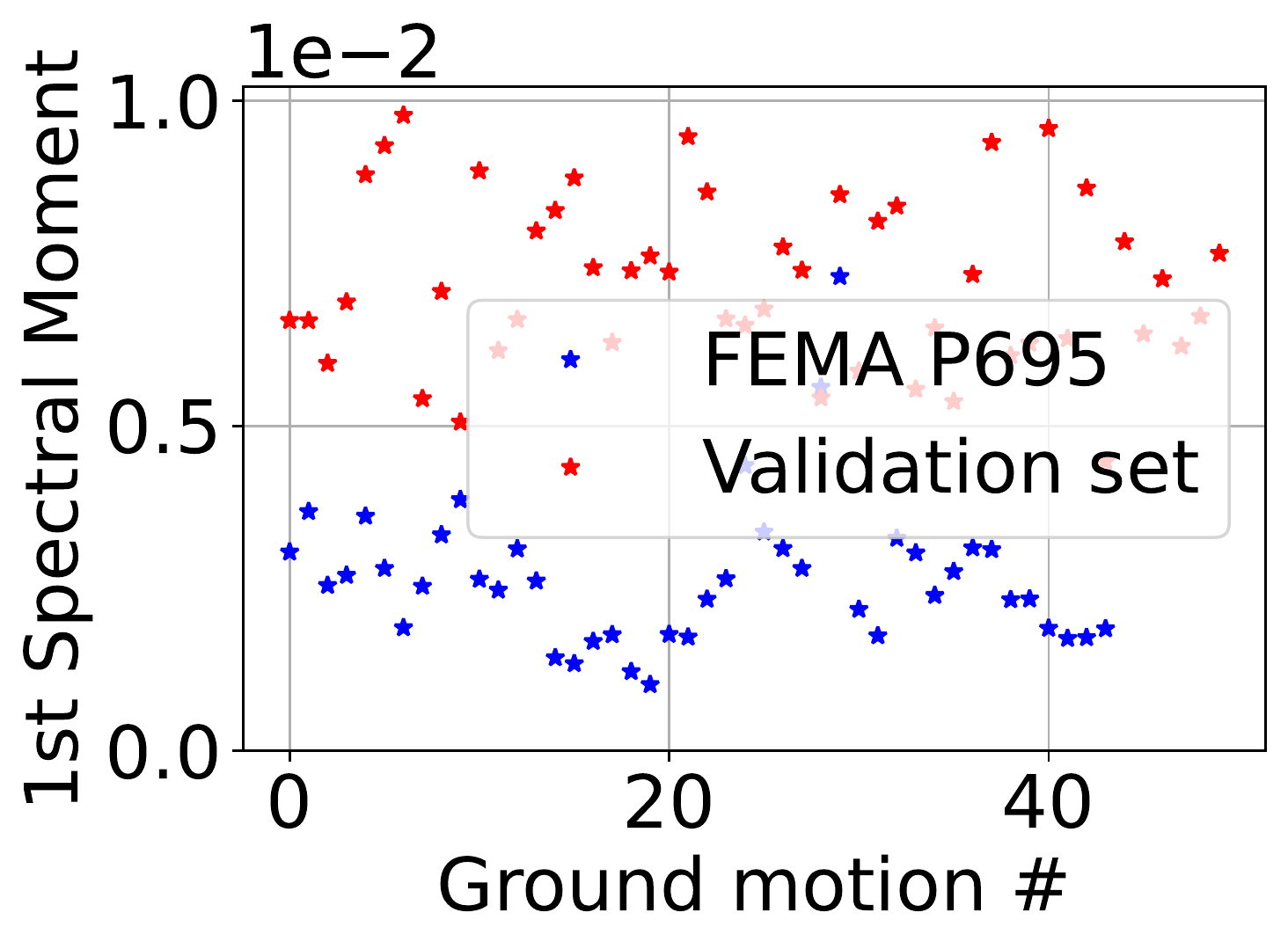}} 
\caption{Comparison of ground motion intensity measures of (a) PGA, (b) Arias intensity and (c)1st spectral moment of validation dataset with FEMA P695 ground motion suite.}
\label{IMS}
\end{center}
\end{figure}

The validation effort starts with generating 50 realizations of $\Teta$ which results in 50 ground motions ($\Teta_1$) and 50 material parameters ($\Teta_2$).  To ensure that the generated earthquakes are different from the initial suite of 22 ground motion pairs,  various earthquake intensity measures are shown in Figure \ref{IMS}. In Figure \ref{IMS} (a), (b) and (c), the peak ground acceleration (PGA), Arias Intensity (IA) and first spectral moment, respectively,  of the generated earthquake motion are compared with the original suite of earthquakes. As it can be noted that, these intensity measures are quite different from the original earthquakes and hence the generated earthquakes can be considered as new unseen earthquakes. 

\begin{figure}[h!]
\begin{center}
\subfigure[\label{1d_train} ]{\includegraphics[width=0.45\textwidth]{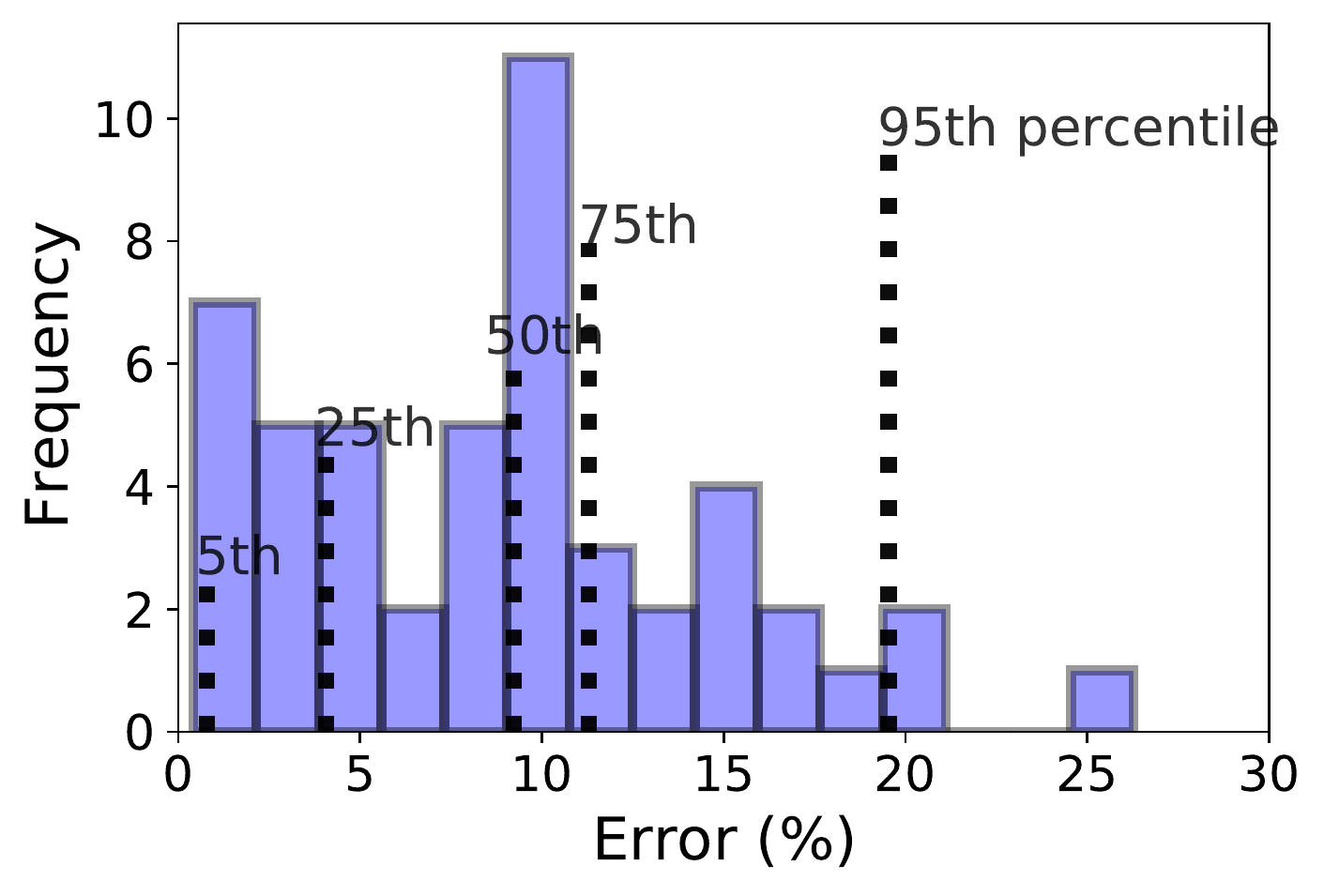}} 
\hspace*{0.1truecm}
\subfigure[\label{1d_test} ]{\includegraphics[width=0.45\textwidth]{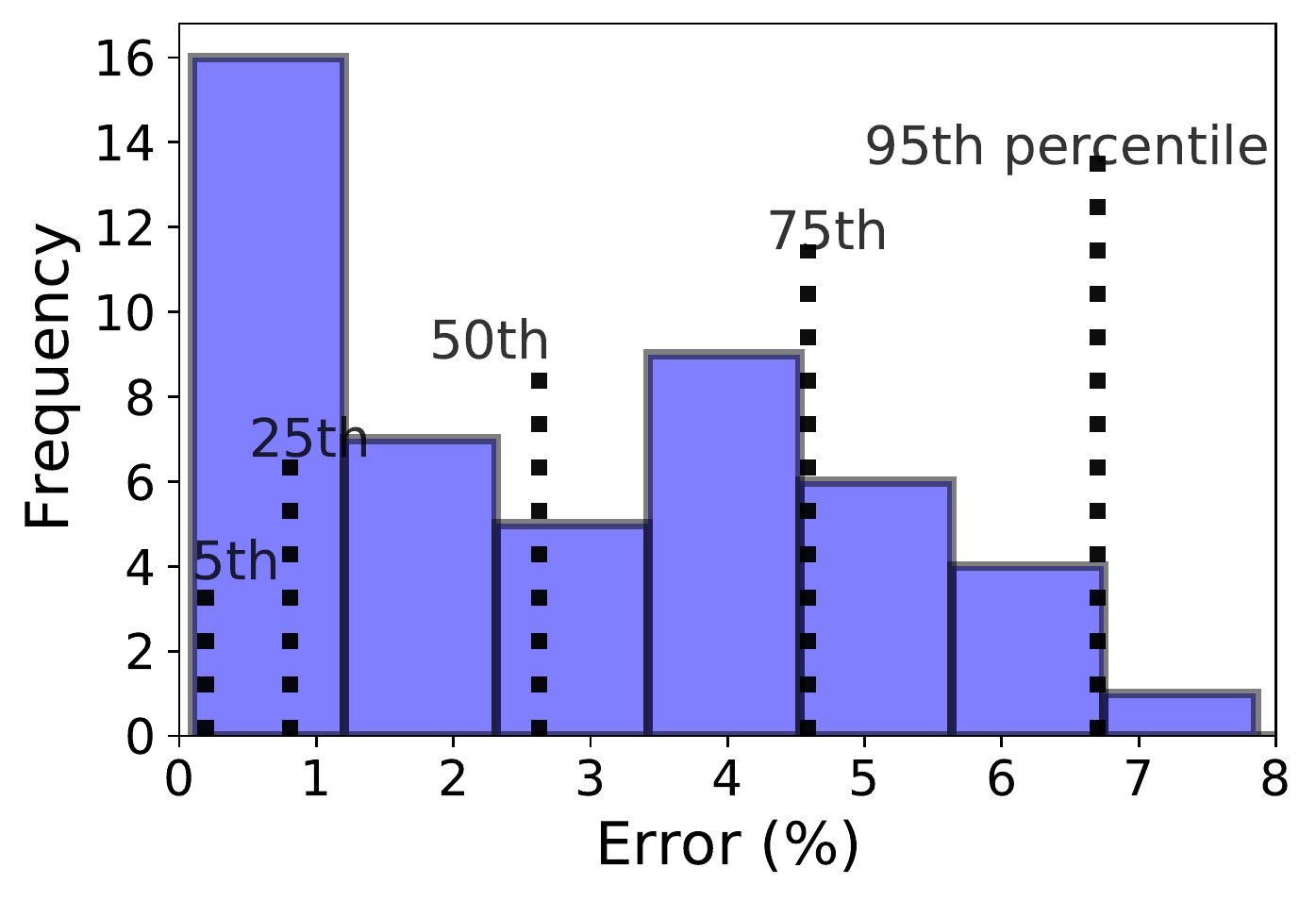}} 
\caption{Histogram of prediction error (\%) between DNN and FE estimate for (a) peak roof displacement and (b) peak floor acceleration of the one story  building when using generated earthquake.}
\label{histo_valid1_prd}
\end{center}
\end{figure}

\begin{figure}[h!]
\begin{center}
\subfigure[\label{1d_train} ]{\includegraphics[width=0.45\textwidth]{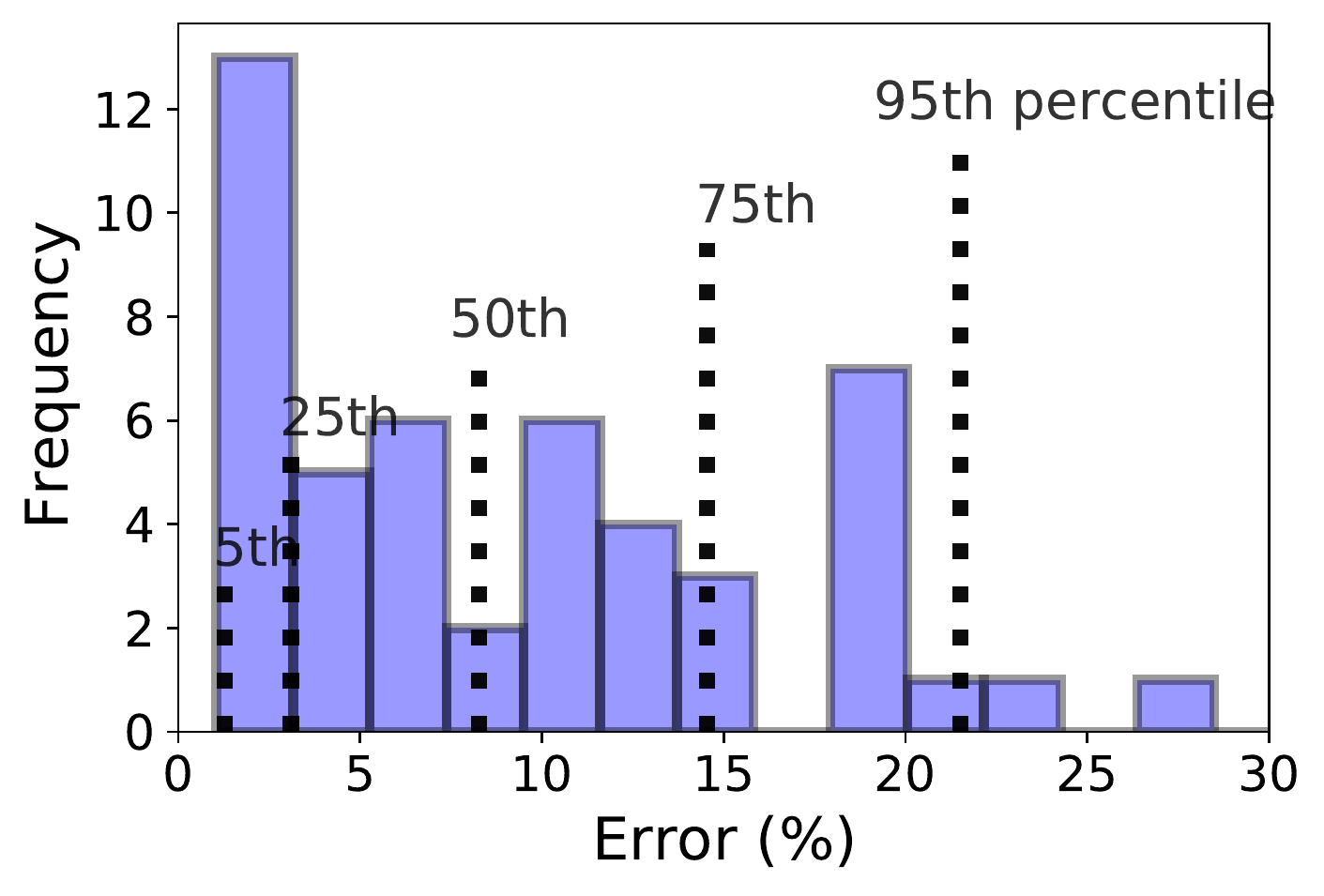}} 
\hspace*{0.1truecm}
\subfigure[\label{1d_test} ]{\includegraphics[width=0.45\textwidth]{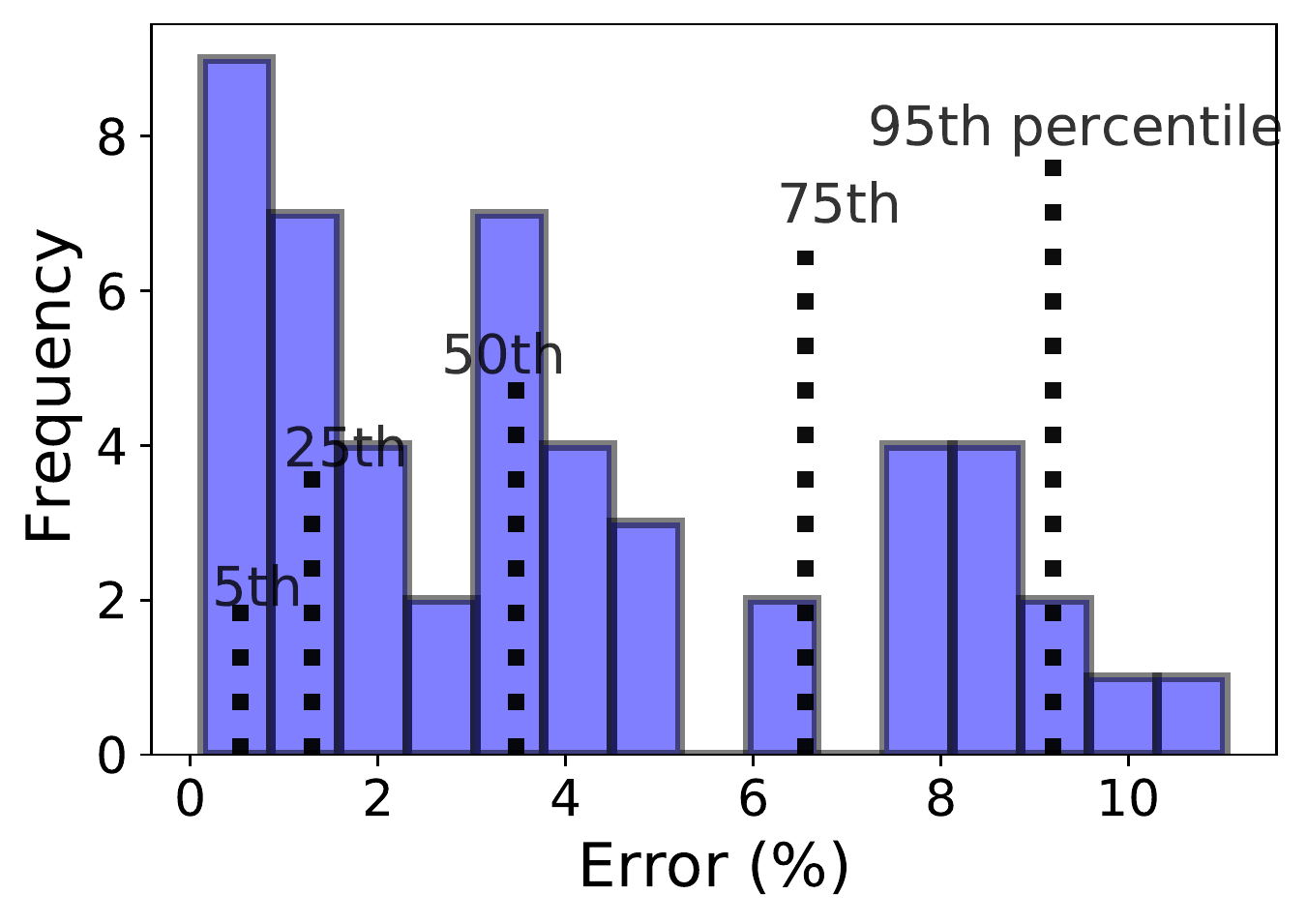}} 
\caption{Histogram of prediction error (\%) between DNN and FE estimate for (a) peak roof displacement and (b) peak floor acceleration of the three story building when using generated earthquake.}
\label{histo_valid1_pfa}
\end{center}
\end{figure}

The FE models were simulated for all the realization sets and the output $Y$ was obtained. The machine learning model was used to predict the output $\hat{Y}$  using the input $\Teta$.   The median error with respect to the true value of $Y$ as obtained from the FE model was 9\% for peak roof displacement and 3\% for peak floor acceleration for the 1-DoF case.  For the 3-DoF case, a median error of 8\% was observed for peak roof displacement and 3\% for peak floor acceleration. The histograms of the errors in predicting the response are shown in Figures \ref{histo_valid1_prd} and \ref{histo_valid1_pfa}, for the 1-DoF and 3-DoF, respectively. It can be observed that for both the structural systems, the 95th percentile error is less than 20\% for peak displacement and less than 10\% for peak floor acceleration. These low values of prediction errors were very promising, given the machine learning model is predicting the response for unseen generated earthquakes.

\subsection{Prediction for Loma Prieta earthquake}

\begin{figure}[h!]
\begin{center}
\subfigure[\label{normal} ]{\includegraphics[width=0.45\textwidth]{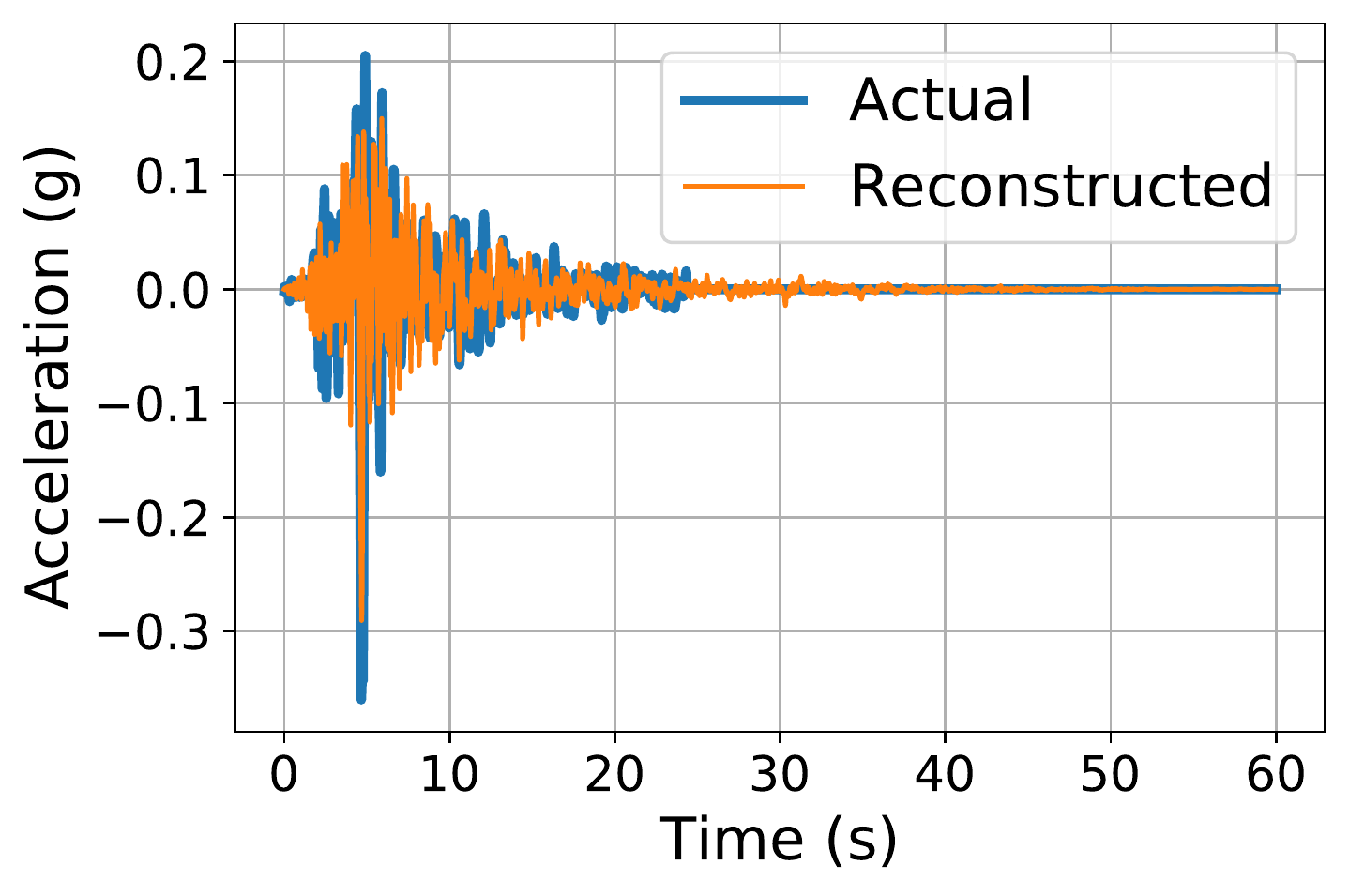}} 
\hspace*{0.2truecm}
\subfigure[\label{zoomed} ]{\includegraphics[width=0.45\textwidth]{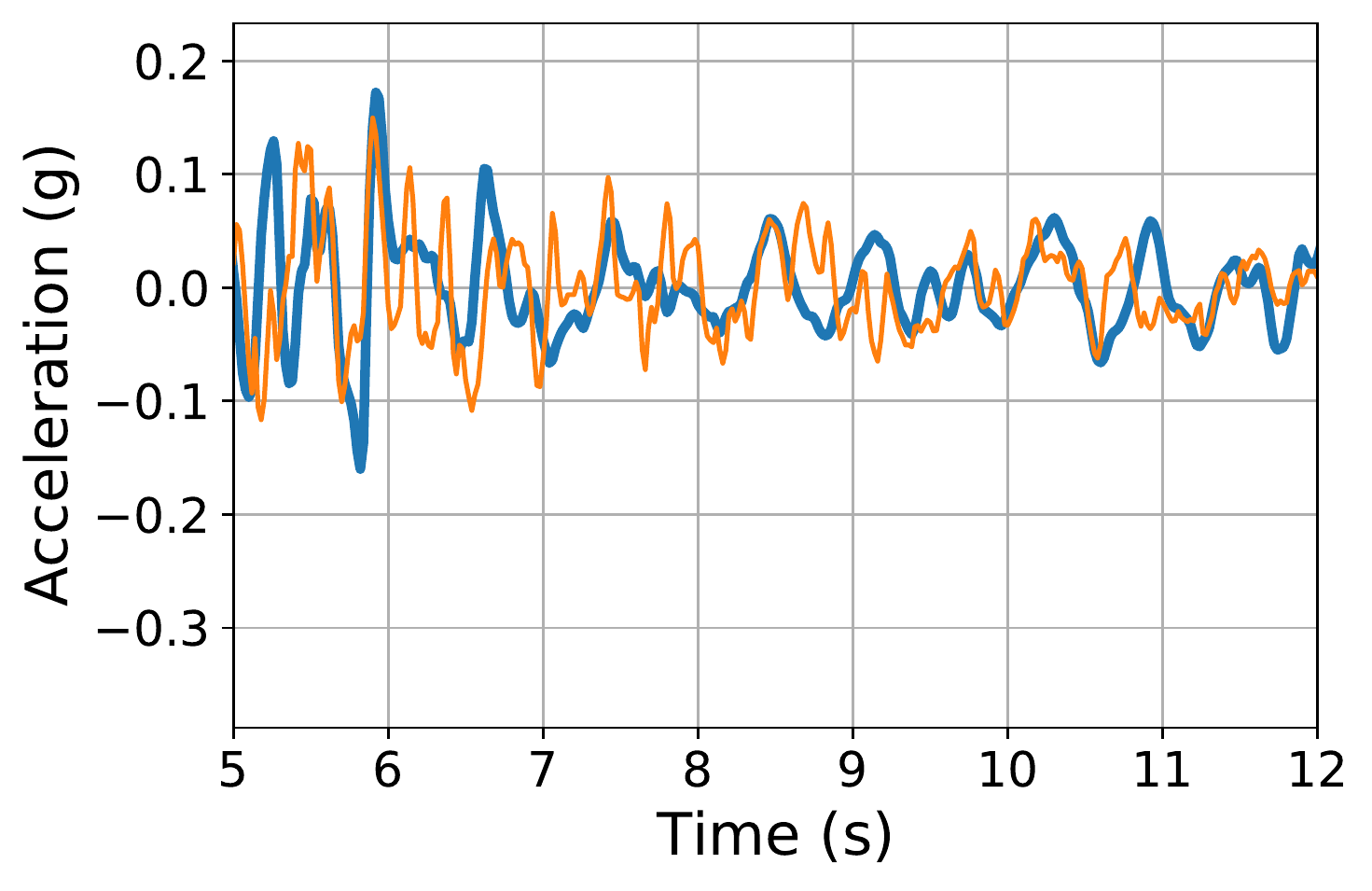}} 
\caption{(a) Comparison of Loma-Prieta earthquake and reconstructed earthquake using $U$ basis and weights obtained by projecting the earthquake onto $U$ basis.(b) Zoomed in strong motion portion of the time history.}
\label{loma_prieta}
\end{center}
\end{figure}

\begin{figure}[h!]
\begin{center}
\subfigure[\label{1d_train} ]{\includegraphics[width=0.45\textwidth]{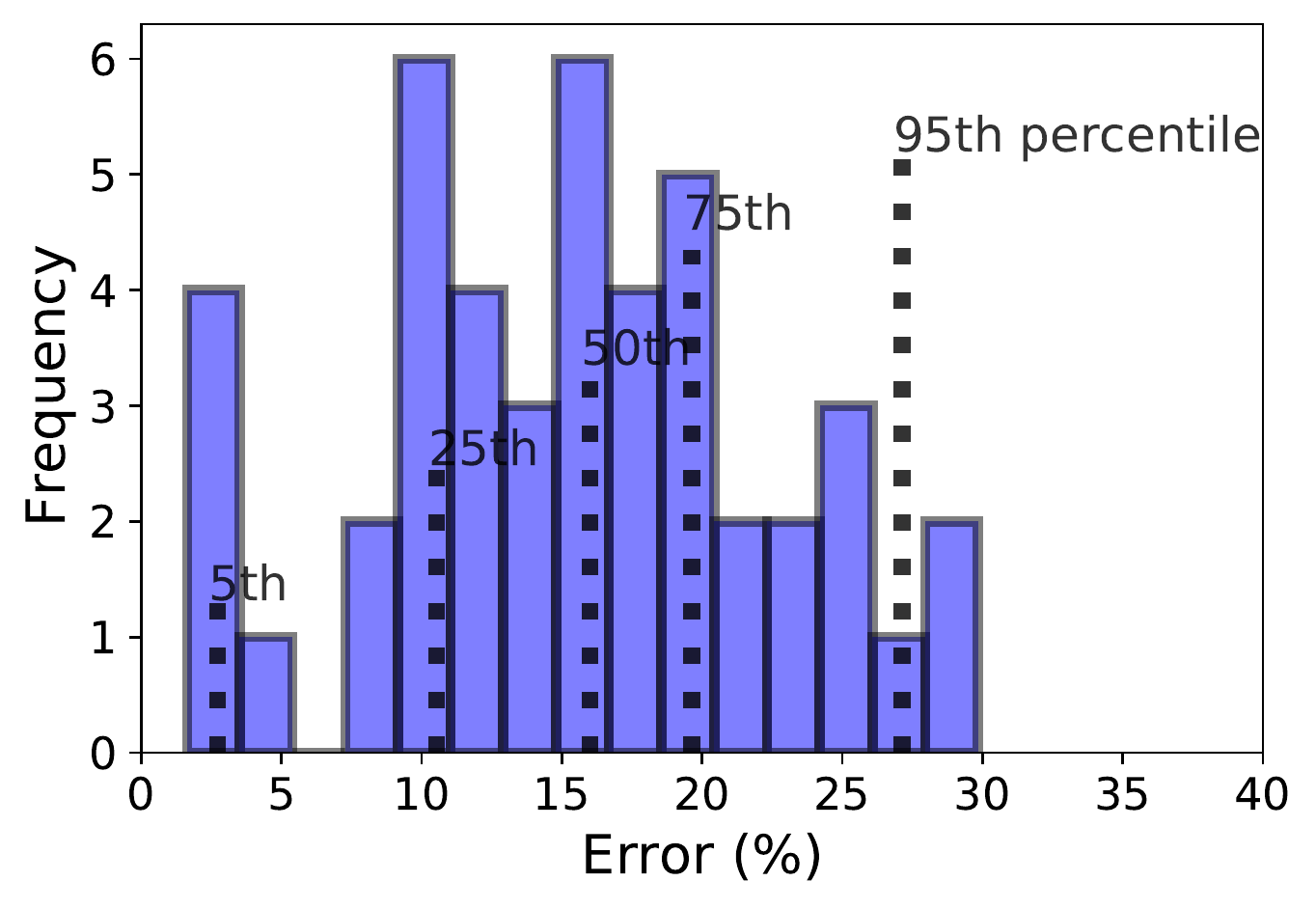}} 
\hspace*{0.1truecm}
\subfigure[\label{1d_test} ]{\includegraphics[width=0.45\textwidth]{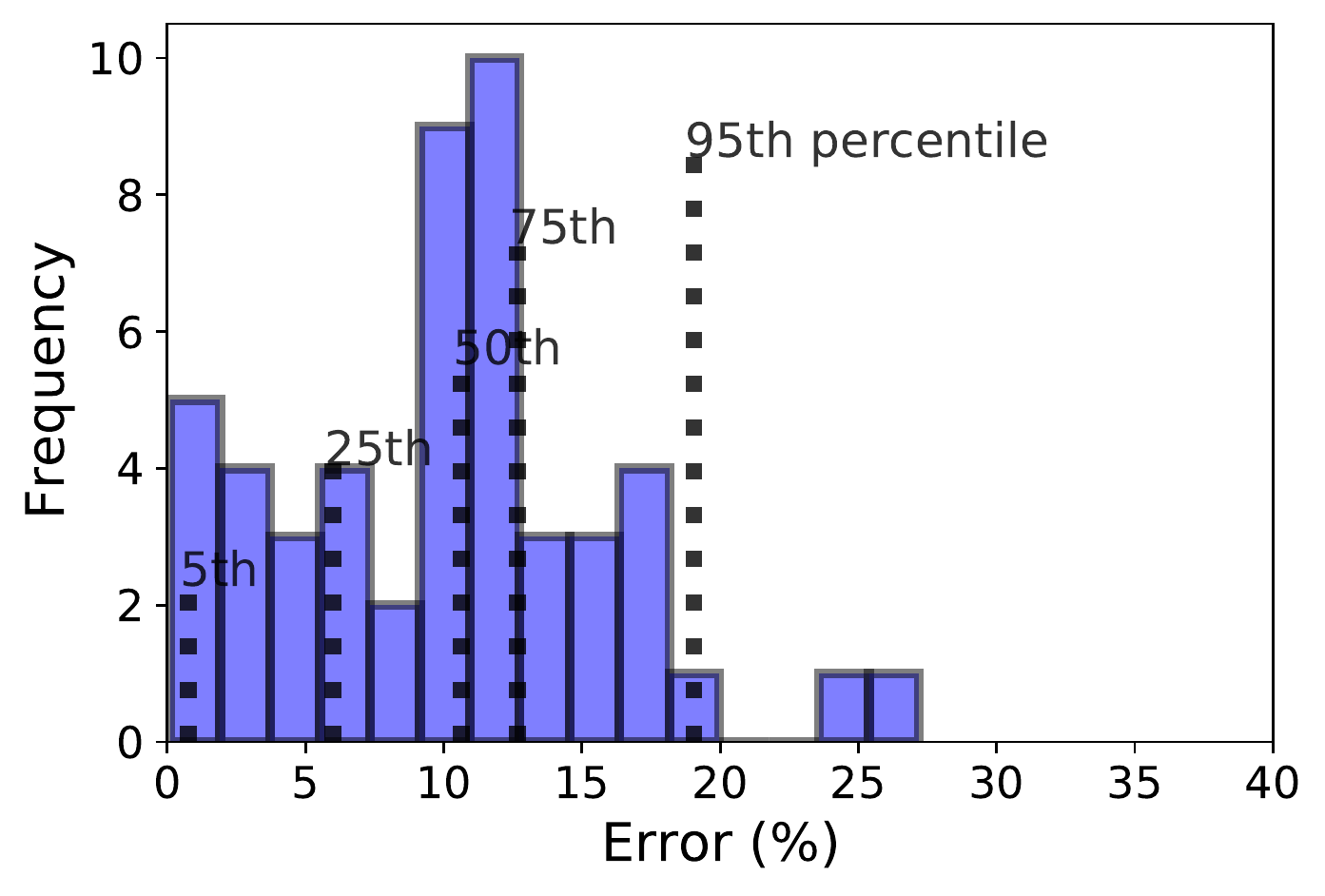}} 
\caption{Histogram of prediction error (\%) between DNN and FE estimate for (a) peak roof displacement and (b) peak floor acceleration of the one story building when using Loma-Prieta earthquake.}
\label{histo_validlp_prd}
\end{center}
\end{figure}

\begin{figure}[h!]
\begin{center}
\subfigure[\label{1d_train} ]{\includegraphics[width=0.45\textwidth]{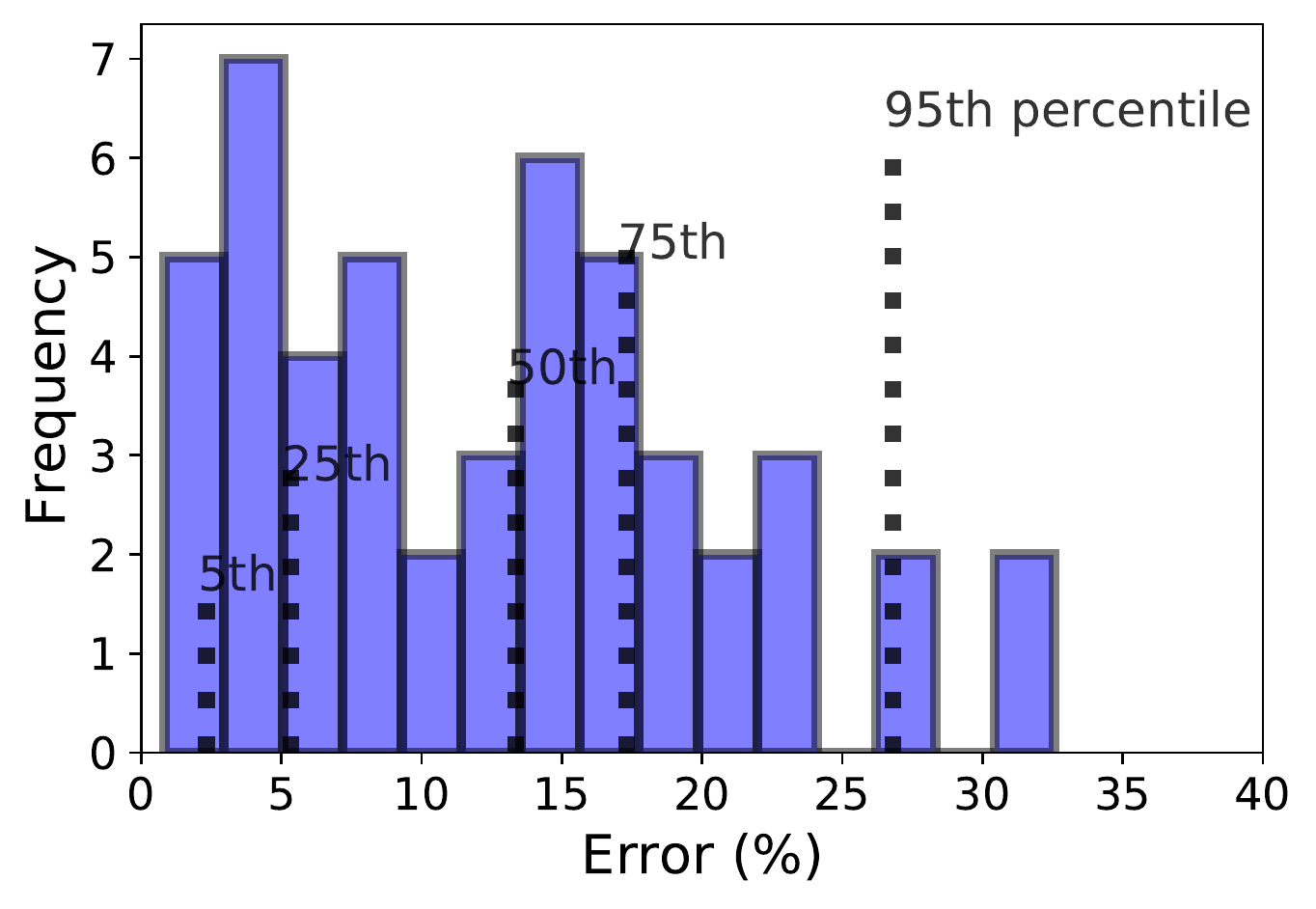}} 
\hspace*{0.1truecm}
\subfigure[\label{1d_test} ]{\includegraphics[width=0.45\textwidth]{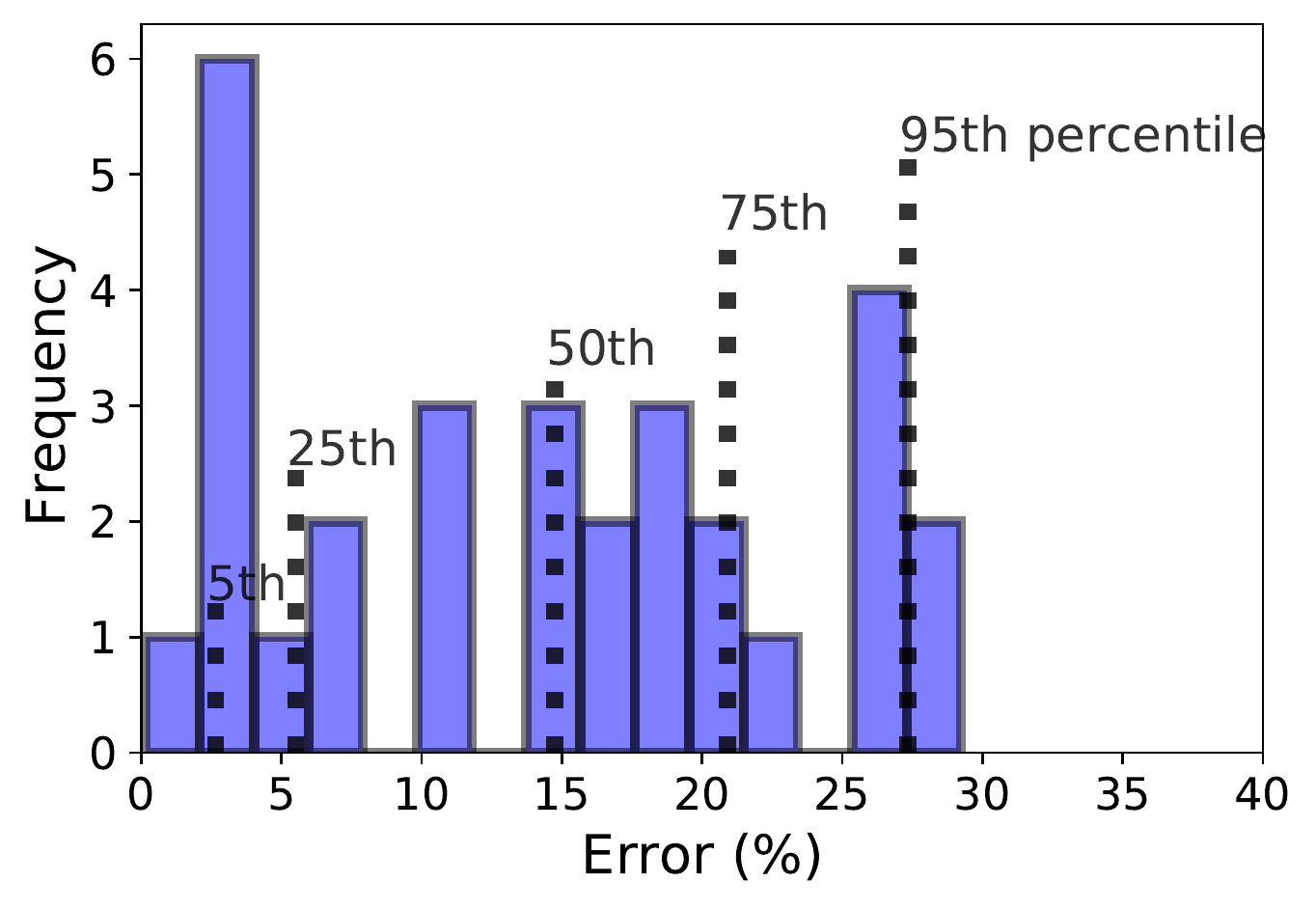}} 
\caption{Histogram of prediction error (\%) between DNN and FE estimate for (a) peak roof displacement and (b) peak floor acceleration of the three story building when using Loma-Prieta earthquake.}
\label{histo_validlp_pfa}
\end{center}
\end{figure}

Next,  the DNN surrogate models of both the 1-DoF and 3-DoF systems were validated by predicting the response for the Loma Prieta earthquake recorded at the station Hollister - South and Pine.  This ground motion is not a part of the initial suite of 22 earthquakes, and therefore, successful prediction for this record will demonstrate the robustness of the surrogate model developed through the proposed framework.  The earthquake is first projected onto the $U$ basis and its weights were obtained according to Equation \ref{PCA2}. The reconstructed time history obtained by multiplying the weights with the $U$  basis is superimposed on the original ground motion as shown in Figure \ref{loma_prieta}. It can be seen that earthquake constructed using basis functions and weights can reproduce the actual record with considerable accuracy and can capture the overall features of the ground motion.  The intensity measures for both the original earthquake and reconstructed earthquake also showed a close match with 5\%, 4\%, and 8\% error for PGA,  Arias intensity and spectral moment respectively. Since the earthquake is fixed,  $\Teta_1$ is deterministic.  50 realizations of $\Teta_2$ were obtained to produce 50 realization of the input vector $\Teta$.  The FE model was simulated with the original earthquake and $\Teta_2$ as material parameters to obtain the output $Y$.  The surrogate model was fed to $\Teta$ to obtain the output $\hat{Y}$.  For the 1-DoF system, it was observed that the surrogate model could predict the peak displacement and the acceleration with a median error of 16\% and 10\%, respectively,  while that for the 3-DoF system, the median errors were 14\% and 15\%, respectively.  The histograms of the prediction errors are provided in Figures \ref{histo_validlp_prd} and \ref{histo_validlp_pfa} for the 1DoF and 3DoF, respectively. As observed, the 95th percentile error is less than 30\% for both peak roof displacement and peak floor acceleration. Given that this ground motion was not considered a part of the initial suite of the 22 pairs,  these predictions are very encouraging and based on this, the authors are confident about the application of the proposed framework to predict structural response to far-field motions recorded at firm rock site.


\section{Conclusion}\label{section6}

Performance based earthquake engineering framework is a benchmark decision making tool. However, it is often limited by its high computational cost, incurred by the probabilistic non-linear finite element analysis step in the framework, while taking into account uncertainties in both earthquakes and constitutive parameters.  In this study,  the authors have proposed a machine learning based surrogate model framework to accurately replicate finite element model response of a building for a given set of material parameters and far field ground motions recorded on firm rock site.  Challenge lies in training the surrogate model to predict for unseen earthquakes because performance depends on the characteristic earthquake features used to train them.  Traditional characterization techniques using intensity measures,  like Arias intensity,  PGA,  PGV, among others,  are relevant in giving information about a particular earthquake.  However,  due to the lack of an inverse relationship,  these measures cannot be used to generate ground motions that can be used in training, which is important to predict response for unseen earthquakes.  Moreover,  given a set of ground motion characteristics and model features,  different machine learning models perform differently. Therefore,  one needs to select a machine learning model that is specific to the input-output dataset.  To this end,  the authors propose a framework that a) characterizes ground motion in a fashion that can be used for training machine learning models and b) selects the best machine learning model that accurately replicates FE model output.  

The proposed framework  starts with selecting a representative suite of far field ground motions recorded on firm rock site.  Next,  an orthonormal basis is chosen using singular value decomposition that spans the space of the ground motion suite.  Assuming the weights of the basis vectors as random vectors and the constitutive parameters as random variables,  a large set of earthquakes and constitutive parameters are generated by random sampling.  The finite element model is simulated using these earthquakes and constitutive parameters.  Randomly generated weights and material parameters values along with the finite element model output serve as training data-set for the machine learning model.  Competing machine learning models are used such as decision trees,  random forests,  support vector regression and deep neural networks,  to map the input (weights of the basis vectors and constitutive model parameters) to the output (finite element model response).  The best model is then proposed based on some performance metric ($R^2$ error).  The proposed framework is illustrated on one-story and three-story buildings represented by spring-mass-damper systems, subjected to far field ground motions to predict peak roof displacement and peak floor acceleration.  Among the competing set of machine learning models, it was shown that deep neural networks can be used to accurately estimate non-linear response of the buildings subject to unseen earthquakes. This was validated by using the model to estimate building response for earthquakes and constitutive model parameters that were not a part of the training set,  for which the surrogate models could predict the response of the FE models with reasonable accuracy (median error less than 20\%). Thus, the proposed framework bolstered by validation results on unseen earthquakes provides a firm basis for the validity and applicability of the machine learning based surrogate model in predicting non-linear building response. 
 
\section{Future work}

Despite the very good results, this framework has certain limitations that should be brought to the notice of the reader and would be addressed in future studies. These are summarized in the following.
\begin{itemize}
\item The authors are characterizing the far-field motions in this study for an initial demonstration of the proposed framework. These records are good representation of the characteristics of far-field motions on firm soil representing soil classes C and D and originating from strike slip and reverse thrust faults. In case one has to predict structural response for near-field motions or motions with long-period pulses (expected in soft soil), the initial ground motion suite needs to be expanded.
\item The illustrated framework uses SVD to span the space of the selected ground motion suite. This is a linear representation. Non-linear representation like radial basis functions , fourier basis, and autoencoders \cite{autoencoder} would be used in future studies.
\item Implementation of this framework would be done to more complicated, high fidelity finite element models. Computational judiciousness would be accomplished by using multi-fidelity deep neural network surrogate models where the surrogate for a low fidelity model would aid in directing the solution direction for the surrogate of a high fidelity model. This would help in achieving convergence for a low number of training data points. 
\end{itemize}

\bibliographystyle{elsarticle-num}
\bibliography{SFEM,SoilProperties,allrefs,alex,Parida,Parida2,sysid,alexPaper,references}

\end{document}